\documentclass[10pt, a4paper, twocolumn]{article}
\usepackage[a4paper,margin=2cm]{geometry}
\usepackage[utf8]{inputenc}
\usepackage[normalem]{ulem}
\usepackage[dvipsnames]{xcolor}
\usepackage[hidelinks]{hyperref} %
\usepackage{glossaries}
\usepackage{xspace}
\usepackage{tikz}
\usepackage{amsmath}
\usepackage{physics} 
\usepackage{caption}
\usepackage{abstract}
\usepackage{booktabs}

\usepackage[raggedright]{titlesec}

\usepackage{array,multirow} %
\usepackage{pifont} %

\usepackage{amssymb}
\usepackage{float}
\usepackage{amsfonts} 
\usepackage{algorithm2e}
\RestyleAlgo{ruled}
\usepackage{algpseudocode}
\usepackage{pgfplots}
    \pgfplotsset{
        compat=1.3,
    }
\usepackage{subfig}
\usepackage{multirow}
\usepackage{afterpage}

\usepackage{bookmark}
\usepackage{chngcntr}

\glsdisablehyper %
\usepackage{xspace}
\usepackage{xr}
\usepackage{csquotes}

\newcommand{\edited}[2]{{}{{#2}}} %

\newacronym{ai}{AI}{Artificial Intelligence}
\newacronym{amax}{ActMax}{Activation Maximization}
\newacronym{ch}{CH}{Clever Hans}
\newacronym{cnn}{CNN}{Convolutional Neural Network}
\newacronym{crc}{CRP}{Concept Relevance Propagation}
\newacronym{dl}{DL}{Deep Learning}
\newacronym{dnn}{DNN}{Deep Neural Network}
\newacronym{ga}{GA}{Gradient Ascent}
\newacronym{gan}{GAN}{Generative Adversarial Network}
\newacronym{hitl}{HITL}{Human in the Loop}
\newacronym{lrp}{LRP}{Layer-wise Relevance Propagation}
\newacronym{ml}{ML}{Machine Learning}
\newacronym{mlp}{MLP}{Multilayer Perceptron}
\newacronym{rmax}{RelMax}{Relevance Maximization}
\newacronym{rnn}{RNN}{Recurrent Neural Network}
\newacronym{tcav}{TCAV}{Testing With Activation Vectors}
\newacronym{xai}{XAI}{eXplainable Artificial Intelligence}

\newacronym{classspecific}{CS}{class-specific}
\newacronym{samplespecific}{SS}{sample-specific}
\newacronym{localized}{L}{localized}
\newacronym{examples}{E}{examples}
\newacronym{partially}{PP}{partially}
\newacronym{features}{F}{features}
\newacronym{architecture}{A}{architecture} 
\newacronym{dataNlabels}{D\&L}{data and labels}
\newacronym{training}{T}{training}

\newcommand{\ie}{{i.e.}\xspace}
\newcommand{\sm}{Supplementary Note}
\newcommand{\sms}{Supplementary Notes}

\newcommand{\cf}{{cf.}\xspace}
\newcommand{\wrt}{{wrt.}\xspace}
\newcommand{\eg}{{e.g.}\xspace}

\newcommand{\x}{\mathbf{x}}
\newcommand{\act}{a}
\newcommand{\preact}{z}
\newcommand{\weight}{\mathbf{W}}
\newcommand{\comment}[1]{}

\DeclareMathOperator*{\argmax}{argmax}
\usepackage[normalem]{ulem}

\title{\textbf{From Attribution Maps to Human-Understandable Explanations through Concept Relevance Propagation}} %

\author{Reduan Achtibat$^{1,\ast}$ \and
Maximilian Dreyer$^{1,\ast}$ \and 
Ilona Eisenbraun$^1$ \and
Sebastian Bosse$^1$ \and
Thomas Wiegand$^{1,2,3}$ \and
Wojciech Samek$^{1,2,3,\dagger}$ \and
Sebastian Lapuschkin$^{1,\dagger}$}

\date{
\footnotesize
$^1$ Fraunhofer Heinrich-Hertz-Institute, 10587 Berlin, Germany\\
$^2$ Technische Universität Berlin, 10587 Berlin, Germany\\
$^3$ BIFOLD – Berlin Institute for the Foundations of Learning and Data, 10587 Berlin, Germany\\
$^\ast$ contributed equally\\
$^\dagger$ corresponding authors: \texttt{\{sebastian.lapuschkin,wojciech.samek\}@hhi.fraunhofer.de}
}

\newcommand\blfootnote[1]{
    \begingroup
    \renewcommand\thefootnote{}\footnote{#1}
    \addtocounter{footnote}{-1}
    \endgroup
}

\begin{document}
\twocolumn[
  \begin{@twocolumnfalse}
    \maketitle
    \begin{abstract}
    The field of \gls{xai} aims to bring transparency to today's powerful but opaque deep learning models. While local \gls{xai} methods explain individual predictions in form of attribution maps, thereby identifying \emph{where} important features occur (but not providing information about what they represent), global explanation techniques visualize \emph{what} concepts a model has generally learned to encode. Both types of methods thus only provide partial insights and leave the burden of interpreting the model's reasoning to the user. In this work we introduce the Concept Relevance Propagation (CRP) approach, which combines the local and global perspectives and thus allows answering both the ``where'' and ``what'' questions for individual predictions. We demonstrate the capability of our method in various settings, showcasing that CRP leads to more human interpretable explanations and provides deep insights into the model's representation and reasoning through concept atlases, concept composition analyses, and quantitative investigations of concept subspaces and their role in fine-grained decision making.
    \newline
    \end{abstract}
  \end{@twocolumnfalse}
]

\blfootnote{Paper published in Nature Machine Intelligence (year 2023, volume 5, pages 1006–1019, DOI: \url{https://doi.org/10.1038/s42256-023-00711-8}).}

\section{Introduction} %
\label{sec:introduction}
Considerable advances have been made in the field of \gls{ml}, with especially \glspl{dnn}~\cite{lecun2015deep} achieving impressive performances on a multitude of domains \cite{dai2021coatnet,senior2020improved,jaderberg2019human}. However, the reasoning of these highly complex and non-linear \glspl{dnn} is generally not obvious \cite{rudin2019stop,samek2021explaining}, and as such, their decisions may be (and often are) biased towards unintended or undesired features \cite{stock2018convnets, lapuschkin2019unmasking, schramowski2020making, anders2022finding}. This in turn hampers the transferability of \gls{ml} models to many application domains of interest, \eg, due to the risks involved in high-stakes decision making~\cite{rudin2019stop}, or the requirements set in governmental regulatory frameworks~\cite{goodman2017european} and guidelines brought forward~\cite{eu2019building}.

\begin{figure*}[th]
    \centering
    \includegraphics[width=1.0\linewidth]{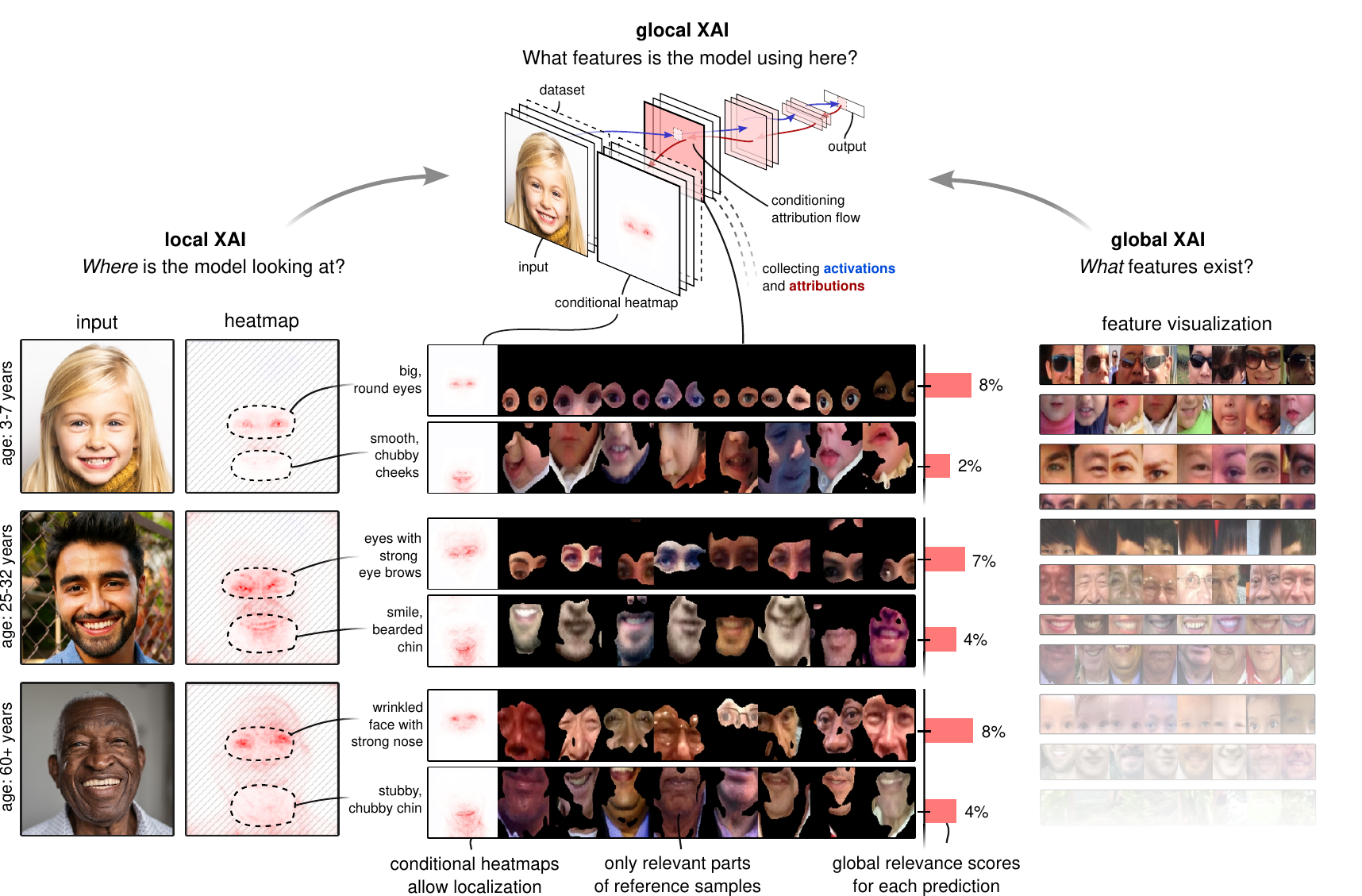}
    \caption{
    Glocal \gls{xai} can tell which features exist and how they are used for predictions by unifying local and global \gls{xai}. (\textit{Left}): Local explanations visualize which input pixels are relevant for the prediction. Here, the model focuses on the eye region for all three predictions. However, what features in particular the model has recognized in those regions remains open for interpretation by the user. (\emph{Right}): By finding reference images that maximally represent particular (groups of) neurons, global \gls{xai} methods give insight into the concepts generally encoded by the model. However, global methods alone do not inform which concepts are recognized, used and combined by the model in per-sample inference. (\emph{Center}): Glocal \gls{xai} can identify the relevant neurons for a particular prediction (property of local XAI) and then visualize the concepts these neurons encode (property of global XAI). Further, by using concept-conditional explanations as a filter mask, the concepts' defining parts can be highlighted in the reference images, which largely increases interpretability and clarity. Here, the topmost sample has been predicted into age group (3-7) due to the sample's large irides and round eyes, while the middle sample is predicted as (25-32), as more of the sclera is visible and eyebrows are more apparent. For the bottom sample, the model has predicted class (60+) based on its recognition of heavy wrinkles around the eyes and on the eyelids, and pronounced tear sacs next to a large knobby nose.
    }
    \label{fig:introduction:overview}
\end{figure*}

In order to alleviate the ``black box'' problem and gain insights into the model and its predictions, the field of \glsfirst{xai} has recently emerged. In fact, multitudes of \gls{xai} methods have been developed that are able to provide explanations of a model's decision while approaching the subject from different angles, \eg, based on gradients \cite{morch1995visualization, sundararajan2017axiomatic}, as modified backpropagation processes \cite{bach2015pixel, springenberg2015simplicity, shrikumar2017learning, murdoch2018beyond}, by probing the model's reaction to changes in the input \cite{zeiler2014visualizing, ribeiro2016lime,blucher2022preddiff} or visualizing stimuli specific neurons react strongly to~\cite{erhan2009visualizing, olah2017feature}. The field can roughly be divided into \emph{local \gls{xai}} and \emph{global \gls{xai}}. Methods from local \gls{xai} commonly compute attribution maps in input space highlighting input regions or features, which carry some form of importance to the individual  prediction process (\ie, with respect to a specific sample). However, the visualization of important input regions is often of only limited informative value on its own as it does not tell us what features in particular the model has recognized in those regions, as Figure \ref{fig:introduction:overview} illustrates. 

Furthermore, attribution maps can be understood as a superposition of many different model-internal decision sub-processes (\eg, see~\cite{Kindermans2018LearningHT}), working through various transformations of the same input features and culminating to the final prediction. Many intricacies are lost with local explanation techniques producing only a singular attribution map in the input space per prediction outcome. The result might be unclear, imprecise or even ambiguous explanations.

Assuming for example an image classification setting and an attribution map computed for a specific prediction, it might be clear \emph{where} (in terms of pixels) important information can be found, but not \emph{what} this information is, \ie, what characteristics of the raw input features the model has extracted and used during inference, or whether this information is a singular characteristic or an overlapping plurality thereof. This introduces many degrees of freedom to the interpretation of attribution maps generated by local \gls{xai}, rendering a precise understanding of the models' internal reasoning a difficult task.

Global \gls{xai}, on the other hand, attempts to address the very issue of understanding the \emph{what} question, \ie, which features or concepts have been learned by a model or play an important role in a model's reasoning in general. Some approaches from this category synthesize example data in order to reveal the concept a particular neuron activates for \cite{erhan2009visualizing, szegedy2013intriguing,mahendran2015understanding, mordvintsev2015inceptionism,olah2017feature}, but do not inform which concept is in use in a specific classification or how it can be linked to a specific output. From these approaches, we can at most obtain a global understanding of all possible features the model can use, but how these features interact with each other given some specific data sample and how the model infers a decision remains hidden. Other branches of global \gls{xai} propose methods, \eg, to test a model's sensitivity to a priori known, expected or pre-categorized stimuli \cite{kim2018interpretability, rajalingham2018large, bau2017network, bau2020understanding}. These approaches require labeled data, thus limiting, and standing in contrast to, the exploratory potential of local \gls{xai}.

Some recent works have begun to bridge the gap between local and global \gls{xai} by, for example, drawing weight-based graphs that show how features interact in a global, yet class-specific scale, but without the capability to deliver explanations for individual data samples \cite{hohman2019summit, liu2016towards}. Others plead for creating inherently explainable models in the hope of replacing black box models \cite{rudin2019stop}. These methods, however, either require specialized architectures, data and labels, or training regimes (or a combination thereof)~\cite{chen2019looks,chen2020concept} and do not support the still widely used off-the-mill end-to-end trained \gls{dnn} models with their extended explanation capabilities. A detailed discussion of related work can be found in \sm~\ref{sec:appendix:relatedwork}.
  
In this work, we connect lines of local and global \gls{xai} research by introducing \gls{crc} and \gls{rmax}, a set of next-generation XAI techniques that explain individual predictions in terms of localized and human-understandable concepts. Other than the related state-of-the-art, \gls{crc} and \gls{rmax} answer both the “where” and “what” questions of \gls{ml} model inference, thereby providing deep insights into the model’s reasoning process. As post-hoc XAI methods, \gls{crc} and \gls{rmax} can be applied to (almost) any \gls{ml} model with no extra requirements on the data, model or training process. We demonstrate on multiple datasets, model architectures and application domains, that CRP-based analyses allow one to (1) gain insights into the representation and composition of concepts in the model as well as quantitatively investigate their role in prediction, (2) identify and counteract Clever Hans filters \cite{lapuschkin2019unmasking} focusing on spurious correlations in the data, and (3) analyze whole concept subspaces and their contributions to fine-grained decision making.

Analogously to Activation Maximization \cite{nguyen2016synthesizing}, our proposed \glsdesc{rmax} approach searches for the most important (in terms of relevance, not activation) examples for latent encodings in, \eg, the training dataset.
Together, \gls{crc} and \gls{rmax} show their advantages in a conducted user study comparing our proposed techniques to various traditional attribution map-based approaches.
Finally, where transparency on unique samples are promptly required, the computational efficiency and ease of application of \gls{crc} and \gls{rmax} quickly provide valuable insights into the model’s representation and decision-making to the human user.

In summary, by lifting XAI to the concept level, \gls{crc} and \gls{rmax} open up new ways to analyze, debug and interact with ML models, which can be particularly beneficial for safety-critical applications and ML-supported investigations in the sciences.

\section{Methods in Brief}
\label{sec:methods_in_brief}

This section provides a brief overview over the methodological contributions of this work, \ie the \glsdesc{crc} and \glsdesc{rmax}.
A more detailed description of our methods can be found in Section~\ref{sec:methods}.

\noindent

\begin{figure*}[t!] 
    \centering
    \includegraphics[width=1\textwidth]{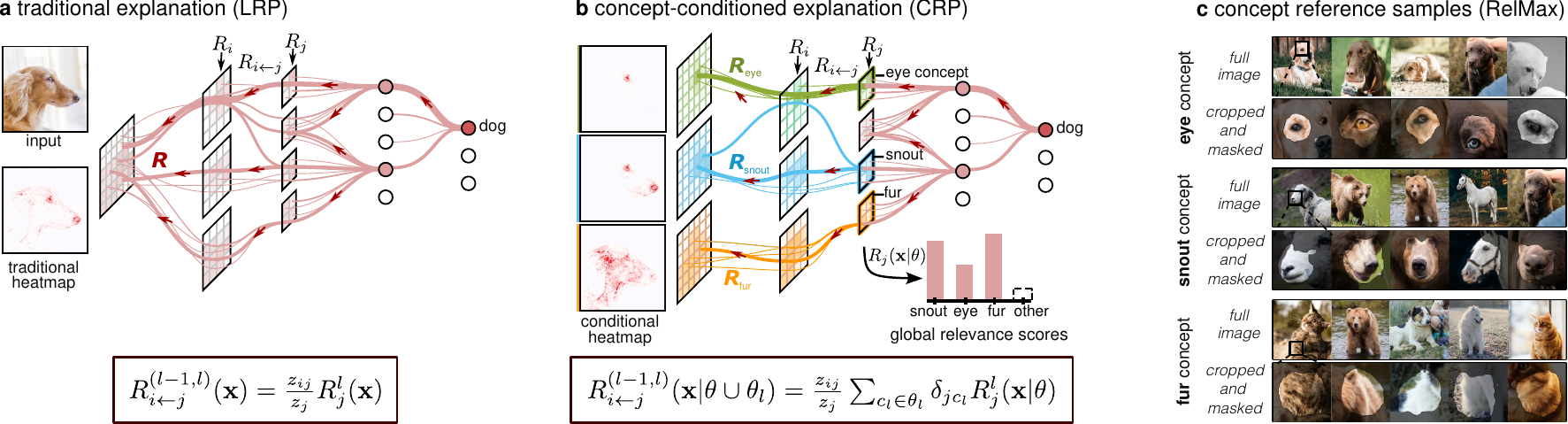}
    \caption{
    Brief overview over the methodological contributions of this work. 
    \textbf{a)} Traditional backpropagation-based methods such as \gls{lrp} propagate relevance scores backwards through the network
    culminating into single attribution map.
    \textbf{b)} By conditioning on a concept encoded by a hidden layer channel of the network, \gls{crc} allows to compute concept-conditional explanations. 
    \textbf{c)} To provide a semantic meaning for latent model structures, 
    we propose with \gls{rmax} to visualize input samples where the latent structure was strongly relevant for a prediction.
    We can further highlight the semantics by only displaying the relevant input parts according to concept-specific explanations as introduced in \textbf{b}.
    }
    \label{fig:methods_in_brief}
\end{figure*}

\textbf{Concept Relevance Propagation in brief}.
\gls{lrp} \cite{bach2015pixel} is a popular method for explaining predictions of a neural network by attributing relevance values to individual input dimensions (\eg, pixels of images). In this process, relevance is propagated backwards through the network, starting from the output until the input layer (see Figure~\ref{fig:methods_in_brief}a), which also provides relevance values for each intermediate element of the model (\eg, channel of an intermediate layer).
Since  literature suggests that latent structures of neural networks are encoding abstract human-understandable concepts with distinct semantics, especially in higher layers \cite{zhou2015object, olah2017feature, radford2017learning, bau2020understanding, cammarata2020thread, goh2021multimodal}, the channel-wise relevance values can be interpreted as scores quantifying the importance of the corresponding concepts in the inference process.

\gls{crc} is an extension of \gls{lrp}, which disentangles the relevance flows associated with concepts learned by the model via conditional backpropagation. Thus, it allows to compute concept-conditional relevance maps $R(\x|\theta)$, where $\x$ represents the data point the model has predicted for and $\theta$ describes a set of conditions (\ie, specifying the explained output category (\eg, ``dog'') and concepts as learned and distinctly encoded by model components (\eg, ``fur'')),  determining the flow of relevance via controlled masking operations in the backward process (see Section~\ref{sec:methods} for technical details). These concept-conditional explanations show us, e.g., in which part of the image the concepts (encoded in hidden layer channels) ``fur'' or ``eye'' are present (the \emph{where} question) and how much they contribute to the prediction. For the example in Figure~\ref{fig:methods_in_brief}b it turns out that the concept ``fur'' is more relevant than the concept ``eye'' for the prediction ``dog'', which is not obvious when looking at explanations from \gls{lrp} (Figure~\ref{fig:methods_in_brief}a) or other local attribution methods (\eg \cite{selvaraju2017grad, smilkov2017smoothgrad, shrikumar2017learning, lundberg2017Shap}), where the contributions of all concepts are superimposed into a single attribution map.

Please note that condition sets $\theta$ can be chosen by the human stakeholder (i.e., depending on the task), or as we prefer to do in this paper, they can be configured automatically: Per layer we configure $\theta$ algorithmically by ranking the network units in descending order of their relevance values for the current prediction, while choosing layer indices uniformly and arbitrarily from the higher, middle or bottom parts of the models throughout the paper for illustration out of simplicity.

\noindent{\bf Relevance Maximization in brief}.
Although \gls{crc} allows to compute concept-specific attribution maps by disentangling the backward flow, our understanding of the semantics of latent model structures largely remains elusive with local attributions alone. In other words, the channel-wise relevance values and the concept-conditional relevance maps do not provide the full answer to which specific concept a particular channel is actually encoding (the \emph{what} question). A canonical approach for gaining insight into the meaning and function of latent model structures is \gls{amax} \cite{zhou2015object, olah2017feature, radford2017learning, cammarata2020thread} for generating or selecting samples as representations for concepts encoded in hidden space.
We find, however, that (maximizing) the activation of a latent encoding by a given data point does not always  correspond to its \emph{utility} to the model in an inference context (see \eg \cite{becking2021ecq} or Supplementary Figure~\ref{fig:appendix:rel-vs-act-general-examples}), putting the faithfulness of activation-based example selection for latent concept representation into question.

We therefore introduce \gls{rmax},
as an alternative measure to \gls{amax}, with the objective to maximize the relevance criterion %
for the selection of representative samples for latent model features (see Section~\ref{sec:methods} for technical details).
For each of the observed concepts, Figure~\ref{fig:methods_in_brief}c (right) shows 12 image segments from a holdout set, for which channel 274 of layer \texttt{features.28} of a pre-trained VGG-16 (from \cite{lufficc2018ssd}) encoding the concept ``fur'' becomes maximally relevant for the prediction of class ``dog''.
Since relevance is, other than activation, directly linked to a model's prediction output, the obtained example sets per latent feature are also highly outcome specific. That is, for a latent feature, we may obtain multiple sets of examples, each illustrating how the model is using a particular feature for the prediction of different outcomes, \eg classes.

\section{Results}
\label{sec:experiments}
Beginning with Section~\ref{sec:experiments:local:analysis}, we present approaches to study the role of learned concepts in individual predictions using our glocal \gls{crc}- and \gls{rmax}-based approach. The understanding of hidden features and their function then allows to interact with the model and to test its robustness against feature ablation in Section~\ref{sec:experiments:quantitative:imageretrieval}. In Section~\ref{sec:experiments:quantitative:clusters}, we study concept subspaces in order to identify (dis)similarities and roles of concepts in fine-grained decision making. Finally, Section~\ref{sec:results:study} examines the benefits of \gls{crc} over traditional local \gls{xai} methods in a user study.

More detailed investigations can be found in \sms~\ref{sec:appendix:conceptunderstanding}, \ref{sec:appendix:conceptlocalization} and \ref{sec:experiments:quantitative}. In addition, \sm~\ref{sec:appendix:fairness} provides an example on how \gls{crc} can be leveraged to identify systematically learned biases in male vs.\ female face classification and \sm~\ref{sec:appendix:timeseries} demonstrates the applicability of \gls{crc} to time series data.

\begin{figure*}[h!]
    \centering
    \includegraphics[width=.65\textwidth]{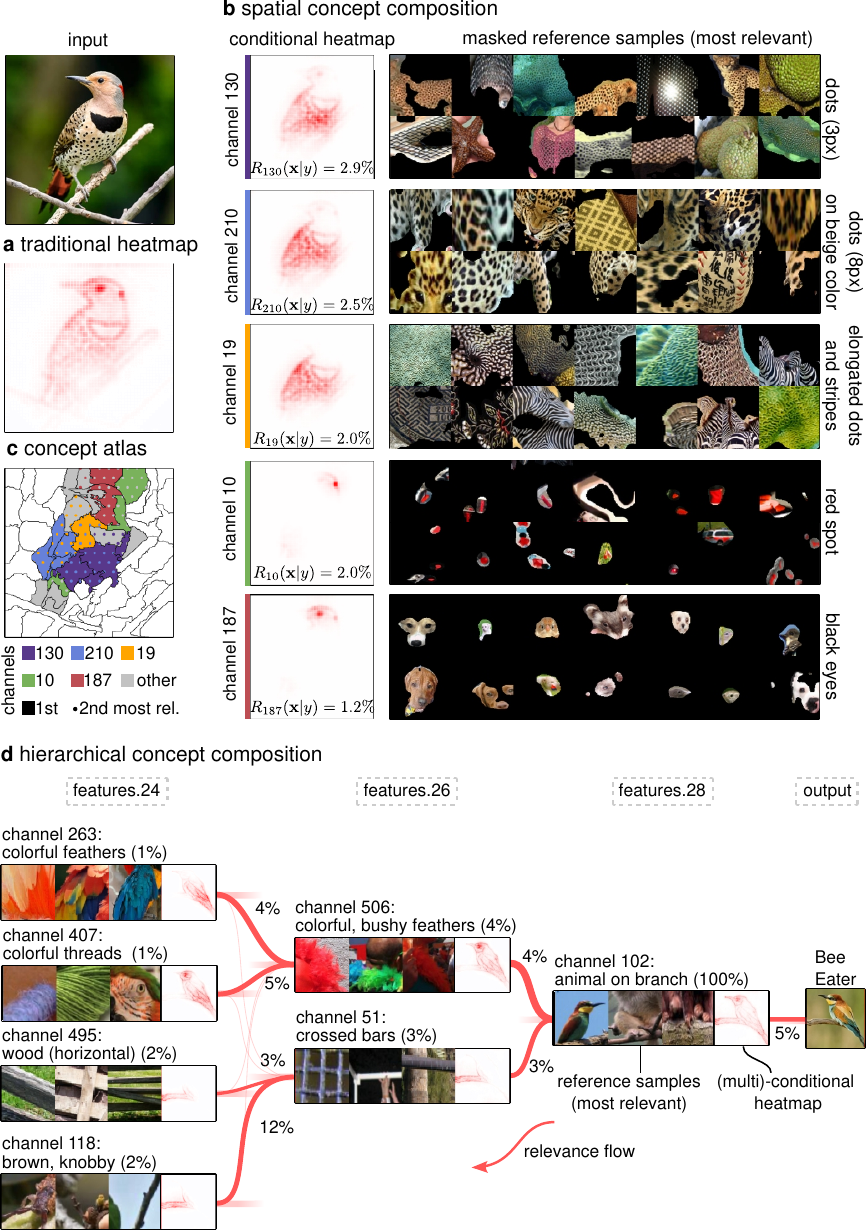}
    \caption{
    Understanding concepts and concept composition with CRP. \textbf{a)} An attribution map indicates that various body parts of the bird are relevant for the prediction. \textbf{b)} Channel-conditional explanations computed with \gls{crc} help to localize and understand channel concepts by providing masked reference samples (explaining by example). \textbf{c)} \gls{crc} relevances can further be used to construct a concept atlas, visualizing which concepts dominate in specific regions in the input image defined by super-pixels. Here, the most relevant channels in layer \texttt{layer3.0.conv2} can be identified with concepts ``dots'' (channel 210 and 130), ``red spot'' (10), ``black eyes'' (187) and ``stripes-like'' (19). \textbf{d)} Concept Composition Graphs decompose a concept of interest given a particular prediction into lower-layer concepts, thus improving concept understanding. Shown are relevant (sub)-concepts in \texttt{features.24} and \texttt{features.26} for concept ``animal on branch'' in \texttt{features.28} for the prediction of class ``Bee Eater''. The relevance flow is highlighted in red, with the relative percentage of relevance flow to the lower-level concepts. For each concept, the channel is given with the relative global relevance score (\wrt channel 102 in \texttt{features.28}) in parentheses. Following the relevance flow, concept ``animal on branch'' is dependent on concepts describing the branch (\eg, ``wood (horizontal)'' and ``brown, knobby'') and colorful plumage (\eg, ``colorful feathers'' and ``colorful threads''). Additional examples can be found in \sm~\ref{sec:appendix:conceptlocalization}.
    }
    \label{fig:experiments:localize-concepts:concept-atlas}
\end{figure*}
\subsection{Understanding Concept Composition Leading to Prediction}
\label{sec:experiments:local:analysis}
Attribution maps provide only partial insights into the decision making process as they only show where the model is focusing on and not which concepts are actually being used. Figure \ref{fig:experiments:localize-concepts:concept-atlas}a shows an attribution map computed for the prediction ``Northern Flicker''. In this case, the bird's head --- in particular the black eye and red stripe --- can be identified as the most relevant part of the image. However, it remains unclear from the explanation whether the color or the shape (or both) of the eye and stripe were the decisive features for the model to arrive at its prediction, and how much these body parts contribute, e.g., compared to the bird's feathers. Furthermore, as shown in Supplementary Figure~\ref{fig:appendix:methods:lrp:entangle}b attribution maps almost always point to the head or the upper body of a bird, irrespective of the bird explained. Thus, the non-trivial task of interpreting what particular feature of the bird (e.g., color, texture, body part shape or relative position of the body parts) actually led to the decision is put onto the human user, which can result in false conclusions.

By conditioning the explanation on relevant hidden-layer channels via \gls{crc}, we can assist in concept understanding and overcome the interpretation gap. Figure \ref{fig:experiments:localize-concepts:concept-atlas}b shows the result of the CRP analysis. The conditional heatmaps help to localize regions in input space for each relevant concept, and at the same time reveal \emph{what} the model has picked up in those regions by providing reference samples (i.e., explaining by example). Here, the concepts we identified as ``red spot'' and ``black eyes'' (based on our subjective understanding of the representative examples) can be assigned to the head of the Northern Flicker bird. These concepts play a crucial role in the classification of the bird, although, e.g., the ``black eyes'' concept naturally also occurs in images of cats and dogs. Furthermore, both ``dots'' concepts affecting the prediction can be assigned to the bird's torso and the ``elongated dots and stripes'' concept to the bird's wings. Note that CRP also allows to quantitatively determine the individual contribution of each concept to the final classification decision by summation of the conditional relevance scores (see Section~\ref{sec:methods:crp}). This additional information is very valuable as it indicates, e.g., that the dotted texture is the most relevant feature for this particular prediction, or that color is a very relevant cue (e.g., the masked reference samples for channel 10 are all red and for channel 187 contain only black/brown eyes). 

The concept atlas shown in Figure \ref{fig:experiments:localize-concepts:concept-atlas}c further eases to comprehend the relevant concepts. Technically, the atlas visualizes which concepts are most relevant (and here, second most relevant) in specific input image regions (for details see \sm~\ref{sec:appendix:methodsindetail:lrp:conceptual:atlas}). By choosing super-pixels as regions of interest, we can aggregate the channel-conditional relevances per super-pixel into regional relevance scores, as discussed in the extended Methods Section in \sm~\ref{sec:appendix:methodsindetail:local_conceptual_importance}. Here, the concept atlas indicates that the ``red spot'' and ``black eye'' concepts are most relevant at the bird's head, while the two ``dots'' concepts mostly fill the upper body part. Interestingly, a stripe of red color in the tail feathers of the bird is detected and used by the model, as indicated by the ``red spot'' concept being second-most relevant in this region. In \sm~\ref{sec:appendix:conceptlocalization:atlas}
an alternative way for constructing concept atlases using single pixels instead of super-pixels is also discussed. 

Alternatively to investigating the most relevant channels overall as in Figure~\ref{fig:experiments:localize-concepts:concept-atlas}b, a region of interest, \eg a super-pixel, can be chosen and its most relevant concepts studied. A comparison of relevant concepts regarding two regions of unrelated visual features is shown in Supplementary Figures \ref{fig:appendix:localize-concepts:concepts-in-region0}, \ref{fig:appendix:localize-concepts:concepts-in-region} and \ref{fig:appendix:localize-concepts:concepts-in-region2}.

With the selection of  a specific neuron or concept, \glsdesc{crc} allows investigating how relevance flows from and through the chosen network unit to lower-level neurons and concepts, as is discussed in Section~\ref{sec:methods:crp}. This gives information about which lower-level concepts carry importance for the concept of interest and how it is composed of more elementary conceptual building blocks, which may further improve the understanding of the investigated concept and model as a whole. In Figure~\ref{fig:experiments:localize-concepts:concept-atlas}d, we visualize and analyze the backward flow of relevance scores. The graph-like visualization reveals how concepts in higher layers are composed of lower layer concepts. Here, we show the top-2 concepts influencing our concept of choice, the ``animal on branch'' concept encoded in \texttt{features.28} of a VGG-16 model trained on ImageNet. Edges in red color indicate the flow of relevance \wrt to the particular sample from class ``Bee Eater'' shown to the very right between the visualized filters with corresponding examples and (multi)-conditional heatmaps. The width of each red edge describes the relative strength of contribution of lower layer concepts to their upper layer neighbors. This example shows that different paths in the network potentially activate the filter. Here, concepts that encode feathers, threads or fur together with horizontal structures are responsible for the activation of filter 102 in the observed layer in this particular case. In \sm~\ref{sec:appendix:timeseries} an example on time series data is illustrated.

\subsection{Understanding Concept Impact and Reach}
\label{sec:experiments:quantitative:imageretrieval}
In this section, we demonstrate how \glsdesc{crc} can be leveraged as a \gls{hitl} solution for dataset analysis. In a first step, we will uncover a \glsdesc{ch} artifact \cite{lapuschkin2019unmasking}, suppress it by selectively eliminating the most relevant concepts in order to assess its decisiveness for the recognition of the correct class of a particular data sample. Then, we utilize class-conditional reference sampling (\cf~Section~\ref{sec:methods}) to perform an inverse search to identify multiple classes making use of the filter encoding the associated concept, both in a benign and a \glsdesc{ch} sense.

\begin{figure}[t]
    \centering
    \includegraphics[width=0.88\linewidth]{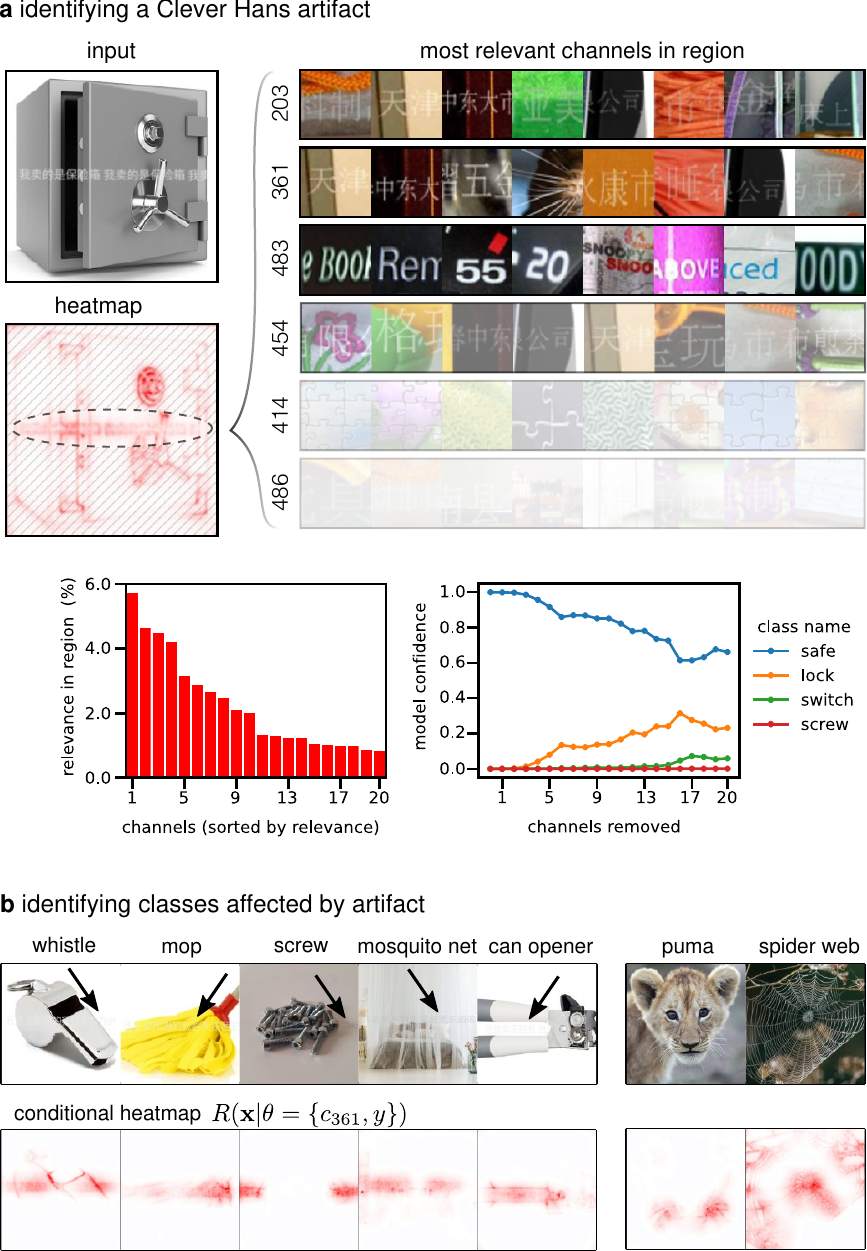}
    \caption{
    From concept-level explanations to model and data debugging. \textbf{a)} Local analysis on the attribution map reveals several channels (203, 361, 483, 454, 414, 486 and more) in layer \texttt{features.30} of a VGG-16 model with BatchNorm pretrained on ImageNet that encode for a \glsdesc{ch} feature exploited by the model to detect the \text{safe} class. (\textit{Left}): Input image and heatmap. (\textit{Center}): Reference samples ${\mathcal{X}^{*}_{8}}^{\text{rel}}_{\text{sum}}$ for the 6 most relevant channels in the selected region in descending order of their relevance contribution. (\textit{Right}): Relevance contribution of 20 most relevant filters inside the region. These filters are successively set to zero and the change in prediction confidence of different classes is recorded. \textbf{b)} The previously identified \glsdesc{ch} filter 361 plays a role for samples of different classes (most relevant reference samples shown). Here, black arrows point to the location of a \gls{ch} artifact, \ie, a white, delicate font overlaid on images (best to be seen in a digital format). In the case of class ``puma'' or ``spiderweb'', the channel is used to recognize the puma's whiskers or the web itself, respectively. Below the reference samples, the \gls{crc} heatmaps conditioned on filter 361 and the respective true class $y$ illustrate, which part of their attribution map would result from filter 361.
    }
    \label{fig:hitl:safe}
\end{figure}

In Figure~\ref{fig:hitl:safe}a we analyze a sample of the ``safe'' class of ImageNet in a pretrained VGG-16 BN model. Initially, we obtain an input attribution map highlighting a centered horizontal band of the image, where a watermark is located. If we take a closer look at layer \texttt{features.30} and perform a local analysis (\cf~Section~\ref{sec:methods}) on the watermark, we notice that the five most relevant filters are 203, 361, 483, 454, 414, 486. Visualizing them using \gls{amax} as illustrated in Supplementary Figure \ref{fig:appendix:act_max_hitl},
we conclude that they approximately encode for white strokes. Using our proposed \gls{rmax} approach, which uses \gls{crc} for identifying the most relevant samples, we gain a deeper insight into the model's preferred usage of the filters and discover that the model utilizes them to detect white strokes in ``written characters''. A detailed comparison between  \gls{amax} and \gls{rmax} can be found in \sm~\ref{sec:appendix:experiments:selecting_reference_samples}. To test the robustness of the model against this \glsdesc{ch} artifact, we successively set the activation output map of the 20 most relevant filters activating on the watermark to zero. In Figure~\ref{fig:hitl:safe}a (right), we record the change of classification confidence of four classes with the highest prediction confidence for this sample. From the graph, it can be inferred that the \glsdesc{ch} filters focussing on the watermark help the model in prediction, but they are not decisive for correct classification. Thus, the model relies on other potential non-\glsdesc{ch} features to detect the safe, verifying the correct functioning of the model in cases of samples without watermarks. Another example with strong dependency on \glsdesc{ch} artifacts is found in \sm~\ref{sec:appendix:experiments:quantitative:whatif}.

In an inverse search, we can now explore for which samples and classes these filters also generate high relevance. This allows us to understand the behavior of the filter in more detail and to find other possible contaminated classes. In Figure~\ref{fig:hitl:safe}b are the seven most relevant classes for filter 361 illustrated. Surprisingly, many classes including ``whistle'', ``mop'', ``screw'', ``mosquito net'', ``can opener'' and ``safe'' (among others) in the ImageNet Challenge 2014 data are contaminated with similar watermarks encoded via filter 361  of \texttt{features.30} which is used for the correct prediction of samples from those classes. To verify our finding, we locate via \gls{crc} the source of the filters' relevance \wrt\ the true classes in input space and confirm that these filters indeed are used to recognize the characters. This implies that the model has learned a shared \glsdesc{ch} artifact spanning over multiple classes to achieve higher accuracy in classification. The high number of contamination of samples with the identified artifactual feature could be explained by the fact that watermarks are sometimes difficult to see with the naked eye (location marked with a black arrow) and thus slip any quality ensuring data inspection. The impact of this image characteristic can, however, be clearly marked using the \gls{crc} heatmap. Although the filter is mainly used to detect characters, there are also valid use cases for the model, such as for the puma's whiskers or the spider's web. This suggests that the complete removal of \glsdesc{ch} concepts through pruning may harm the model in its ability to predict other classes which make valid use of the filter, and that a class-specific correction~\cite{anders2022finding} might be more appropriate.
\begin{figure*}[!ht]
    \centering
    \includegraphics[width=0.7\textwidth]{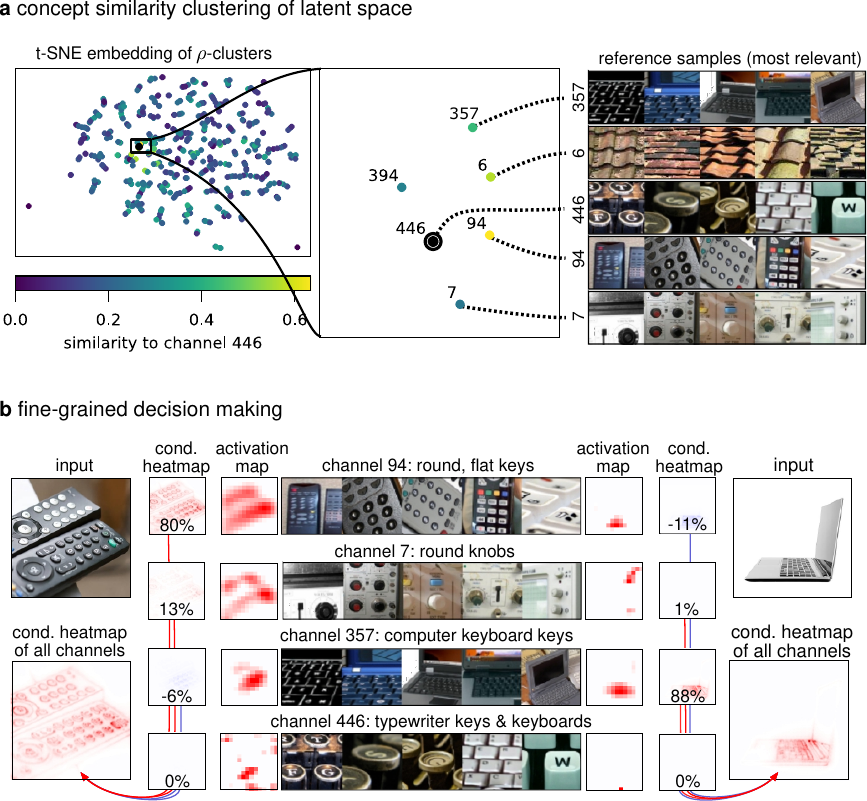}
    \caption{
    Similarity of concepts and analysis of fine-grained decision making. \textbf{a)} \emph{(Left)}: Channels from layer \texttt{features.40} of a VGG-16 with BatchNorm, clustered and embedded according to $\rho$-similarity (see Section~\ref{sec:methods:understanding:comparing_latent_filters}). Markers are colored according to their $\rho$-similarity to filter 446. (\emph{Center and right}): One particular cluster around channel 446 is shown in more detail with five similarly activating channels and their reference images ${\mathcal{X}^{\star}_{8}}^{\text{rel}}_{\text{sum}}$ obtained via \gls{rmax}. As per the reference images, the over-all concept of the cluster seems to be related to keyboard keys, round buttons as well as rectangular roofing shingles. \textbf{b)} Relevance-based investigation of the previously identified similarly activating channels. \textit{(Center)}: Reference examples for the identified filters with similar underlying theme. \textit{(Left)}: Exemplary input from class ``remote control'' with per-channel activation maps and respective ground truth \gls{crc} relevance maps, as well as their aggregation $\theta=\{L: \{y\}, \texttt{features.40}: \{c_{94}, c_{7}, c_{357}, c_{446} \}\}$ (bottom left). \textit{(Right)}: Exemplary input from class ``laptop computer''  with per-channel activation maps and respective true class \gls{crc} relevance maps, as well as their aggregation. Conditional relevance attributions $R(x|\theta)$ are normalized \wrt~the common maximum amplitude. Similarly activating channels do not necessarily encode redundant information, but might be used by the model for making fine-grained distinctions, which can be observed from the attributed relevance scores.
    } 
    \label{fig:experiments:tsne_buttons}
\end{figure*}
\subsection{Understanding Concept Subspaces, (Dis)Similarities and Roles}
\label{sec:experiments:quantitative:clusters}
So far in our experiments we have treated single filters as functions assumed to (fully) encode learned  concept. Consequently, we have visualized examples and quantified effects based on per-filter granularity. While previous work suggest that individual neurons or filters often encode for a single human comprehensible concepts, it can generally be assumed that concepts are encoded by sets of filters (see \sm~\ref{sec:appendix:relatedwork}). The learned weights of potentially multiple filters might correlate and thus redundantly encode the same concept, or the directions described by several filters situated in the same layer might span a concept-defining subspace. In this section, we now aim to investigate the encodings of filters of a given neural network layer for similarities in terms of activation and use within the model.

Figure~\ref{fig:experiments:tsne_buttons}a shows an analysis result focusing on a cluster around filter $446$ from \texttt{features.40} of a VGG-16 network with BatchNorm layers trained on ImageNet. The reference samples show various types of typewriter and rectangular laptop keyboard buttons and roofing shingles photographed in oblique perspective, as well as round buttons of typewriters, remote controls for televisions, telephone keys and round turnable dials of various devices and machinery. Thus, the filters around filter 446 seem to cover different aspects of a shared ``button'' or ``small tile'' concept. The filters located in this cluster have been identified as similar due to their similar activations over  sets of analyzed reference samples (see Section~\ref{sec:methods:understanding:comparing_latent_filters}).
Assuming redundancy based on the filter channels' apparently similar activation behavior, a human could merge them to one encompassing concept, thereby simplifying interpretation by reducing the number of filters in the model. We therefore further investigate the filters 7, 94, 446 and 357 (all showing buttons or keys) in order to find out (1) whether they encode a concept collaboratively, (2) whether they are partly redundant, or (3) whether the cluster serves some discriminative purpose.

Figure \ref{fig:experiments:tsne_buttons}b visualizes the reference samples of these four filters for the most-relevant classes ``laptop computer'' and ``remote control''. We compute filter activations during a forward pass through the model using instances of both classes as input, as well as filter-conditioned CRP maps for the samples' respective ground truth class label. Regardless of whether an instance from class ``laptop'' or ``remote control'' is chosen as input, the activation maps across the observed channels are in part similar per image, e.g., they all activate on the center diagonal part for the left input image. 
The per-channel \gls{crc} attribution map, however, reveals that while all filters react to similar stimuli in terms of activations, the model seems to use the subtle differences among the observed concepts to distinguish between the classes ``laptop'' and ``remote control''. In both cases, buttons are striking and defining features, and all observed filters activate for button features. However, when computing the conditional heatmaps with CRP for class ``remote control'', the activating filters representing round buttons (filters 7 and 94) dominantly receive positive attribution scores, while filter 357 clearly representing typical keyboard button layouts receives negative relevance scores and filter 446 does not receive any relevance despite being reactive to the given input. For samples of class ``laptop'', the computation of relevance scores \wrt their true class yields almost opposite attributions, indicating that filters encoding round buttons and dials (filters 94 and 7) provide evidence against class ``laptop'', while the activation of channel 357 clearly speaks for the analyzed class as visible in the conditional heatmaps. In both relevance analyses, however, filter 446 receives weak negative to no attributions, presumably as it represents a particular expression of both round and angular buttons which fits (or contradicts) neither of the compared classes particularly well. In fact, filter 446 is highly relevant for class ``typewriter keyboard'' instead.

In conclusion, we report that although several filters may show signs of correlation in terms of output activation, they are not necessarily encoding redundant information or are serving the same purpose. Conversely, using our proposed \glsdesc{crc} in combination with the \glsdesc{rmax}-based process for selecting reference examples representing seemingly correlating filters, we are able to discover and  understand the subtleties a neural network has learned to encode in its latent representations. See \sm~\ref{sec:appendix:experiments:quantitative:clusters} for additional results in extension to this section.

\subsection{Human Evaluation Study}
\label{sec:results:study}
\begin{figure*}[h]
    \centering
    \includegraphics[width=1\linewidth]{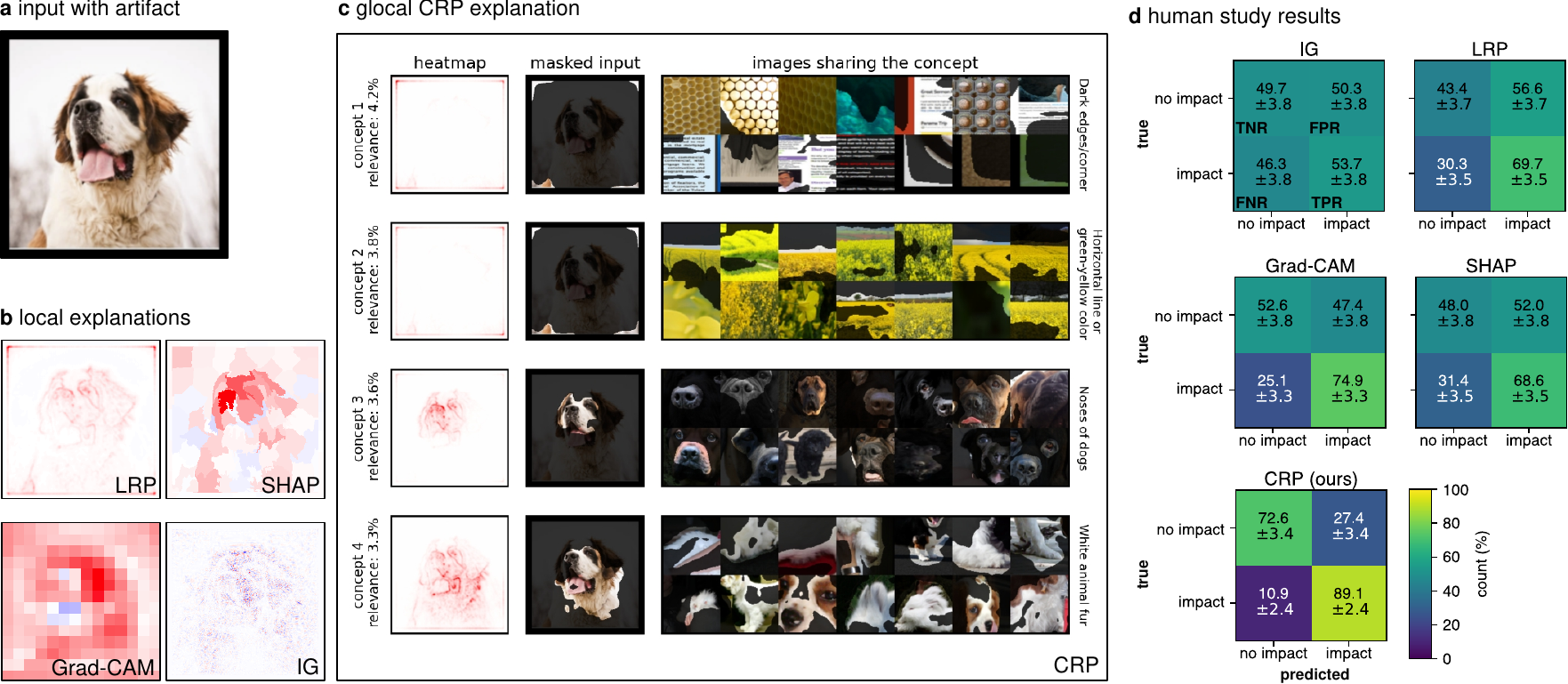}
    \caption{
    User study evaluating the informativeness of different explanation methods wrt.\ the model's reliance on a border artifact.
    \textbf{a)} An example input image with exemplarily added black border artifact.
    \textbf{b)} Attribution maps computed via the  methods LRP, SHAP, Grad-CAM, IG.
    \textbf{c)}
    An explanation derived with \gls{crc} and \gls{rmax}.
    In both \textbf{b} and \textbf{c}, the model has been trained to be affected by the artifact shown in \textbf{a} during inference.
    \textbf{d)}
    Human evaluation results of participant predictions whether the border artifact has an impact on the prediction outcome, based on the given explanation.
    Confusion matrices illustrating true positive (TPR), false positive (FPR), true negative (TNR), and false negative (FNR) rate per \gls{xai} method.
    }
    \label{fig:main:study:confusion_matrix}
\end{figure*}

This section presents the results of a human evaluation study, which we performed to assess the practical utility of the CRP- and RelMax-based explanations to (non-expert) end users for understanding ML model behavior. Human subjects were asked to decide --- based on explanations --- whether the model's prediction has been influenced by the presence of a particular and known data artifact or not. We trained two image classifiers, of which one has learned to utilize a data artifact --- a thick black border around the image (see Figure~\ref{fig:main:study:confusion_matrix}a). For both models, we then generate explanations (see Figure~\ref{fig:main:study:confusion_matrix}b and Figure~\ref{fig:main:study:confusion_matrix}c) on images containing the artifact, using the proposed \gls{crc} maps with \gls{rmax} examples as well as four popular XAI methods, namely Integrated Gradients (IG) \cite{sundararajan2017axiomatic}, SHAP \cite{lundberg2017Shap}, Grad-CAM \cite{selvaraju2017grad} and \gls{lrp} \cite{bach2015pixel}. In the primary task, the participants are asked to assess, whether the black border impacts the model prediction according to the explanation (binary answer: yes or no). Furthermore, we ask secondary questions on how confident they are in their answer and about the perceived clarity of the presented explanations.
For more details on the study setup, we refer the reader to  Section \ref{sec:methods:human_evaluation_study_details}.

The results of the study consistently show that participants were reliably able to detect whether the prediction was impacted by the border artifact when exposed to \gls{crc}- and \gls{rmax} explanations (see Figure~\ref{fig:main:study:confusion_matrix}d). \gls{crc} shows the highest true positive (TPR) and true negative (TNR) rates of $(89.1 \pm 2.4)\,\%$ and $(72.6\pm 3.4)\,\%$, respectively, and thus results in an accuracy (wrt.\ the primary task) which is significantly higher than that of all other methods (two-sample $t$-test $p$-values of less than $8\cdot10^{-4}$),
as shown in the
Supplementary Table~\ref{tab:appendix:study:results}.
The study participants obtained the second best results when exposed to Grad-CAM explanations with a TPR and TNR of $(74.9 \pm 3.3)\,\%$ and $(52.6\pm 3.8)\,\%$, respectively.
Participants exposed to explanations from IG performed worst with $(53.7 \pm 3.8)\,\%$ and $(49.7\pm 3.8)\,\%$, respectively.
It is to note, that random guessing would correspond to values of 50\,\%, 
indicating that the insight obtainable from IG explanations is limited in this study.
We note that in general it seems to be easier to detect a model's use of the introduced border artifact through all evaluated \gls{xai} approaches, than it is to correctly reject an impact of the artifact, since for all methods we observe TPR $>$ TNR.

When inspecting the proclaimed confidence with which the participants made their prediction, a direct link does not seem to exist to the measured prediction performance when assessing the artifact's impact.
Interestingly, participants exposed to the IG explanations report the highest confidence (approx. 77\,\%) --- while at the same time performing worst in the primary task ---
followed by \gls{crc} (approx. 76\,\% self-rated confidence).
These results support the findings of \cite{kim2018interpretability}, that \emph{traditional} (\ie single per sample and class, as shown in Figure~\ref{fig:main:study:confusion_matrix}b) saliency- or attribution maps alone might be misleading and insufficient for understanding the reasoning of a machine learning predictor, e.g., see the IG heatmap.
Regarding clarity of the explanations as perceived by the participants, the fine-grained attribution maps of IG and \gls{lrp} receive the highest scores. \gls{crc} and \gls{rmax} interestingly result in the lowest reported clarity, which might be linked to the more complex nature of the method, potentially leaving some of the participants overwhelmed with the increased amount of information to process, and time required to do so. This result is consistent with the observation of Hacker and Passoth \cite{hacker2022varieties} that addressees prefer simple and concise explanations. Despite that, our results demonstrate that our proposed approach is the most effective option for the participants to solve the primary task of the study.

\section{Discussion}
\label{sec:conclusion}
In this work we have introduced \glsdesc{crc}, a post-hoc explanation method, which not only indicates which part of the input is relevant for an individual prediction, but also communicates the meaning of involved latent representations by providing human-understandable examples. Since \gls{crc} combines the benefits of the local and global \gls{xai} perspective, it computes more detailed and contextualized explanations, considerably extending the state-of-the-art. Among its advantages are the high computational efficiency (within the order of a backward pass to compute near-instantaneous local explanations for the most relevant concepts, and a complete concept atlas visualization within the order of seconds) and the out-of-the-box applicability to (almost) any model without imposing constraints on the training process, the data and label availability, or the model architecture. Furthermore, \gls{crc} introduces the idea of conditional backpropagation tied to a single concept or a combination of concepts as encoded by the model, within or across layers. Via this ansatz, the contribution of all neurons' concepts in a layer can be faithfully attributed, localized in the input space, and finally their interaction can be studied. As shown in this work, such an analysis allows one to disentangle and separately explain the multitude of in-parallel partial forward processes, which transform and combine features and concepts before culminating into a prediction. Finally, with \glsdesc{rmax} we move beyond the decade-old practice of communicating latent features of neural networks based on examples obtained via maximized activation. In particular, we show that the examples which stimulate hidden features maximally are not necessarily useful for the model in an inference context, or representative for the data the model is familiar and confident with. By providing examples based on relevance, however, the user is presented with data with characteristics which actually play an important role in the prediction process. Since the user can select examples wrt.\ any (i.e., not necessarily the ground truth) output class, our approach constitutes a new tool to systematically investigate latent concepts in neural networks.

Our experiments have qualitatively and quantitatively demonstrated the additional value of the CRP approach for common datasets and end-to-end trained models. Specifically, we showed that reference samples selected with relevance-based criteria, concept heatmaps and atlases, as well as concept composition graphs  open up the ability to understand model reasoning on a more abstract and conceptual level.  These insights then allowed us to identify \glsdesc{ch} concepts, to investigate their impact, and finally to correct for these misbehaviors. Further, using our relevance-based reference sample sets, we were able to identify concept themes spanned by sets of filters in latent space. Although channels of a cluster have a similar function, they seem to be used by the model for fine-grained decisions regarding details in the data, such as the particular type of buttons to partially decide whether an image shows a laptop keyboard, a mechanical typewriter or a TV remote control. In addition, we have demonstrated the usefulness of \gls{crc} in non-image data domain, where traditional attribution maps are often difficult to interpret and comprehend by the user. Our experiments on time series data have shown that as long as a visualization of the data can be found, the meaning of latent concepts can be communicated via reference examples. Finally, we did conduct a user study which validates a significant increase in utility of our glocal \gls{crc}- and \gls{rmax}-based approach above traditional post-hoc local \gls{xai} methods for understanding a model's inference behavior by human assessors. For completeness, we would like to make the reader aware of two factors possibly affecting the outcome of our study, namely the potentially varying degree of technical and in-domain training of the study participants, as well as the given prior knowledge about the nature of the data artifact potentially affecting the model. 
Both factors should therefore be addressed and evaluated individually in future work, \eg,  in order to assess the potential of (g)local \gls{xai} approaches for assessing yet unexplored model behavior based on feedback for single instance predictions, across different levels of expert knowledge.

Overall, we believe that the tools we have proposed in this work, and the resulting increase in semantics and detail to be found in sample-specific neural network explanations, will advance the applicability of post-hoc \glsdesc{xai} to novel or previously difficult to handle models, problems and data domains.

\section{Methods}
\label{sec:methods}
This section presents the techniques used and introduced in this paper in brief.
For an elaborate introduction and discussion, please refer to \sms~\ref{sec:appendix:methodsindetail} and~\ref{sec:appendix:methodsindetail:understanding}.
For an estimation of run time requirements,
the computational steps involved, as well as for guidelines on the interpretation of the output obtained by our techniques, please refer to \sm~\ref{sec:appendix:workflow}.

\subsection{\texorpdfstring{\glsdesc{crc}}{Concept Relevance Propagation}}
\label{sec:methods:crp}

In the following, we introduce \glsfirst{crc}, a backpropagation-based attribution method extending the framework of \glsfirst{lrp}~\cite{bach2015pixel}.
As such, \glsdesc{crc} inherits the basic assumptions and properties of \gls{lrp}.\\

\noindent {\bf LRP Revisited}. Assuming a predictor with $L$ layers
\begin{equation} \label{eq:lrp_nn}
    f(\x)=f_L \circ \dots \circ  f_1(\x)~,
\end{equation}
\gls{lrp} follows the flow of activations computed during the forward pass through the model in opposite direction, from the final layer $f_L$ back to the input mapping $f_1$. Given a particular mapping $f_\ast(\cdot)$, we consider its pre-activations $\preact_{ij}$ mapping inputs $i$ to outputs $j$ and their aggregations $\preact_j$ at $j$. 
Commonly in neural network architectures such a computation is given with
\begin{align}
    \preact_{ij} & = \act_iw_{ij}\label{eq:nn_forward_zij}\\
    \preact_j & = \sum_i \preact_{ij}\label{eq:nn_forward_zj}\\
    \act_j & = \sigma(\preact_j)\label{eq:nn_forward_a}~,
\end{align}
where $\act_i$ are the layer's inputs and $w_{ij}$ its weight parameters. Finally, $\sigma$ constitutes a (component-wise) non-linearity producing input activation for the succeeding layer(s). The LRP method distributes relevance quantities $R_j$ corresponding to $a_j$ and received from upper layers towards lower layers proportionally to the relative contributions of $\preact_{ij}$ to $\preact_j$, i.e., 
\begin{equation} \label{eq:lrp_basic}
    R_{i \leftarrow j} = \frac{\preact_{ij}}{\preact_j}R_j.
\end{equation}
Lower neuron relevance is obtained by losslessly aggregating all incoming relevance messages $R_{i \leftarrow j}$ as
\begin{equation} \label{eq:lrp_aggregation}
    R_i = \sum_j R_{i \leftarrow j}.
\end{equation}
This process ensures the property of relevance conservation between a neuron $j$ and its inputs $i$, and thus adjacent layers. \gls{lrp} is mathematically founded in Deep Taylor Decomposition~\cite{montavon2017explaining}.\\
 
\noindent {\bf Disentangling Explanations with CRP}. \glsdesc{crc} extends the formalism of \gls{lrp} by introducing conditional relevance propagation determined by a set of conditions $\theta$. Each condition $c \in \theta$ can be understood as an identifier for neural network elements, such as neurons $j$ located in some layer, representing latent encodings of concepts of interest. One such condition could, for example, represent a particular network output to initiate the backpropagation process from. Within the \gls{crc} framework, the basic relevance decomposition formula of \gls{lrp} given in Equation~\eqref{eq:lrp_basic} then becomes
\begin{align}
    R^{(l-1,l)}_{i \leftarrow j}(\x|\theta \cup \theta_{l}) = \frac{\preact_{ij}}{\preact_j} \cdot \sum_{c_l \in \theta_l}\delta_{jc_l} \cdot R^l_j(\x|\theta)~,
    \label{eq:lrp:masked}
\end{align}
following the potential for a ``filtering'' functionality briefly discussed in~\cite{montavon2018methods}. Here, $R^l_j(\x|\theta)$ is the relevance assigned to layer output $j$ given from the \gls{crc} process performed in upper layers under conditions $\theta$, to be distributed to lower layers. The sum-loop over $c_l \in \theta_l$ then ``selects'' via the Kronecker-Delta $\delta_{jc_l}$ neurons $j$ of which the relevance is to be propagated further, given $j$ corresponds to concepts as specified in set $\theta_l$ specific to layer $l$. The result is the concept-conditional relevance message $R^{(l-1,l)}_{i \leftarrow j}(\x|\theta \cup \theta_{l})$ carrying the relevance quantities \wrt the prediction outcome on $\x$ conditioned to $\theta$ and $\theta_l$. Note that the sum is not particularly necessary in Equation~\eqref{eq:lrp:masked}, but serves as a means to compare all possible $c_l$ for identity to the current $j$. In practice
, \gls{crc} can be implemented efficiently as a single backpropagation step by binary masking of relevance tensors, and is compatible to the recommended rule composites for relevance backpropagation~\cite{montavon2019layer,kohlbrenner2020towards}.
We provide an efficient implementation of \gls{crc} based on Zennit~\cite{anders2021software} at~\url{https://github.com/rachtibat/zennit-crp}.

The effect of \gls{crc} over \gls{lrp} and other attribution methods is an increase in detail of the obtained explanations.
Given a typical image classification \gls{cnn},
one may assume the computation of three-dimensional latent tensors,
where the first two axes span the application coordinates of $n$ spatially invariant convolutional filters, which generate output activations stored in the $n$ channels of the third axis. For simplicity, one can further assume that each filter channel is associated to exactly one latent concept.
Neurons $j$ can thus be grouped into spatial and channel axes in order to restrict the application of \gls{crc} conditions $\theta_l$ to the channel axis only, i.e.,
\begin{equation}
    R^{(l-1,l)}_{i \leftarrow \left(p,q,j\right)}(\x|\theta \cup \theta_{l}) = \frac{\preact_{i\left(p,q,j\right)}}{\preact_{\left(p,q,j\right)}} \cdot \sum_{c_l \in \theta_l} \delta_{jc_l} \cdot R^l_{\left(p,q,j\right)}(\x|\theta).
    \label{eq:lrp:masked_regrouped}
\end{equation}
Here, the tuple $\left(p,q,j\right)$ uniquely addresses an output voxel of the activation tensor $\preact_{\left(p,q,j\right)}$ computed during the forward pass with $p$ and $q$ indicating the spatial tensor positions and $j$ the channel.
Figure~\ref{fig:methods_in_brief}a exemplarily contrasts the attribution-based explanation \wrt class ``dog'' only (which also is possible with \gls{lrp} and other attribution methods) as $\theta_\text{d} = \lbrace\texttt{dog}^L\rbrace$, to the attributions for, \eg,  ``dog $\land$ fur'' as $\theta_\text{df} = \lbrace \texttt{dog}^L, \texttt{fur}^l\rbrace$
(possible with \gls{crc} only) by conditionally masking channels responsible for fur pattern representations. 
Alternatively, conditions can be notated in the form of $\theta_\text{df} = \lbrace L:\{\texttt{dog}\}, l:\{\texttt{fur}\}\rbrace$, to provide a more explicit notation specifying the affiliation of concepts to distinct layers.
Here we use the terms \texttt{fur} and \texttt{dog} describing latent or labelled concepts, respectively, as proxy representations for network element identifiers $c$. We further assume that in any layer $l'$ without explicit designation of conditions all $\delta_\ast$-operators always evaluate to $1$ to not restrict the flow of attributions through these layers.

Due to the conservation property of \gls{crc} inherited from \gls{lrp}, the global relevance of individual concepts to per-sample inference can be measured by summation over input units $i$ as
\begin{equation}
    R^l(\x|\theta) = \sum_{i}R^l_i(\x|\theta)~,
    \label{eq:lrp:masked_aggregated}
\end{equation}
in any layer $l$ where $\theta$ has taken full effect. This can easily be extended to a localized analysis of conceptual importance, by restricting the relevance aggregations to regions of interest $\mathcal{I}$
\begin{equation}
    R^l_\mathcal{I}(\x|\theta) = \sum_{i\in\mathcal{I}}R^l_i(\x|\theta)~,
    \label{eq:lrp:masked_aggregated_local}
\end{equation}
as also illustrated in Supplementary Figure~\ref{fig:appendix:lrp-hidden-spatial-disentanglement}. Additionally, as shown in Supplementary Figure~\ref{fig:appendix:lrp-concept-composition}, an aggregation of the relevance messages may be utilized to identify dependencies of a concept $c$ encoded by channels $j$, to concepts encoded by channels $i$ in a lower layer, in context of the prediction of a sample $\x$ and \gls{crc}-conditions $\theta$. With an expansion of the indexing of downstream target voxels \wrt Equation~\eqref{eq:lrp:masked_aggregated} as
\begin{equation}
    R^{(l-1,l)}_{(u,v,i) \leftarrow (p,q,j)}(\x|\theta)  =  \frac{\preact_{(u,v,i)(p,q,j)}}{\preact_{(p,q,j)}}R_{(p,q,j)}^{l}(\x|\theta)~,
    \label{eq:lrp:superexpandend:decomposition}
\end{equation}
the tuple $(u,v,i)$ addresses the spatial axes with $u$ and $v$, and the channel axis $i$ at layer $l-1$. An aggregation over spatial axes with
\begin{equation}
    R^{(l-1,l)}_{i\leftarrow j}(\x|\theta) = \sum_{u,v}\sum_{p,q} R^{(l-1,l)}_{(u,v,i) \leftarrow (p,q,j)}(\x|\theta) 
    \label{eq:lrp:superexpandend:aggregation}
\end{equation}
communicates the dependency between channel $j$ to lower-layer channel $i$, and thus related concepts, in terms of relevance in the prediction context of sample~$\x$. 
Following the LRP methodology, an adaptation of the CRP approach beyond CNN, e.g., to recurrent~\cite{arras2017explaining} or graph~\cite{schnake2021higher} neural networks, is possible. Further details on our proposed \glsdesc{crc} method are given in \sm~\ref{sec:appendix:methodsindetail}.

\subsection{Selecting Reference Examples} \label{sec:methods:understanding}
In the following, we discuss the widely-used Activation Maximization approach to procuring representations for latent neurons, and present our novel CRP-based Relevance Maximization technique to improve concept identification and understanding. An in-depth introduction to all details of our proposed technique is given in \sm~\ref{sec:appendix:methodsindetail:understanding},
with various modes of application and analyses being discussed in \sm~\ref{sec:appendix:conceptunderstanding}.

\subsubsection{Activation Maximization}
\label{sec:methods:understanding:selecting_reference_samples:activation}
A large part of feature visualization techniques rely on \glsfirst{amax}, where in its simplest form, input images are sought that give rise to the highest activation value of a specific network unit. Recent work \cite{yeh2020completeness,chen2020concept} proposes to select reference samples from existing data for feature visualization and analysis. In the literature, the selection of reference samples for a chosen concept $c$ manifested in groups of neurons is often based on the strength of activation induced by a sample. For data-based reference sample selection, the possible input space $\mathcal{X}$ is restricted to elements of a particular finite dataset $\mathcal{X}_d\subset\mathcal{X}$. The authors of \cite{chen2020concept} assume convolutional layer filters to be spatially invariant. Therefore, entire filter channels instead of single neurons are investigated for convolutional layers. One particular choice of maximization target $\mathcal{T}(\x)$ is to identify samples $\mathbf{x}^{\star} \in \mathcal{X}_d$, 
which maximize the sum over all channel activations, \ie, 
\begin{equation}
{\mathcal{T}}^{\text{act}}_{\text{sum}}(\x) = \sum_{i} \preact_{i}(\x)~.
\label{eq:sum-max-short}
\end{equation}
resulting in samples ${\x^{\star}}^{\text{act}}_{\text{sum}}$ which are likely to show a channel's concept in multiple (spatially distributed) input features, as maximizing the entire channel also maximizes ${\mathcal{T}}^{\text{act}}_{\text{sum}}$. However, while targeting all channel neurons, reference samples including both concept-supporting and contradicting features might result in a low function output of ${\mathcal{T}}^{\text{act}}_{\text{sum}}$,  as negative activations are taken into account by the sum. Alternatively, a non-linearity can be applied on $z_i(\x)$,  e.g., ReLU, to only consider positive activations. A different choice is to define maximally activating samples by observing the maximum channel activation
 \begin{equation}
{\mathcal{T}}^{\text{act}}_{\text{max}}(\x) = \max_{i} \preact_{i}(\x),
\label{eq:max-short}
 \end{equation}
leading to samples ${\x^{\star}}^{\text{act}}_{\text{max}}$ with a more localized and strongly activating set of input features characterizing a channel's concept. These samples ${\x^{\star}}^{\text{act}}_{\text{max}}$ might be more difficult to interpret, as only a small region of a sample might express the concept. 

In order to collect multiple reference images describing a concept, the dataset $\mathcal{X}_d$ consisting of $n$ samples is first sorted in descending order according to the maximization target $\mathcal{T}(\mathbf{x})$, \ie
\begin{equation}
    \mathcal{X}^{\star}
    = {\left\{ {\x^{\star}_1}, \dots, {\x^{\star}_n} \right\}}
    = \underset{\x \in \mathcal{X}_d}{\text{argsort}^\text{desc}} \mathcal{T}(\x)
    .
\end{equation}
Subsequently, we define the set 
\begin{equation}
    {\mathcal{X}^{\star}_{k}} 
    = {\left\{ {\x^{\star}_1}, \dots, {\x^{\star}_k} \right\}}
     \subseteq \mathcal{X}^{\star}
\end{equation}
containing the $k \leq n$ samples ranked first according to the maximization target to represent the concept of the filter(s) under investigation. We denote the set of samples obtained from ${\mathcal{T}}^{\text{act}}_{\text{sum}}$ as ${\mathcal{X}^{\star}_{k}}^{\text{act}}_{\text{sum}}$ and the set obtained from ${\mathcal{T}}^{\text{act}}_{\text{max}}$ as ${\mathcal{X}^{\star}_{k}}^{\text{act}}_{\text{max}}$.

\subsubsection{Relevance Maximization}
\label{sec:methods:understanding:selecting_reference_samples:relevance}
We introduce the method of \glsfirst{rmax} as a complement to \glsdesc{amax}. Regarding \gls{rmax}, we do not search for images that produce a maximal activation response. Instead, we aim to find samples, which contain the relevant concepts for a prediction. In order to select the most relevant samples, we define maximization targets ${\mathcal{T}}^{\text{rel}}_{\ast}(\x)$ by using the relevance $R_i(\mathbf{x}|\theta)$ of neuron $i$ for a given prediction, instead of its activation value $\preact_i$. Specifically, the maximization targets are given as
\begin{equation}
{\mathcal{T}}^{\text{rel}}_{\text{sum}}(\x) = \sum_{i} R_{i}(\x|\theta)
\quad \text{and} \quad
{\mathcal{T}}^{\text{rel}}_{\text{max}}(\x) = \max_{i} R_{i}(\x|\theta).
\label{eq:max-rel-target}
\end{equation}

By utilizing relevance scores $R_i(\mathbf{x}|\theta)$ instead of relying on activations only, the maximization target ${\mathcal{T}}^{\text{rel}}_{\text{sum}}$ or ${\mathcal{T}}^{\text{rel}}_{\text{max}}$ is class- (true, predicted or arbitrarily chosen, depending on $\theta$), model-, and potentially concept-specific (depending on $\theta$), as is also illustrated in Supplementary figure~\ref{fig:appendix:rel-vs-act-general-examples}a.
The resulting set of reference samples thus includes only samples which depict facets of a concept that are actually useful for the model during inference (see figure~\ref{fig:appendix:rel-vs-act-general-examples}b). How differences in resulting reference sets ${\mathcal{X}_k^{\star}}^{\text{act}}$ and ${\mathcal{X}_k^{\star}}^{\text{rel}}$ can occur, is depicted in figure~\ref{fig:appendix:rel-vs-act-general-examples}c-d. One can see that relevances are not strictly correlated to activations, because they also depend on the downstream relevances propagated from higher layers affected by feature interactions at the current and following layers. For further details and evaluations, we would like to refer the interested reader to the \sms~\ref{sec:appendix:methodsindetail:understanding} and \ref{sec:appendix:conceptunderstanding}.

\subsection{Comparing Feature Channels with Averaged Cosine Similarity on Reference Samples}
\label{sec:methods:understanding:comparing_latent_filters}
We propose a simple but qualitatively effective method for comparing filters in terms of activations based on reference samples, for grouping similar concepts in \gls{cnn} layers. Based on the notation in previous sections,  ${\mathcal{X}^{*}_{(k, q)}}$ denotes a set of $k$ reference images for a channel $q$ in layer $l$ and $\textbf{z}_q^l(\textbf{W},\textbf{x}_m)$ the ReLU-activated outputs of channel $q$ in layer $l$ for a given input sample $\textbf{x}_m$ with all required network parameters $\textbf{W}$ for its computation. Specifically, for each channel $q$ and its associated full-sized (\ie~not cropped to the channels' filters' receptive fields, \cf~\sm~\ref{sec:appendix:methodsindetail:understanding:scaling}) reference samples $\textbf{x}_m \in {\mathcal{X}^{\star}_{k, q}}^{\text{rel}}_{\text{sum}}$ we compute $\textbf{z}^q_m = \textbf{z}_q^l(\textbf{W},\textbf{x}_m)$, as well as
$\textbf{z}^p_m = \textbf{z}_p^l(\textbf{W},\textbf{x}_m)$ for all other channels $p \neq q$, by executing the forward pass, yielding activation values for all spatial neurons for the channels. We then define the averaged cosine similarity $\rho_{qp}$ between two channels $q$ and $p$ in the same layer $l$ as
\begin{align}
\rho_{qp} & = \frac{1}{2} \left( \cos(\phi)_{qp} + \cos(\phi)_{pq} \right) \label{eq:rho}\\
 \text{with} \qquad \cos(\phi)_{qp} & =  \frac{1}{k} \sum_{\textbf{x}_m \in {{\mathcal{X}^{\star}_{(k, q)}}^{\text{rel}}_{\text{sum}}}} \frac{\textbf{z}^q_m \cdot \textbf{z}^p_m}
{||\textbf{z}^q_m|| \cdot ||\textbf{z}^p_m||}~. \label{eq:cos}
\end{align}
Note that we symmetrize $\rho_{qp}$ in Equation~\eqref{eq:rho} as the cosine similarities $\cos(\phi)_{qp}$ and $\cos(\phi)_{pq}$ are in general not identical, due to the potential dissimilarities in the reference sample sets ${\mathcal{X}^{*}_{(k, q)}}$  and ${\mathcal{X}^{*}_{(k, p)}}$. Thus, $\cos(\phi)_{qp}$ measures the cosine similarity between filter $q$ and filter $p$ \wrt the reference samples representing filter $q$. The from Equation~\eqref{eq:rho} resulting symmetric similarity measures $\rho_{qp} = \rho_{pq} \in [0,1]$ can now be clustered, and visualized via a transformation into a distance measure $d_{qp} = 1 - \rho_{qp}$ serving as an input to t-SNE \cite{van2008visualizing} which visually clusters similar filters together in, typically, $\mathbb{R}^2$.
Note that normally, the output value of the cosine distance covers the interval $[-1,1]$, where for $-1$ the two measured vectors are exactly opposite to one another, for 1 they are identical and for 0 they are orthogonal. In case output channels of dense layers are analyzed, \ie scalar values, the range of output values reduces to the set $\lbrace -1,0,1 \rbrace$, as both values are either of same or different signs, or at least one of the values is zero. Since we are processing layer activations \emph{after} the ReLU nonlinearities of the layer, which yields only positive values for $\textbf{z}^q_m$ and $\textbf{z}^p_m$. This results in $ \rho_{pq} \in [0,1]$, and a conversion to a canonical distance measure $d_{qp} \in [0,1]$

\subsection{Human Evaluation Study Details}
\label{sec:methods:human_evaluation_study_details}

In the following, 
we provide further details on the conduction of the human study in Section~\ref{sec:results:study}. 
All participants have been recruited on the Amazon Mechanical Turk platform, representing people from all backgrounds that do not necessarily have any background knowledge from the field of \glsdesc{ai}. 
As such, the participants reflect the general non-expert population in interaction with (X)AI.
It is to note, however,
that on this platform, participants might work on other unrelated studies for several 
hours, which can have a negative impact on their performance.
The study did not consider the sex, gender, race, ethnicity, or other socially relevant groupings of the participants, as they were not relevant to the research. Consequently, no corresponding data has been collected.

The study was conducted using a between-subject design from September 19th to September 26th, 2022.
Each participant has been assigned randomly to one of the groups (25 participants per group) associated with one of the XAI methods. 
The sample size of 25 is chosen such that the differences in terms of accuracy between our method and the other methods becomes significant (according to two sample t-test probabilities).
For the analysis, we only considered studies fully finished by the participants.

Regarding the computation and visualization of explanations,
we used the publicly available ImageNet~\cite{russakovsky2015imagenet} dataset,
and fine-tuned two VGG-16~\cite{simonyan2015very} \glspl{dnn},
with parameters pre-trained on ImageNet as obtained from the PyTorch~\cite{paszke2017automatic} model zoo.
The interested reader may find additional details about the design and the evaluation of the conducted study in \sm~\ref{sec:appendix:study} and on GitHub (\url{https://github.com/maxdreyer/crp-human-study}) providing Python code for generating explanations as well as HTML templates for Amazon Mechanical Turk.\\

\noindent {\bf Ethics Approval}
The Ethics Commission Faculty IV TU Berlin provided guidelines for the study procedure and determined that no protocol approval is required. Informed consent has been obtained from all participants.

\section*{Declarations}

\subsection*{Data Availability}
The study was conducted using the publicly available ImageNet \cite{russakovsky2015imagenet} dataset. Code, models and samples used for the execution of our user study can be found at \url{https://github.com/maxdreyer/crp-human-study}. More information about data and models utilized in other experiments can be found in \sm~\ref{sec:appendix:dataset_and_models}.

The license to re-use and reproduce have been granted for the images shown in the figures of this manuscript and its supplementary material to the authors by the respective copyright holders by iStock, Shutterstock, Pixabay and Pexels.

\subsection*{Code Availability}
We provide an open-source CRP toolbox for the scientific community written in Python and based on PyTorch \cite{paszke2017automatic} and Zennit \cite{anders2021software}. The GitHub repository containing our implementations of \gls{crc} and \gls{rmax} is publicly available on \url{https://github.com/rachtibat/zennit-crp} \cite{achtibat2023zennit}. All experiments were conducted with Python 3.8, zennit-crp v0.6, Zennit v0.4.6 and PyTorch v1.13.1.

\subsection*{Acknowledgements}
We would like to express our gratitude to Alessa Angerschmid --- associated to the Human-Centered AI Lab at the University of Natural Resources, Vienna, and the Medical University of Graz --- for fruitful discussions and feedback.

\subsection*{Authors' Contributions}
Conceptualization and methodology: SL, RA, MD, SB, TW, WS;
Design of experiments: RA, MD, SL, SB, WS, TW;
Data analysis: RA, MD, SL;
Software: RA, IE, MD, SL;
Supervision and funding acquisition: SL, WS, TW;
Writing - original draft and revision: RA, MD, SL, WS, IE, SB, TW;

\subsection*{Conflict of Interest / Competing Interests}
The authors have no conflicts of interest to declare.

\clearpage
\bibliographystyle{plain}
\bibliography{bibliography.bib}

\cleardoublepage

\counterwithout{figure}{section}
\counterwithout{table}{section}
\setcounter{figure}{0}
\renewcommand{\figurename}{Supplementary Fig.}
\renewcommand{\tablename}{Supplementary Table}
\renewcommand{\thefigure}{\arabic{figure}}
\renewcommand\refname{Supplementary References}

\appendix

\section*{Supplementary Material}
This article has supplementary files providing additional details and information, descriptions, experiments and figures. \sm~\ref{sec:appendix:relatedwork} offers a detailed survey of current related work important to our contribution, contrasting several related explainability techniques to our proposed technical contribution regarding different criteria in Supplementary Table~\ref{tab:related:overview}. \sm~\ref{sec:appendix:methodsindetail} describes our first technical contribution, the \glsdesc{crc} technique, in increased detail and provides additional background. \sm~\ref{sec:appendix:methodsindetail:understanding} dives deep into \glsdesc{rmax} and expands on additional refinements for the presentation of representative examples compatible to sample sets selected from both \glsdesc{rmax} and \glsdesc{amax} procedures. Our methods' capabilities are demonstrated individually in \sm~\ref{sec:appendix:conceptunderstanding}. \sm~\ref{sec:appendix:conceptlocalization} then introduces the Concept Atlas and the idea of localized concept-based analyses of model predictions, as well as the investigation of relevance flows over concepts in different layers.
In \sm~\ref{sec:appendix:workflow} we summarize the computational steps involved in generating \gls{crc}- and \gls{rmax}  explanations, as well as the user workflow of user interactions with which explanations may be interpreted.
\sm~\ref{sec:appendix:study} provides additional details and results on the user study presented in Section~\ref{sec:results:study}, of which the outline was given in Section~\ref{sec:methods:human_evaluation_study_details} .
In \sm~\ref{sec:experiments:quantitative} we leverage the \gls{xai}-derived insights from \gls{crc} to evaluate several models' sensitivities regarding potentially spurious learned concepts in order to assess their individual effects on prediction making, and dive into concept subspaces to explore sets of filters covering various semantic topics learned from data.
\sm~\ref{sec:appendix:fairness} gives an example on how our methods can be utilized to spot systematically learned biases from single explanations given in a male vs.\ female face classification task, while 
\sm~\ref{sec:appendix:timeseries} discusses the application of our method in a medical time series use case.
Current challenges and an outlook to future work are discussed in \sm~\ref{sec:appendix:limitations}.
\sm~\ref{sec:appendix:dataset_and_models} concludes with a description of  the datasets and models used in our experiments, as well as implementation details for \gls{crc}.

\section{Survey of Related Work}
\label{sec:appendix:relatedwork}
With the increased awareness that \gls{ml} predictions need to be transparent, numerous approaches to \gls{xai} have emerged within the last decade (\cf \cite{samek2021explaining, das2020opportunities, nguyen2019understanding,Adadi2018Peeking}). In this section, we reiterate specific landmark approaches which have influenced and motivated our own \gls{crc} and the extended, exemplary explanation pipeline enabled by our \gls{rmax} paradigm. The current \gls{xai} landscape can roughly be divided into \emph{local} and  \emph{global} \gls{xai}. However, we are fully aware that a precise categorization is not always possible and the boundaries between have grown fuzzy with the ongoing development of the field, specifically with advances towards \emph{glocal} \gls{xai}. Supplementary Table~\ref{tab:related:overview}  provides a summary of the related work discussed in Appendices~\ref{sec:appendix:relatedwork:local} to~\ref{sec:appendix:relatedwork:glocal}  in terms of requirements and capabilities of recent methods of \gls{xai}.

\definecolor{OliveGreen}{rgb}{0.23,0.49,0.15} %
\newcommand{\yes}{{\color{OliveGreen}\checkmark}} %
\newcommand{\kinda}{{\color{orange}$\bigcirc$}}
\newcommand{\redx}{{\color{red}$\boldsymbol \times$}}
\newcommand{\softx}{{\color{orange}$\boldsymbol \times$}}

\begin{table*}[h]
    \centering
    \caption{A comparison of selected local, global and glocal \gls{xai} approaches at a glance, considering the explanatory insight they provide and  their specific requirements restricting the application. As explanatory capabilities it is considered whether explanations are \gls{classspecific}, \gls{samplespecific}, \gls{localized} in input space, providing \gls{examples} for learned latent concepts, or whether they can provide sample-specific feedback regarding latent \gls{features}. The table indicates if an explainer exhibits specific explanatory capabilities partially (\kinda) or fully (\yes). We assess specific requirements in terms of necessity to control the \gls{training} process, the need for specific \gls{dataNlabels} (or just labels) for the method to be applicable, and the restriction to a certain \gls{architecture}. Corresponding requirements restricting the applicability of an explainer are denoted by \redx\ for hard, and \softx\ for soft requirements.}
    \label{tab:related:overview}
    \resizebox{\textwidth}{!}{ %
    \begin{tabular}{|c  || l ||  c|c|c|c|c ||  c|c|c| }
            \hline
                \gls{xai} & \multicolumn{1}{c||}{Method} & \multicolumn{5}{c||}{Explaining Capabilities} & \multicolumn{3}{c|}{Requirements} \\
            \hline
              & & \gls{classspecific}   & \gls{samplespecific}     &  \gls{localized}  & \gls{examples} &  \gls{features}                  & \gls{training} & \gls{dataNlabels} & \gls{architecture} \\
            \hline
            \hline
            \parbox[t]{2mm}{\multirow{3}{*}{\rotatebox[origin=c]{90}{Local}}}
                & Gradient-based
             \cite{simonyan2015very,shrikumar2017learning, sundararajan2017axiomatic, zeiler2014visualizing}                
                & \yes & \kinda & \yes & & \yes            & & & \\
                & Mod.-Backpropagation 
                \cite{zeiler2014visualizing,bach2015pixel,montavon2017explaining,selvaraju2017grad,shrikumar2017learning}    
                & \yes & \kinda & \yes & &  \yes           & & & \softx \\
                & Perturbation / Surrogates \cite{guidotti2018local,ribeiro2016lime,ribeiro2018anchors, 
                lundberg2017Shap}            
                & \yes & \yes & \yes & &             & & & \\

            \hline
            \parbox[t]{2mm}{\multirow{4}{*}{\rotatebox[origin=c]{90}{Global}}}
                & Summit \cite{hohman2019summit} 
                & \kinda &  &  & \yes &   \kinda     &  &  &  \\
                & Network Dissection \cite{bau2017network}  
                & & \yes & \kinda & \yes  &     & & \redx & \redx  \\
                & TCAV \cite{kim2018interpretability} 
                &\kinda & \kinda & & \yes  & \kinda              & & \redx &  \\
                & Feature Visualization \cite{erhan2009visualizing,olah2017feature,olah2018building}  & & \kinda & \kinda & \yes & \yes             & & & \\
            \hline
            \parbox[t]{2mm}{\multirow{4}{*}{\rotatebox[origin=c]{90}{Glocal}}}
                
                & TCAV + IG \cite{schrouff2021best} 
                &\yes & \kinda &  & \yes &  \kinda                           & & \redx &  \\
                & StylEx \cite{lang2021explaining}
                & \yes & \yes & \yes & \yes  & \kinda                            & \redx & &   \\
                & ProtoPNet \cite{chen2019looks}& \yes & \yes & \kinda & \yes  & \kinda                          & \redx &  & \redx \\
            \cline{2-10}
                & \textbf{Ours} & \yes & \yes & \yes & \yes & \yes               & & & \softx \\
            \hline
        \end{tabular}
    } %
\end{table*}

\subsection{Local \texorpdfstring{\gls{xai}}{XAI}}
\label{sec:appendix:relatedwork:local}
Local post-hoc techniques aim to explain the classification of individual samples with no or little assumptions about the model's architecture or its training procedure. For this purpose, attribution maps are usually generated in the input domain, highlighting  from \emph{where} the model is deriving its inference outcome. In our view, the current diversity of post-hoc techniques can be categorized as follows: Interpretable Local Surrogates and Perturbation Analysis and Gradient-based and (modified) Backpropagation-based methods.

Regarding explanations for neurons in the hidden space,
(modified) gradient-based methods result in latent attributions as a by-product, which is not the case for methods based on learned explainable surrogate methods or input perturbation.
However, (modified) gradient-based methods may require the model to be differentiable or contain particular layers,  and thus have a soft requirement regarding the architecture.

\paragraph{Interpretable Local Surrogates \& Perturbation Analysis} 
Interpretable Local Surrogates replace the black-box model by a local surrogate model (\eg, a linear model) that locally approximates the model function for which an explanation is sought for. Since the surrogate has low complexity, interpretability is facilitated. Prominent methods include LIME \cite{ribeiro2016lime}, LORE \cite{guidotti2018local} and Anchors \cite{ribeiro2018anchors}. While these approaches are model-agnostic, they have a high computational cost in terms of forward passes, since each explanation requires sampling and prediction of several data points as well as fitting a local surrogate. Moreover, explanations are no longer generated on the original model, which is why internal subprocesses and representations of the original model cannot be traced back and explained.

In perturbation analysis, such as Occlusion-based attribution~\cite{zeiler2014visualizing}, the input features are repeatedly perturbed while the effect on the model output is measured~\cite{zeiler2014visualizing, zintgraf2017visualizing, fong2017interpretable,blucher2022preddiff}. Some methods form a combination of occlusion and surrogates, such as SHAP \cite{lundberg2017Shap}. Usually only the input is perturbed to calculate the influence on the output, but it is conceivable that intermediate layers are also perturbed to investigate the influence of single neurons. However, repeated perturbation leads to high computational cost, whereas backpropagation-based methods rely only on modified backward-passes to attribute all neurons. 

\paragraph{Gradient-based \& Modified Backpropagation-based Methods}
Early works \cite{morch1995visualization,baehrens2010explain} show, that the gradient of the model predictor with respect to the input features can be used to visualize the sensitivity of the model regarding the input features. Specifically, saliency maps can be constructed that show in the input space the attribution of each input feature. In order to stabilize the resulting heatmaps, gradients can be multiplied with the input, thereby approximating the model function in first order Taylor approximation \cite{shrikumar2017learning, montavon2017explaining}. Integrated Gradients \cite{sundararajan2017axiomatic} and SmoothGrad~\cite{smilkov2017smoothgrad} further increases robustness against gradient noise by averaging the gradient through multiple attribution steps.

Modified backpropagation-based methods further modify the backpropagation process in order to improve explainability. These methods therefore view calculations made by a \gls{dnn} as sequences of smaller layer-wise operations. Starting at the output, attributions are assigned layer after layer until the input is reached. These calculations may be implemented efficiently as a (modified) gradient backward pass. Prominent examples are DeepLIFT~\cite{shrikumar2017learning}, Deconvolution~\cite{zeiler2014visualizing}, Contribution Propagation~\cite{landecker2013interpreting}, \gls{lrp}~\cite{bach2015pixel} and Deep Taylor Decomposition~\cite{montavon2017explaining}. Notably, \gls{lrp} has been shown to achieve faithful intermediate attribution, as can be seen in \cite{becking2021ecq, yeom2021pruning}. These methods have the advantage that they efficiently provide attributions for intermediate neurons ``for free'' as a by-product, without any additional algorithmic extensions of the aforementioned. However, this by-product has usually been ignored in the literature except for a few works that use this information to directly improve specific aspects of deep models~\cite{molchanov2017pruning,yeom2021pruning,sun2022explain,becking2021ecq}, or regard them as proxy representations of explanations for the identification and eradication of systematic \glsdesc{ch} behavior~\cite{anders2022finding}. While we assume differentiability as a given for \glspl{dnn} in context of Supplementary Table~\ref{tab:related:overview},  modified backpropagation methods might require particular neural network building blocks to be present in order to be applicable, such as Global Average Pooling layers need to be part of the network for (Grad-)CAM~\cite{zhou2016learning,selvaraju2017grad} to be applicable.

\subsection{Global \texorpdfstring{\gls{xai}}{XAI}}
\label{sec:appendix:relatedwork:global}
Methods of global \gls{xai} aim at identifying learned concepts, \eg, by visualizing them and making their general interaction within the model comprehensible. Unlike local \gls{xai}, global \gls{xai} focuses on representing the overall behavior of the model without revealing the exact decision process for individual samples. Note, that all concept discovery and visualization techniques listed here are compatible with the \gls{crc} framework proposed in this work.

\paragraph{Neurons as Elementary Building Blocks for Explanations}
\label{sec:appendix:relatedwork:neurons}
It has been observed in numerous works \cite{zhou2015object, olah2017feature, radford2017learning, bau2020understanding, cammarata2020thread, goh2021multimodal} that stable, human-understandable concepts emerge in neurons, although the \gls{dnn} was not explicitly trained to use them. While low level neurons encode features like textures and edges, high level neurons conceptualize notions from simple objects to abstract emotions \cite{zhou2015object, olah2017feature}. Even the existence of multimodal neurons was reported that respond to a concept regardless whether it is shown as a photograph, a cartoon or just an image of the letters of a concept's name \cite{goh2021multimodal}. Although literature \cite{ilyas2019adversarial} that examines adversarial examples suggests that models might use features imperceptible to humans, practical experience and experiments show that neurons in general convey human understandable concepts. In this regard we highlight the work of \cite{bau2017network, bau2020understanding} where the authors present an analytical framework called ``Network Dissection'' through which the semantic meaning of a filter kernel in a \gls{cnn} layer can be identified, given densely labelled input data. In addition, the interested reader is referred to the work of Cammarata et al. \cite{cammarata2020thread} where an illustrative discussion of the foregoing topic is given.

\paragraph{Feature Visualization}
\label{sec:appendix:relatedwork:feature_visualization}
A large part of feature visualization techniques relies on \glsdesc{amax} of single neurons or a linear combination thereof~\cite{zhou2015object, olah2017feature, radford2017learning, bau2020understanding, cammarata2020thread, goh2021multimodal}, where in its simplest form, input images are sought that give rise to the highest activation value of a specific unit. Reference images can be generated synthetically using gradient descent, or alternatively found from a sample dataset by collecting neuron activations during predictions. While conceptually simple, preventing the emergence of adversarial synthetic examples became a main research endeavor. Several priors were proposed to guide optimization into realistic looking images, \eg, Transformation Robustness \cite{mordvintsev2015inceptionism}, Frequency Penalization \cite{mahendran2015understanding}, Preconditioning \cite{olah2017feature}, and learning a natural prior from data distribution \cite{nguyen2016synthesizing, yin2020dreaming} (more details in  Section~\ref{sec:appendix:methodsindetail:understanding:generative}). However, simply because a stimulus leads to high activation does not necessitate that the stimulus is representative of a neuron’s function --- adversarial examples are a prime example of this. Our proposed relevance-based sample selection (see~Section~\ref{sec:appendix:methodsindetail:understanding:selecting_reference_samples:relevance}) mitigates this issue by taking into account the context in which a neuron preferably activates. All feature visualization methods such as \cite{erhan2009visualizing, olah2017feature} generate human-understandable images that describe the concept of a neuron, but do not focus on concept interactions, or local analysis.

\paragraph{Projection-based Visualization \& Weight-based Graphs}
The Activation Atlas \cite{carter2019activation} unites dimensionality reduction with feature visualization. Activation vectors for millions of input images are computed and projected onto a low-dimension space via UMAP \cite{mcinnes2018umap} or t-SNE \cite{van2008visualizing}. Feature visualization is then applied on averaged activation vectors to get a general sense of what concepts the network is utilizing. While t-SNE and UMAP may introduce neighborhood distortions, TopoAct \cite{rathore2021topoact} differs by utilizing a tool from topological data analysis. This way, they obtain a topological summary by preserving the clusters as well as relationships between them in the original high-dimensional activation space revealing branches and loops. Focusing on the image domain, Summit and others \cite{hohman2019summit, liu2016towards} draw weight-based graphs that show how features interact in a global, yet class-specific scale, but without the capability to deliver explanations for individual data samples: SUMMIT \cite{hohman2019summit} measures on the one hand, which input samples lead to high activations, and on the other, which class is attributed to that input sample. Therefore, Summit allows to condition explanations on classes. Further, mean interactions between neurons of consecutive layers are found by investigating the activation flow. Thus, Summit allows inspecting the composition of features, but is thereby limited to a global analysis.

\paragraph{Concept Discovery}
\label{sec:appendix:relatedwork:concept_discovery}
Concept discovery methods rely on a priori known stimuli (via labeled data) to find concepts the model is sensitive to or is utilizing~\cite{rajalingham2018large}. However, the use of labeled data carries the risk of a human confirmation bias, that is tried to be minimized with the help of statistical calculations. Network Dissection \cite{bau2020understanding} and Net2Vec \cite{Fong2018Net2VecQA} find concepts by measuring the alignment between upsampled convolutional channel map activation and a ground truth segmentation mask, and thus require densely labeled data and a \gls{cnn} architecture. Moreover, upsampling might result in imprecise localization. While Network Dissection focuses on single channels, concept vectors are searched for in Net2Vec. 

\gls{tcav} \cite{kim2018interpretability} introduces concept activation vectors that are obtained by training linear classifiers to distinguish between pre-categorized data of a priori known concepts, \eg, encoding conceptual differences. TCAV thus requires concept data and labels. These vectors are subsequently used to compute directional derivatives to measure the model's prediction sensitivity \wrt a specific class. These sensitivities can be used for a local analysis as well, leading to TCAV explanations being class and sample specific, and requiring a differentiable model. The local explainability is however limited, as the sensitivity values are typically not stable due to the noisy gradients of very deep \glspl{dnn}. While \gls{tcav} relies on human-defined labels, the follow-up work ACE \cite{ghorbani2019automatic} clusters segments of the input image in an unsupervised way to generate proposal concept vectors. 

\subsection{Glocal \texorpdfstring{\gls{xai}}{XAI}}
\label{sec:appendix:relatedwork:glocal}
Glocal \gls{xai} aims to combine local and global \gls{xai} viewpoints in order to further improve explainability by minimizing the observer's interpretation workload.

\paragraph{Extending Concept Activation Vectors}
Since sensitivity scores only measure the degree of change in output, and not the actual contribution to the final prediction as attribution methods do, Schrouff et al. \cite{schrouff2021best} combine TCAV with Integrated Gradients and hereby enable local attribution, while preserving global explanation capabilities. As far as we understand, no heatmaps are computed for latent features in order to localize the concepts in input space. Moreover, ConceptSHAP \cite{yeh2020completeness} improves on the idea of TCAVs by defining a completeness-score with which concept vectors can be selected that contribute the most to a specific class. However, in both cases, the internal interaction between different concepts across layers and their localization in the input domain is not observable.

\paragraph{Attributing Internal Subprocesses}
To the best of our knowledge, the possibility of using model internal attributions for sample-specific explanations of \gls{dnn} subprocesses and feature interaction has been outlined first in the description of the Deep Taylor Decomposition Algorithm \cite{montavon2017explaining}, which serves as the theoretical basis for \gls{lrp}~\cite{bach2015pixel}. Later, Interpretable Basis Decomposition \cite{zhou2018Interpretable} was introduced that decomposes the weight matrix of a higher layer into semantically interpretable basis sets. Multiplying with activation vectors, the individual contribution to the final classification can be computed. Grad-CAM \cite{selvaraju2017grad}, an extension of CAM \cite{zhou2016learning}, is used to generate low-resolution heatmaps for individual latent components. However, this approach is limited to an application to the last convolutional layer in connection with a global pooling operation. In addition, basic components are not learned in an unsupervised fashion, but need to be identified via predefined label sets. Finally, Olah et al. \cite{olah2018building} have provided an informal proof-of-concept where the interaction between internal subprocesses and the localization of concepts via attribution maps have been demonstrated. However, the attribution algorithm is limited to a linear approximation between the potentially nonlinear hidden representation and input, and the visualization of concepts is restricted to \glsdesc{amax}.

\paragraph{Learning Generative Explainers}
A different glocal approach based on \glspl{gan} is presented in the work of StyleEx~\cite{lang2021explaining}.
Here, the authors propose to train a generative model to specifically explain multiple attributes that underlie classifier decisions in context of a particular input data point. To ensure that the learned \gls{gan}-based explainer uses similar representations as the classifier to be explained, the authors propose a training procedure, which tightly incorporates the classifier model. As a result, the features available to the \gls{gan} can therefore be manipulated in order to measure and understand their effect on the model's decision making. Further, a comparison of the original to the StyleEx-augmented image allows for localization of the manipulated feature in input space. In summary, for StyleEx to be applied, the training process must be adapted. Manipulating the latent space of the GAN, synthetic images are generated, which capture the change in concepts due to the manipulation. Therefore, explanations are sample-specific and conditioned on a class output of interest. Further, concepts can be localized by the visible change of pixel-values in the input space.

\paragraph{Analyzing Datasets of Attributions}
The work of Chan et al. \cite{chan2020Melody} proposed Melody, which groups similar local explanations together using techniques from information theory, thus generating a global overview of local explanations and consequently systematically learned concepts. The Spectral Relevance Analysis \cite{lapuschkin2019unmasking,anders2022finding} follows a similar idea and is built around the Spectral Clustering \cite{meilua2001random,von2007tutorial} of a dataset of attribution maps for any layer of representation in the network. Both methods are independent of the choice of local attribution algorithm. Another line of work processes attribution maps with methods from Inductive Logical Programming to identify and define relational concepts from post-hoc attributions \cite{rabold2019enriching, rabold2020expressive}.

\paragraph{Self-Explainable Models}
In contrast to post-hoc methods that apply to any \gls{dnn} model irrespective of architecture or training procedure, self-explainable models are specifically built or trained to maximize interpretability.  Examples of self-explainable models include deep network architectures with an explicit top-level sum-pooling structure \cite{chen2019looks, koh2020concept, poulin2006visual, lin2013network}, that reduces the attribution to a linear combination of features, making the reasoning process comprehensible. In particular, ProtoPNet~\cite{chen2019looks} proposes to build a model based on a convolutional architecture and introduces a loss during the training process to force the generation of human-understandable concepts/prototypes. Using activation maps, concepts can be localized --- resolution-wise limited by the size of the map. Thereby, single samples can be investigated as well. Other architectures involve attention mechanisms \cite{larochelle2010learning, bahdanau2014neural} whose inspection resembles attribution visualizations, or replacing a network’s final linear layer with a decision tree \cite{wan2020nbdt}, as well as specifically controlling the flow and representation of features throughout the model~\cite{rieckmann2020causes}. Another line of work \cite{finzel2021explanation} creates explanation trees for models based on learnable predicate logic, providing access to explanations at different levels of abstraction in the prediction process. While self-explainable models mitigate the requirement to use post-hoc methods for gaining insight (for further arguments for explainable models see \cite{rudin2019stop}), these methods can not be applied for explaining the inference of the more widely-spread and heavily used end-to-end trained black-box neural network models due to additional constraints imposed for the architecture or training procedures.

In contrast to aforementioned related work, our proposed method aims at making no additional assumptions about the model architecture, the availability of specific data- or label sets, or the structure of the learned latent space, while simultaneously providing human-aligned multifaceted explanations beyond the current state of the art. Conversely, this means that our contribution should be applicable to most purpose-built self-explaining model variants described in literature: Our method uses a modified backpropagation approach (\gls{crc}) in order to efficiently calculate attributions in the latent space,  and is thus weakly restricted to most of the currently used model architectures. For any sample, traditional heatmaps (conditioned on a class of interest) or concept-conditional heatmaps can be computed (localizing the concept in high resolution). Collecting the most relevant (via \gls{rmax}) or activating samples for latent neurons from a source dataset of choice serves for obtaining visualizations of concepts in the hidden space. By conditioning the relevance flow on (multiple) concepts, compositions and superpositions of concepts can be analyzed as well.

\cleardoublepage
\clearpage
\section{Concept Relevance Propagation}
\label{sec:appendix:methodsindetail}
In our work, we contribute technically by leveraging and extending the capabilities of the  \glsdesc{lrp} framework \cite{bach2015pixel}. After reiterating the general technical approach behind \gls{lrp} in Section~\ref{sec:appendix:methodsindetail:lrp}, we introduce our Concept Relevance Propagation (CRP) approach in  Section~\ref{sec:appendix:methodsindetail:lrp:disentangled}. Section~\ref{sec:appendix:methodsindetail:lrp:conceptual} follows with an overview of the novel and insightful analyses, which can be performed with CRP.

\subsection{\texorpdfstring{\glsdesc{lrp}}{Layer-wise Relevance Propagation}}
\label{sec:appendix:methodsindetail:lrp}
\glsdesc{lrp}~\cite{bach2015pixel} is a white-box attribution method grounded on the principles of flow conservation and proportional decomposition. Its application is aligned to the layered structure of machine learning models. Assuming a model with $L$ layers
\begin{equation} \label{eq:lrp_nn_appendix}
    f(\x)=f_L \circ \dots \circ  f_1(\x)~,
\end{equation}
\gls{lrp} follows the flow of activations and pre-activations computed during the forward pass through the model in opposite direction, from the final layer $f_L$ back to the input mapping $f_1$. Let us consider some (internal) layer or mapping function $f_\ast(\cdot)$ in the model. Within such a layer, \gls{lrp} assumes the computation of pre-activations $\preact_{ij}$, mapping inputs $i$ to outputs $j$, which are then aggregated as $\preact_j$ at $j$, \eg, by summation. Commonly, in neural network architectures, such a computation is given with
\begin{align}
    \preact_{ij} & = \act_iw_{ij}\label{eq:appendix:nn_forward_zij}\\
    \preact_j & = \sum_i \preact_{ij}\label{eq:appendix:nn_forward_zj}\\
    \act_j & = \sigma(\preact_j)\label{eq:appendix:nn_forward_a}~,
\end{align}
where $\act_i$ are the input activations passed from the previous layer and $w_{ij}$ are the layer's learned weight parameters, mapping inputs $i$ to layer outputs $j$. Note that the aggregation by summation to $\preact_j$ can be generalized, \eg, to also support max-pooling by formulating the sum as a $p$-means pooling operation~\cite{bach2015pixel}. Finally, $\sigma$ constitutes a (component-wise) non-linearity producing input activation for the succeeding layer(s). In order to be able to perform its relevance backward pass, \gls{lrp} assumes the relevance score of a layer output $j$ as given as $R_j$. The algorithm usually starts by using any (singular) model output of interest as an initial relevance quantity. In its most basic form, the method then distributes the quantity $R_j$ towards the neuron's input as
\begin{equation} \label{eq:appendix:lrp_basic}
    R_{i \leftarrow j} = \frac{\preact_{ij}}{\preact_j}R_j~,
\end{equation}
\ie, proportionally \wrt\ the relative contribution of $\preact_{ij}$ to $\preact_j$. Lower neuron relevance is obtained by simply aggregating all incoming relevance messages $R_{i \leftarrow j}$ without loss:
\begin{equation} \label{eq:appendix:lrp_aggregation}
    R_i = \sum_j R_{i \leftarrow j}
\end{equation}
This proportionality simultaneously ensures a conservation of relevance during decomposition as well as between adjacent layers, \ie,
\begin{equation}\label{eq:appendix:lrp_equality}
    \sum_i R_i = \sum_i\sum_j R_{i \leftarrow j} = \sum_j \sum_i \frac{\preact_{ij}}{\preact_j}R_j = \sum_j R_j~.
\end{equation}
Note, that above formalism, at the scope of a layer, introduces the variables $i$ and $j$ as the inputs and outputs of the whole layer mapping, and assumes $\preact_{ij} = 0$  for unconnected pairs of $i$ and $j$, as it is the case in single applications of filters in, \eg, convolutional layers. For component-wise non-linearities $\sigma$ in Equation~\eqref{eq:appendix:nn_forward_a}, commonly implemented by, \eg, the tanh or ReLU functions which (by \gls{lrp}) typically are treated as a separate layer instances, this results in $\preact_{ij} = \delta_{ij}\preact_{j}$ (with $\delta_{ij}$ being the Kronecker-Delta representing the input-output connectivity between all $i$ and $j$) and consequently in an identity backward pass through $\sigma$. This principle of attribution computation by relevance decomposition can be implemented and executed efficiently as a modification of gradient backpropagation~\cite{montavon2019layer}, \cf, \eg~\cite{alber2019innvestigate,anders2021software}.

In order to ensure robust decompositions and thus stable heatmaps and explanations, several purposed LRP rules have been proposed in literature (see \cite{montavon2019layer,samek2021explaining} for an overview), for which Equations~\eqref{eq:appendix:lrp_basic} and~\eqref{eq:appendix:lrp_aggregation} serve as a conceptual basis. Recent works further recommend a composite strategy, mapping different rules to different parts of a neural network~\cite{montavon2019layer,kohlbrenner2020towards,samek2021explaining}, which qualitatively and quantitatively increases attribution quality for the intent of explaining prediction outcomes. In the following analysis, different composite strategies are therefore used following recent recommendations from literature. See Appendix~\ref{sec:appendix:technical} for the technical details regarding the computation of attributions used in this work.

\subsection{Disentangling Explanations with \texorpdfstring{\glsdesc{crc}}{Concept Relevance Propagation}}
\label{sec:appendix:methodsindetail:lrp:disentangled}
\glsdesc{lrp}, like other backpropagation-based methods (\eg, \cite{zeiler2014visualizing,springenberg2015simplicity,shrikumar2017learning}), computes attribution scores for all hidden units of a neural network model in order to allot a score to the model input. While in some recent works those hidden layer attribution scores have been used as a (not further semantically interpreted) means to improve deep models~\cite{molchanov2017pruning,yeom2021pruning,sun2022explain,becking2021ecq}, or as proxy representations of explanations for the identification of systematic Clever Hans behavior~\cite{anders2022finding}, they are usually disregarded as a ``by-product'' for obtaining per-sample explanations in input space. The reason is fairly simple: End-to-end learned representations of data in latent space are usually difficult (or impossible) to interpret, other than the samples in input space, \eg, images. Using attribution scores for rating the importance of individual yet undecipherable features and their activations does not provide any further insight about the model's inference process. We further discuss this problem, and propose a solution to this uninterpretability of latent representations in Section~\ref{sec:appendix:methodsindetail:understanding}.

Let us for now assume that we do have an understanding of the distinct roles of latent filters and neurons in end-to-end learned \glspl{dnn}. Then, another problem emerges for interpreting model decisions in input space, rooted in the mathematics of modified backpropagation approaches. As Equation~\eqref{eq:appendix:lrp_equality} summarizes for intermediate layers, \gls{lrp} (and related approaches) propagates quantities from all layer outputs $j$ simultaneously to all layer inputs $i$. This leads at a layer input to a weighted superposition of attribution maps received from all upper layer representations, where detailed information about the individual roles of interacting latent representations is lost. What remains is a coarse map identifying the (general) significance of an input feature (\eg, a pixel) to the preceding inference step. A notable difference to this superimposing backpropagation procedure within the model is the initialization of the backpropagation process,  where usually only one network output, of which the meaning (\eg, representation of categorical membership) generally is known, is selected for providing an initial relevance attribution quantity, and all others are masked by zeroes. This guarantees that an explanation heatmap represents the significance of (input) features to only the model output of choice. Let us call this heatmap representation a (class or output) \emph{conditional relevance} map $R(\x|y)$ specific to a network output $y$ and a given sample $\x$. Would one backpropagate all network outputs simultaneously, as it is demonstrated in Supplementary Figure~\ref{fig:appendix:methods:lrp:entangle}a, class-specificity would be lost, and again only information about ``some'' not further specified feature importance could be obtainable.
Still, even class-specific attribution maps can be uninformative, as shown in Supplementary Figure~\ref{fig:appendix:methods:lrp:entangle}b.
Here, attributions tend to mark the same body parts for all bird species, as a result of attribution scores specific to latent concepts superposed in input space.
In all explanations, we see that, \eg, the bird's head seems to be of importance. We do not know, however, whether the animal's eyes or beak carry some individual characteristics recognized and utilized by the model.

\begin{figure*}[t]
    \centering
    \includegraphics[width=\textwidth]{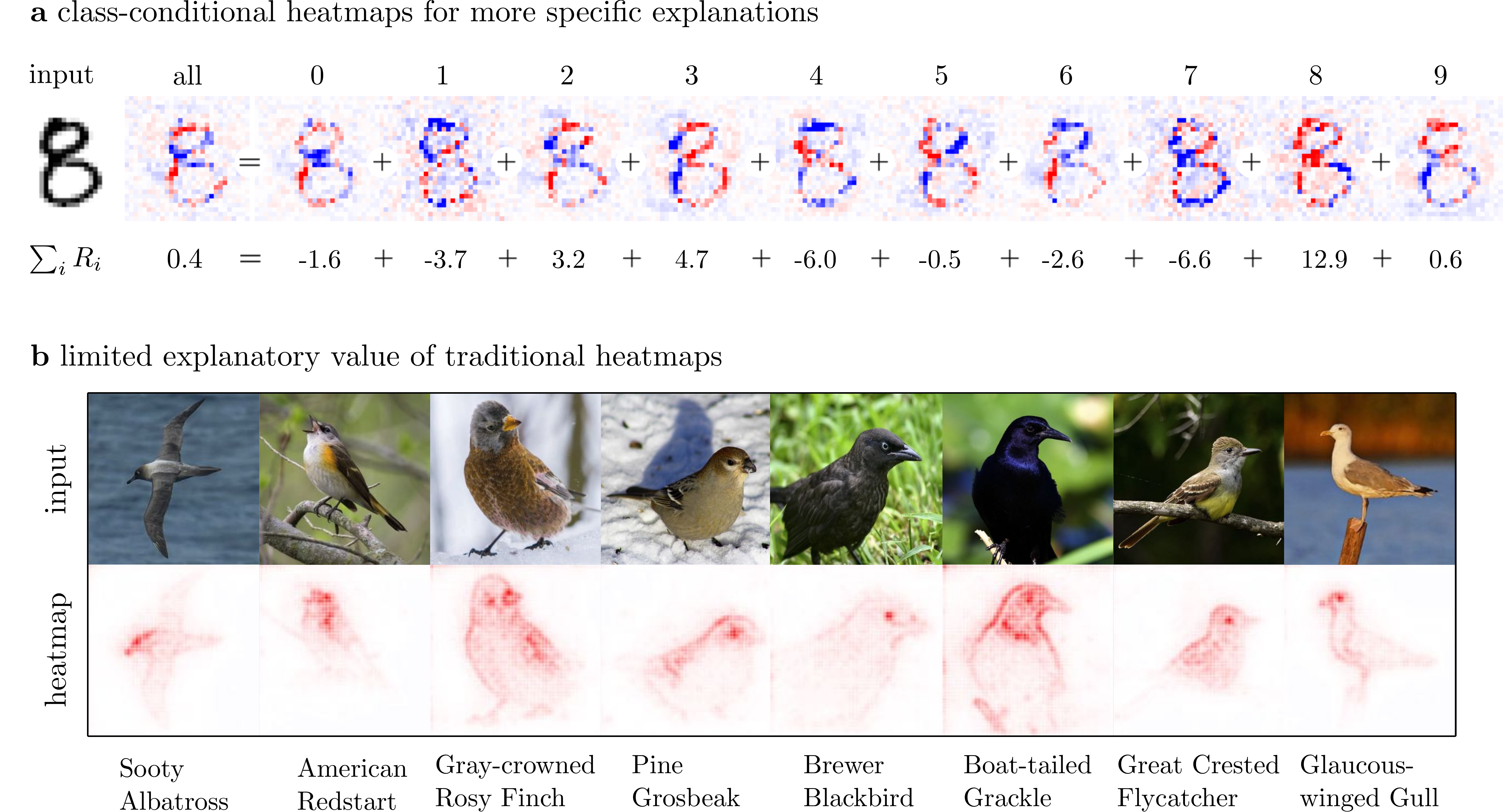}
    \caption{\textbf{a)} Class-conditional heatmap example for MNIST using \gls{lrp}: Selecting one specific output class (see numbers above) for the heatmap calculation leads to explanations conveying precise meaning, namely ``which features in their given state support (red) or contradict (blue) the class output'', compared to choosing all output classes at once, where the meaning of the class-specific sign of the attribution is lost. For \gls{lrp}, the heatmap computed for the whole output at once is a superposition of all class-specific heatmaps. 
    \textbf{b)} In some cases, traditional heatmaps can still be rather uninformative despite being class-specific, as is shown for bird species classification examples. Here, heatmaps only hint at the location of relevant body parts, without specifying the (different) concepts, such as, \eg, species-specific beak shapes, being recognized and considered by the model.
    A reference colormap for heatmaps shown in this paper is given in \edited{Appendix~\ref{sec:appendix:technical:colormap}}{Supplementary Figure~\ref{fig:appendix:color_map}}.}
    \label{fig:appendix:methods:lrp:entangle}
\end{figure*}
In the following, we propose strategies for disentangling attribution scores for latent representations in order to increase the semantic fidelity of explaining heatmaps via \glsfirst{crc}. We introduced the notion of a class- or output-conditional relevance quantity $R(\x|y)$ for describing the use of knowledge about the meaning of particular neural network neurons and filters and their represented concepts --- here the categories represented by the neurons at the model output.  The key idea for obtaining $R(\x|y)$ is the masking of unwanted network outputs prior to the backpropagation process via a multiplication with zeroes. Perpetuating the notation introduced in the previous Section~\ref{sec:appendix:methodsindetail:lrp}, obtaining the attribution scores $R^1_i(\x|y)$ for input units $i$ corresponding to the individual components, features or dimensions $x_i$ of input sample $\x$ at layer $l=1$ and model output category $y$ is achievable by initializing the layer-wise backpropagation process with the initial relevance quantity $R^L_j(\x|y) = \delta_{jy}f^L_j(\x)$, with $f^L_j(\x)$ being the model output of the $j$-th neuron at output layer $L$. Using the Kronecker-Delta $\delta_{jy}$, only the output of the neuron corresponding to the output category $y$ is propagated in the backward pass. Let us uphold our assumption of knowledge about the concepts encoded by each filter or neuron within a \gls{dnn}. We generalize the principle of masking or selecting the model output for a particular outcome to be explained by introducing the variable $\theta$ describing a set of conditions $c_l$ %
bound to representations of concepts and applying to layers $l$. Multiple such conditions in combination might extend over multiple layers of a network. Note that we use natural numbers as identifying indicators for neural network filters (or elements in general) in compliance to the Kronecker Delta. Here, $\theta$ then allows for a (multi-) concept-conditional computation of relevance attributions $R(\x|\theta)$.

We therefore extend the relevance decomposition formula in Equation~\eqref{eq:appendix:lrp_basic}, following the potential for a ``filtering'' functionality briefly discussed in~\cite{montavon2018methods} to
\begin{align}
    R^{(l-1,l)}_{i \leftarrow j}(\x|\theta \cup \theta_{l}) = \frac{\preact_{ij}}{\preact_j} \cdot \sum_{c_l \in \theta_l}\delta_{jc_l} \cdot R^l_j(\x|\theta) 
    \label{eq:appendix:lrp:masked}
\end{align}
where $\delta_{jc_l}$ ``selects'' the relevance quantity $R^l_j$ of layer $l$ and neuron $j$ for further propagation, if $j$ meets the condition(s) $c_l$ tied to concepts we are interested in. Note that for layers $l'$ without explicitly selected conditions, our notation assumes all conditions applicable in that layer to be valid, \ie, if $\theta_{l'} = \varnothing$ is an empty set, we define $\forall j~\exists ! c_{l'}\in \theta_{l'}: c_{l'}=j$ and therefore $\forall j~\sum_{c_{l'}\in\theta_l}\delta_{jc_{l'}} = 1$. Thus, without conditions, attribution flow is not constrained, as no masking is applied. Furthermore, due to our approach based on binary masking, which controls the flow of backpropagated quantities through the model, this assumption illustrates that a combination of conditions within a layer $l$ behaves similarly to a logical OR $(\lor)$ operator, and combinations of conditions across layers behave similar to logical AND ($\land$) operators.

\subsection{Possible Analyses with CRP}
\label{sec:appendix:methodsindetail:lrp:conceptual}
The most basic form of relevance disentanglement is the masking of neural network outputs for procuring class-specific heatmaps, as is shown in Supplementary Figure~\ref{fig:appendix:methods:lrp:entangle}a. Here, heatmaps gain (more) detailed meaning by specifying a class output for attribution distribution, answering the question of ``which features are relevant for predicting (against) a chosen class''. Backpropagation-based XAI methods also assign attribution scores to neurons of intermediate layers, and thus further reveal the relevance of hidden neurons for a prediction. Regarding DNNs, these hidden neurons can represent human-understandable concepts. It has been shown that the meaning of filters in a neural network is hierarchically organized within its sequence of layered representations, meaning that an abstract latent representation within the model is based on (weighted) combinations of simpler concepts in lower layers~\cite{zeiler2014visualizing,bau2017network,hohman2019summit}. Such concepts can be allocated to individual neurons or groups of neurons, \eg, a filter or filters of a convolutional layer of a \gls{dnn}. By introducing (multi-)conditional \gls{crc}, \ie, via a  masking of hidden neurons, the relevance contribution of individual concepts used by a neural network can be, in principle, disentangled and individually investigated as well. This expands the information horizon to questions such as ``how relevant a particular concept is for the prediction'', or ``which features are relevant for a specific concept''.

\subsubsection{Global Concept Importance}
\label{sec:appendix:global_concept_importance}
The possibility to isolate and identify the global (wrt.\ the whole sample processed by the model) importance of concepts with \gls{crc} is illustrated in Supplementary Figure~\ref{fig:appendix:disentangle:conceptual}, and presumes the ability to meaningfully group the neurons of a hidden (\eg, convolutional) layer as per their function. While this will in practice be an arguably difficult task semantically (which is explored in Section~\ref{sec:experiments:quantitative:clusters} of the main paper), we can exploit prior knowledge about the structure of the data and the architecture of the model at hand.

\begin{figure*}[t]
    \centering
    \includegraphics[width=\textwidth]{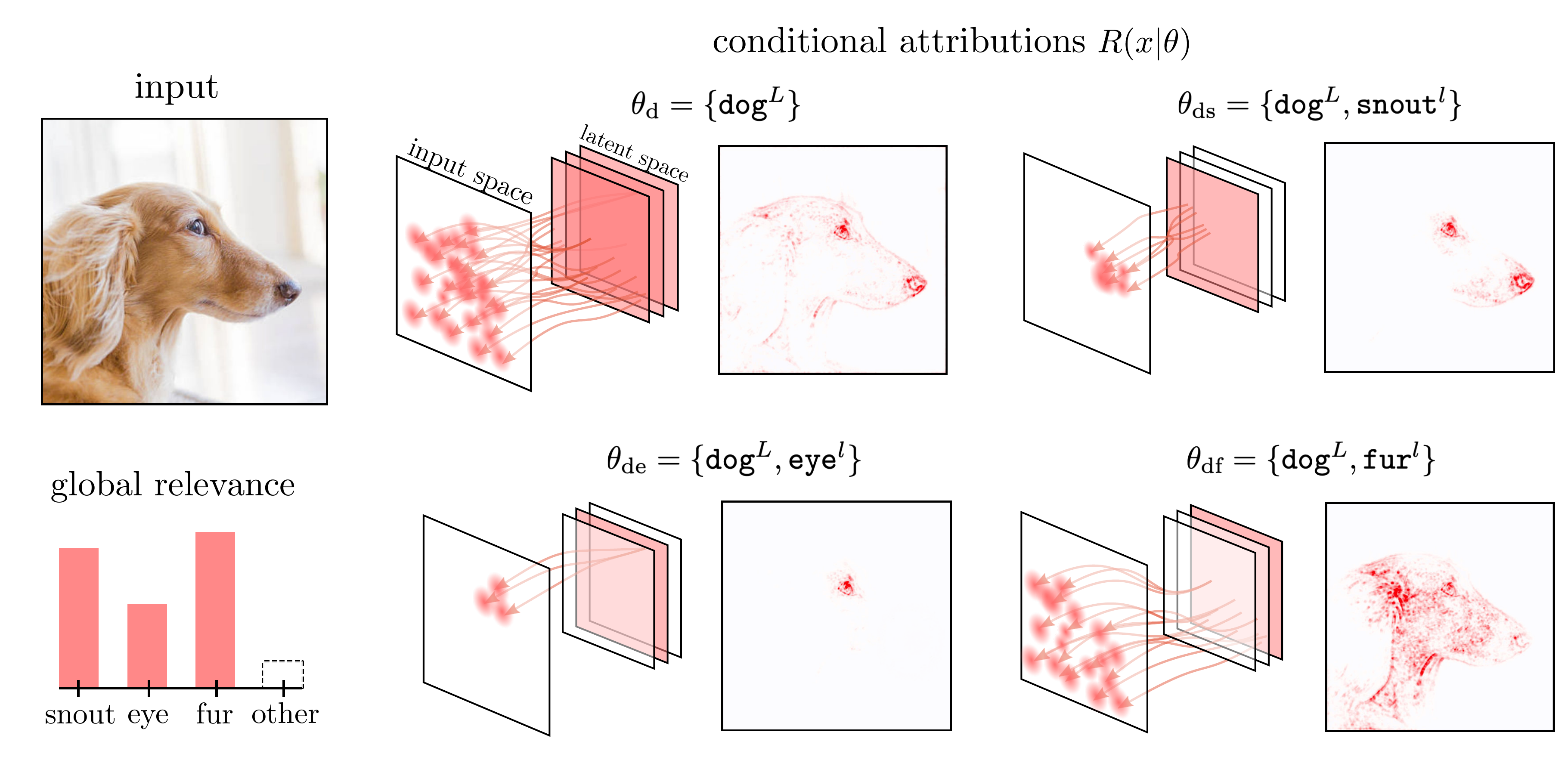}
    \caption{Explanation disentanglement via \glsdesc{crc}. Target concept ``dog'' as expressed by an output neuron in layer $L$ is described by a combination of lower-level concepts such as ``snout'', ``eye'' and ``fur'' encoded in a lower layer $l$. \gls{crc} heatmaps regarding individual concepts, and their contribution to the prediction of ``dog'', can be generated by applying masks to filter-channels in the backward pass. Global (in the context of an input sample) relevance of a concept \wrt\ to the explained prediction can thus not only be measured in latent space, but also precisely visualized, localized and measured in input space. The concept-conditional computation of $R(\x|\theta_{df})$ reveals the relatively high importance of the spatially distributed ``fur'' feature for the prediction of ``dog'', compared to the feature ``eye''. The attribution of $R(\x|\theta_{df})$ in the input space visualization of $R(\x|\theta_{d})$ (which was computed jointly over all concepts), however, is dominated by $R(\x|\theta_{de})$ and $R(\x|\theta_{ds})$ which both concentrate more strongly on smaller image regions and attribute both to the dog's eye. Here, the visualization of $R(\x|\theta_{d})$ alone does not represent the relative importance of the concepts' contributions to the ``dog'' outcome.}
    \label{fig:appendix:disentangle:conceptual}
\end{figure*}

In the example shown in Supplementary Figure~\ref{fig:appendix:disentangle:conceptual},
we use a feed-forward \gls{cnn} based on the VGG-16~\cite{simonyan2015very} model trained for object recognition on image data, which typically consists of blocks of convolutional and fully connected layers. The 2D-convolutional layers of the model consist of multiple learned filters sliding over the two spatial axes of the output activation of the previous layer (and transitively, over the input image), measuring the presence of specific structures in input space. Specifically, the tensor of output activations in a convolutional layer $l$ is of shape $(h_l, w_l, n_l)$, where $n_l$ is the number of filters applied to each of the possible $h_l \times w_l$ vertical and horizontal positions in the input tensor, depending on filter size and stride. Since convolutional layers share the same set of weights among all applications along the spatial axes, and thus measure the same features, it can be assumed that each of the $n_l$ output channels (or coactive group of output channels) encodes one particular concept. Note, that for fully connected layers, each output neuron receives inputs from \emph{all} preceding input neurons. Thus, each output neuron (or coactive group of output neurons) can be regarded as the encoding of a concept. Consequently, the concept-conditional masking of the attribution backpropagation introduced in Equation~\eqref{eq:appendix:lrp:masked} can be applied ``as is'' to the concepts encoded via fully connected layers. For convolutional layers, we group the output neurons $j$ into spatial and channel axes, and apply the masking to the channel axis only for the selection of concepts for backpropagation:
\begin{equation}
    R^{(l-1,l)}_{i \leftarrow \left(p,q,j\right)}(\x|\theta \cup \theta_{l}) = \frac{\preact_{i\left(p,q,j\right)}}{\preact_{\left(p,q,j\right)}} \cdot \sum_{c_l\in\theta_l}\delta_{jc_l} \cdot R^l_{\left(p,q,j\right)}(\x|\theta)
    \label{eq:appendix:lrp:masked_regrouped}
\end{equation}
Here, the tuple $\left(p,q,j\right)$ uniquely addresses an output voxel of the activation tensor $\preact_{\left(p,q,j\right)}$ computed during the forward pass through $l$, with $p$ indicating the first and $q$ the second spatial axis of the tensor, and $j$ the channel- or (as per our assumptions) the concept-axis. Note that this notion of neuron groups based on prior knowledge about the data and model architecture can trivially be adapted to other cases, such as models using 1D-convolutions to process time series data (see Section~\ref{sec:appendix:timeseries}), etc. Our knowledge about the structure of the data and the architecture of the model allows the attribution-based selection of specific filters along the channel axis of a hidden convolutional layer, which can be identified with concepts such as ``snout'', ``eye'' and ``fur''. Here, we are then able to compare attribution maps \wrt\ class ``dog'' by selecting the respective model output at final layer $L$ to, \eg, the attributions for ``dog $\land$ fur'' by additionally selecting channels responsible for fur pattern, animal snout or eye-like features encoded in a lower layer $l$. Specifically, in Supplementary Figure~\ref{fig:appendix:disentangle:conceptual} we choose
$\theta_\text{d} = \lbrace L:\lbrace\texttt{dog}\rbrace\rbrace$,
$\theta_\text{df} = \lbrace l:\lbrace\texttt{fur}\rbrace,L:\lbrace\texttt{dog}\rbrace\rbrace$,
$\theta_\text{ds} = \lbrace l:\lbrace\texttt{snout}\rbrace,L:\lbrace\texttt{dog}\rbrace\rbrace$ and 
$\theta_\text{de} = \lbrace l:\lbrace\texttt{eye}\rbrace,L:\lbrace\texttt{dog}\rbrace\rbrace$, for the explanations of ``dog'', ``dog $\land$ fur'', ``dog $\land$ snout'' and ``dog $\land$ eye'', respectively. To uphold readability, however, we will define the following equivalent shorthand notation of \mbox{$\theta_\text{df} = \lbrace \texttt{dog}^L,\texttt{fur}^l\rbrace$} throughout the rest of the paper at our convenience. Therefore, the unique placement of concept-encoding feature descriptors within the model and the application of conditions at the correct location is assumed. At this point, the relevance $R^l(\x|\theta_c)$ of a concept (combination) $c$ can be calculated by summation of relevances $R^l_i(\x|\theta_c)$ in a given layer, \eg, at or below the lowest layer $l$ where a part of the concept (combination) is expressed in latent space, as 
\begin{equation}
    R^l(\x|\theta_c) = \sum_{i}R^l_i(\x|\theta_c)~,
    \label{eq:appendix:lrp:masked_aggregated}
\end{equation}
or any other lower layer (\eg, the input space) to which the masking of the relevance flow \wrt\ $c$ has taken full effect. Here, the conservation laws of \gls{crc} inherited from \gls{lrp} (\cf Equation~\eqref{eq:appendix:lrp_equality}) ensure that the amount of relevance $R^l(\x|\theta_c)$ corresponds to the impact a concept combination selected via $\theta_c$ has had to inference. The relevance scores assigned to specific concepts then give insight into the decision process of the model. Regarding a particular prediction, relevance scores aggregated for each concept illuminate which concept is used by the model during inference and how much it contributes to the prediction outcome.

Given the example shown in Supplementary Figure~\ref{fig:appendix:disentangle:conceptual}, we can see that the concept ``fur'' is more important for the prediction of class ``dog'' than for example the concept ``eye'', which is not apparent from the heatmap visualization of $R(\x|\theta_\text{d})$ alone. Here, the importance of ``eye'' is more explicit, despite its comparatively lower measured relevance sum, as this lower total quantity of relevance concentrates on a considerably smaller set of pixels (\cf $R(\x|\theta_\text{de})$ vs. $R(\x|\theta_\text{df})$), increasing the per-pixel relevance magnitude reflecting in visualized heatmap color space in bright red color.
\begin{figure*}[th!]
    \centering
    \includegraphics[width=0.7\textwidth]{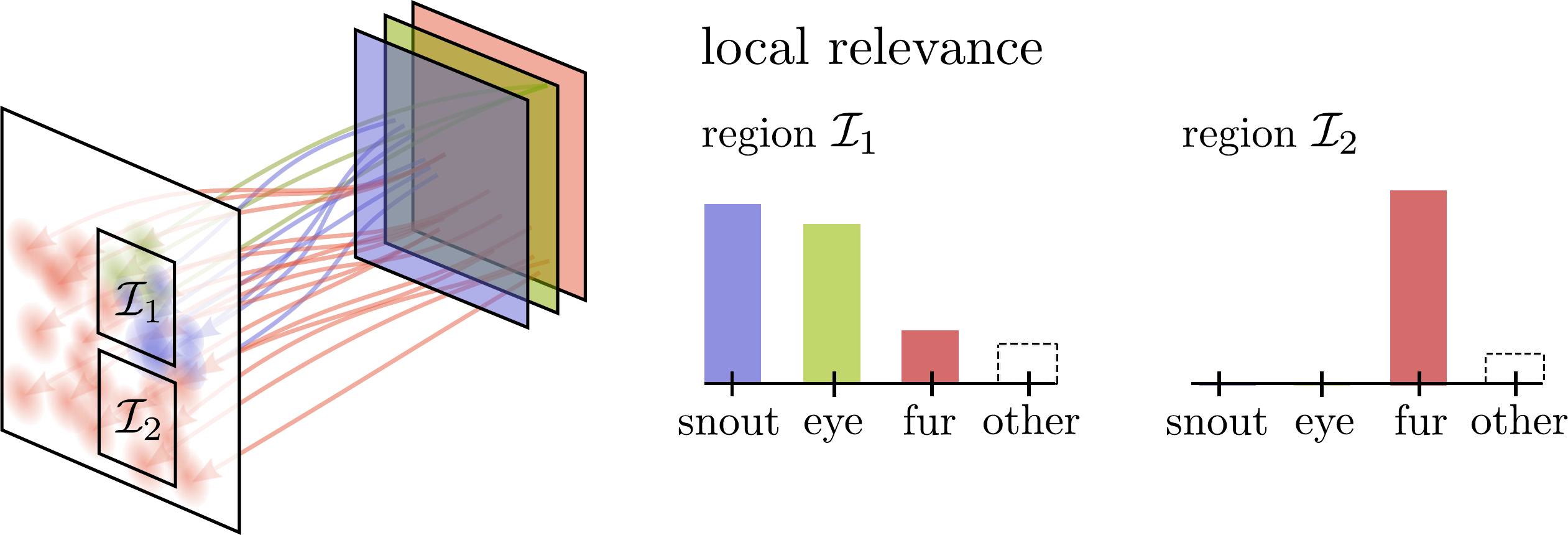}
    \caption{\glsdesc{crc}, in combination with local aggregation of relevance scores $R(\x|\theta)$ over regions $\mathcal{I}_1$ and $\mathcal{I}_2$, based on the example given in Supplementary Figure~\ref{fig:appendix:disentangle:conceptual}.
    Specifically, we compute $R_i^1(\x|\theta_\text{ds})$, $R_i^1(\x|\theta_\text{de})$, $R_i^1(\x|\theta_\text{df})$ and others in order to assess their local contribution to the prediction outcome for class ``dog'', and the origin of their contributions among image regions $\mathcal{I}_1$ and $\mathcal{I}_2$.}
    \label{fig:appendix:lrp-hidden-spatial-disentanglement}
\end{figure*}
\subsubsection{Local Concept Importance}
\label{sec:appendix:methodsindetail:local_conceptual_importance}
Visualizations of concept-conditional relevance in heatmaps show where concepts contributing to a chosen network output are localized and recognized by the model in the input space. Typically, as discussed in Section~\ref{sec:appendix:methodsindetail:lrp:disentangled}, an explanation heatmap regarding a singular specific output class is described by a combination of interactions of individual concepts. The bar chart in the bottom left of Supplementary Figure~\ref{fig:appendix:disentangle:conceptual} illustrates how class-conditional attribution maps (here, for class ``dog'') can be separated into the individual contributions of learned concepts using further refinement of the attribution backpropagation \wrt\ conceptual conditioning, on a global scale in context of the given sample.  At this point, given $R_i^l(\x|\theta_c)$, conceptual disentanglement allows measuring the individual importance of the in $\theta_\text{c}$ selected concepts $\text{c}$ on a local scale, \ie, over a subset $\mathcal{I}$ of neurons $i$ in layer $l$. Specifically, relevance scores $R_i^l(\x|\theta_\text{c})$ in, \eg, input space, can be aggregated meaningfully over image regions for the concept $\text{c}$ to a localized relevance score
\begin{equation}
    R_{\mathcal{I}}^l(\x|\theta_c) = \sum_{i\in\mathcal{I}} R_i^l(\x|\theta_c)
    \label{eq:appendix:lrp:masked_aggregated_local}
\end{equation}
measuring the importance of a concept to the prediction on a set of given input features, \eg, pixels. Extending the notation introduced in Equation~\eqref{eq:appendix:lrp:masked_regrouped}, for a convolutional layer with $J$ channels,  local relevance aggregation along the spatial axis is given by
\begin{equation}
    R_{\mathcal{I}}^l(\x|\theta_c) = \sum_{(p, q) \in \mathcal{I}}\sum_{j=1}^{J} R_{(p, q, j)}^l(\x|\theta_c)~,
    \label{eq:appendix:lrp:masked_aggregated_local_conv}
\end{equation}
aggregating over all positions $(p, q)$ defined in the set $\mathcal{I}$.

For methods adhering to a conservation property such as \gls{crc}, this property permits the comparison of multiple local image regions $\mathcal{I}$ and/or sets of concepts $c$ in terms of (relative) importance of learned latent concepts, as illustrated in Supplementary Figure~\ref{fig:appendix:lrp-hidden-spatial-disentanglement}. In the given example we compute relevance scores according to $\theta_\text{ds}$, $\theta_\text{de}$, $\theta_\text{df}$, et cetera, individually in input space. Then we aggregate the resulting attribution maps according to Equation~\eqref{eq:appendix:lrp:masked_aggregated_local} over two input regions $\mathcal{I}_1$ and $\mathcal{I}_2$ in order to locally measure the relative importance of by the model perceived concepts. As seen later, the capabilities of localized \gls{crc} can be utilized to visualize a ``Concept Atlas'' which demonstrates which concepts models perceive and use locally for their decision making process.
\begin{figure*}[t]
    \centering
    \includegraphics[width=0.65\textwidth]{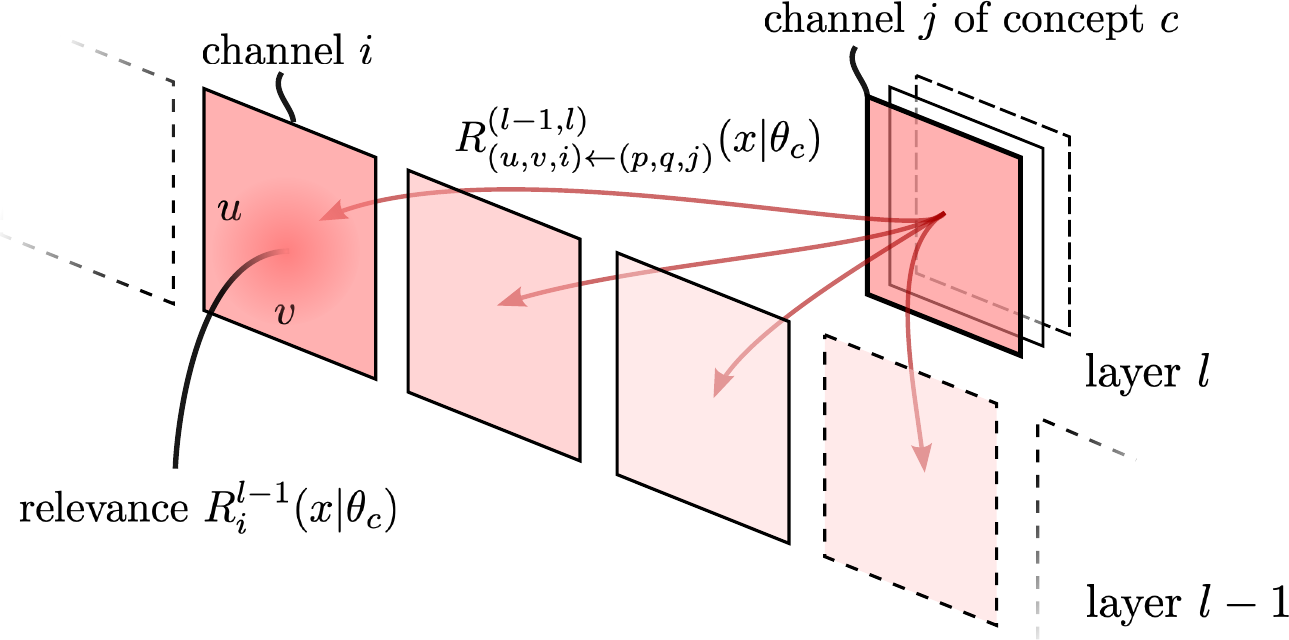}
    \caption{Given a particularly interesting concept $c$ encoded by channels $j$ in layer $l$, as well as $R^l_i(\x|\theta_c)$, adequately aggregated relevance quantities computed during a \gls{crc} backward pass can be utilized to identify how $R_j^l(\x|\theta_c)$ distributes across lower layer channels representing concepts. The identification of the most influential concepts in layer $l-1$ contributing to concept $c$ in layer $l$ \wrt\ an explanation computed regarding $\theta_c$ is then just a matter of ranking $R_i^{l-1}(\x|\theta_c)$, \eg, by (signed) magnitude.
    \label{fig:appendix:lrp-concept-composition}
    }
\end{figure*}
\subsubsection{Hierarchical Composition of Relevant Concepts}
Let us assume we have identified a particularly relevant concept $c$ located in layer $l$ via the computation of concept-conditional relevance attributions, resulting in a relevance quantity $R^{l}_{(p,q,j)}(\x|\theta_c)$. Let us further assume that we are interested in the composition of concept $c$ in terms of sub-concepts in preceding network layers, as used by the model for the purpose of inferring the explained outcome. Leveraging \gls{crc}, we are able to go beyond the capabilities of related work \cite{hohman2019summit, liu2016towards}, which often is limited to the identification of per-model (\ie,~static) or per-class interactions of concepts via the analysis of weights and/or activations.  Specifically, using \gls{crc}, we can compute \emph{interactions of concepts} encoded by the predictor $f$ in context of the prediction of a particular sample $\x$, via the corresponding model outcome $f(\x)$ (as well as the conditions $\theta_c$). Let us further uphold the assumption that our model is a convolutional \gls{dnn} for image categorization, \ie, the activation and attribution tensors we have access to have three axes. This assumption can of course be adapted to any other structuring of layer activations. Analogously to the tuple $(p,q,j)$ indexing the spatial and channel axes of the upstream activations and relevance tensors at layer $l$, we introduce the tuple $(u,v,i)$ applying to downstream tensors at preceding layer $l-1$, with the spatial coordinates $u$ and $v$, as well as the channel coordinate $i$ addressing the different features/concepts encoded by the model. We can now further extend Equation \eqref{eq:appendix:lrp:masked_regrouped} to
\begin{equation}
    R^{(l-1,l)}_{(u,v,i) \leftarrow (p,q,k)}(\x|\theta_c)  =  \frac{\preact_{(u,v,i)(p,q,k)}}{\preact_{(p,q,k)}}R_{(p,q,k)}^{l}(\x|\theta_c)~,
    \label{eq:appendix:lrp:superexpandend:decomposition}
\end{equation}
where $R_{(p,q,k)}^{l}(\x|\theta_c)$ is the already masked upstream relevance for concept $c$ at layer $l$ (and above), \ie, all channels $j$ of layer $l$ are masked except channel $k$. Further, the resulting $R^{(l,l+1)}_{(u,v,i) \leftarrow (p,q,j)}(\x|\theta_c)$ represents the flow of downstream relevance messages from all upstream voxels $(p,q,j)$ to downstream voxels $(u,v,i)$.
In order to find the most relevant lower layer concepts in layer $l-1$ given our specific sample $\x$ and conditions $\theta_c$, we can simply aggregate the downstream relevance into per-channel/concept relevances at layer $l-1$ over the spatial coordinates $u$ and $v$:
\begin{equation}
    R^{l-1}_i(\x|\theta_c) = \sum_{u,v}\sum_{p,q,j} R^{(l-1,l)}_{(u,v,i) \leftarrow (p,q,j)}(\x|\theta_c)
    \label{eq:appendix:lrp:superexpandend:aggregation}
\end{equation}
Optionally, the spatial coordinates $(p,q)$ or $(u,v)$ can trivially be selected over regions $\mathcal{I}$ as in Equation \eqref{eq:appendix:lrp:masked_aggregated_local_conv}, should a more localized analysis of layer- and concept interaction be desired.

Supplementary Figure~\ref{fig:appendix:lrp-concept-composition} visualizes the approach taken in Equations~\eqref{eq:appendix:lrp:superexpandend:decomposition} and~\eqref{eq:appendix:lrp:superexpandend:aggregation}. The property of proportionality ensures that the relative importance of each $R^{l-1}_i(\x|\theta_c)$ for concept $c$ in layer $l$ is represented. The properties of conservation and conditionality ensure that the here measurable interactions between concepts in adjacent layers are specific to the data sample $\x$, the respective model outcome $f(\x)$ as well as the conditions $\theta_c$ set for the computation of relevance. The composition of $c$ in $l$ is given by the most relevant lower layer concepts $b$ in layer $l-1$ and can then (simply) be obtained by sorting, \eg, via
\begin{equation}
    \mathcal{B}
    = {\left\{ {b_1}, \dots, {b_n} \right\}} = \mathrm{argsort}_i^\text{desc} R^{l-1}_{i}(\x|\theta_c)
\end{equation}
for selecting the concepts supporting $c$ the most, in context of the sample $\x$ and the desired explanation specified via $\theta_c$. Sorting ascendingly would return the least supporting and potentially contradicting concepts in context of $\x$ instead. Subsequently, we define ${\mathcal{B}_{k}}$ as the set containing the $k$ most (least) relevant concepts $b_i$. 
Specifically, ${\mathcal{B}_{k}}$ is defined as
\begin{equation}
    {\mathcal{B}_{k}} 
    = {\left\{ {b_1}, \dots, {b_k} \right\}}
     \subseteq \mathcal{B}
\end{equation}
In Supplementary Figure \ref{fig:experiments:localize-concepts:concept-atlas}d of the main manuscript, we use this information to visualize an ontology-like semantic composition of concepts from sub-concepts specific to individual predictions.

\subsubsection{Concept Visualization through Reference Samples}
In the previous sections, we have introduced \gls{crc} assuming that neurons can be identified with specific concepts, which are used by a neural network in the prediction process. In fact, several works in the field of Feature Visualization \cite{erhan2009visualizing, szegedy2013intriguing, zeiler2014visualizing, mahendran2015understanding, mordvintsev2015inceptionism, olah2017feature} have shown different functions of individual neurons using visualizations in the input space.  One common approach in this field is to select reference samples from existing data for feature visualization and analysis. In the literature, the selection of reference samples for a chosen concept $c$ manifested in a neuron (group) is often based on the strength of activation induced by a sample. Here, in principle, samples leading to a strong activation of the corresponding neuron (group) are assumed to likely incorporate the concept $c$ of interest.

However, many facets in the input sample can activate a neuron, from which only few manifestations might be significantly used for a neural network's predictions. Simply because an image leads to high activation does not mean that the image is representative of the neuron's function in a larger inference context. Adversarial examples are a prime example of this. Thus, in order to select the most relevant samples, we propose to use the relevance $R_i(\mathbf{x}|\theta)$ of neuron $i$ for a given prediction, instead of its activation value $\preact_i$ (see Section~\ref{sec:appendix:methodsindetail:understanding:selecting_reference_samples:relevance} for details). By utilizing relevance scores $R_i(\mathbf{x}|\theta)$ we naturally take into account (in contrast to  activation maximization approaches) the class/concept specificity (depending on $\theta$) and thus guide the selection process towards reference samples, which depict facets of a concept that are \emph{actually useful} for the model during inference. In Section~\ref{sec:appendix:experiments:qualitative:variety} we will show how conditioning the relevance computation to different classes provides more clarity about a concept's functioning.
\begin{figure*}[t]
    \centering
    \includegraphics[width=1.0\textwidth]{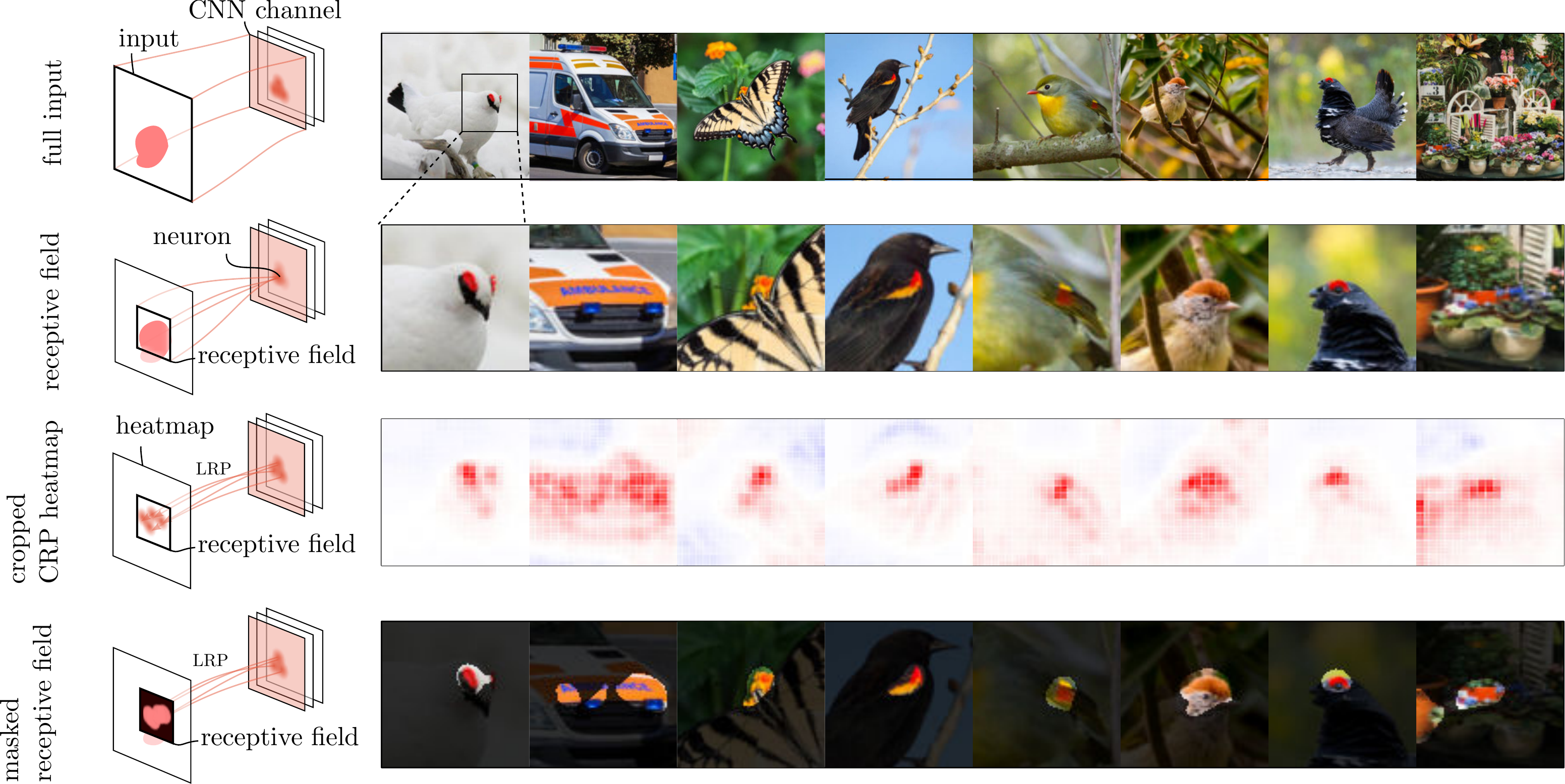}
    \caption{Improving concept visualization and localization using reference samples. \emph{(Top row)}: Full sized reference samples  leading to the largest activation values for a channel of interest (Activation Maximization). \emph{(2nd row)}: The same reference samples cropped to the receptive field of the most activating neuron. \emph{(3rd row)}: \gls{crc} heatmap \wrt\ the filter output of the  cropped reference samples. \emph{(Bottom row)}: Cropped reference samples masked \wrt \gls{crc} heatmap and a threshold of 40 \% of the maximal heatmap value. The concept encoded by a filter can be much smaller than an input image. Cropping the input image to the receptive field of the most activating neuron does increase the focus on the most defining sub-image matching the observed filter's receptive field. A combination with a \gls{crc} backward pass of the reference sample's filter activation can further highlight the core concept. Optionally, distracting image information of which the relevance attribution falls below a given threshold and may be masked out, leaving only the concept-defining parts of the cropped samples for observation. Understanding the function of a filter thus becomes considerably easier in comparison to only using whole reference example images. Another example can be found in Supplementary Figure~\ref{fig:appendix:enhancing-samples-receptive-field-heatmap2}.}
    \label{fig:appendix:enhancing-samples-receptive-field-heatmap}
\end{figure*}
Furthermore, in Section \ref{sec:appendix:methodsindetail:understanding:scaling} we discuss how  concept/channel-wise \gls{crc} heatmaps can aid in the process of concept localization and understanding. Supplementary Figure~\ref{fig:appendix:enhancing-samples-receptive-field-heatmap} demonstrates the advantages of the \gls{crc}-based localization of relevant concepts in reference samples obtained through Activation Maximization (see Section \ref{sec:appendix:methodsindetail:understanding:selecting_reference_samples:activation}). In the set of full-size reference samples shown in the top row, it is not obvious to identify the common concept among the selected reference samples. While using receptive field information helps to crop out the crucial part of the reference images (2nd row), using the \gls{crc} heatmaps (3rd row), additionally highlights the presence of orange/red part of the samples as the definite key component of the concept reference samples have been identified and presented for. In order to further improve concept localization and readability, the reference images can be masked using heatmap information, by removing distracting background objects. To achieve this, a Gaussian filter can be utilized to smooth the heatmap in the first step. In the second step,  a threshold can be chosen below which image parts are masked with black color (or any other suitable background), as shown in the last row of Supplementary Figure~\ref{fig:appendix:enhancing-samples-receptive-field-heatmap}. More information about the concept visualization and localization can be found in Section \ref{sec:appendix:methodsindetail:understanding}.

\begin{figure*}[h]
    \centering
    \includegraphics[width=.9\textwidth]{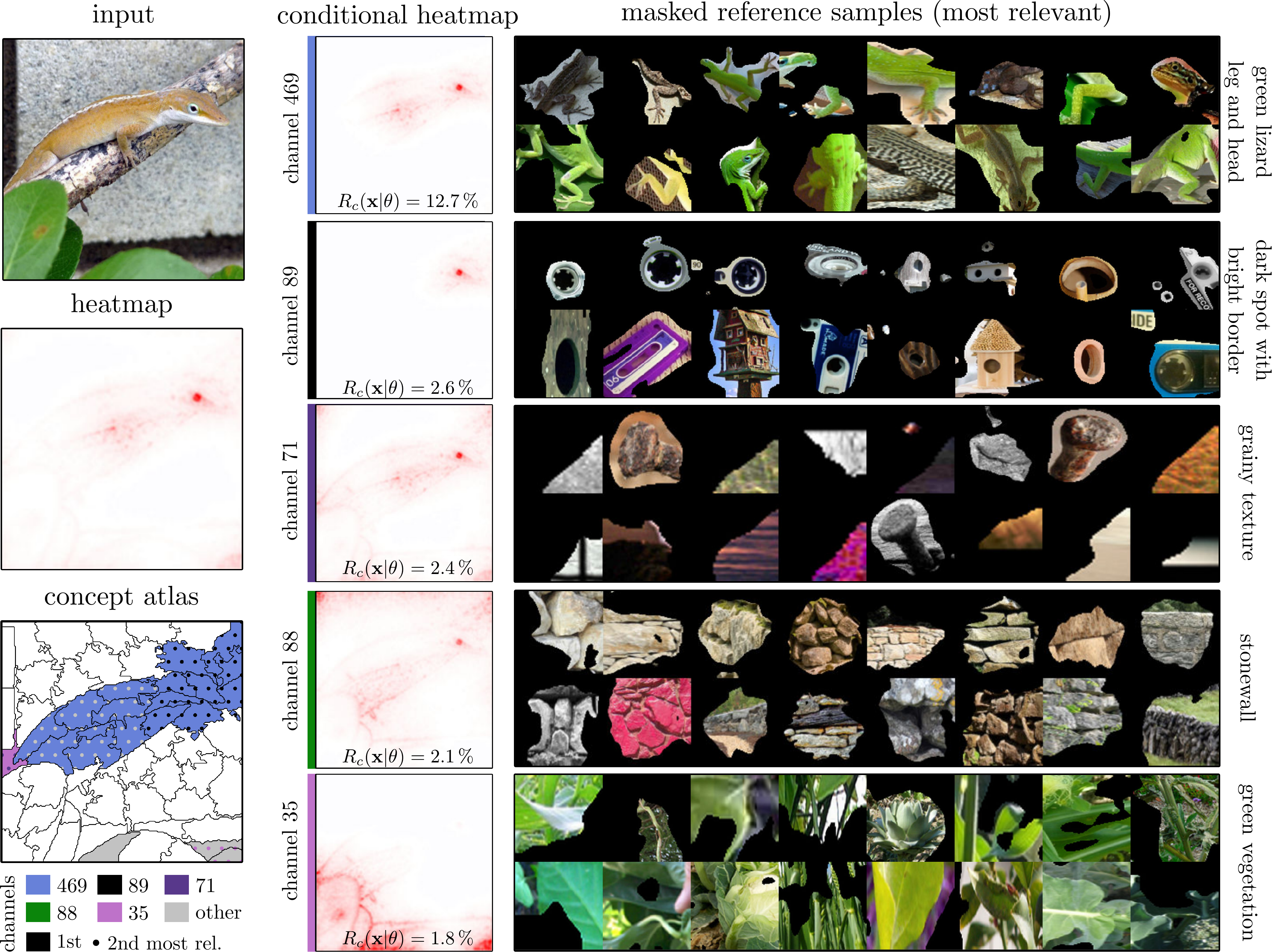}
    \caption[Concept Atlas of a sample of Lizard class.]{Channel-conditional heatmaps enable the localization and understanding of channel concepts. \gls{crc} relevances can further be used to construct a concept atlas, visualizing which concepts dominate in specific regions in the input image defined by super-pixels. For the prediction of a \edited{}{``Green Anole''} using a VGG-16 model with BatchNorm layers, the most relevant channels of \texttt{features.40} can be identified with concepts concerning lizard \edited{}{legs and heads (channel 469), ``dark spot with bright border'' (89) activating on the eye, ``stonewall'' (88), ``green vegetation'' (35) and ``grainy texture'' present on the stonewall and lizard scales (71).}}
    \label{fig:appendix:localize-concepts:concept-atlas}
\end{figure*}

\subsubsection{Concept Atlases}
\label{sec:appendix:methodsindetail:lrp:conceptual:atlas}
Per-channel heatmaps computed through \gls{crc} describe in input space (1) which input features were relevant for the channel and its concept during prediction in context of an explained model output, and (2) where those concepts are located in pixel space. We can now combine multiple concept-specific attributions in order to visualize them together in the form of a ``concept atlas'', motivated by the idea of an Activation Atlas from~\cite{carter2019activation}. As is shown in Supplementary Figure~\ref{fig:appendix:localize-concepts:concept-atlas} (and Supplementary Figure~\ref{fig:experiments:localize-concepts:concept-atlas}b-c in the main manuscript), the concept atlas then visualizes in input space, which concepts were most relevant (and here, second most relevant) in specific input image regions. Choosing superpixels as regions of interest, we can aggregate the channel-conditional relevances per superpixel into regional relevance scores. Specifically, given superpixel $k$ described by the index set $\mathcal{I}_k$ selecting some input features, we aggregate the conditional relevance scores $R_{i}^1(\x|\theta_{c})$ for channel $c$ and obtain $R_{\mathcal{I}_k}^1(\x|\theta_{c})$, which assesses the importance of concept $c$ in superpixel $\mathcal{I}_k$. All the resulting scores $R_{\mathcal{I}_k}^1(\x|\theta_{c})$ applying to region $k$ can then be sorted in descending fashion for each superpixel in order to get a sorted set $\mathcal{B}_k$
\begin{equation}
\label{eq:appendix:localize-concepts:most_rel_channel_set}
    \mathcal{B}_k = \mathrm{argsort}_c^\text{desc} R^{1}_{\mathcal{I}_k}(\x|\theta_c)~,
\end{equation}
determining a regional ranking of concepts and respective importance, which can be used to construct the concept atlas. Regarding concept atlas construction,
each channel can, \eg, be assigned a specific color for visualization purposes. Each super-pixel $k$ is filled with the color of the most relevant concept given by $\mathcal{B}_k$. The second(/third/fourth/etc.) most relevant concept defines the color of the texture/overlay (dots in our case).

In Supplementary Figure \ref{fig:appendix:localize-concepts:concept-atlas}, channel heatmaps help to localize relevant regions in input space, and at the same time reveal \emph{what} the model has picked up in those regions. Here, a \edited{}{``Green Anole''} is predicted by a VGG-16 BN model trained on ImageNet. Regarding the channels located in the layer \texttt{features.40}, our analysis shows, that concepts exist which refer to \edited{}{lizard legs and heads (channel 469), ``dark spot with bright border'' (89) activating on the eye, ``stonewall'' (88), ``green vegetation'' (35) and ``grainy texture'' present on the stonewall and lizard scales (71).}

\cleardoublepage
\clearpage

\section{Understanding Concepts in Latent Representations}
\label{sec:appendix:methodsindetail:understanding}
The emergence of human-understandable concepts in neurons of a \gls{dnn} has been observed in several works \cite{zhou2015object, olah2017feature, radford2017learning, bau2020understanding, cammarata2020thread, goh2021multimodal}, although the networks were not explicitly trained to use them. While lower-layer neurons encode features like textures and edges, higher-layer neurons have been observed to conceptualize notions from simple objects to abstract emotions \cite{zhou2015object, olah2017feature}. Even the existence of multimodal neurons was reported, that respond to a concept regardless of whether it is shown as a photograph, cartoon or written in textual form \cite{goh2021multimodal}. While the gradual formation of abstract concepts is widely known, it is surprising to see that a non-negligible amount of neurons encode features, that resemble the way humans categorize the world. Our work joins a whole series of papers showing that single neurons not only encode invariant concepts useful for solving the task, but that these can also be understood by humans and thus enable interpretability of a DNN's reasoning.  Nevertheless, assigning meaning to individual units is often subjective and depends on the human observer. Further, generative representations might not be necessarily human-interpretable. Thus, misinterpretation or misunderstanding of the inner functioning of a chosen model is possible. In the following, we therefore discuss current feature visualization techniques and present novel methods to improve concept identification and understanding. 

\subsection{Generative Approaches to Understanding Latent Features}
\label{sec:appendix:methodsindetail:understanding:generative}
A large part of feature visualization techniques rely on \gls{amax}, where in its simplest form, input images are sought that give rise to the highest activation value of a specific unit. In general, \gls{amax} can be seen as an optimization problem, \ie,  maximizing a neuron's activation, which can be solved using \gls{ga}. Let $\preact^{l}_{i}(\weight, \x)$ be the activation of neuron $i$ in layer $l$. Here, $\weight$ denotes the parameters of a neural network and $\x$ denotes the input being an element of the input space $\mathcal{X}$. Then formally, we seek to find the input $\x^{\star}$ leading to a maximal activation:
\begin{equation}
\label{eq:appendix:activation_max}
\x^{\star} = \underset{\x \in \mathcal{X}}{\argmax}\,\preact^{l}_{i}(\weight, \x).
\end{equation}
While conceptually simple,  preventing the emergence of adversarial examples became a main research endeavor. Several priors were proposed to guide optimization into realistic looking images, \eg, Transformation Robustness \cite{mordvintsev2015inceptionism}, Frequency Penalization \cite{mahendran2015understanding}, Preconditioning \cite{olah2017feature}, and learning a natural prior from data distribution \cite{nguyen2016synthesizing}. 
Despite progress made, the generation of high-resolution and natural images from deep networks remains challenging. User studies \cite{borowski2020exemplary} also found, that synthetic images tend to be abstract and more difficult for people to interpret than real example images. Therefore, without limitation of generalization, we will focus on data-based approaches in the following Sections.

\subsection{Activation-based Reference Sample Selection}
\label{sec:appendix:methodsindetail:understanding:selecting_reference_samples:activation}
Alternatively to generative sample synthesis, other works \cite{yeh2020completeness,chen2020concept} propose to select reference samples from existing data for feature visualization and analysis.  In the literature, the selection of reference samples for a chosen concept $c$ manifested in a neuron (group) is often based on the strength of activation induced by a sample. Here, in principle, samples leading to a strong activation of the corresponding neuron (group) are assumed to likely incorporate the concept $c$ of interest. For data-based reference sample selection, the input space is restricted to elements of a particular dataset $\mathcal{X}_d\subset\mathcal{X}$. Thus, we seek to find data samples $\x$ of a finite set $\mathcal{X}_d$ that result in the highest activation values for a specific unit. Without limitation of generality, the dataset is in the following assumed to consist of images. However, the same principles apply to time series data, tabular data or any other arbitrary data format for feed-forward neural networks\footnote{The same analysis for Recurrent Neural Networks is more challenging as time dependency must be taken into account.}.

As discussed in Section~\ref{sec:appendix:methodsindetail:lrp:conceptual}, all neurons of a convolutional layer's channel of a \gls{dnn} can be identified with the same concept due to the learned filters' (assumed) spatial invariance. Thus, in literature \cite{chen2020concept}, entire filters instead of single neurons are investigated for convolutional layers. Here, in principle, different maximization targets $\mathcal{T}(\x)$ can be specified, that define a ``maximum'' channel activation. One particular choice of $\mathcal{T}(\x)$ is to identify samples $\mathbf{x}^{\star} \in \mathcal{X}_d$, which maximize the sum over all channel activations, \ie, 
\begin{equation}
{\mathcal{T}}^{\text{act}}_{\text{sum}}(\x) = \sum_{i} \preact_{i}(\x)
\label{eq:appendix:sum-max-short}
\end{equation}
resulting in samples ${\x^{\star}}^{\text{act}}_{\text{sum}}$ which are likely to show a channel's concept in multiple input features, as the entire channel is maximized. However, while targeting all channel neurons, reference samples including both concept-supporting and contradicting features might result in a low score ${\mathcal{T}}^{\text{act}}_{\text{sum}}$,  as negative activations are taken into account by the sum.  Alternatively, a non-linearity can be applied additionally,  e.g., ReLU, focusing only on positive activations.

A different choice is to define maximally activating samples by looking at the maximum channel activation, \ie,
 \begin{equation}
{\mathcal{T}}^{\text{act}}_{\text{max}}(\x) = \max_{i} \preact_{i}(\x)
\label{eq:appendix:max-short}
 \end{equation}
leading to samples ${\x^{\star}}^{\text{act}}_{\text{max}}$ with a more localized and strongly activating set of input features characterizing a channel's concept. These samples ${\x^{\star}}^{\text{act}}_{\text{max}}$ are more difficult to interpret, as only a small region of a sample might contain the concept. Thus, concepts might be more difficult to grasp for the human observer. Chen et al. \cite{chen2020concept} have observed, that ${\mathcal{T}}^{\text{act}}_{\text{sum}}$ is better suited to find samples for lower-level concepts, and ${\mathcal{T}}^{\text{act}}_{\text{max}}$ for higher-level concepts. Thus, they propose to use a max-pooling operation followed by summation for the maximization target representing a mixture out of both ${\mathcal{T}}^{\text{act}}_{\text{sum}}$ and ${\mathcal{T}}^{\text{act}}_{\text{max}}$. However, we argue that ${\mathcal{T}}^{\text{act}}_{\text{max}}$ could also be utilized by handling the problem of localization with subsample selection and conceptual heatmaps presented in the following sections.

In order to collect multiple reference images describing a concept, the dataset $\mathcal{X}_d$ consisting of $n$ samples is first sorted in descending order according to the maximization target $\mathcal{T}(\mathbf{x})$, \ie
\begin{equation}
    \mathcal{X}^{\star}
    = {\left\{ {\x^{\star}_1}, \dots, {\x^{\star}_n} \right\}}
    = \underset{\x \in \mathcal{X}_d}{\text{argsort}^\text{desc}} \mathcal{T}(\x)
    .
\end{equation} 
Subsequently, we define ${\mathcal{X}^{\star}_{k}}$ as the set containing the $k$ most maximizing samples $\x^{\star}$. Specifically, ${\mathcal{X}^{\star}_{k}}$ is defined as
\begin{equation}
    {\mathcal{X}^{\star}_{k}} 
    = {\left\{ {\x^{\star}_1}, \dots, {\x^{\star}_k} \right\}}
     \subseteq \mathcal{X}^{\star}
\end{equation}
In the following, we denote the set of reference images obtained by taking the summation of activation as ${\mathcal{X}^{\star}_{k}}^{\text{act}}_{\text{sum}}$ and the set obtained by taking the maximal value of activation as ${\mathcal{X}^{\star}_{k}}^{\text{act}}_{\text{max}}$. From Equations~\eqref{eq:appendix:sum-max-short} and~\eqref{eq:appendix:max-short} it then generally follows, that
\begin{equation}
{\mathcal{X}^{\star}_{k}}^{\text{act}}_{\text{sum}} \neq {\mathcal{X}^{\star}_{k}}^{\text{act}}_{\text{max}}~.
\end{equation}
Due to the different maximization targets $\mathcal{T}^{\text{act}}_{\ast}(\mathbf{x})$ an experimental comparison between both variants and our relevance-based reference sample selection approach (presented in the next section) motivated by \cite{chen2020concept}, can be found in Section~\ref{sec:appendix:experiments:selecting_reference_samples}.

\subsection{Relevance-based Reference Sample Selection}
\label{sec:appendix:methodsindetail:understanding:selecting_reference_samples:relevance}
The hierarchical structure in DNNs encourages concepts in higher layers to be invariant to transformations in the input domain. This leads to the emergence of multifaceted features that respond to different manifestations\footnote{One definition of a concept is the abstraction obtained by generalization from particulars~\cite{spritzer1975concept}. For example, the concept \emph{roundness} can be explained by reference images of a clock or a ball. It is then the task of the human observer to abstract the universal concept from several reference images. In the following, we will use ``manifestation'' or ``facet'' also as synonyms for particular.} of a concept: For instance, a neuron that encodes for \textit{apples} might activate for different colors (green, yellow, red) or other shapes (cut in half, with or without peel). See Supplementary Figure~\ref{fig:appendix:experiments:visual:rel:vs:act:refsamples:scatter} for a visual example. Likewise, the emergence of polysemantic neurons was reported, that fire for different concepts \cite{olah2017feature, bau2020understanding, cammarata2020thread, goh2021multimodal}. 

Multifaceted neurons \cite{nguyen2016multifaceted} are particularly challenging to illustrate, as (potentially conflicting) different facets cannot be easily visualized in a single image. Gradient based methods either illustrate all facets at once, resulting in unintelligible images,  or reveal only one facet, ignoring other important ones \cite{nguyen2016multifaceted, goh2021multimodal}. When using real images instead, the sheer volume of images makes it difficult to filter out facets. Thus, typically only the $k$ most important samples are displayed. To enforce more variety in reference samples for \gls{amax}, the authors of \cite{nguyen2016multifaceted} introduce the \textit{Multifaceted Feature Visualization} algorithm, which projects the latent representation into a 2D embedding using t-SNE and applies $k$-means to extract clusters of reference images. Other works initialize gradient ascent with random seeds \cite{erhan2009visualizing}, or regularize the optimization process with cosine similarity to enforce more variety \cite{olah2017feature}. Although these techniques enhance sample variety and increase understanding, it is still an open question, which manifestations are preferred by the network in practice. In theory, many facets can activate a neuron, however, only few manifestations might be significantly used for a neural network's predictions. Simply because an image leads to high activation does not mean that the image is representative of the neuron's function in a larger inference context. Adversarial examples are a prime example of this.

In this sense, we take up the authors' remark in \cite{olah2017feature} \textit{``The truth is that we have almost no clue how to select meaningful directions [of activations], or whether there even exist particularly meaningful directions.''} and introduce the method of \glsfirst{rmax} based on our \gls {crc} approach as a complement to \glsdesc{amax}. Regarding \gls{rmax}, we do not search for images that produce a maximal activation response. Instead, we ask: \emph{``How, and for which samples, does the model use the neuron in practice?''}

Thus, in order to select the most relevant samples, we define maximization targets ${\mathcal{T}}^{\text{rel}}_{\ast}(\x)$ by using the relevance $R_i(\mathbf{x}|\theta)$ of neuron $i$ for a given prediction, 
instead of its activation value $\preact_i$. 
Specifically, the maximization targets are given as

\begin{equation}
{\mathcal{T}}^{\text{rel}}_{\text{sum}}(\x) = \sum_{i} R_{i}(\x|\theta)
\quad \text{and} \quad
{\mathcal{T}}^{\text{rel}}_{\text{max}}(\x) = \max_{i} R_{i}(\x|\theta).
\label{eq:appendix:max-rel-target}
\end{equation}

\begin{figure*}[t]
\centering
\includegraphics[width=0.85\textwidth]{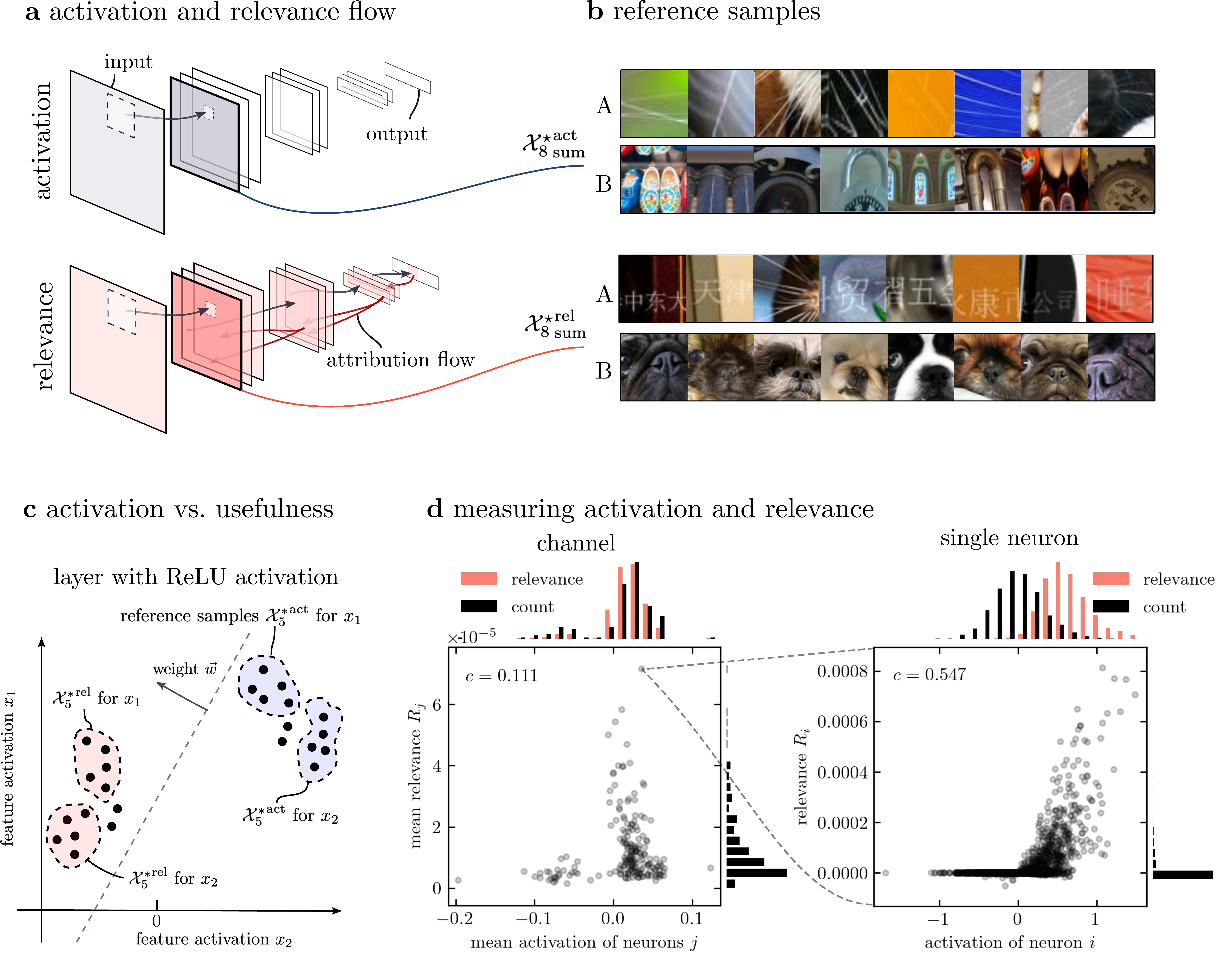}
\caption{
Activation- and relevance-based sample selection. \textbf{a)} Activation scores only measure the stimulation of a latent filter without considering its role and impact during inference. Relevance scores are contextual to distinct model outputs and describe how features are utilized in a \gls{dnn}'s prediction of, \eg, a specific class. \textbf{b)} As a result, samples selected based on \glsdesc{amax} only represent maximized latent neuron activation, while samples based on \glsdesc{rmax} represent features which are actually useful and representative for solving a prediction task. \textbf{c)} Assume we wish to find representative examples for features $x_1$ and $x_2$. Even though a sample leads to a high activation score in a given layer and neuron (group) --- here $x_1$ and $x_2$ --- it does not necessarily result in high relevance or contribution to inference: The feature transformation $\Vec w$ of a linear layer with inputs $x_1$ and $x_2$, which is followed by a ReLU non-linearity, is shown. Here, samples from the blue cluster of feature activations lead to high activation values for both features $x_1$ and $x_2$, and would be selected by \gls{amax}, but receive zero relevance, as they lead to an inactive neuron output after the ReLU, and are thus of no value to following layers. That is, even though the given samples activate features $x_1$ and $x_2$ maximally strong, they can not contribute meaningfully to the prediction process through the context determined by $\Vec w$, and samples selected as representative via activation might not be representative to the overall decision process of the model. Representative examples selected based on relevance, however, are guaranteed to play an important role in the model's decision process. \textbf{d)}: Correlation analyses are shown for an intermediate ResNet layer's channel and neuron. Neurons that are on average highly activated are not, in general, also highly relevant, as a correlation coefficient of $c=0.111$ shows, since a specific combination of activation magnitudes is important for neurons to be representative in a larger model context. 
}
\label{fig:appendix:rel-vs-act-general-examples}

\end{figure*}

It is to note, that the activation-based maximization targets are only specific to the sample (and filter). Class-specificity may only be obtained in data collection-based approaches, given sufficient label information, by pre-selecting samples from classes of interest~\cite{hohman2019summit}.  Utilizing relevance scores $R_i(\mathbf{x}|\theta)$ instead, the maximization target is class- (true, predicted or arbitrarily chosen, depending on $\theta$), model-, and potentially concept-specific (depending on $\theta$),  as illustrated in Supplementary Figure~\ref{fig:appendix:rel-vs-act-general-examples}. The resulting set of reference samples thus includes only samples which depict facets of a concept that are \emph{actually useful} for the model during inference. In Section~\ref{sec:appendix:experiments:qualitative:variety} we will show how conditioning the relevance computation to different classes provides more clarity about a concept's functioning.

How different resulting reference sets ${\mathcal{X}_k^{\star}}^{\text{act}}$ and ${\mathcal{X}_k^{\star}}^{\text{rel}}$ can occur, is depicted in Supplementary Figure~\ref{fig:appendix:rel-vs-act-general-examples} \emph{(bottom left)}. Here, a hypothetical transformation within a linear layer $l$ followed by a ReLU non-linearity in a two-dimensional activation space is shown. Two clusters of points are projected to different sides of the hyperplane described by the weighted normal vector $\Vec w = (w_1, w_2)$ with $w_1 > 0$, $w_2 < 0$ and bias term $b$. The output $y$ is thus given by $y=\text{ReLU}(w_1 x_1 + w_2 x_2 + b)$, leading to relevances for features $x_i$ of

  \begin{equation}
      R_i^l = 
      \begin{cases}
        \frac{w_i x_i}{w_1 x_1 + w_2 x_2 +b} R^{l+1}\,&\text{if $y > 0$} \\
        0 \,&\text{else.}
      \end{cases}
      \label{eq:appendix:lrp-general-hyperplane-example}
  \end{equation}

Here, inactive neurons (with $y=0$) cannot be attributed with relevance scores $R \neq 0$. Whereas samples from the blue clusters lead to a high activation for both features $x_1$ and $x_2$, they are not relevant (or used) by the model in the following computation, as they lead to an output of zero.  Note, that due to the ReLU non-linearity, negative outputs are set to zero. Samples from the red cluster result in smaller activations, but are actually depicting concepts, which can be used in the remaining prediction steps as they cause the output neuron represented by $\Vec w$ to activate. Experiments showing the differences in reference sample sets are located in Section~\ref{sec:appendix:experiments:selecting_reference_samples}. Relevance attributions depend on the activations $x_i$, as Equation~\eqref{eq:appendix:lrp-general-hyperplane-example} illustrates. However, attributions are not strictly correlated to activations,  because they also depend on the downstream relevances propagated from higher layers affected by feature interactions at the current and following layers. In Supplementary Figure~\ref{fig:appendix:rel-vs-act-general-examples} \emph{(bottom center and right)}, correlation analyses are shown for an intermediate ResNet layer. In the left of both panels, mean activations and relevances are shown for all neurons of a CNN channel. Here it is shown, that neurons that are on average highly activated are not, in general, also highly relevant, as a correlation coefficient of $c=0.111$ demonstrates. The right plot depicts activation and relevance values of a strongly activating neuron. As has been discussed previously, samples leading to a high feature activation do not necessarily also result in high feature relevance. Here, a correlation coefficient of $c=0.547$ supports however, that relevance is dependent on activation.

To summarize, reference samples from \gls{amax}-sorting illustrate feature interactions that lead to a high excitation value, and thereby important (to higher layers) but yet weaker activating concepts may not be taken into account. \glsdesc{rmax} takes all higher layer activation distributions into account by relying on \glsdesc{crc} and is consequently a middle way between assigning semantics only to orthogonal activation basis vectors (\ie, single neurons) or a combination of them, by focussing on activations which see further use in a larger inference context. 

\subsection{Zooming into the Reference Sample by Using Receptive Field Information}
\label{sec:appendix:methodsindetail:understanding:scaling}
The reference sample selection approach described in the last section visualizes whole data samples based on the activations or the relevances of (parts of) latent representations. However, depending on the maximization target concepts of interest might take up only a small part of the input sample. Further, selected samples might include multiple concepts, hampering the process of identifying the right concept. 

One step towards better concept localization in input samples is to visualize the input wrt.\ the receptive field of a maximally activated or relevant neuron, as is also discussed in \cite{yeh2020completeness}. Concept localization and identification is made easier due to the masking of all other (possibly distracting) concepts outside the receptive field. Moreover, the receptive field represents an upper bound to the true scale of the concept of interest. Specifically, the receptive field of a neuron is the portion of the input space that can contribute to the neuron's activation, which is determined by the neuron's connectivity to lower layers. Since each neuron in a channel can be assumed to encode features representing the same concept, the receptive field of a singular neuron can be used to identify the input area responsible for the excitation and thus the scale limit of the by the filter/neuron learned concept. As receptive field computations for \glspl{dnn} are not straight forward, libraries specialized on this task, such as \cite{araujo2019computing}, have been developed.

In the following, we present an alternative for receptive field computation using the $\flat$-rule \cite{lapuschkin2017understanding, kohlbrenner2020towards} of \gls{lrp}. By choosing one specific neuron activation as the starting point for relevance decomposition in the backward pass, we can apply the LRP $\flat$-rule to calculate the receptive field of the particular neuron in intermediate layers until the input layer is reached (see Supplementary Figure~\ref{fig:appendix:channel-upsampling-relevance-receptive-field}). 
\begin{figure*}[t]
    \centering
    \includegraphics[width=0.9\textwidth]{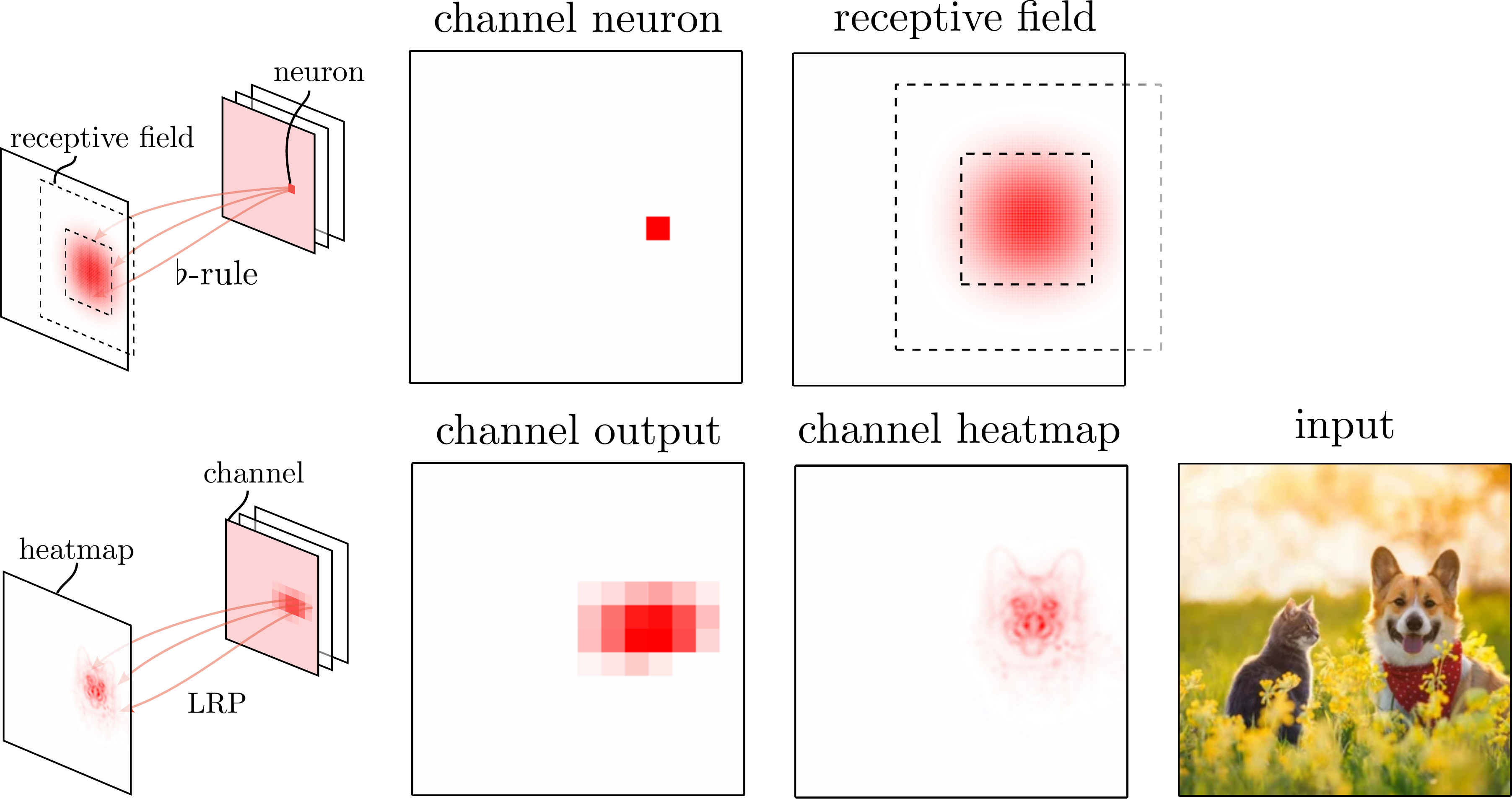}
    \caption{\emph{(Top)} \gls{crc} readily offers receptive field calculation using the \gls{lrp} $\flat$-rule starting at the neuron of interest. \emph{(Bottom)} \gls{crc} can further be used to improve resolution of activation maps for (convolutional) filters by beginning relevance propagation at the actual channel activations. %
    }
    \label{fig:appendix:channel-upsampling-relevance-receptive-field}
\end{figure*}
Using the $\flat$-rule, relevance of upper-level neuron $j$ is equally distributed to all contributing lower-level neurons, \ie, for all connected neuron pairs, we set $\forall i,j~\preact_{ij} = 1\, $ in Equation~\eqref{eq:appendix:lrp_basic}, resulting in $\preact_j = \sum_i \preact_{ij} = \sum_i 1$, thus disregarding any influence of learned weights or input features. Layer after layer, any neuron that contributed computationally by being (transitively) connected to the neuron starting the backpropagation, receives attribution scores $>0$. The more often an input neuron contributes, the higher the attribution score it receives.

In general, at a layer $l$, the set $\mathcal{I}_{\text{rec}}$ of neurons (or pixels in $l = 1$) inside the receptive field of neuron $j$ located in layer $l' > l$ is given as
\begin{multline}
    \mathcal{I}_{{\text{rec}(}j\in l')}^l = \left\{ i{\in l}~|~R_i^l(\x|\theta_{j\in l'}) \neq 0 \right\},\\
    \text{ with }\theta_{j\in l'}=\{l':\{j\}\}. 
    \label{eq:appendix:receptive-field-with-lrp}
\end{multline}
Here, the condition $\theta_{j\in l'}=\{l':\{j\}\}$ ensures the LRP initialization at layer $l'$ with neuron $j$, \ie, $R_k^{l'}(\x) = \delta_{kj}$. It is to note, that this method can also be applied to non-image data. Having calculated the set $\mathcal{I}_{\text{rec}(j\in l')}^{1}$ of input features inside the receptive field of neuron $j$, the input sample can be cropped to incorporate solely the pixels inside $\mathcal{I}_{\text{rec}(j\in l')}^1$, \eg, by fitting a bounding box around the ``blob'' of attribution scores exceeding a given threshold. As is illustrated in Supplementary Figure~\ref{fig:appendix:channel-upsampling-relevance-receptive-field} (top), the size of the receptive field might differ for neurons at different spatial positions inside a convolutional layer's channel. In this particular case, a channel neuron's receptive field close to the (input-dependent) spatial channel edge results in a rectangular receptive field shape in pixel space, which becomes more squared, if positioned closer to the channel's spatial center. Application of convolutions with input padding (\eg, with zeros) during the forward pass are leading to the effect in this case, due to filters placed partially ``outside'' the input-connected spatial extent of the channel. Therefore, neurons close to the spatial edge of a channel filter can result in a receptive field which extends into the padded areas. The resulting differences in receptive field sizes (\wrt the possible maximum) can be handled by, \eg, padding with a contrastive or transparent color, if a uniform size is favored.

In summary, restricting the analysis to a filter's or neuron's receptive field can help with the identification and localization of the core manifestation of the encoded concept. For the purpose of receptive field computation, we use efficient tools already available in the \gls{lrp} toolset. The effects of our technique are visualized in Supplementary Figures~\ref{fig:appendix:enhancing-samples-receptive-field-heatmap} and~\ref{fig:appendix:enhancing-samples-receptive-field-heatmap2} and further extended in the following Section~\ref{sec:appendix:methodsindetail:understanding:visualization}.

\subsection{Finding the Relevant Concepts within the Receptive Field by using CRP}
\label{sec:appendix:methodsindetail:understanding:visualization}
Receptive fields of neurons in later stages of a network can cover larger input regions, or even the whole input, as is the case for dense layer neurons. Thus, encoded concepts can be much smaller than a receptive field, creating the need for further localization (i.e., within the receptive field). For convolutional layers, an intuitive way to localize important regions is up-sampling of channel activations,  as is done with \textit{Network Dissection} \cite{bau2020understanding}. This is sensible, as convolutions are always applied locally in the spatial dimension and thus, layer after layer, activations tend to stay in the same spatial position. However, this approach lacks precision, as up-sampling introduces blurriness \cite{zhou2016learning}, and is further limited to convolutional layers.

Application of \gls{lrp} and in extension \gls{crc} allows to go beyond channel up-sampling by \emph{precisely} resolving important input features and further being applicable to dense layer neurons as well. Specifically, \gls{crc} can be utilized to calculate heatmaps for any concept of interest and thus aid in concept localization. Therefore, two cases can be distinguished: (1) If one is interested in which input feature of a given data sample a specific filter is activated regardless of any class, then \gls{crc} is initialized starting at the filter output of choice, backpropagating with the actual activations, \ie, the score output by the observed latent concept detector. This yields an indication of \emph{where} the concept is localized in the input space. (2) Alternatively, relevance can be conditioned to a specific class output. Then, the heatmap depicts, where the concept is \emph{used} in the prediction \emph{of a chosen class}. The precise localization of relevant input features for the activation of \gls{dnn} channels is shown in Supplementary Figure~\ref{fig:appendix:channel-upsampling-relevance-receptive-field} \emph{(bottom)}.

\begin{figure*}[t]
\label{fig:heatmap_on_examples}
  \centering
    \includegraphics[width=1\textwidth]{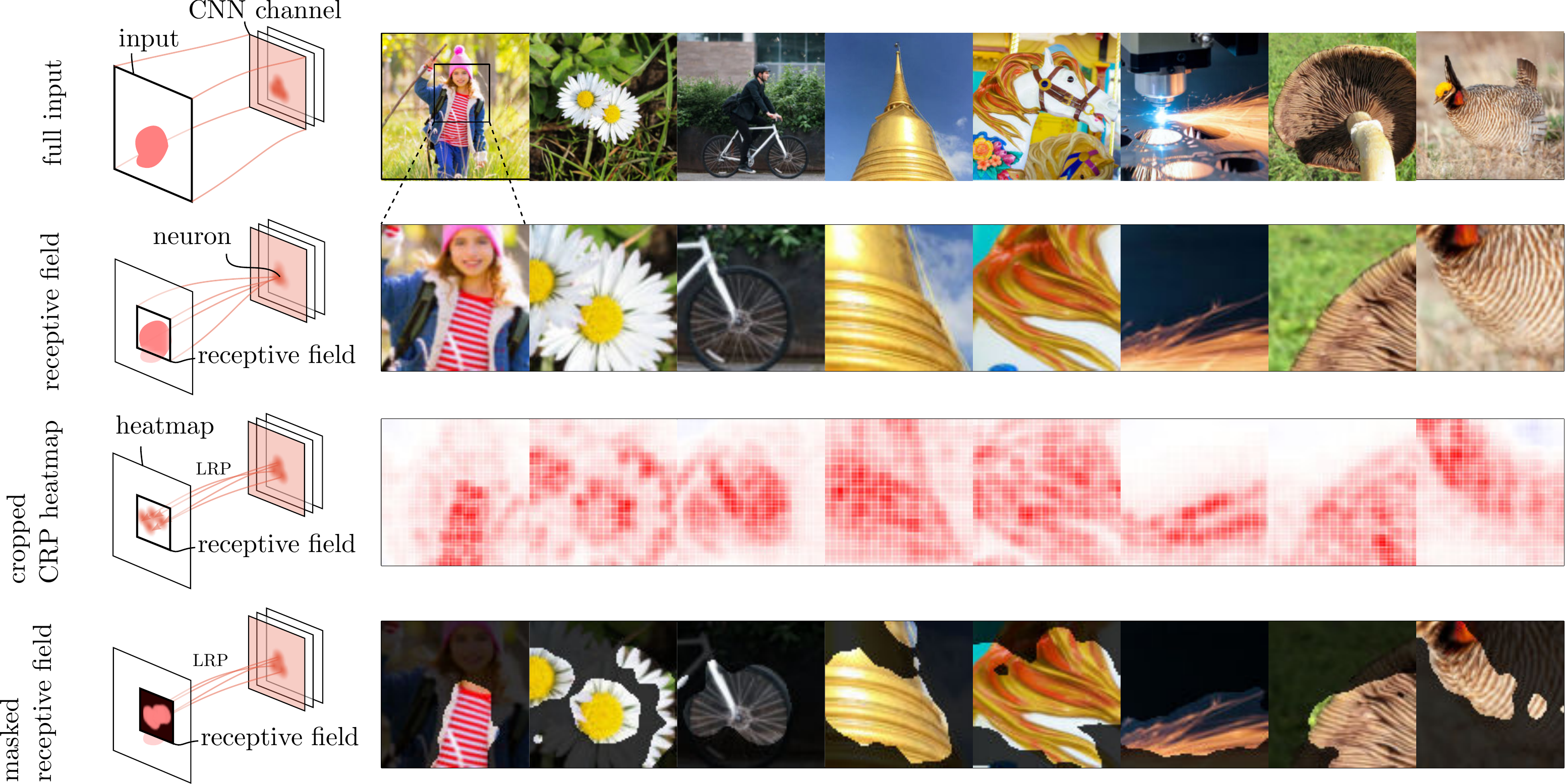}
    \caption{Improving concept visualization and localization using reference samples. \emph{(Top row)}: Full sized reference samples in ${\mathcal{X}^{\star}_{8}}^{\text{act}}_{\text{max}}$ sorted accordingly to their activation. \emph{(2nd row)}: The same reference samples cropped to the receptive field of the most activating neuron. \emph{(3rd row)}: \gls{crc} heatmap for filter output of cropped reference samples. \emph{(Bottom row)}: Cropped reference samples masked \wrt \gls{crc} heatmap and a threshold $\tau$ of 40 \% of the maximal heatmap value. This figure shows additional results corresponding to Supplementary Figure \ref{fig:appendix:enhancing-samples-receptive-field-heatmap}.
    }  
    \label{fig:appendix:enhancing-samples-receptive-field-heatmap2}
\end{figure*}

Here, the activations of channel $c$ in layer $l$ during the prediction of the input $\mathbf{x}$ are collected. Thereafter, the heatmap computation is initialized with the activations ${\preact}^l_{(p, q, j)}$, \ie, $R^l_{(p, q, j)}(\mathbf{x}|\theta) = {\preact}^l_{(p, q, j)} \delta_{jc}$ for all channels $j$ and spatial coordinates $p$ and $q$.  Analogously, initialization for dense neurons is given as $R^l_j(\mathbf{x}|\theta) = {\preact}^l_j \delta_{jc}$ for all neurons $j$. Using \gls{crc}, we can thus visualize, where a concept activates in high-precision independent of the layer architecture. This way, we are not limited to convolutional layers, but applications to fully connected layers (and others) commonly built into \gls{mlp} and \gls{rnn} models \cite{Arras2019EvaluatingRN} are conceivable as well. Further, by defining additional conditions, information can be extracted at a higher degree of detail. For example, beginning relevance decomposition at a specific class output $y$, we can localize where a concept of channel $c$ (in layer $l$) is used for the prediction of class $y$. This follows the approach presented in Section~\ref{sec:appendix:methodsindetail:lrp:disentangled} with relevance propagation using the condition $\theta_{cy} = \left\{l:\{\text{neurons $j$ of channel $c$}\}, L: \{y\}\right\}$.

In Supplementary Figures~\ref{fig:appendix:enhancing-samples-receptive-field-heatmap} and~\ref{fig:appendix:enhancing-samples-receptive-field-heatmap2}, we see, how concept/channel-wise \gls{crc} heatmaps can aid in the process of concept localization and understanding. In the set of full-size reference samples shown in the top row, it is not obvious to identify the common concept. While using receptive field information to crop out the crucial part of the reference images \emph{(2nd row)}, using the \gls{crc} heatmaps \emph{(3rd row)}, additionally highlights the presence of orange/red parts of the samples as the definite key component of the concept reference samples have been identified and presented for. In order to further improve concept localization and readability, the reference images can be masked using heatmap information, by removing distracting background objects. To achieve this, a Gaussian filter can be utilized to smooth the heatmap in the first step. In the second step,  a threshold can be chosen below which image parts are masked with black color (or any other suitable background of choice), as shown in the last row of Supplementary Figures~\ref{fig:appendix:enhancing-samples-receptive-field-heatmap} and~\ref{fig:appendix:enhancing-samples-receptive-field-heatmap2}.

Naturally, in the given full-sized example images alone in Supplementary Figure~\ref{fig:appendix:enhancing-samples-receptive-field-heatmap2}, a human observer may involuntarily focus on the striking red dot on the bird's \edited{}{head (last} reference example in top row of Supplementary Figure~\ref{fig:appendix:enhancing-samples-receptive-field-heatmap2}) or the \edited{}{striking blue spark on the drilling machine (6th} reference example) and be distracted from the actual concept summarized by the presented examples\footnote{the authors certainly did and were}. A cropping of the most activating part of the sample, followed by relevance map computation however reveals that in fact the model uses the analyzed filter to focus on the \edited{}{stripes on the bird's breast and the hair-like sparks on the right of the drilling machine.} A similar observation can be made for the field of flowers shown in the rightmost reference example \edited{}{of Supplementary Figure~\ref{fig:appendix:enhancing-samples-receptive-field-heatmap}. The sample represents the red color}, which is not apparent from looking at the full sized input image in the top row. By first cropping the local image patch yielding the highest latent channel activation, and then explaining the observed filter's reaction to the cropped reference sample via \gls{crc} it becomes clear that the model is stimulated by \edited{}{the red/orange pots} near the bottom left corner of the image and ignores all others. These examples show that human perception can be misguided when only observing whole images for the purpose of understanding latent representations of \gls{ml} predictors.

\cleardoublepage
\section{Detailed Analysis: Explaining Concepts with Examples}
\label{sec:appendix:conceptunderstanding}
The visual and qualitative support for understanding predictions provided by CRP-based explanations are shown in this section. In particular, we first contrast our relevance-based reference example selections to activation-based reference example selection on popular real world image datasets in Section~\ref{sec:appendix:experiments:selecting_reference_samples}. We then follow up by establishing how concept understanding can be improved by focusing onto the common denominator of presented reference samples via  class-conditionally selecting relevance examples and extending  reference dataset variety in Section~\ref{sec:appendix:experiments:qualitative:variety}. The insights discussed in this section together with the tools established in Section \ref{sec:appendix:methodsindetail:lrp:conceptual} are leveraged in Section~\ref{sec:appendix:conceptlocalization} for building localizing concept maps and identifying conceptual compositions along the sequence of neural network layers. Later, Section \ref{sec:appendix:timeseries} then demonstrates that our concept-based approach to explaining neural networks predictions can be applied to arbitrary data, not only images, at hand of an experiment on time series data.
    \begin{figure*}[t]
      \centering
       \includegraphics[width=1\linewidth]{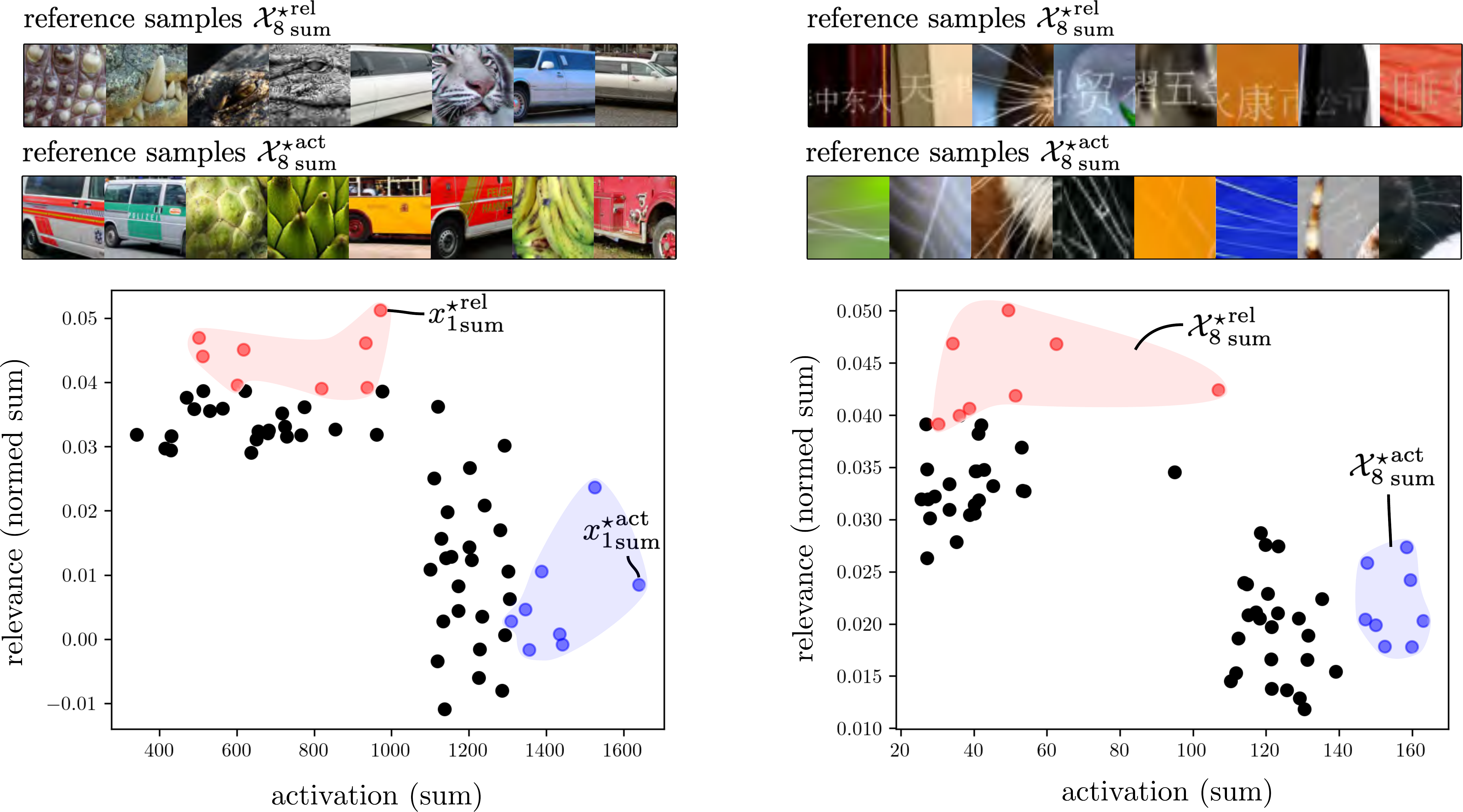}
      \caption{Relevance-based reference samples illustrate how concepts are \emph{used} in a global inference context of the model, activation-based samples show the strongest manifestations along the direction or dimension describing a concept in latent space. Both methods can produce different reference samples for the same filter. (\textit{Bottom}): Each scatter plot shows two sets of 32 samples maximizing the \gls{rmax} or \gls{amax} objective along two axes measuring their summed activation and normalized relevance values (see Section \ref{sec:appendix:technical} for details). (\textit{Top}): The respective reference examples for ${\mathcal{X}^{\star}_8}^\text{rel}_\text{sum}$ (shaded red in scatter plots) and ${\mathcal{X}^{\star}_8}^\text{act}_\text{sum}$ (shaded blue). \emph{(Left)}: A particular channel (VGG-16, \texttt{features.26}, filter index 350) encodes for a polysemantic concept including cars and coarse patterns. Activation-based samples present cars of intense color and vegetables where the coarse pattern is dominant. The relevance-based reference samples however expose, that strong color values are not as important, and the concept seems to be centered around texture and the color gray. Further, the coarse pattern detector is mainly utilized by the model to classify alligators instead of fruits. \emph{(Right)}: For another channel and model (VGG-16 with BatchNorm layers, \texttt{features.30}, filter index 361), activation-based reference samples show thin white stripes. The stripe concept however is used to detect white letters, as relevance-based samples suggest.}
      \label{fig:appendix:experiments:visual:rel:vs:act:refsamples:scatter}
    \end{figure*}
\subsection{Activation- vs.\ Relevance-based Sample Selection}
\label{sec:appendix:experiments:selecting_reference_samples}
\begin{figure*}[t]
        \centering
        \includegraphics[width=1\textwidth]{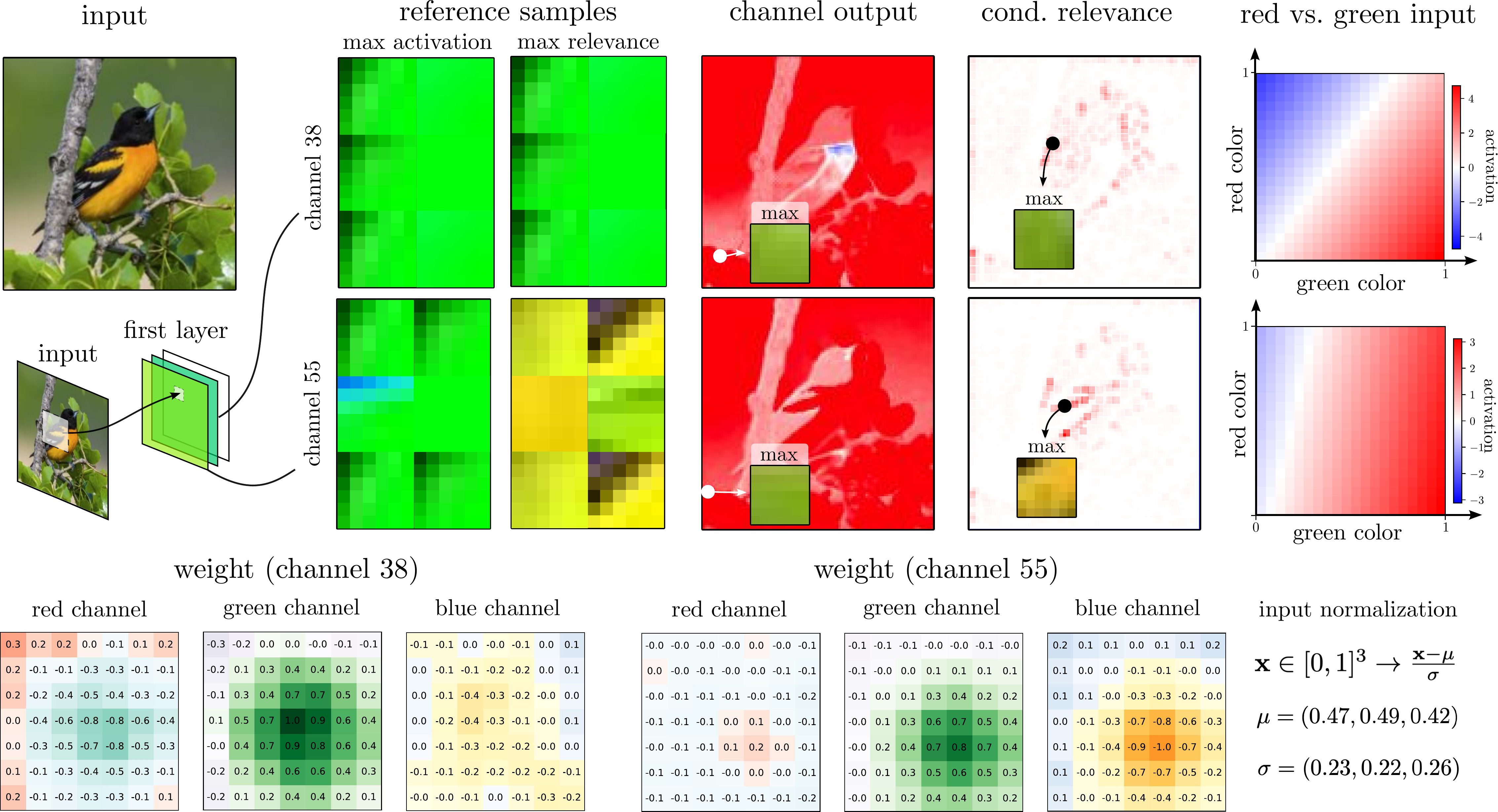}
        \caption{Difference in relevance and activation-based reference samples for low-level color concepts in the first layer of a ResNet architecture for bird classification. Two filters (55 and 38) activate strongly for green colored inputs, as activation-based reference samples and channel output maps indicate. However, one channel (55) is relevant for yellow color detection, as is visible in the relevance-based reference sample set and conditional heatmap that points to the yellow feathers of the bird. Thus, both channels activate for green color, but are used differently by the model. In the bottom, the weight values of each filter are further given, indicating which input tensors lead to a strong activation. This is also illustrated in heatmaps showing the activation values induced by red, green or a mixed-colored inputs. It is to note, that RGB inputs are first preprocessed to be in $[0, 1]^3$ by division with value 255 and thereafter normalized with mean (0.47, 0.49, 0.42) and standard deviation (0.23, 0.22, 0.26). Yellow in raw RGB space is (225, 255, 0).}
        \label{fig:appendix:experiments:birds:green:yellow}
    \end{figure*}

The criteria after which reference samples are chosen impact the resulting sample sets for concept identification. We have stated that relevance-based reference samples show how concepts are \emph{used} in practice. Activation-based samples, on the other hand, illustrate the strongest (potentially adversarial) manifestations along the direction or dimension describing a concept in latent space. In general, both sample sets can overlap, as relevance and activation correlate to some extent (see Supplementary Figure~\ref{fig:appendix:rel-vs-act-general-examples}). However, this is in general not the case, causing the respective sets of reference samples to (potentially highly) differ. Thus, it is a sensible option to observe both relevance and activation-based reference sample sets complementarily as they can provide different perspectives on the function of an analyzed neuron or filter. However, the potentially adversarial nature of activation-based reference sample selection needs to be considered carefully, as will be discussed in context of the second case study and Supplementary Figure~\ref{fig:appendix:experiments:birds:green:yellow}. In order to illustrate the differences in samples sets selection that can occur, we present and discuss the following case studies.

\paragraph{Case Study 1: Higher-Level Concepts}
    Two examples of reference sets for higher-level concepts extracted from two upper convolutional layers of a pre-trained VGG-16 network (see Section~\ref{ap:vgg16}) are shown in Supplementary Figure~\ref{fig:appendix:experiments:visual:rel:vs:act:refsamples:scatter}. Here, scatter plots show different examples positioned in $\mathbb{R}^2$ according to their relevance attributions and layer activations of the sample sets ${\mathcal{X}^{\star}_8}^\text{rel}_\text{sum}$ (red shaded) and ${\mathcal{X}^{\star}_8}^\text{act}_\text{sum}$ (blue shaded). 

In the left plot, we identified a polysemantic channel in layer \texttt{features.26} of a VGG-16 model, that encodes for cars and coarse patterns. 

Activation-based samples present cars of intense color and vegetables where a coarse pattern is dominant. Specifically, 8 out of the 10 classes for which the neuron activates strongest on average are ``car''-specific\footnote{classes: ``ambulance'', ``minibus'', ``trolleybus'', ``police van'', ``minivan'', ``beach wagon'', ``fire truck'' and ``pickup''}. The two other classes are ``banana'' and ``artichokes''. 

Reference images obtained via \gls{rmax}-sorting tell a different story: Whereas vehicles seem to be important on one hand, the top three classes in terms of relevance attributed to the channel are, on the other hand: ``alligator'', ``limousine'' and ``crocodile''. These three classes are all having in common a gray color --- contrary to the colorful images from \gls{amax}-sorting. Further, the coarse pattern detector is by the model mainly utilized to classify reptiles instead of fruits. In hindsight, the bananas and artichokes obtained via \glsdesc{amax} structurally match the scale pattern of reptiles, however the color saturation ``overshoots''. In fact, for sample $x_{1 \text{sum}}^{*\text{rel}}$  (``alligator'' class) the second most relevant filter in layer \texttt{features.26} is indeed this unit accounting for 5.1\% of total absolute relevance. Analyzing the sample $x_{1 \text{sum}}^{*\text{act}}$ (``ambulance'' class), the filter ranks only at position 27 accounting for 0.8\% of total absolute relevance, thus, having no significant contribution to the decision process, as the perturbation experiment in Section~\ref{sec:appendix:experiments:quantitative:filterperturbation} demonstrates. This exemplifies that a maximally strong activating filter cannot be equated with its usefulness for classification.
    
The right plot of Supplementary Figure \ref{fig:appendix:experiments:visual:rel:vs:act:refsamples:scatter} shows the reference sample sets for channel 361 in \texttt{features.30} of a VGG-16 model with BatchNorm layers (see Appendix~\ref{ap:vgg16}). As \gls{amax}-sorting suggests, the neuron's function consists of the detection of thin white lines, since spider webs or whiskers are visible in the images. On the other hand, samples drawn to maximize relevance mostly depict written characters. Thus, activation indicates which general pattern a filter activates, whereas relevance clarifies its specific usage. In this particular case, \gls{rmax}-sorting has revealed a Clever Hans \cite{lapuschkin2019unmasking} feature, which extends to several classes of the ImageNet dataset, as further analysis in Section~\ref{sec:experiments:quantitative:imageretrieval} in the main manuscript verifies.

\paragraph{Case Study 2: Lower-Level Concepts}
    
Differences in relevance and activation-based reference samples can also occur for lower-level concepts, as is shown in Supplementary Figure~\ref{fig:appendix:experiments:birds:green:yellow}. In this case, two filters, \ie, channels 55 and 38 located in the first layer \texttt{conv1} of a ResNet34 model trained on the Caltech-UCSD Birds 200 dataset, activate strongly for green colored inputs. However, channel 55 is relevant for yellow color detection, as is visible in the relevance-based reference sample set. For a particular sample consisting of green and yellow colored parts, activation maps show, that both channels activate strongly for ``green''. However, the difference in function becomes apparent by investigating the channel-conditional relevance heatmaps. Here, the maximum value of relevance can be assigned to a green patch for channel 38 and to a yellow patch for channel 55.
    
Additionally, when investigating the kernel weights of the filters (see bottom of Supplementary Figure~\ref{fig:appendix:experiments:birds:green:yellow}). we see that both filters favor green inputs, represented in large weights for the green channel. Whereas channel 38 additionally strongly suppresses red colored inputs (the opposite of cyan in RGB color space), channel 55 suppresses blue color (the opposite of yellow, RGB value of yellow is (225, 255, 0)). Thus, both filters activate most strongly for green color, but channel 55 also activates strongly for yellow color --- compared to channel 38 --- yet inputs extremely strong in the green spectrum without red color influence can cause the activations of channel 55 to saturate, obfuscating the true purpose of the channel's filter kernels. This is also supported by heatmaps in the very right part of Supplementary Figure~\ref{fig:appendix:experiments:birds:green:yellow}, showing which ($7 \times 7$)-sized input of uniform color varying in the red and green channels  and matching the receptive field of the neurons of the first layer leads to the highest activation values. Strengthening red color in the input suppresses activation strongly for channel 38, whereas activation for channel 55 barely decreases.

\paragraph{Case Study 3: Maximization Target Sensitivity}
    The previous two case studies illustrated general differences in activation and relevance-based reference sample selection. In the following, motivated by an experiment from Chen et al. \cite{chen2020concept}, the influence of the maximization target, \eg, summation over the channel values or taking the maximum, on the sample sets is investigated. First, two examples are shown, and subsequently the general behavior is analyzed.

    For the first example, reference sample sets $\mathcal{X}^{\star}_{8}$ for \gls{rmax} and \gls{amax} are shown in Supplementary Figure~\ref{fig:appendix:experiments:visual:rel_vs_act:maximization_targets}~(\emph{top}) for different maximization targets, illustrating the concept of channel 323 in layer \texttt{features.28} of a VGG-16 network trained on ImageNet. Here, all reference sets contain images of a maritime theme except from the set ${\mathcal{X}^{*}_{8}}^{\text{act}}_{\text{max}}$, which mostly includes images of dog legs with long fur. A global analysis shows, that the channel's concept is most relevant on average for the classes ``reef coral'', ``cramp fish'', ``scuba diver'', ``sea snake'' and ``tiger shark''. Thus, ${\mathcal{X}^{*}_{8}}^{\text{act}}_{\text{max}}$ spuriously presents potentially misleading examples for the channel's encoded concept.
    
    \begin{figure*}[t]
      \centering
      \includegraphics[width=1\linewidth]{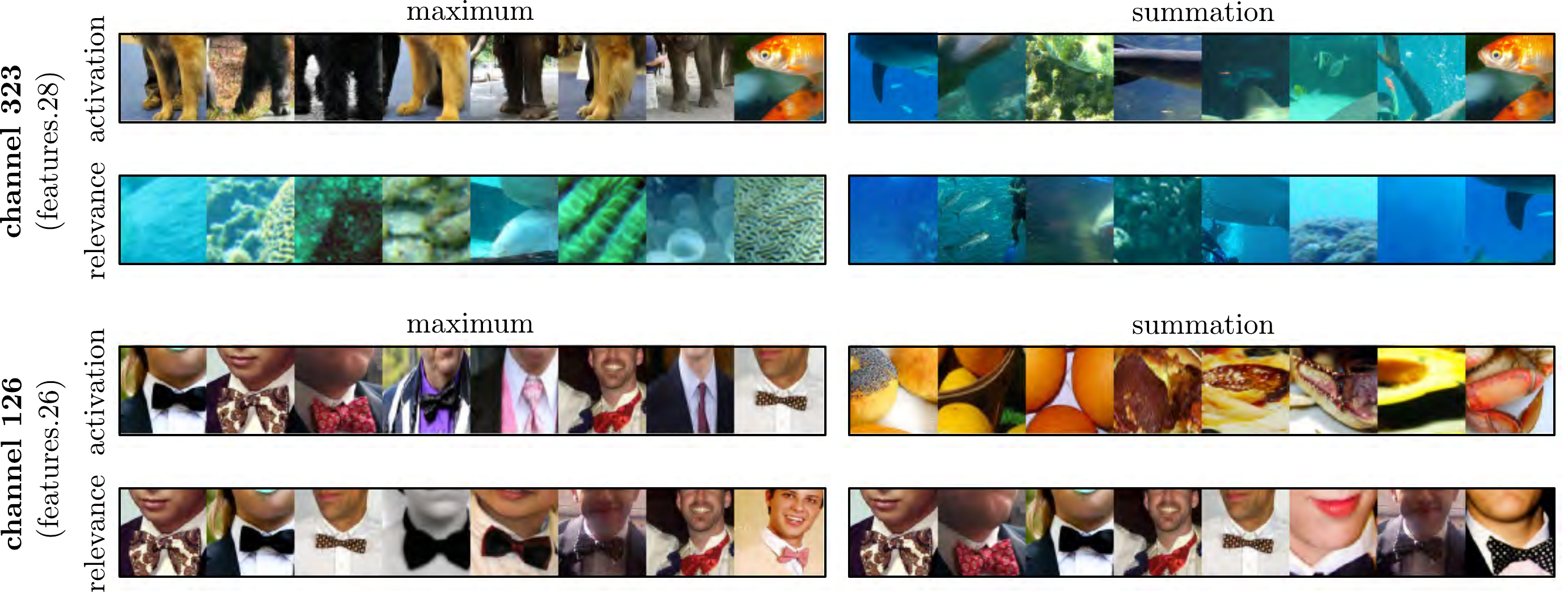} 
      \caption{Reference images (cropped \wrt\ the receptive field of the most active/relevant neuron inside channel, see Section~\ref{sec:appendix:methodsindetail:understanding:scaling}) collected using \glsdesc{amax} and \glsdesc{rmax} via summation, or taking the maximum,  within a convolutional channel. Whereas relevance-based reference samples show similar themes, activation-based reference samples are more sensitive to the chosen maximization target. (\emph{Top}): All reference sets contain images of a maritime theme except from ${\mathcal{X}^{*}_{8}}^{\text{act}}_{\text{max}}$, which mostly includes images of dog legs with long fur. (\emph{Bottom}): All reference sets contain images of bow or neck ties except from ${\mathcal{X}^{*}_{8}}^{\text{act}}_{\text{sum}}$, which mostly includes images of rounded shapes and orange color. See Supplementary Figure~\ref{fig:appendix:visual:rel_vs_act:maximization_targets} for relevance and activation maps corresponding to the examples shown here, and Supplementary Figure~\ref{fig:appendix:visual:rel_vs_act:maximization_targets2} for additional example filters analyzed.}
      \label{fig:appendix:experiments:visual:rel_vs_act:maximization_targets}
    \end{figure*}

    \begin{figure*}[h]
      \centering
      \includegraphics[width=1\linewidth]{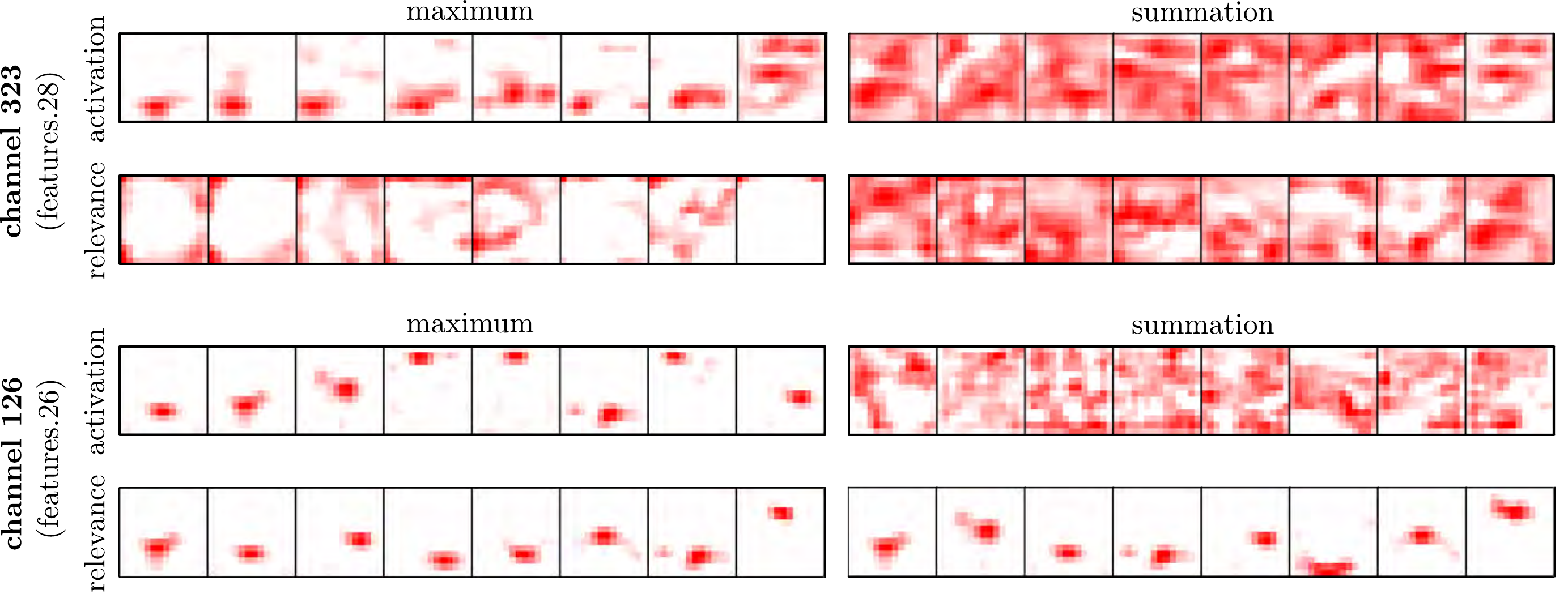} 
      \caption{Complementary information to Supplementary Figure~\ref{fig:appendix:experiments:visual:rel_vs_act:maximization_targets} showing the distribution of activation scores or relevance inside the channel for samples collected using \glsdesc{rmax} and the activation distribution for samples collected using \glsdesc{amax}, \ie, the quantities serving as respective criterions for reference sample selection. (\emph{Top}): Channel 323 from layer \texttt{features.28} is mostly used in a maritime setting based on relevance-sorted reference samples. It can be seen that the concept is hereby spatially extended (covers large parts of the input sample). Sorting regarding maximum activation value, on the other hand, results in edge-case samples. (\emph{Bottom}): Channel 126 from layer \texttt{features.26} is most often used to detect bow ties or neck ties based on the relevance-sorted reference samples. It can be seen that the concept is spatially localized (covers small parts of the input sample). Sorting regarding summation activation, on the other hand, results in edge-case samples.}
      \label{fig:appendix:visual:rel_vs_act:maximization_targets}
\end{figure*}

The second example is shown in Supplementary Figure~\ref{fig:appendix:experiments:visual:rel_vs_act:maximization_targets} (\emph{bottom}), depicting reference sample sets for filter channel 126 in \texttt{features.26} of the same VGG-16 network. Here, all sample sets contain images of bow or neck ties except from the set ${\mathcal{X}^{*}_{8}}^{\text{act}}_{\text{sum}}$, which includes images depicting round shapes and orange color. Globally most relevant classes on average for this filter are ``bow-tie'', ``neck brace'', ``suit'', ``military uniform'', ``academic robe'' and ``grenadier guards'' (known from Buckingham Palace). This time, ${\mathcal{X}^{*}_{8}}^{\text{act}}_{\text{sum}}$ does not lead to clear visualizations of the concept and its examples do not align well to the contexts of the class labels for which the model is using the channel.

This observation is due to two effects: First, \gls{rmax} considers only a limited subset of samples, for which the filter was \emph{actually helpful} for classification. Secondly, summation indirectly favors samples where the activation/relevance is spatially spread out inside the channel, as for example in images showing multiple instances of a concept, or a texture covering large parts of the input. When taking the maximum value instead of sum-aggregating, this is not the case. Since each neuron in a convolutional layer performs the same convolution operation, spatially widespread activation within the channel suggests that the pattern, for which the kernel encodes for, occurs repeatedly in the input. On the contrary, if a filter encodes for a spatially confined concept, \eg, such as a bow-tie in Supplementary Figure~\ref{fig:appendix:experiments:visual:rel_vs_act:maximization_targets} (\emph{bottom}), then the resulting channel activation is also locally restricted, as shown in the corresponding Supplementary Figure~\ref{fig:appendix:visual:rel_vs_act:maximization_targets}. Confusion occurs, when the concept's size of the filter does not match the computation operation, especially when no images exist that contain multiple of these small-scale concepts. In Supplementary Figure~\ref{fig:appendix:visual:rel_vs_act:maximization_targets} (\emph{top}),  ${\mathcal{X}^{*}_{k}}^{\text{act}}_{\text{max}}$ results in spurious edge case samples that cause spatially confined activation, although the kernel actually encodes for water texture, which is usually widespread. The opposite is the case in Supplementary Figure~\ref{fig:appendix:visual:rel_vs_act:maximization_targets} (\emph{bottom}) for ${\mathcal{X}^{*}_{k}}^{\text{act}}_{\text{sum}}$. These examples illustrate, that it is generally not trivial to choose the correct calculation method for activation since a concept's size is not known beforehand. Relevance, on the other hand, seems to be more robust against different computation modalities.
    
\begin{figure*}[h]
    \centering
    \includegraphics[width=1\textwidth]{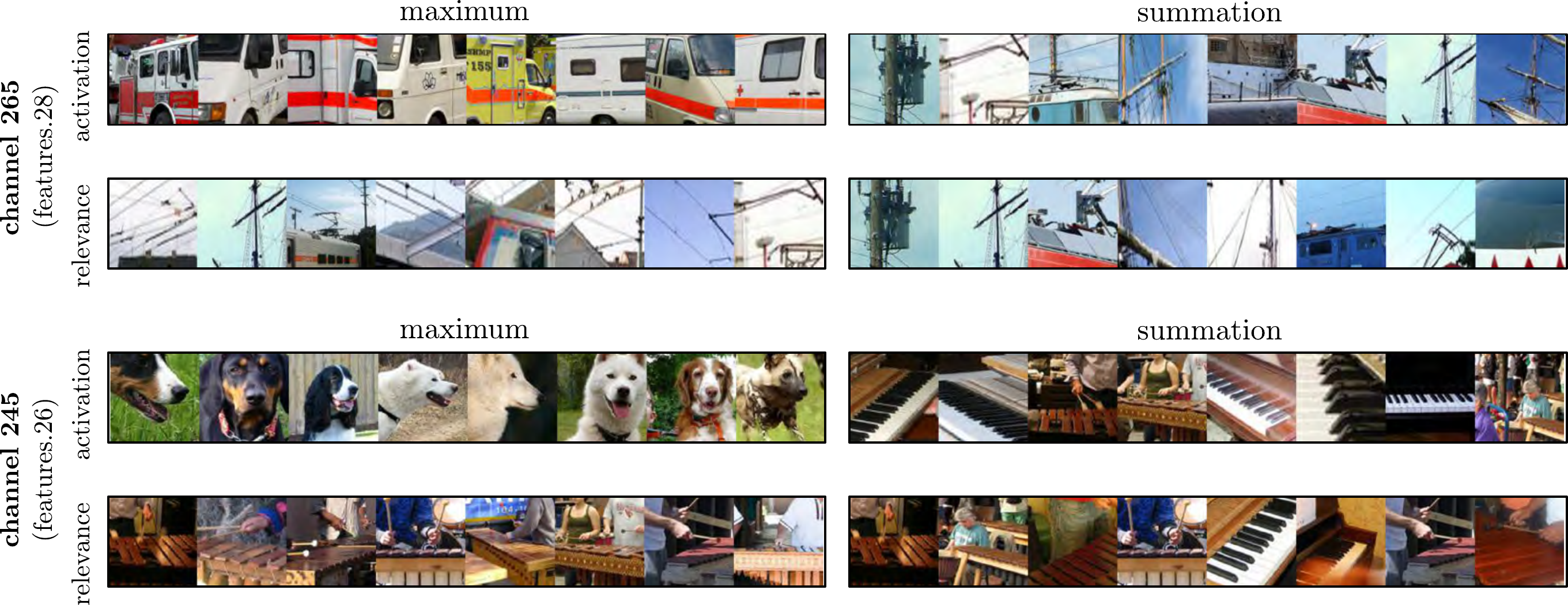}
    \caption{Reference images collected using \glsdesc{amax} and \glsdesc{rmax} via summation, or taking the maximum activations with additional channel maps. (\emph{Top}): The relevance-based reference samples suggest, that channel 265 is used to detect thin lines as found in electric masts or ship ropes. Such lines are also found in vans, as maximum activation-based sorting shows. (\emph{Bottom}): Channel 245 is used to detect repeating patterns of keys from, \eg, pianos or xylophones. This is shown by relevance-based sorting. The keys in pianos and xylophones are separated by a darker gap (xylophone) or black key (piano). Such a contrastive blending in color is also found in dogs with bright fur, as can be seen in maximum-activation-based samples. Supplementary Figure~\ref{fig:appendix:visual:rel_vs_act:maximization_targets2_maps} shows the to these examples corresponding activation- and relevance attribution maps.}
  \label{fig:appendix:visual:rel_vs_act:maximization_targets2}
\end{figure*}

\begin{figure*}[h]
    \centering
    \includegraphics[width=1\textwidth]{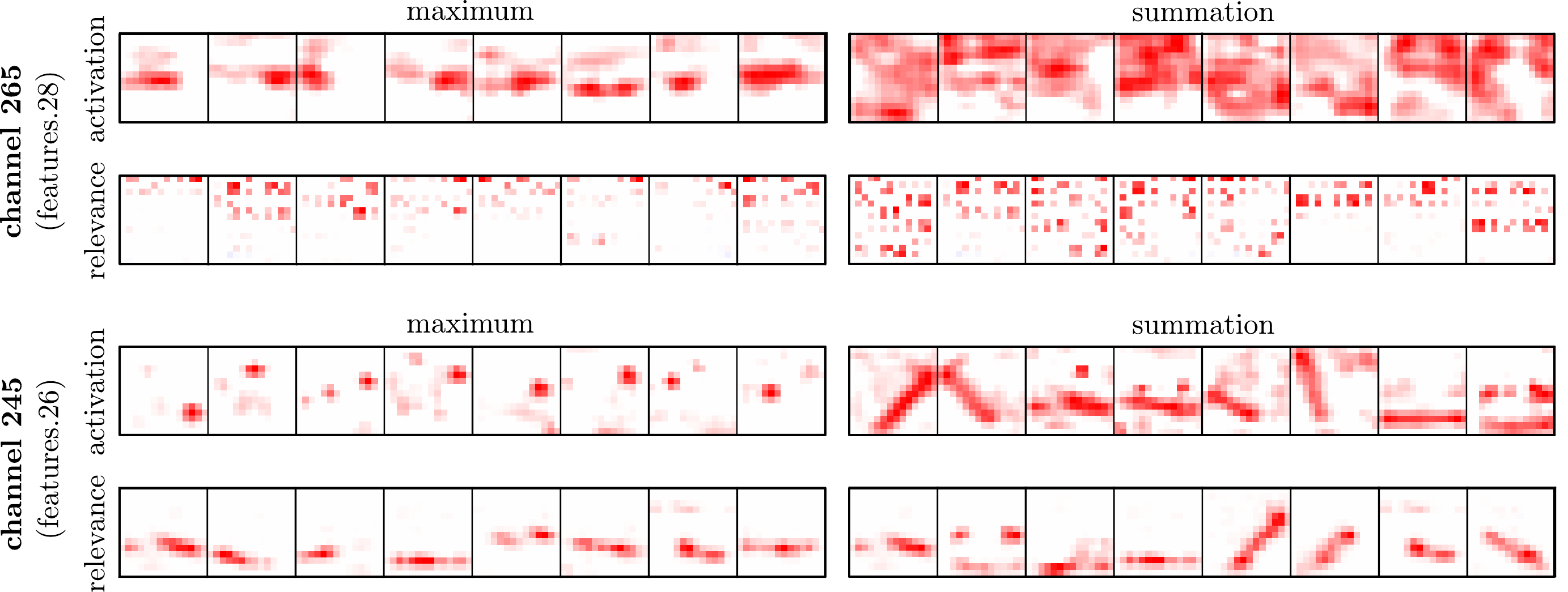}
    \caption{
    Extension of Supplementary Figure~\ref{fig:appendix:visual:rel_vs_act:maximization_targets2} showing the relevance distribution inside the channel for samples collected using \glsdesc{rmax} and the activation distribution for samples collected using \glsdesc{amax}. }
    \label{fig:appendix:visual:rel_vs_act:maximization_targets2_maps}
 \end{figure*}
 
An additional example is discussed with Supplementary Figures~\ref{fig:appendix:visual:rel_vs_act:maximization_targets2} and~\ref{fig:appendix:visual:rel_vs_act:maximization_targets2_maps}, showing two channels of a VGG-16 model trained on ImageNet in layer \texttt{feature.28}
    
Channel 265 is used to detect thin lines as found in electric masts or ship ropes, as relevance-based reference samples suggest. Such lines are also found in images of vehicles, as maximum activation-based sorting shows. Here also, activation-based sampling strongly differs qualitatively in the set of reference samples. Channel 245 is used to detect repeating patterns of keys, found in pianos or xylophones. This is shown by relevance-based sorting. The keys in pianos and xylophones are separated by a darker gap (xylophone) or black key (piano). Such a contrastive blending in color is also found in dogs with bright fur, as can be seen in maximum-activation-based samples. All in all, these examples further show, that relevance-based criteria are more stable regarding the spatial size of a concept.

Since the ranked example images have the purpose to represent \textit{semantic} concepts, it is difficult to make a clear \textit{quantitative} evaluation about which calculation method provides better explainability. One way to quantify the robustness of the explanations, however, is to calculate the intersection between the sets of reference images generated via summation within the convolutional channel or taking the maximum, and \gls{amax} compared to \gls{rmax}. The greater the overlap, the more similar are the image sets. The results of the computation of the mean set similarity in layer \texttt{features.0} until \texttt{features.28} in VGG-16 are depicted in Supplementary Figure~\ref{fig:appendix:robustness:act_rel}, demonstrating that \gls{rmax}-based example sets are in general less sensitive to the choice of summation or maximization (Supplementary Figure~\ref{fig:appendix:robustness:act_rel}~\emph{(left)}) and thus more robust, as they represent examples which are actually used by the model in a larger context. We additionally can observe that under summation, \gls{amax} and \gls{rmax} example sets are considerably more similar than when comparing example sets which maximize their respective criterion. We attribute this to the stabilizing effect to relevance and summation computations. However, there is still a considerable difference between the respective example sets as shown by the low set intersection of less than 0.3. 
\begin{figure*}[t]
        \centering
        \includegraphics[width=0.46\textwidth]{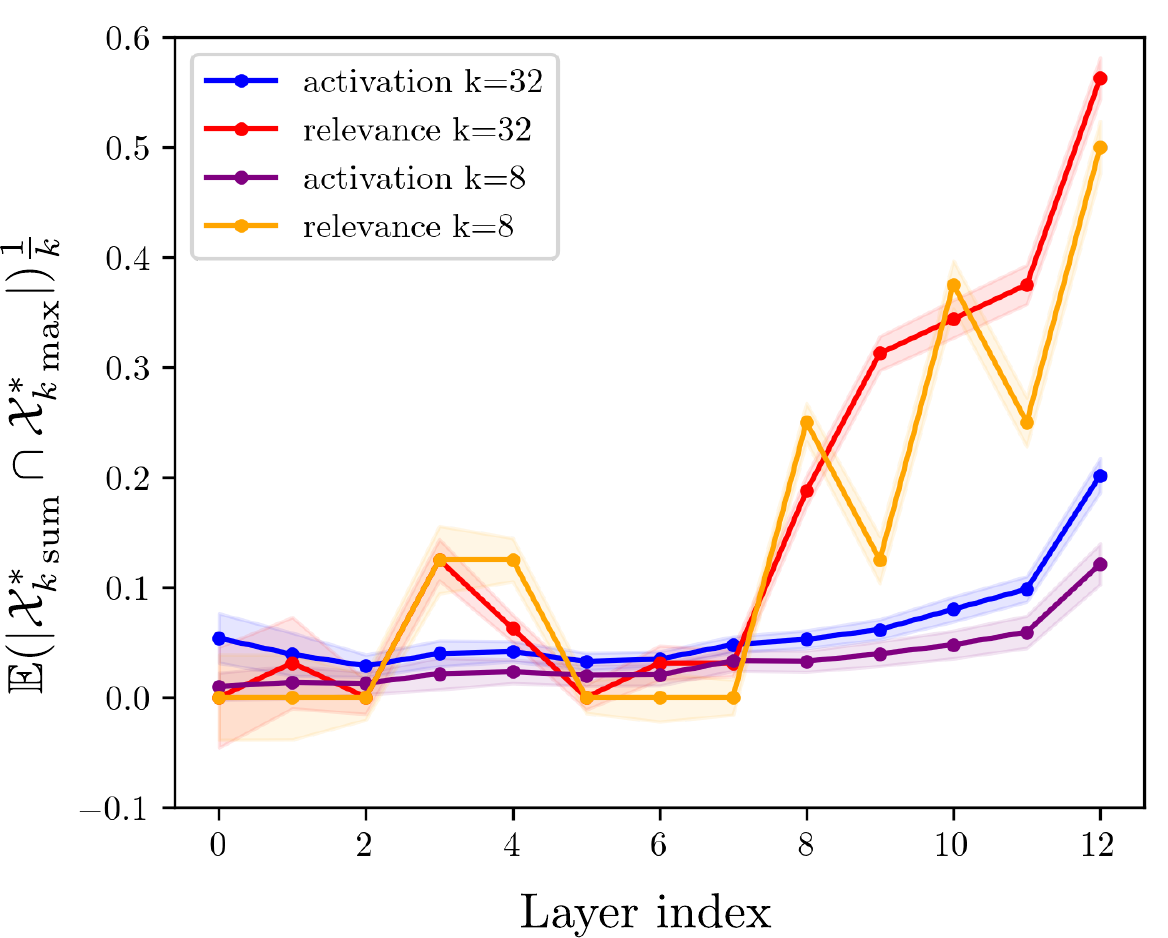}
        \includegraphics[width=0.46\textwidth]{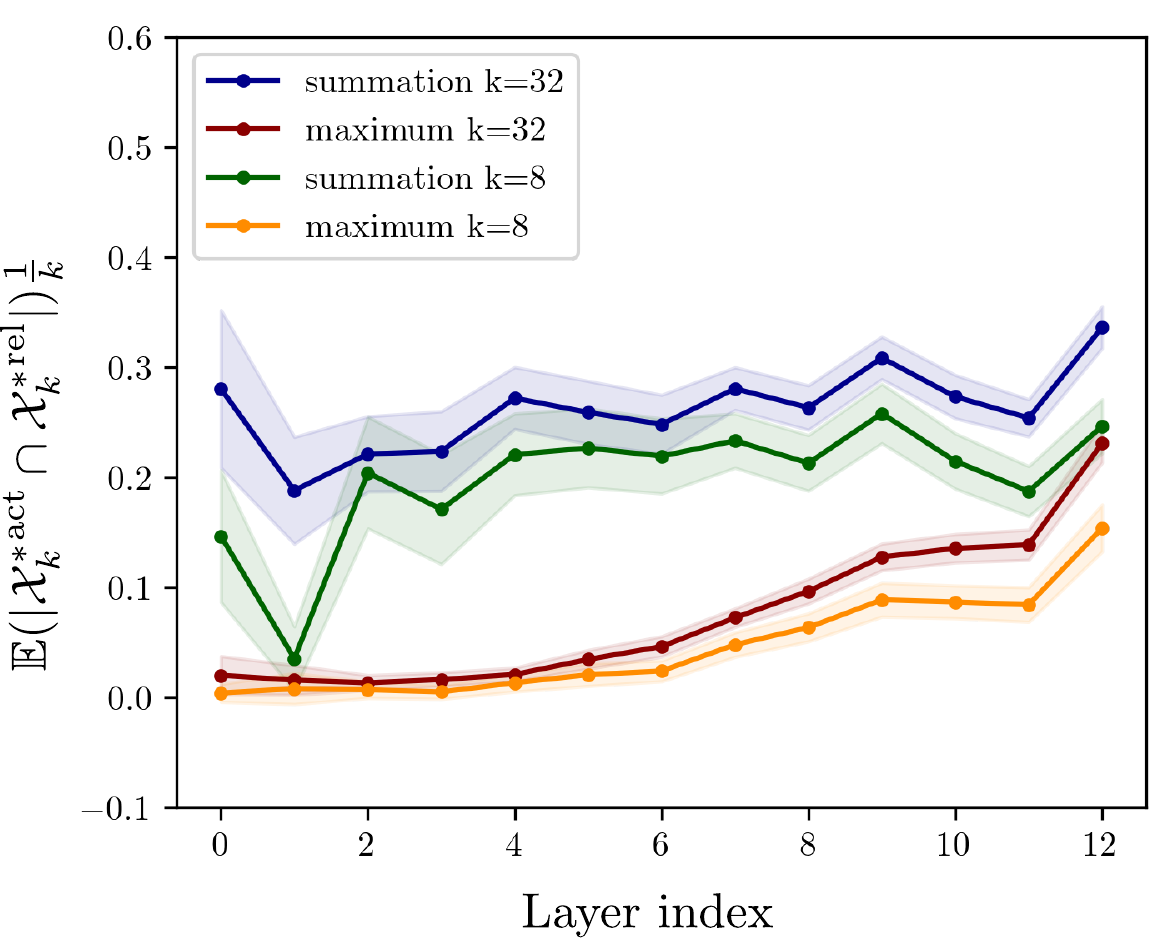}
        \caption{Mean set similarity of each filter channel per convolution layer in a VGG-16 model pretrained on ImageNet with $k=32$ and $k=8$. The standard error of mean is overlaid with a confidence interval of 99.7 \%. The layer index 0 on the x-axis corresponds to layer \texttt{features.0} and index 12 to \texttt{features.28}, since we only compare for convolutional layers with weight kernels (and ignore, \eg, pooling layers). \emph{(Left)}: Comparing the computation modality, channel-summation vs. taking the channel maximum. The plot shows that the selection of relevance-based examples is more robust \wrt\ the choice of maximization vs. summation of values. The low similarity in earlier layers (0-7) is due to these layer's small receptive fields and consequently low probability of selecting same example subsets from the high number of small cropped sub-images, and not properties of \glsdesc{amax} or \glsdesc{rmax}. \emph{(Right)}: Fixing the choice of channel value summation or maximization as a criterion for selecting examples, we here compare \gls{amax} to \gls{rmax} sample sets. Under summation, relevance-based and activation-based sample sets are more similar, \ie, as \gls{amax} example sets become more robust and stable. However, there are still notable differences between the general strategies \gls{amax} and \gls{rmax}.}
        \label{fig:appendix:robustness:act_rel}
    \end{figure*}
Neurons in lower layers typically have a very narrow receptive field. For instance, a neuron may encode for ``red stripes''. This concept is in almost all images present. As a consequence, the probability of finding an intersection of ${\mathcal{X}^{*}_{k}}^{\text{act}}_{\text{sum}} \cap {\mathcal{X}^{*}_{k}}^{\text{act}}_{\text{max}}$ converges to zero due to the increased number of prospective candidate samples and low numbers of $k$, although, \eg,  all images may show red stripes. This is due to the high number of images and the low probability of taking the same subset. This problem does not occur in higher layers. Here, for instance, a neuron encodes for ``trumpets''. The number of images depicting trumpets is much smaller than images containing red stripes. If both variants illustrate ``trumpets'', then an intersection between ${\mathcal{X}^{*}_{k}}^{\text{act}}_{\text{max}} \cap {\mathcal{X}^{*}_{k}}^{\text{act}}_{\text{sum}}$ is more likely, as only a small subset of images contains this higher level concept. If both variants encode for different features, then the intersection should be empty. Finally, this computation is in all cases only an approximation.

The results suggest, that reference image sets obtained via \glsdesc{rmax} are more robust than images obtained through \glsdesc{amax}.

\subsection{Handling High and Low Sample Set Variety}
\label{sec:appendix:experiments:qualitative:variety}
In general, reference samples are collected by measuring activations or attributions during model predictions sample by sample. The higher the variety in the data samples, the more abstractly can the concept be described eventually. A high feature variety is especially crucial for datasets such as, \eg, the Caltech-UCSD Birds 200 dataset with samples mostly consisting of birds. However, if reference sample variety is too high, it can also become difficult to extract the common concept. Further, for polysemantic neurons (\ie, neurons addressing multiple concepts), numerous reference samples might only describe one particular concept/facet, thus missing the other concepts. In order to find the sweet spot between too many and too few potential reference samples, two approaches are presented in the following. The first approach conditions the relevance-based reference sample set towards specific output classes in order to decrease reference set variety within each set, and at the same time better contextualize and organize the selected examples to increase interpretability. The second approach briefly discusses how the reference dataset can be extended in order to increase the sample set variety, which might be of use when all originally identified most relevant examples are near identical and no (singular) common factor among the samples can be easily identified.

\paragraph{Class-Conditional Reference Sample Selection}
\label{sec:appendix:experiments:qualitative:class_conditional}
Activation-based reference sample selection for a neuron of interest only depends on the input sample and the (partially executed) forward pass until the layer of interest. While a per-class selection of activation-based reference examples is possible, this usually requires a class-specific pre-selection of the data samples under ground truth label availability~\cite{nguyen2021explaining, hohman2019summit}. However, assigning a high neuron activation to a ground truth label is only sensible, if one can assume, that merely features regarding the chosen label exist in the input sample. In this case, one implicitly makes the claim, that the high activation value corresponds to a high relevance regarding the chosen class. The neuron can however respond to a different class or might be opposed to the true class, as previously discussed in Section \ref{sec:appendix:methodsindetail:understanding:selecting_reference_samples:relevance}.

On the other hand, using relevance-based attributions instead of activations, a pre-selection based on ground truth labels is also possible. However, in the absence of data labels, it is also feasible to condition reference samples on a class of interest by simply measuring its attribution scores from computing \gls{crc} scores \wrt~the model output category of interest, regardless of the predicted label (\cf Section \ref{sec:appendix:methodsindetail:understanding:selecting_reference_samples:relevance}, Supplementary Figure~\ref{fig:appendix:rel-vs-act-general-examples}). Retrieving relevance-based reference samples for a particular concept, yet conditioned to different output classes of a model, will yield different class (or in extension condition $\theta$) -specific perspectives on the concept, \ie, how it is \emph{actually used} by the model in context of different prediction targets. Consequently, output categories (classes) of a model can be ranked by how much they make use of the observed concept, and class-conditioned reference examples can be obtained by appropriately configuring $\theta$ for \gls{crc}. See Section~\ref{sec:appendix:technical} for details on initializing and computing of relevance.

\begin{figure*}[t]
  \centering
  \includegraphics[width=1\textwidth]{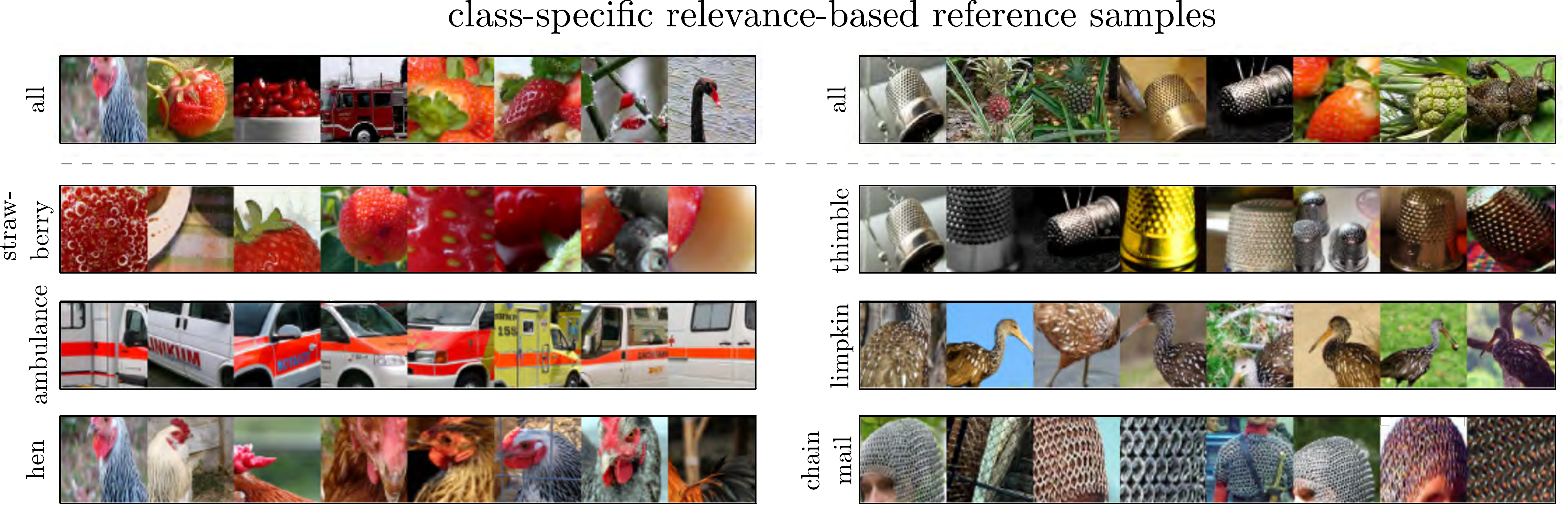}
  \caption{Relevance-based reference samples can be filtered by assessing the relevance of a concept \wrt~target classes $y$ of interest, by adequately setting \glsdesc{crc} conditions $\theta_y$. (\emph{Left column}:) VGG-16, filter 457 of \texttt{features.28}. (\emph{Right column}:) VGG-16, filter 209 of \texttt{features.28}. (\emph{1st row}): Without specifying any further conditions, reference sample sets contain the maximally relevant samples for the observed concept selected from samples of all classes and relevance scores computed for the samples' ground truth labels. (\emph{2nd to last row}): Reference example sets of the top-3 classes (sorted in descending order) for which the analyzed concept is most relevant on average. Conditioned to a specific target class, reference sample sets contain only maximally relevant samples for the concept as used in context of the class of interest. Additional results on the Fashion-MNIST domain can be found in Supplementary Figure~\ref{fig:appendix:reference_sample_selection:class_conditional2}.}
  \label{fig:appendix:class-conditional-reference-sample-selection}
\end{figure*}

Supplementary Figure~\ref{fig:appendix:class-conditional-reference-sample-selection} (\emph{1st row}) shows the top-8 ${\mathcal{X}^{*}_{k}}^{\text{rel}}_{\text{sum}}$  reference examples for convolutional filters in \texttt{features.28} of a VGG-16 network ranked \wrt\ the relevance attributions obtained by computing \gls{crc} for each of the sample's respective ground truth class. (\emph{2nd to last row}) Reference examples for the respective filters are selected based class-conditional relevance scores  computed for \emph{all} samples, for each of the top-3 classes the observed filter has contributed most strongly to on average. We can observe that while in the first row all sorts of objects can appear among the selected examples which express a manifestation of a shared concept, the class variety may distract from the identification of the core of the visualized concept. By selecting the available reference samples with regard to the classes being influenced the most from the observed concept filter, we can neatly select and organize the examples in different ``perspectives'' of the model.

In the left column, filter 457 is analyzed. While the first row with examples, drawn based on relevance from the product space of all samples and classes, already suggests that the color red is important, the second to the last row reveal that classes ``strawberry'', "ambulance" and ``hen'' most prominently make use of this filter during inference, and that here, among the examples chosen based on class-conditional relevance all share red colored features such as berries, stripes or caruncles. Similarly for filter 209 analyzed in the right column, the set of examples with highly diverse class memberships is presented in context of the top-3 benefitting classes ``thimble'', ``limpkin'' and ``chain mail''. Observing all class-conditioned examples together, it immediately becomes clear that the analyzed filter encodes for a concept of regular dot-like textures. The increase in clarity from conditionally retrieved reference examples can naturally be combined with the computation of per-example heatmap explanations, as is described in Section~\ref{sec:appendix:methodsindetail:understanding:visualization} for an exponentiated effect.

\begin{figure*}[h]
    \centering
    \includegraphics[width=1\textwidth]{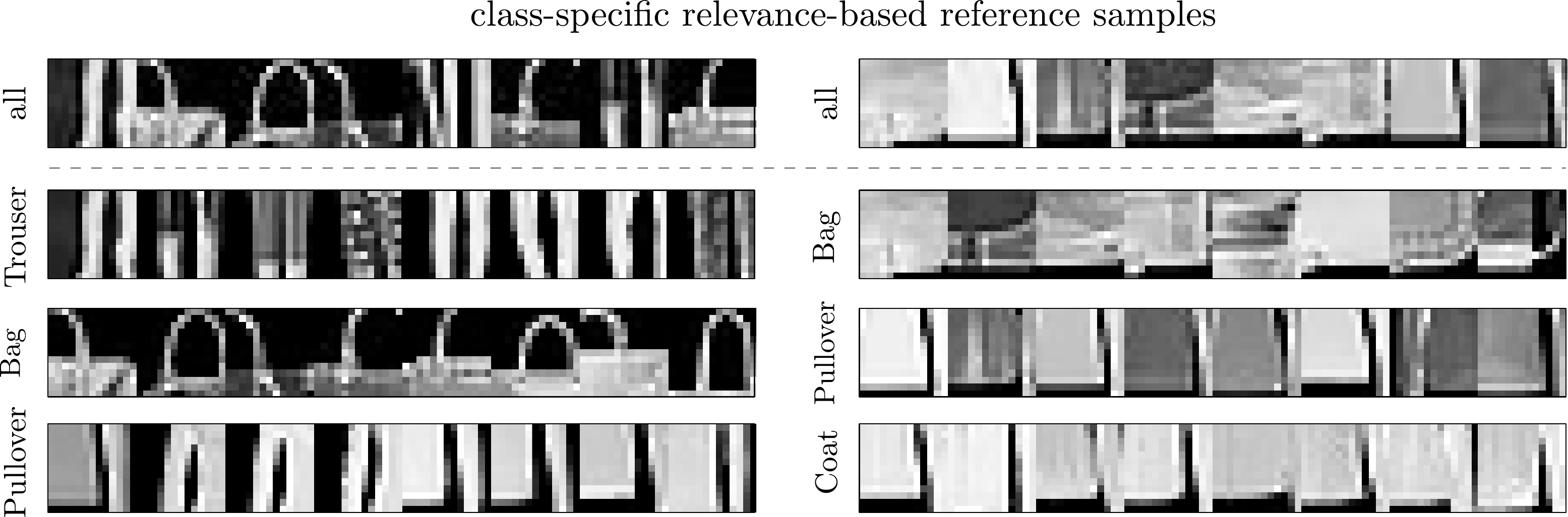}
    \caption[Reference samples based on relevance-sorting can be conditioned on a class of interest. Samples of Fashion-MNIST are shown.]{Reference samples based on relevance-sorting can be conditioned on a class of interest. Shown are reference samples of channel 1 in layer \texttt{l3} (\emph{Left column}) and channel 14 of \texttt{l3} (\emph{Right column}) of a LeNet-5 model trained on Fashion-MNIST (details in Section~\ref{ap:lenet}). (\emph{1st row}): Without specifying any further conditions, reference sample sets contain the maximally relevant samples for the observed concept selected from samples of all classes. Here, channel 1 is used to detect thin vertical lines. Channel 14 is used to detect spatially extended pieces of clothing with a black line at the bottom. (\emph{2nd to last row}): Reference example sets of the top-3 classes for which the analyzed concept is most relevant on average. Here, it can be seen, how the concepts of channel 1 and 14 are used for the respective classes. For example, channel 1 is used to detect thin pant legs of trousers or handles of a bag.}
    \label{fig:appendix:reference_sample_selection:class_conditional2}
\end{figure*}

In a second example depicted in Supplementary Figure~\ref{fig:appendix:reference_sample_selection:class_conditional2}, class-specific relevance-based reference samples are shown for the custom LeNet trained on Fashion-MNIST. Without specifying any further conditions, reference sample sets contain the maximally relevant samples for the observed concept selected from samples of all classes. Here, channel 1 in layer  \texttt{l3} (\emph{Left column}) is used to detect thin lines and curves. Channel 14 of layer \texttt{l3} is used to detect spatially extended pieces of clothing with a black line at the bottom. By conditioning reference samples towards specific classes, the use case of the concepts can be further separated. For example, channel 1 is used to detect thin pant legs of trousers or handles of a bag.

\paragraph{Reference Dataset Extension}
\label{sec:appendix:experiments:qualitative:dataset_extension}
To visualize concepts with reference samples, a first step is to use reference samples from the dataset used for training and validation. This set might be however limited in the manifested variety of concepts. Further, reference samples might share multiple similar concepts, making it hard to find the true common theme throughout these samples. At this point, 
a dataset extension can be sensible, as is shown in Supplementary Figure~\ref{fig:appendix:experiments:dataset:extension}. Here, two channels (10 and 130) of intermediate layer \texttt{3.0.conv2} of a ResNet34 model trained on the Caltech-UCSD Birds 200 dataset (see Appendix~\ref{ap:resnet34})  are investigated. As is visible in the extracted reference samples, higher-level concepts often depict whole birds having multiple common concepts. The identification of the key concept might be ambiguous in such a case, or the data might be difficult to interpret for a layman and, in this case, a non-ornithologist. Extending the reference sample set with images from ImageNet results in a higher variety of reference samples, making it easier to identify and interpret the common concepts individually. The extension then yields an alternative and additional view point on the concepts, by extending the search for concept representations in an out-of (yet related) domain dataset. Specifically, with the extended reference sample set, filter 130 can be assigned to a ``dots''-texture ($\sim$ 3px diameter). Further, channel 10 is used for ``red spot'' detection, as is also highlighted using channel-conditional heatmaps for the samples. It is to note, that \gls{xai}-analysis can be done on arbitrary datasets and thus mitigates potential data privacy issues affecting the applicability of our explanation approach.

\begin{figure*}[t]
    \centering
    \includegraphics[width=1\textwidth]{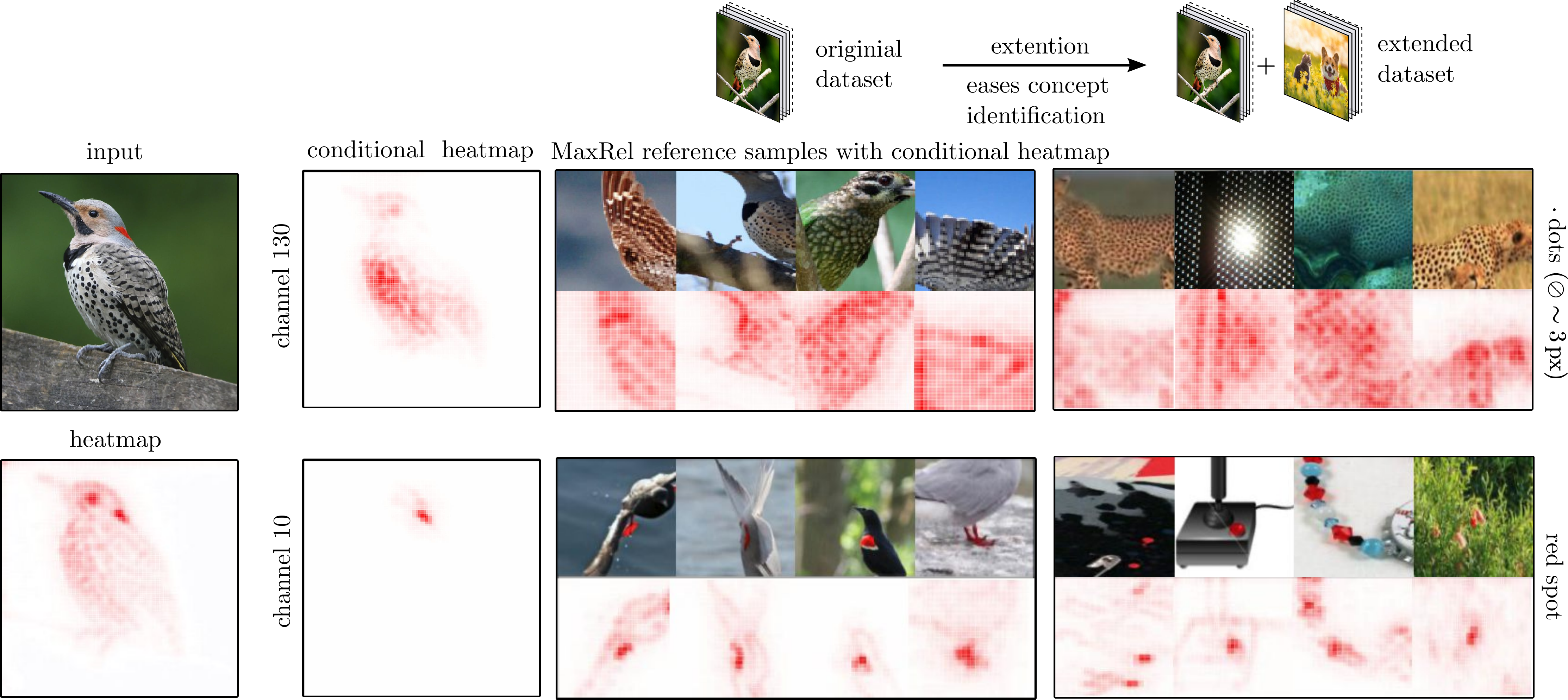}
    
    \caption{Extending the initial reference sample set, \eg, test and training data, with data from a different data domain can make concept understanding easier. A sample set extension yields an alternative and additional view point on the concepts. For the prediction of a ``Northern Flicker'', two relevant channels (10 and 130) of \texttt{layer 3.0.bn2} (Resnet34 model) and their \gls{rmax}-based reference samples from the original Caltech-UCSD Birds 200 dataset (\emph{left}) and with ImageNet extended dataset (\emph{right}) are shown. It might be challenging to comprehend the concept of filters based on the original dataset alone, as image set variety may be limited: Often, birds from the same species with multiple common concepts are shown. However, with thse extended reference sample set, filter 10 and 130 can more easily be assigned to ``$\cdot$ dots'' ($\oslash \sim 3\,$px) and ``red spot'', respectively, as only the relevant concept is shared across reference examples. It is to note, that \gls{xai}-analysis can be done on arbitrary datasets to increase potential data privacy, \eg should the use of the original data be prohibited. %
    }
    \label{fig:appendix:experiments:dataset:extension}
\end{figure*}

\cleardoublepage
\section{Detailed Analysis: Concept Atlas and Concept Composition}
\label{sec:appendix:conceptlocalization}
Attribution maps are ambiguous in the sense that they only show where the model is extracting useful information and not which concepts are being used. Using bird species classification as an example,  input images will generally show various birds that differ in form, texture and color. Heatmaps then usually point to the head or the upper body irrespective of the bird explained, and are in this way almost identical, and not informative (see Supplementary Figure~\ref{fig:appendix:methods:lrp:entangle}b). At this point, in order to  obtain added value from the explanation --- beyond \eg,  ``This bird is a Red Flicker, because \emph{just look at the whole d**n bird!}'' --- and potentially derive knowledge from the model's reasoning, it is important to know whether a particular color, texture, body part shape or relative position of the body parts led to a decision. Using traditional attribution mapping approaches, it can often only be speculated, which concepts are relevant. By visualizing intermediate neuron heatmaps via \gls{crc}, we can assist in concept understanding by localizing individual concepts in heatmaps, as well as reference samples can verify that sensible features have been learned. 

\subsection{From Conditional Heatmaps to Concept Atlases}
\label{sec:appendix:conceptlocalization:atlas}
Here, we present three examples on how channel-conditional heatmaps and concept atlases can aid in the localization and understanding of channel concepts.
\begin{figure*}[h]
    \centering
    \includegraphics[width=.9\textwidth]{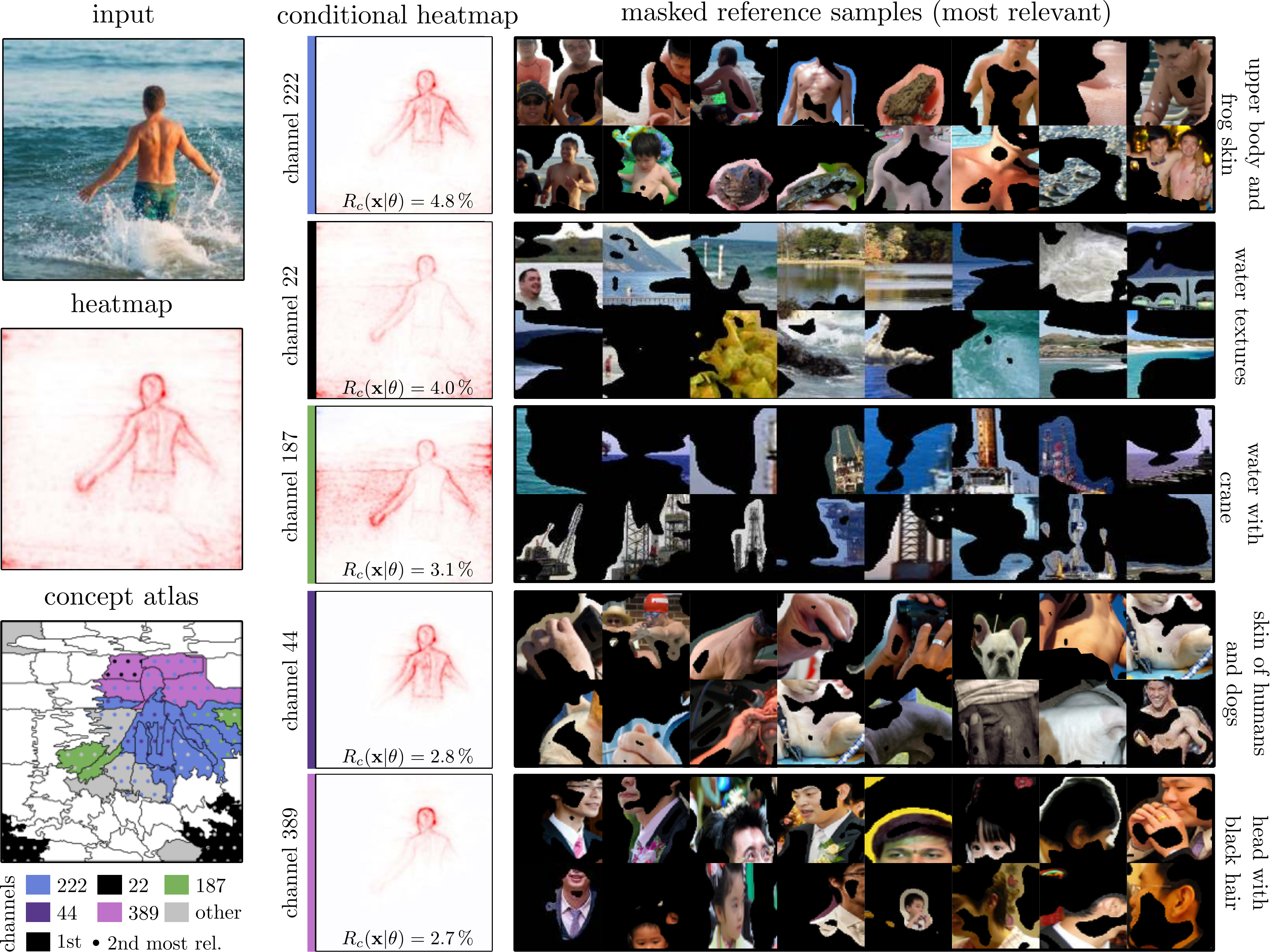}
    \caption[Concept Atlas of a sample of Swimming Trunk class.]{Illustrating the five most relevant filters in layer \texttt{features.28} of VGG-16 trained on ImageNet for the prediction of ``swimming trunk'' with conditional heatmaps and a concept atlas. The input image shows a man standing in the water. \edited{}{Instead of the swimming trunk,} concepts regarding the person's upper body, skin and hair (channels 222, 44 and 389), as well as the water (channels 22 and 187) are primarily used for the prediction --- indicating a \glsdesc{ch} behavior.}
    \label{fig:appendix:localize-concepts:concept-atlas:swimming-trunk}
\end{figure*}
In the first example shown in Supplementary Figure~\ref{fig:appendix:localize-concepts:concept-atlas:swimming-trunk}, class  ``swimming trunk'' is predicted \edited{}{by a VGG-16 model trained on ImageNet. The sample shows a man standing in the water.} CRP-based analysis shows here, that concepts regarding the person's upper body, skin and hair (channels 222, 44 and 389), as well as the water (channels 22 and 187) are primarily used for the prediction.
This is a sign of unreliability, as no concepts regarding the swimming trunk directly seem to be used.

\begin{figure*}[h]
    \centering
    \includegraphics[width=.9\textwidth]{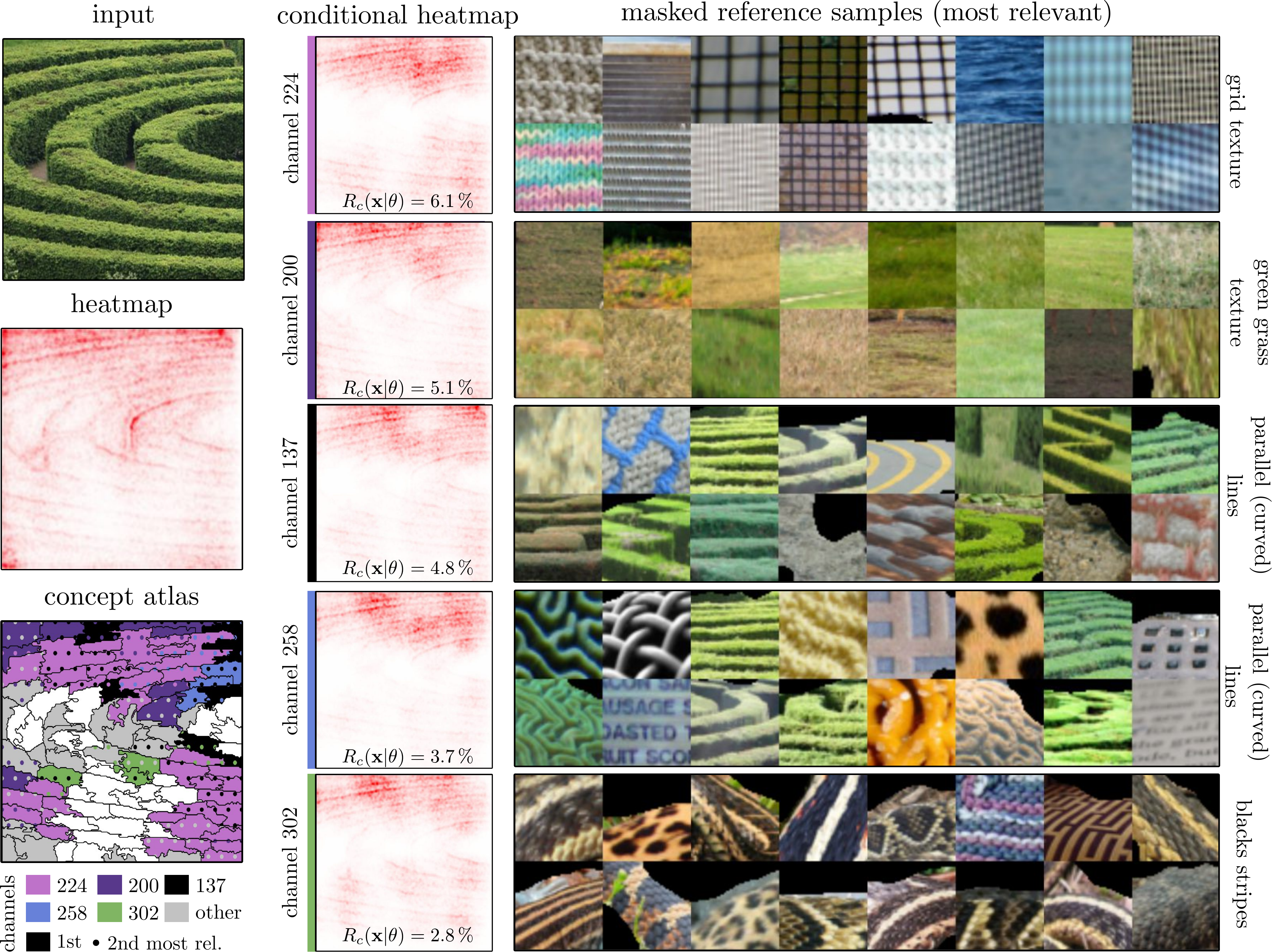}
    \caption{
    Illustrating the five most relevant filters in layer \texttt{features.21} of VGG-16 trained on ImageNet for the prediction of ``maze, labyrinth'' with conditional heatmaps and a concept atlas. The input image shows a maze consisting of green bushes. The model seems to perceive the pathways in the maze through stripe/parallel line concepts \edited{}{(channels 224, 137, 258, and 302).} Also, the fact, that the maze is made up of vegetation seems to be important, indicated by a 'green grass texture' concept of channel 200.
    Looking at the \gls{lrp} heatmap, however,
    it becomes clear, that also other features targeting the bottom left and right corner or the trees in the background are relevant.
    Here, it would be sensible to group similar concepts, or specifically retrieve the most relevant channels for these image regions, as later presented in Supplementary Figure~\ref{fig:appendix:localize-concepts:concepts-in-region0}.
    }
    \label{fig:appendix:localize-concepts:concept-atlas_labyrinth}
\end{figure*}

Another example of a glocal \gls{crc} explanation is shown in Supplementary Figure~\ref{fig:appendix:localize-concepts:concept-atlas_labyrinth},
where a maze is correctly predicted by a VGG-16 model trained on ImageNet.
The maze is made up of green bushes, and the paths can be seen as dark shadows.
For the top-5 most relevant concepts, concept-conditional heatmaps as well as masked reference samples are shown.
It can be seen that four concepts (\eg \edited{}{channels 224, 137, 258, and 302)} target the maze's pathways, as their reference samples visualize stripes and parallel lines.
This can be further verified by the heatmaps and concept atlas, which localize the concepts in the input image.
Channel 200 corresponds here to a green grass texture, which can be found in the green bushes of the maze.
It is to note,
that the \gls{lrp} heatmap indicates,
that also other areas in the input are used by the model, \eg, the vegetation at the top and the lower left and right corner.
Thus,
it might also be sensible to investigate further concepts to understand the model behavior/concepts corresponding to these image parts.

\begin{figure*}[h]
    \centering
    \includegraphics[width=.9\textwidth]{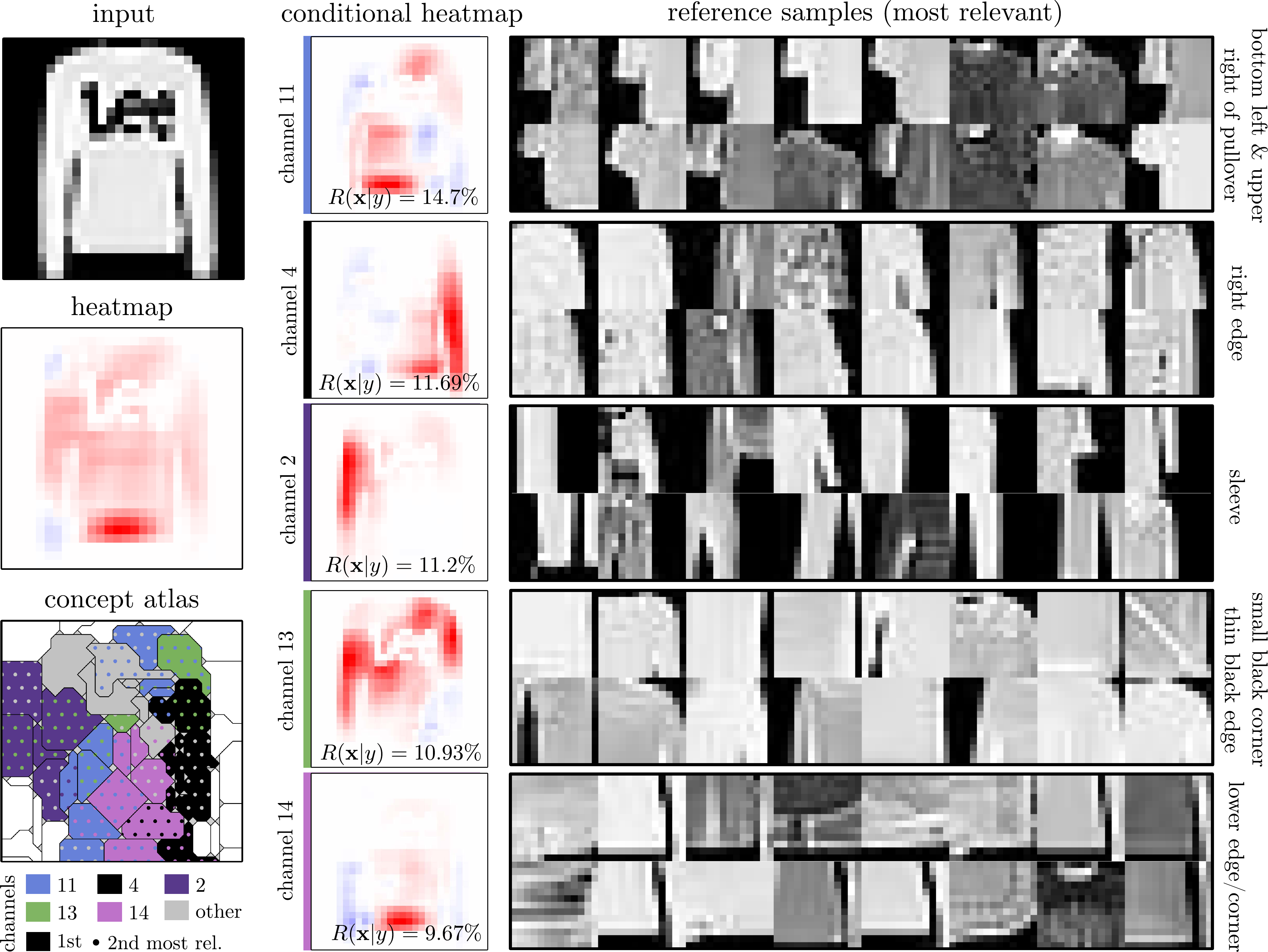}
    \caption[Concept Atlas of a sample of Pullover class.]{Illustrating the five most relevant filters in layer \texttt{l3} of a LeNet model trained on Fashion MNIST for the prediction of class ``pullover''. From the conditional heatmaps and concept atlas it can be seen, that the model has learned concepts for specific spatial parts of the pullover. Channels 4 and 13 activate on the right side of the pullover, whereas channel 2 localizes the left sleeve. Further, channel 14 is used to detect the bottom part of the pullover, and channel 11 for the shoulder region and bottom left.}
    \label{fig:appendix:localize-concepts:concept-atlas:lenet}
\end{figure*}

In a third example shown in Supplementary Figure~\ref{fig:appendix:localize-concepts:concept-atlas:lenet}, class ``pullover'' is predicted by a LeNet-5 model trained on Fashion-MNIST. From the conditional heatmaps and concept atlas it can be seen, that the model has learned concepts for specific spatial parts of the pullover. Channels 4 and 13 activate on the right side of the pullover, whereas channel 2 localizes the left sleeve. Further, channel 14 is used to detect the bottom part of the pullover, and channel 11 for the shoulder region and bottom left. In the previously discussed examples involving concept atlases, concept rankings have been computed for larger input region super-pixels computed from image statistics. Alternatively to using super-pixels, a concept atlas might also be constructed for single pixels, as shown in Supplementary Figure~\ref{fig:appendix:localize-concepts:concept-atlas-comparison}. Here, the first and second-largest relevance scores over the channels are computed for individual pixels. Then, the resulting regions can be color coded according to the largest and second-largest channel importance.
\begin{figure*}[h]
    \centering
    \includegraphics[width=0.91\textwidth]{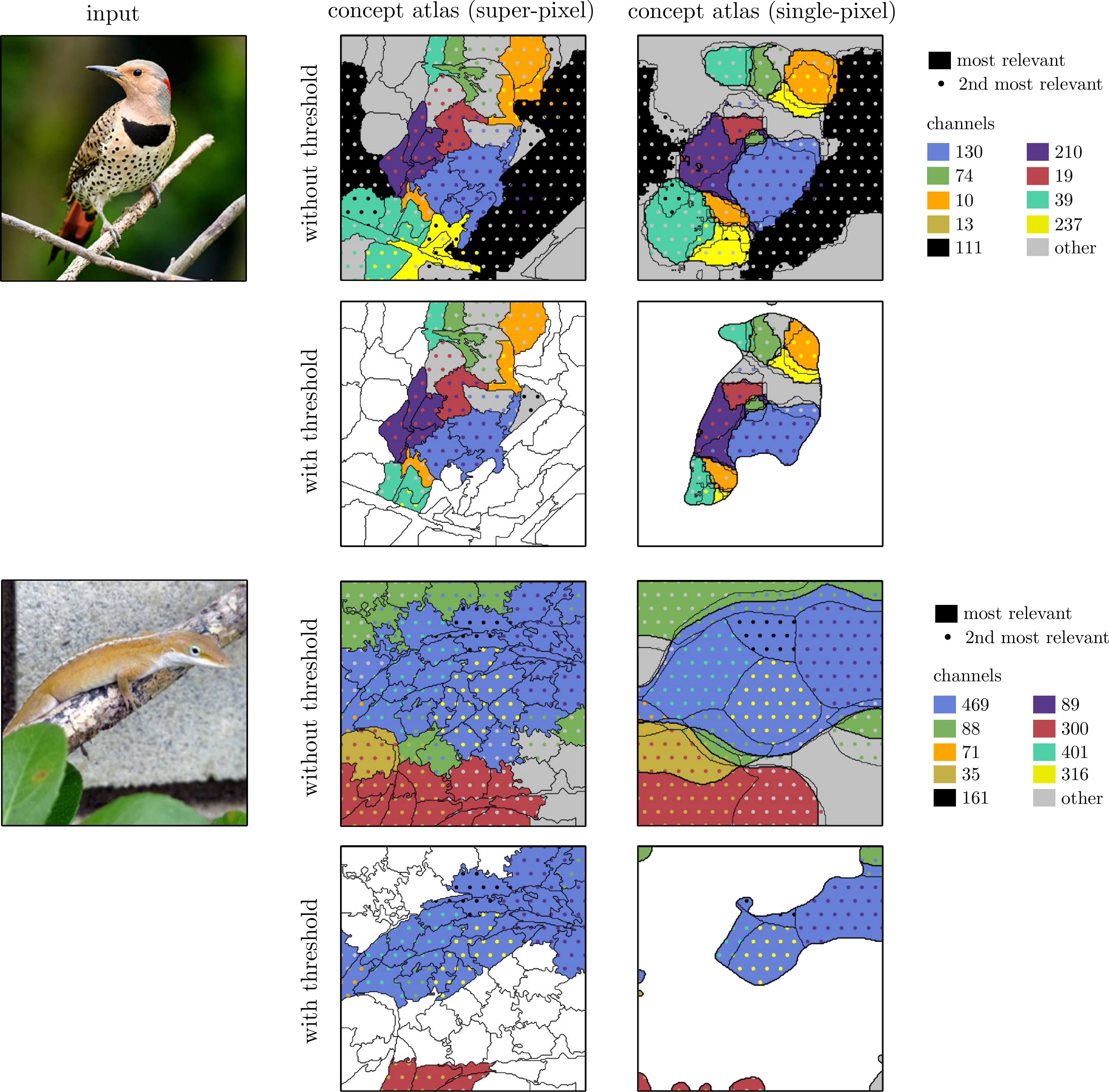}
    \caption{Alternatively to using super-pixels based on image statistics, a concept atlas can also be constructed from single pixels (\ie, single-pixel sized superpixels). For both cases, concept atlases can be filtered with a threshold targeting relevance density, in order to focus on the most relevant input regions. Shown are concept atlases with super-pixels (\emph{center}) using the same samples as shown in Supplementary Figures~\ref{fig:experiments:localize-concepts:concept-atlas} and \ref{fig:appendix:localize-concepts:concept-atlas}, but using a higher number of channels (9 compared to 5) and single-pixels as well (\emph{right}). The single-pixel concept atlases can resolve smaller-scale features. This might however result in noisy atlases if a lot of concepts are used or relevance heatmaps are grainy. Furthermore, the difference of applying thresholding is shown. Here, a threshold results in more localized concept atlases, focusing on the detected objects themselves.}    \label{fig:appendix:localize-concepts:concept-atlas-comparison}
\end{figure*}
Considering only relevant super-pixels for improving visualization clarity, we further introduced a threshold on the relevance density of super-pixels. Specifically, the threshold is set to one tenth of the maximal relevance density. Here the density is calculated for super-pixels as the mean relevance, \ie, the sum of relevance inside a super-pixel divided by the number of pixels. For single pixels, the density is approximated using a Gaussian filter with kernel size of $29\times29\,$px and standard deviation of 4.7. The effect on concept atlas readability and clarity is shown in Supplementary Figure~\ref{fig:appendix:localize-concepts:concept-atlas-comparison}. As can be seen in the figure, the concept atlases with threshold stronger focus on the detected objects, where most relevances are located. It is to note, that densities instead of relevance values are used for the threshold, thereby addressing the issue, that concepts can be similarly important but spatially expanded to different degree. Using relevance values, overall relevant, but spatially expanded concepts are more likely to be hidden, as individual relevance scores are lower.

\subsection{Localized Analysis of Model Predictions}
\begin{figure*}[h]
    \centering
    \includegraphics[width=1\textwidth]{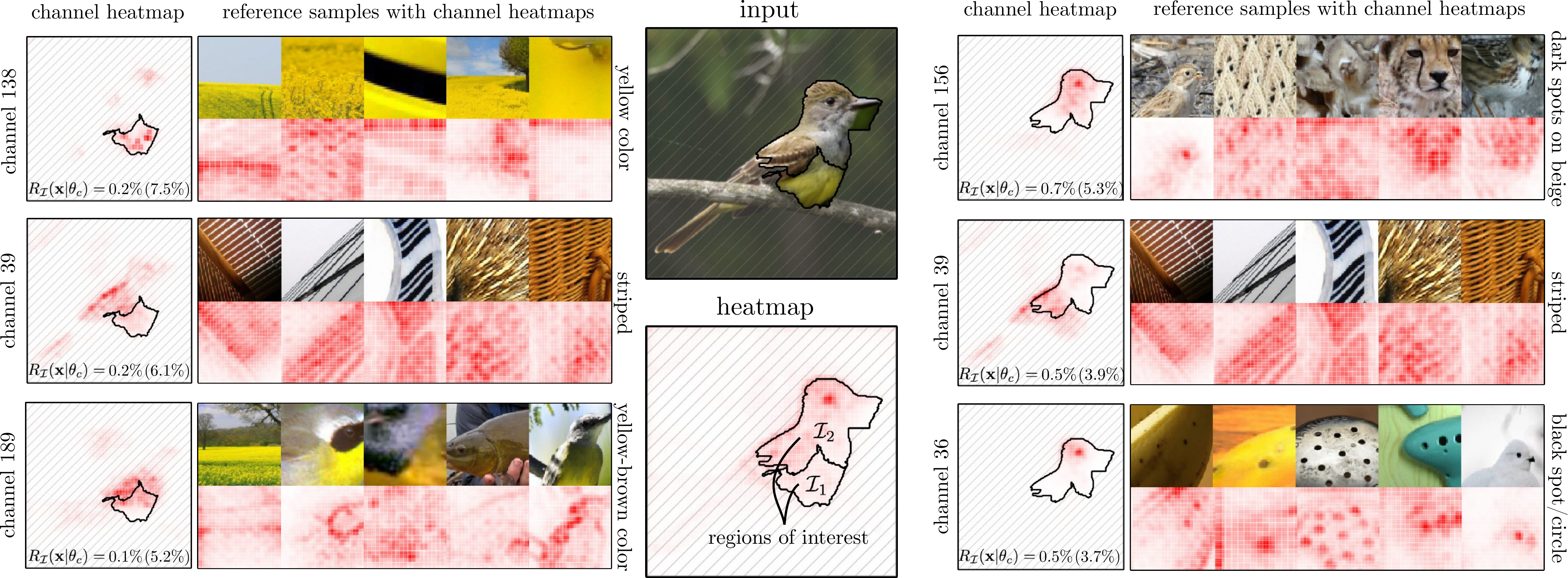}
    \caption[Relevance aggregation inside superpixels allows investigation of the most relevant channels for chosen regions of interest. Sample of \edited{}{Great Crested Flycatcher} class is visualized]{Relevance aggregation inside superpixels allows investigation of the most relevant channels for chosen regions of interest $\mathcal{I}_i$. The top-3 most relevant channels for the prediction of class ``Great Crested Flycatcher'' regarding two superpixels are shown with channel heatmaps and reference samples. (\emph{Left}): Region $\mathcal{I}_1$ covers the bird's belly. Here, ``yellow color'' and concepts for the stripe-like structure of the feathers are most relevant. For each channel, the relevance relative to the global relevance as well as superpixel relevance (in parentheses) is given. (\emph{Right}): Region $\mathcal{I}_2$ covers the bird's eyes and beak. Here, again, the stripe-like texture is important. Further, concepts regarding the round black eye of the Great Crested Flycatcher are relevant.}
    \label{fig:appendix:localize-concepts:concepts-in-region}
\end{figure*}

\begin{figure*}[h]
    \centering
    \includegraphics[width=1\textwidth]{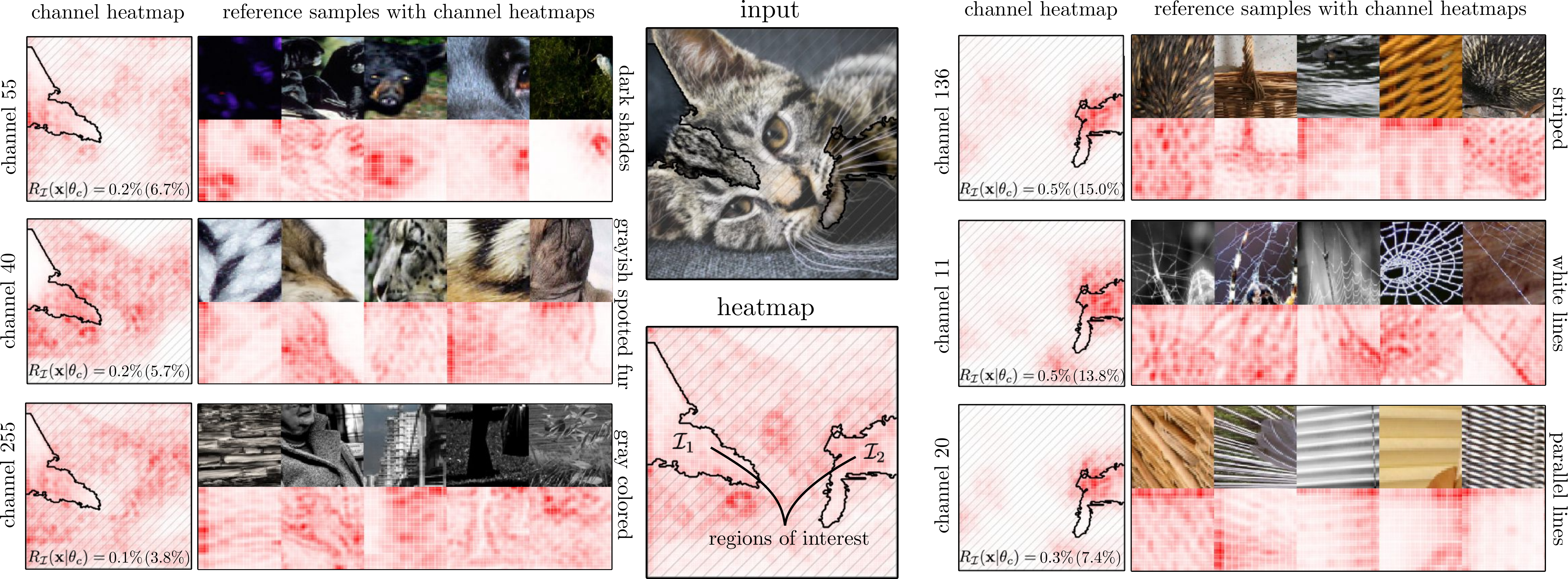}
    \caption[Relevance aggregation inside superpixels allows investigation of the most relevant channels for chosen regions of interest. A sample of tabby cat class is visualized]{Relevance aggregation inside superpixels allows investigation of the most relevant channels for chosen regions of interest $\mathcal{I}_i$. The top-3 most relevant channels for the prediction of class ``tabby cat'' regarding two superpixels are shown with channel heatmaps and reference samples.  (\emph{Left}): Region $\mathcal{I}_1$ covers the cat's head. Here, color-concepts such as ``dark shades'' and ``gray colored'' describing the fur pattern are most relevant. For each channel, the relevance relative to the global relevance as well as superpixel relevance (in parentheses) is given. (\emph{Right}): Region $\mathcal{I}_2$ covers part of the cats's whiskers. Here, the ``striped'', ``white lines'' and ``parallel lines'' concepts capturing the fine white whiskers are most relevant.}
    \label{fig:appendix:localize-concepts:concepts-in-region2}
\end{figure*}

Generally, the meaningful aggregation of relevances allows locally investigating and comparing the most relevant channels or concepts inside different chosen image regions. These regions might depict completely unrelated visual features individually, as is shown in Supplementary Figure~\ref{fig:appendix:localize-concepts:concepts-in-region0}. While concept atlases explain the model's reasoning in a top-down manner, \ie, via the identification and localization of relevant concepts, localized analyses can serve a complementary purpose: By specifying input regions of interest, we can ask the model which particular features it has been influenced by within a constrained set of input dimensions. Defining a region of interest in input space, conditional relevances from \gls{crc} can be aggregated channel-wise in the region and thereafter sorted regarding their relevance value, as discussed above in the context of Concept Atlases. Thus, it is possible, to extract the most relevant channels and their concepts for the region. In the following, we present three examples for analyzing localized model behavior leveraging conditional relevance computation using \gls{crc} in Supplementary Figures~\ref{fig:appendix:localize-concepts:concepts-in-region0},~\ref{fig:appendix:localize-concepts:concepts-in-region} and~\ref{fig:appendix:localize-concepts:concepts-in-region2}. 

Here, the top-3 most relevant channels for two super-pixels in the sorted set $\mathcal{B}_k$, as defined in Equation~\eqref{eq:appendix:localize-concepts:most_rel_channel_set}, are visualized for a local analysis. In the example in Supplementary Figure~\ref{fig:appendix:localize-concepts:concepts-in-region0}, a stork from the ImageNet dataset is classified using a ResNet34 model. One superpixel contains the stork's beak, the other superpixel the stork's right wing. Regarding the beak in $\mathcal{I}_1$, the top-3 most relevant channels located in layer \texttt{layer3.0.conv2} focus on color \edited{}{(``red spot'', ``red-orange'' and ``red')}. Regarding the wing  in $\mathcal{I}_2$ on the other hand, the most relevant channels focus on textures, such as \edited{}{``parallel lines'', ``finely lined'' and ``wooly''}.

\begin{figure*}[t]
    \centering
    \includegraphics[width=1\textwidth]{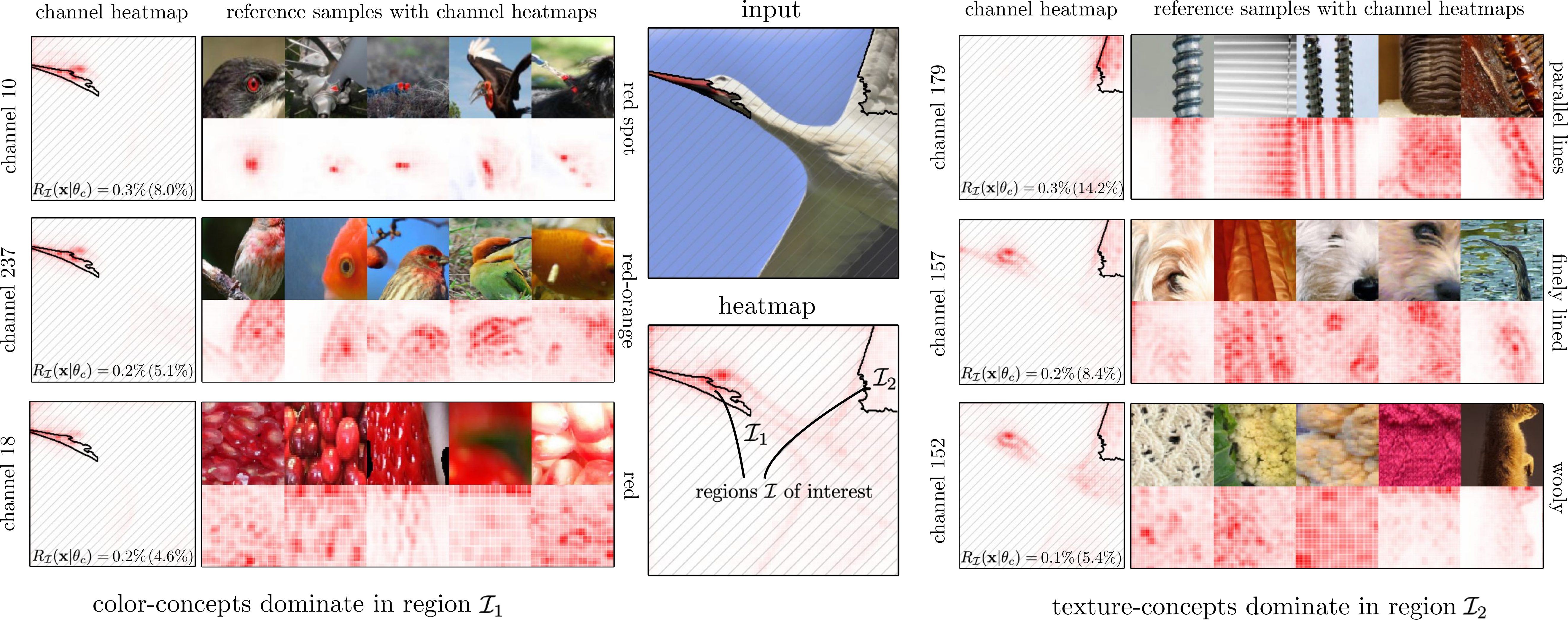}
    \caption{Relevance aggregation inside superpixels allows investigating the locally most relevant channels for chosen regions of interest $\mathcal{I}_k$. The top-3 most relevant channels for the prediction of class \edited{}{``white stork''} regarding two superpixels are shown with channel heatmaps and  reference samples. (\emph{Left}): Region $\mathcal{I}_1$ covers the stork's beak. Here, color-concepts, such as \edited{}{``red spot'', ``red-orange'' and ``red''} are most relevant. For each channel, the relevance relative to the global relevance as well as superpixel relevance (in parentheses) is given. (\emph{Right}): Region $\mathcal{I}_2$ covers the stork's right wing. Here, texture-concepts, such as \edited{}{``parallel lines'', ``finely lined'' and ``wooly''} are most relevant.}
    \label{fig:appendix:localize-concepts:concepts-in-region0}
\end{figure*}

In the second example shown in Supplementary Figure~\ref{fig:appendix:localize-concepts:concepts-in-region} a ``Great Crested Flycatcher'' is predicted by a ResNet34 model trained on the CUB birds dataset. 
Here, the channels of layer \texttt{layer3.0.conv1} are analyzed. 
The first analyzed region $\mathcal{I}_1$ covers the bird's belly (\emph{left}). 
Here, ``yellow color'' and concepts for the stripe-like structure of the feathers are most relevant. 
The second region $\mathcal{I}_2$ covers the bird's eyes and beak. Here, again, ``blue color'' is most important. 
Further, concepts regarding the round black eye of the Great Crested Flycatcher are  relevant.

A ``tabby'' cat is predicted by a ResNet34 model pretrained on ImageNet in the third example in Supplementary Figure \ref{fig:appendix:localize-concepts:concepts-in-region2}. The analyzed layer is \texttt{layer3.0.conv2}. In the first region $\mathcal{I}_1$, concepts for the cat's head are investigated. Here, color-concepts such as ``dark shades'' and ``gray colored'' describing the fur pattern are most relevant. 
The second region $\mathcal{I}_2$ covers the cats's whiskers. 
Here, the ``striped'', ``white lines'' and ``parallel lines'' concepts capturing the fine white whiskers are most relevant.

\subsection{Understanding Hierarchical Concept Composition}
\label{sec:appendix:experiments:local:flow}

\begin{figure*}[t]
    \centering
    \includegraphics[width=1\textwidth]{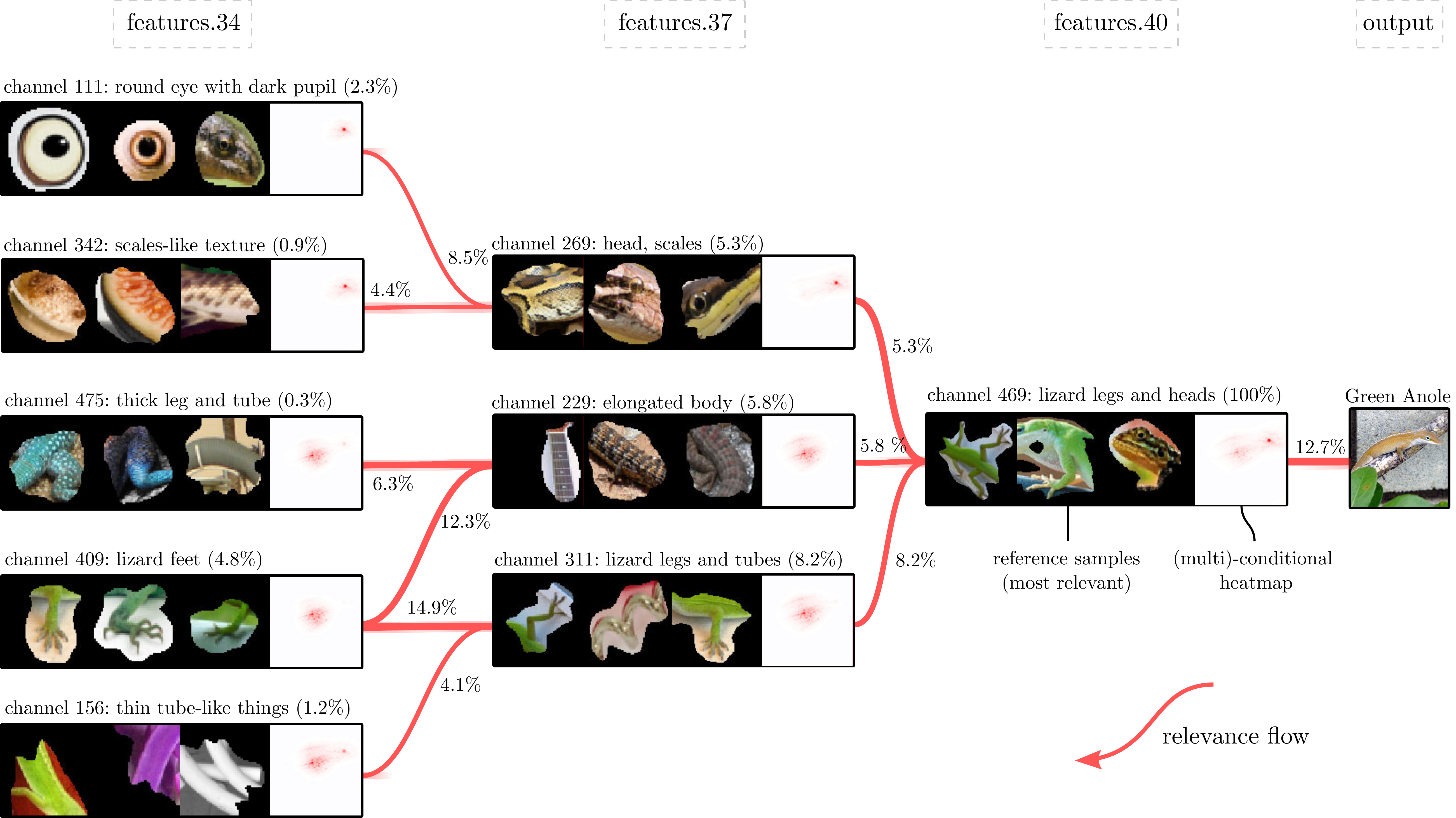}
    \caption{Attribution Graphs can be used to visualize relevant concepts in lower layers for a concept of interest regarding a particular prediction outcome, and thus improve concept understanding. Shown are relevant (sub)-concepts in \texttt{features.34} and \texttt{features.37}
    for concept \edited{}{``lizard legs and heads''} in \texttt{features.40} of a VGG-16 BN model
    trained on ImageNet for the prediction of class \edited{}{``Green Anole''.} See Supplementary Figure~\ref{fig:appendix:localize-concepts:concept-atlas} for a spatial concept composition analysis of the same sample.
    The relevance flow is highlighted in red, with the relative percentage of relevance given that flows to the lower-level concept. For each concept, the channel is given with the relative overall relevance score (\wrt channel 469 in \texttt{features.40}) in parentheses. Following the relevance flow, \edited{}{the concept is dependent on channels encoding for heads, eyes, scales, legs, thin and thick tube-like objects.}
    }
    \label{fig:appendix:experiments:flow}
\end{figure*}  

With the selection of  a specific neuron or concept, \glsdesc{crc} allows to investigate, how relevance flows from and through the chosen network unit to lower-level neurons and concepts. This gives information about which lower-level concepts carry importance for the concept of interest and how it is composed of more elementary conceptual building blocks, which may further improve the understanding of the investigated concept and model as a whole. In Supplementary Figure~\ref{fig:appendix:experiments:flow}, we visualize and analyze the backward flow of relevance scores. The graph-like visualization reveals how concepts in higher layers are composed of lower layer concepts. Here, we show the top-3 concepts influencing our concept of choice, the \edited{}{``lizard legs and head''} concept encoded in \texttt{features.40} of a VGG-16 BN model trained on ImageNet. Edges in red color indicate the flow of relevance \wrt to the particular sample from class \edited{}{``Green Anole''} shown to the very right between the visualized filters with corresponding examples and (multi)-conditional heatmaps. The width of each red edge describes the relative strength of contribution of lower layer concepts to their upper layer neighbors. The conditional relevance $R^{l}(\x|\theta=\{l:\{c_0\}\})$ for a concept $c_0$ \edited{}{(``lizard legs and head''} in Supplementary Figure~\ref{fig:appendix:experiments:flow}) in layer $l$ can be propagated layer by layer until the input space is reached. At each lower-level layer, the computed attribution scores of neurons and filters then describe their respective relevance to the concept $c_0$ for the prediction on $\x$. Now, the top $k$ filters at each layer can be illustrated. While this approach depicts the layer-wise contribution to concept $c_0$, no disentanglement in lower layers is achieved as the relevance in layers lower than $l-1$ is sum-aggregated together and no accurate hierarchical flow of relevance is obtained, which, however, is possible in principle via the evaluation of multiple backward passes specifically conditioned to lower-layer concepts. Precisely, it is not possible to obtain the sole contribution of a concept $c_i$ in layer $l-2$ to a concept in layer $l-1$ in a single backward pass.

To disentangle the attribution graph, we expand on $\theta$ and mask each of the top $k$ filters in the previous layer independently with an additional backward pass. At layer $l$ we set $\theta=\{l:\{c_0\}\}$ and compute an attribution until layer \mbox{$l-1$}. At $l-1$ we iterate over the $k$ most important units and set $\theta=\{l:\{c_0\},~l-1: \{c_1\},~\dots, ~l-1: \{c_k\}\}$. We repeat the steps outlined above for each layer. This way, each lower level concept can be broken down independently and an accurate hierarchical representation achieved. To illustrate the resulting graph, we visualized an attribution flow for filter 469 in layer \texttt{features.40} of a VGG-16 BN model trained on ImageNet given an example image $\x$ showing an instance from the \edited{}{``Green Anole''} class. The filter is specialized in detecting instances of green-colored lizard \edited{}{legs and heads}. This example shows that different paths in the network potentially activate the filter. 
Regarding the lizard prediction, path-specific conditional heatmaps illustrate how the sub-concepts either target the head or the lizard's legs. In summary, we are able to visualize how abstract concepts arise from low-level patterns, and how they are composed specifically to the prediction the flow of relevance as been computed for.

\begin{figure*}[h]
    \centering
    \includegraphics[width=1\textwidth]{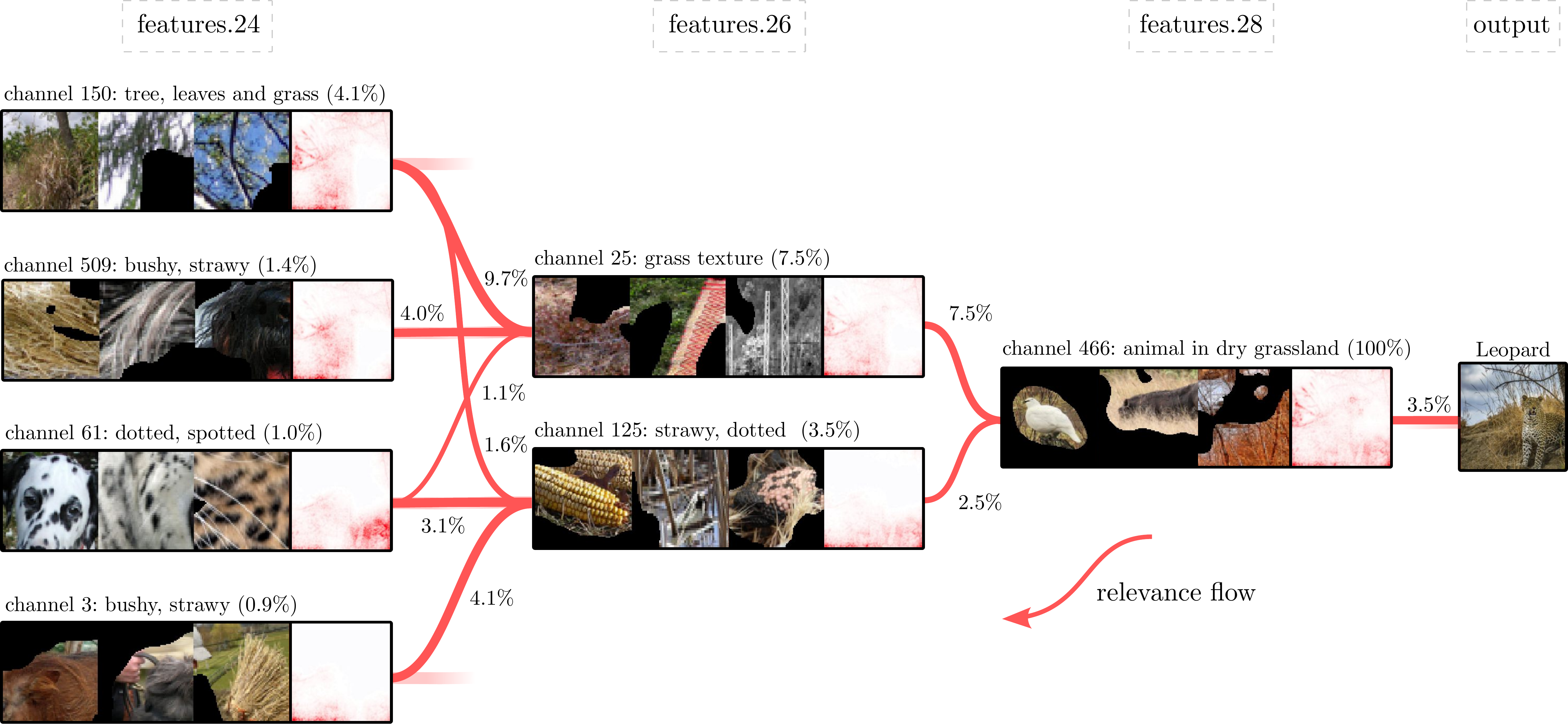}
    \caption{
    Attribution Graphs visualize relevant concepts in lower layers for a concept of interest regarding a particular prediction outcome and thus improve concept understanding. Shown are relevant (sub)-concepts in \texttt{features.24} and \texttt{features.26}
    for concept ``animal in dry grassland'' in \texttt{features.28} of a VGG-16 model
    trained on ImageNet for the prediction of class ``leopard''.
    The relevance flow is highlighted in red, with the relative percentage of relevance given that flows to the lower-level concept. For each concept, the channel is given with the relative overall relevance score (\wrt channel 19 in \texttt{features.28}) in parentheses. Following the relevance flow, concept ``animal in dry grassland'' is dependent on concepts describing the vegetation of the grassland (\eg, ``trees, leaves and grass'' and \edited{}{``bushy, strawy'')}. However,
    part of the upper-level concept ``animal in dry grassland'' also perceives the dotted pattern of the leopard (\eg, \edited{}{``dotted, spotted'')}.
    }
    \label{fig:appendix:experiments:flow_leopard}
\end{figure*}

Another example of an attribution graph is shown in Supplementary Figure~\ref{fig:appendix:experiments:flow_leopard} for concept ``animal in dry grassland'' in \texttt{features.28} of a VGG-16 model trained on ImageNet that is used in a prediction of class ``leopard''.
According to the concept heatmap,
the concept is located on the background vegetation as well as the leopard itself.
Investigating the relevant sub-concepts,
it can be seen that most important for the concept seem to be concepts targeting the vegetation (\eg, ``trees, leaves and grass'' and \edited{}{``bushy, strawy}'').
However,
part of the sub-concepts also perceives the dotted texture of the cheetah (\eg, \edited{}{``dotted, spotted''}).

\subsection{Challenging to Interpret Examples}
\label{sec:appendix:difficult}

In the following,
we will show glocal \gls{crc}- and \gls{rmax}-explanations that are more difficult to understand for humans.
Reasons for low interpretability can include (1) redundant/similar concepts, (2) concepts with low human interpretability as they are very abstract, 
not well defined or poly-semantic, (3) limited quality of feature visualization (e.g. a limited reference sample variety), or (4) ambiguous conceptual localization through heatmaps.
We invite the interested reader to explore further challenging and interesting cases on arbitrary input images via the tutorial notebooks provided with \url{https://github.com/rachtibat/zennit-crp}.

\begin{figure*}[h]
    \centering
    \includegraphics[width=.9\textwidth]{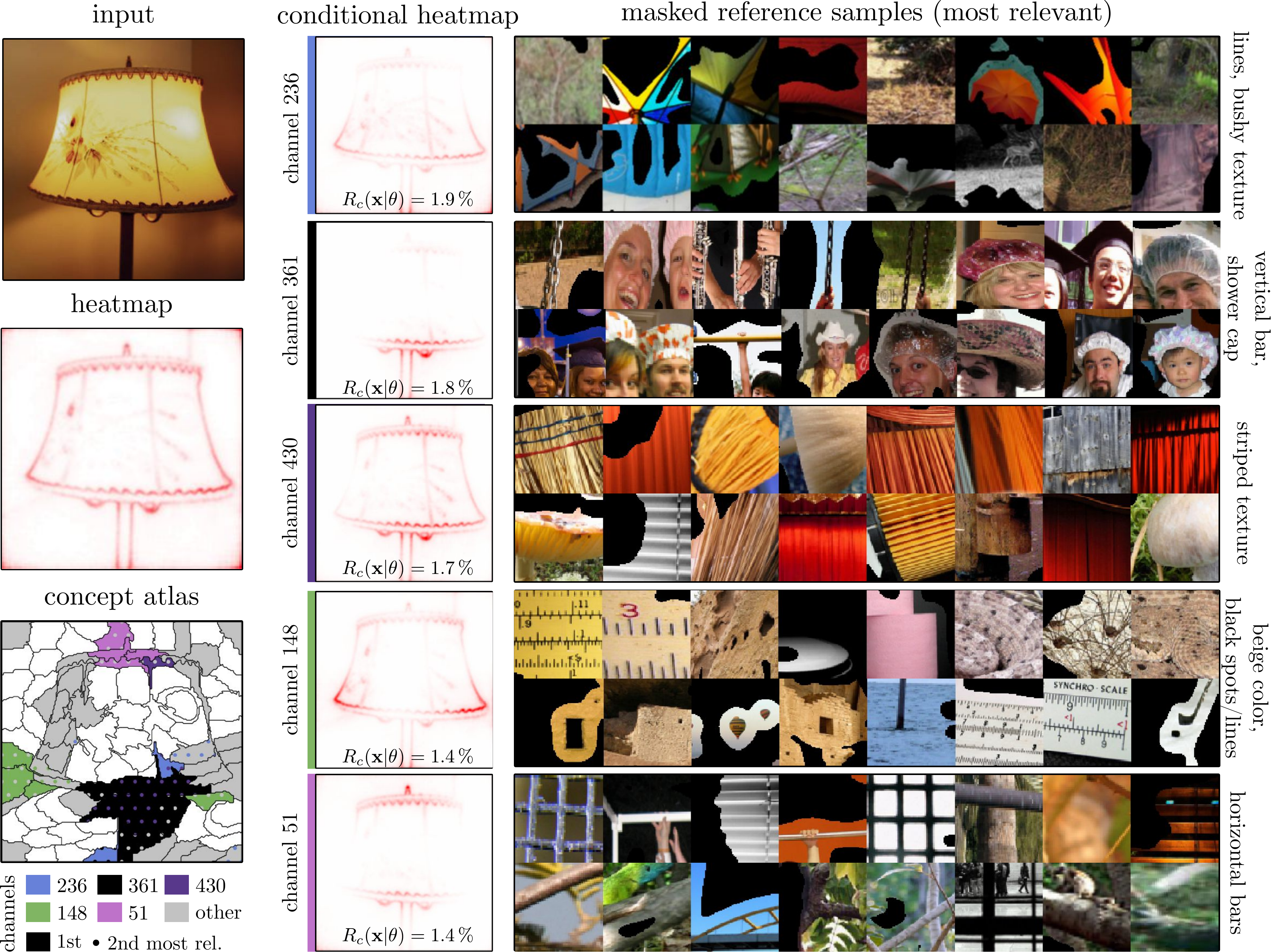}
    \caption{
    Glocal CRP explanation that is challenging to interpret.
    Shown are the top-5 most relevant concepts in layer \texttt{features.26} of VGG-16 model trained on ImageNet for the prediction of class ``lamp shade''.
    Here, channels 236 and 430 seem to address the brushy pattern and warm colors of the lampshade.
    Further,
    channel 148 might address the texture of the lower edge of the lamp,
    whereas channels 51 might directly perceive the edge itself.
    However, the concept visualizations are not very clear, and the channel-conditional heatmaps do not clearly pinpoint the location.
    }
    \label{fig:appendix:localize-concepts:concept-atlas_bad1}
\end{figure*}

A first example is shown in Supplementary Figure~\ref{fig:appendix:localize-concepts:concept-atlas_bad1},
where a \edited{}{``lamp shade''} is classified using a VGG-16 model trained on ImageNet.
Here, the top-5 most relevant concepts in layer \texttt{features.26} are shown with concept-conditional heatmaps and masked reference samples.
Understanding the explanation is difficult,
as some concept visualizations (\eg channels 236, \edited{}{430, 148)} cannot be clearly localized to features of the lamp.
Channels \edited{}{236, 430 and 148} might address the texture of the bottom edge of the lamp
or the brushy texture of the shade. 
The concept-conditional heatmaps indicate that they might be relevant for both the texture of shade and the edge.
At this point,
looking at relevant sub-concepts can help in learning more about how these concepts are actually used, as in Section~\ref{sec:appendix:experiments:local:flow}.

\begin{figure*}[h]
    \centering
    \includegraphics[width=.9\textwidth]{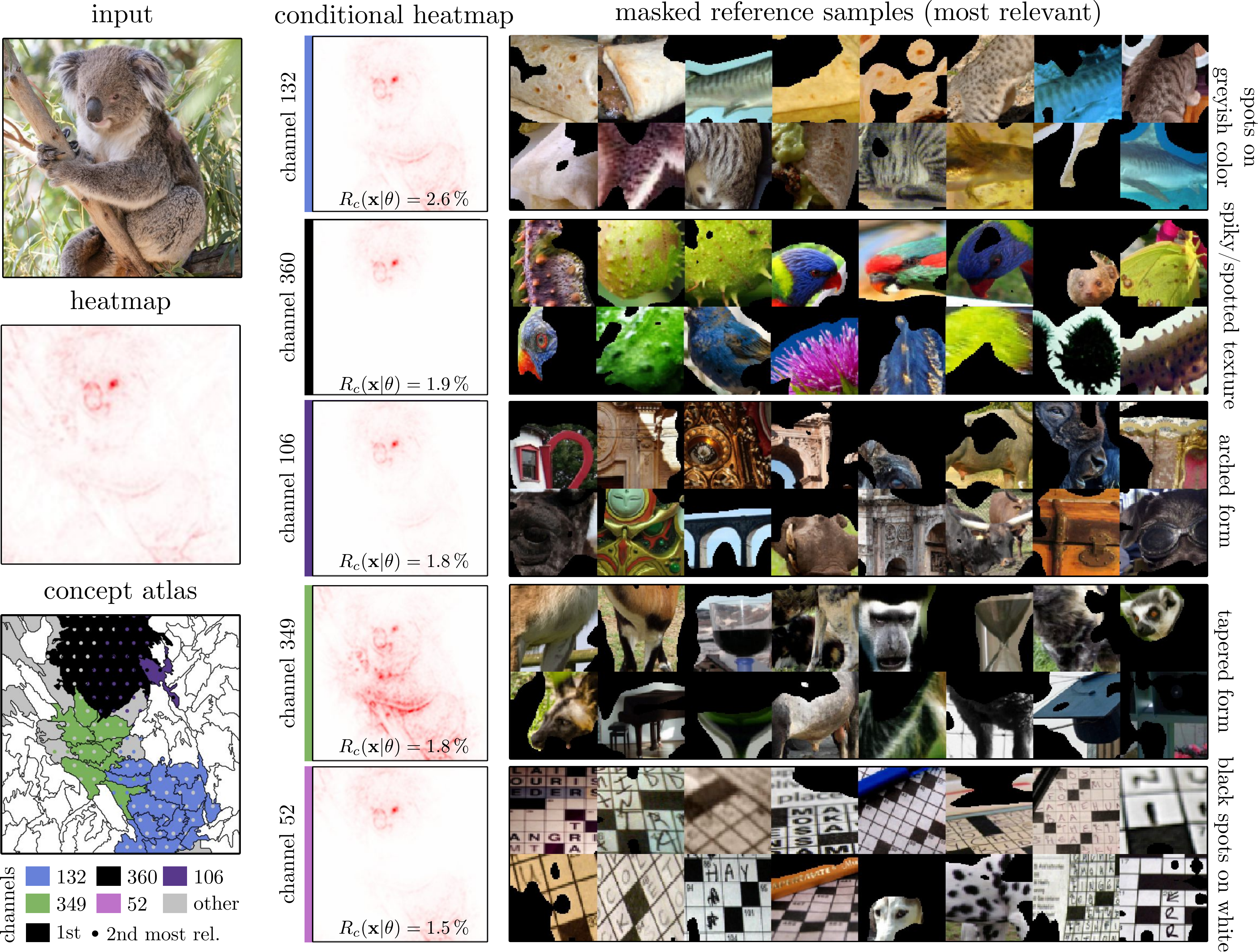}
    \caption{Glocal CRP explanation that is challenging to interpret. 
    Shown are the top-5 most relevant concepts in layer \texttt{features.26} of VGG-16 model trained on ImageNet for the prediction of class ``koala''.
    Here, 
    several channels (132, 360, 106, 52) are located on the koala's face. Their actual function or concept is not obvious \wrt the koala.
    Concretely,
    concepts 132, 360 and 52 might correspond to both the fur texture and the eye.
    Further, concept 349 seems to represent a specific tapered form. However,
    how this applies to the koala is not obvious.}
    \label{fig:appendix:localize-concepts:concept-atlas_bad2}
\end{figure*}

A second example is shown in Supplementary Figure~\ref{fig:appendix:localize-concepts:concept-atlas_bad2},
where a ``koala'' is classified using the same VGG-16 model trained on ImageNet.
Here, the top-5 most relevant concepts in layer \texttt{features.26} are shown with concept-conditional heatmaps and masked reference samples.
Understanding the explanation is difficult,
as several channels (132, 360, 106, 52) are located on the koala's face, but their actual function or concept is not obvious \wrt the koala.
Further, concept 349 seems to represent a specific tapered form.
However, how this applies to the koala is not clearly communicated.

\begin{figure*}[h]
    \centering
    \includegraphics[width=.9\textwidth]{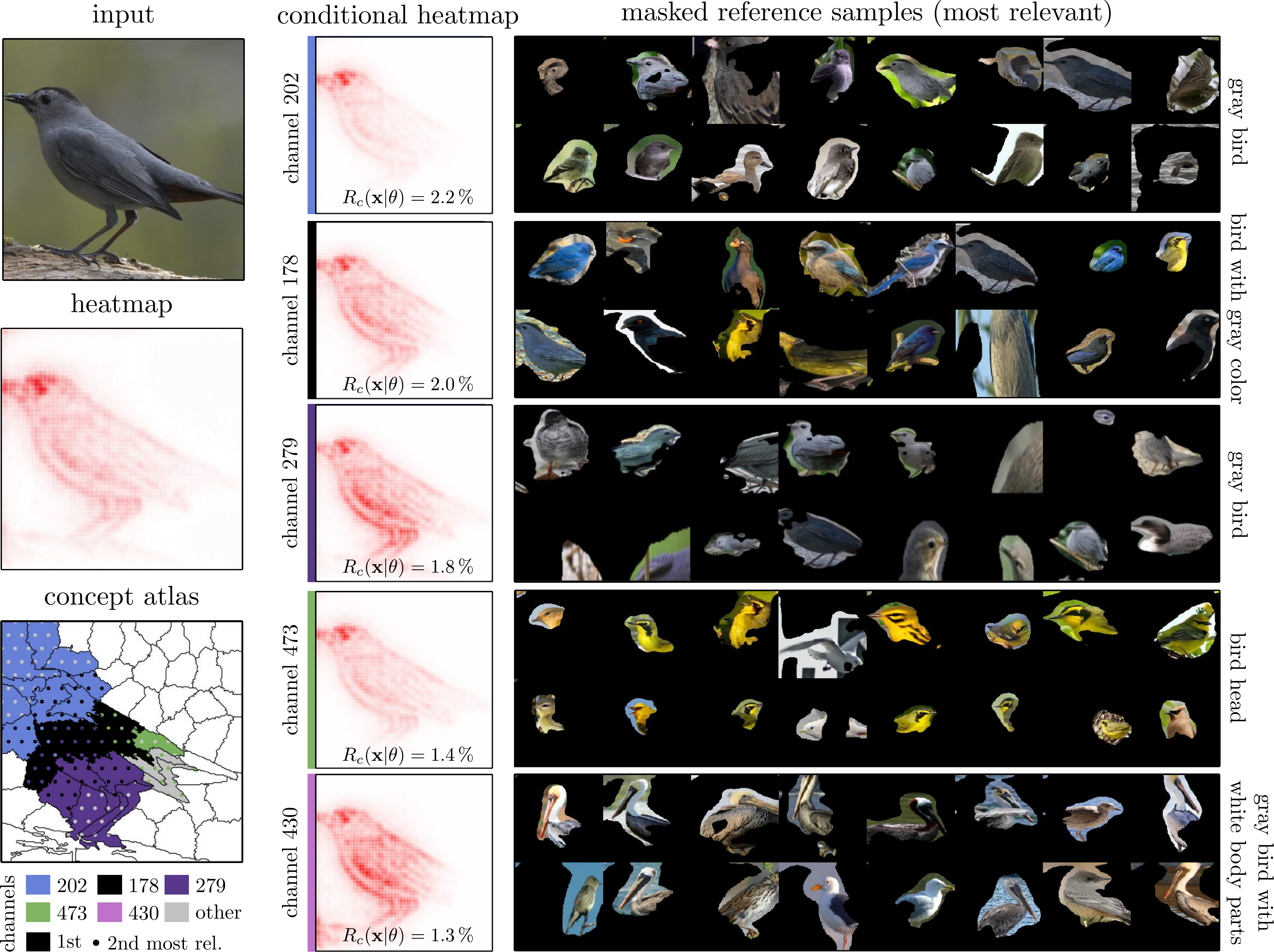}
    \caption{Glocal CRP explanation that is challenging to interpret.
    Shown are the top-5 most relevant concepts in layer \texttt{4.0.conv2} of Resnet-34 model trained on the CUB-200-2011 birds dataset for the prediction of class ``Gray Catbird''.
    Whereas channels 279 and 202 focus on the gray color of the bird, the other channels' concepts are harder to understand.
    It is not easy to see, whether channels 473 and 178 target special types of birds' heads.
    Further, channel 430 seems to perceive gray birds, but also straight lines, as in the reference samples are often shown birds with long and straight beaks.
    }
    \label{fig:appendix:localize-concepts:concept-atlas_bad3}
\end{figure*}

A third example is shown in Supplementary Figure~\ref{fig:appendix:localize-concepts:concept-atlas_bad3},
where a ``Gray Catbird'' is classified using a Resnet-34 model trained on the CUB-200-2011 birds dataset.
Here, the top-5 most relevant concepts in layer \texttt{layer4.0.conv2} are shown with concept-conditional heatmaps and masked reference samples.
The explanation is challenging to understand,
as some concepts (\eg channels 178, 473, 430) have no clear concept visualizations.
This is probably due to the limited reference sample variety (consisting only of birds),
which can be increased by extending the dataset used for the reference images, as discussed in Section~\ref{sec:appendix:experiments:qualitative:variety}.

\cleardoublepage
\clearpage
\section{Walkthrough Demonstration of CRP-based Glocal Explanation Computation and Interpretation}
\label{sec:appendix:workflow}

In this section, we discuss the computation and interpretation of glocal \gls{crc}- and \gls{rmax} explanations.
First,
the computational workflow is introduced and discussed.
Here, all relevant steps for computing explanations --- specifically those for Concept Atlas visualizations (see Section~\ref{sec:appendix:conceptlocalization:atlas})--- including constructing concept atlases or extracting the most relevant concepts in specific regions of the input, are discussed step-by-step.
Thereafter,
we present a guideline on how glocal \gls{crc}-based explanations can be read and understood.

\subsection{Computational Workflow}
\label{sec:appendix:workflow:compute}

The computational steps involved in the generation of glocal \gls{crc}- and \gls{rmax} explanations such as shown in Supplementary Figure~\ref{fig:appendix:localize-concepts:concept-atlas} are outlined in Supplementary Figure~\ref{fig:appendix:workflow:workflow_overview}.
These steps can be roughly divided into a ``pre-processing'' phase (which usually takes minutes to perform), but is only necessary once per trained model, 
and a post-hoc instance-based part (that takes a few seconds). See Supplementary Table~\ref{tab:appendix:workflow:runtime} for exact run-time listings on reference hardware.

\begin{figure*}[h]
    \centering
    \includegraphics[width=.9\textwidth]{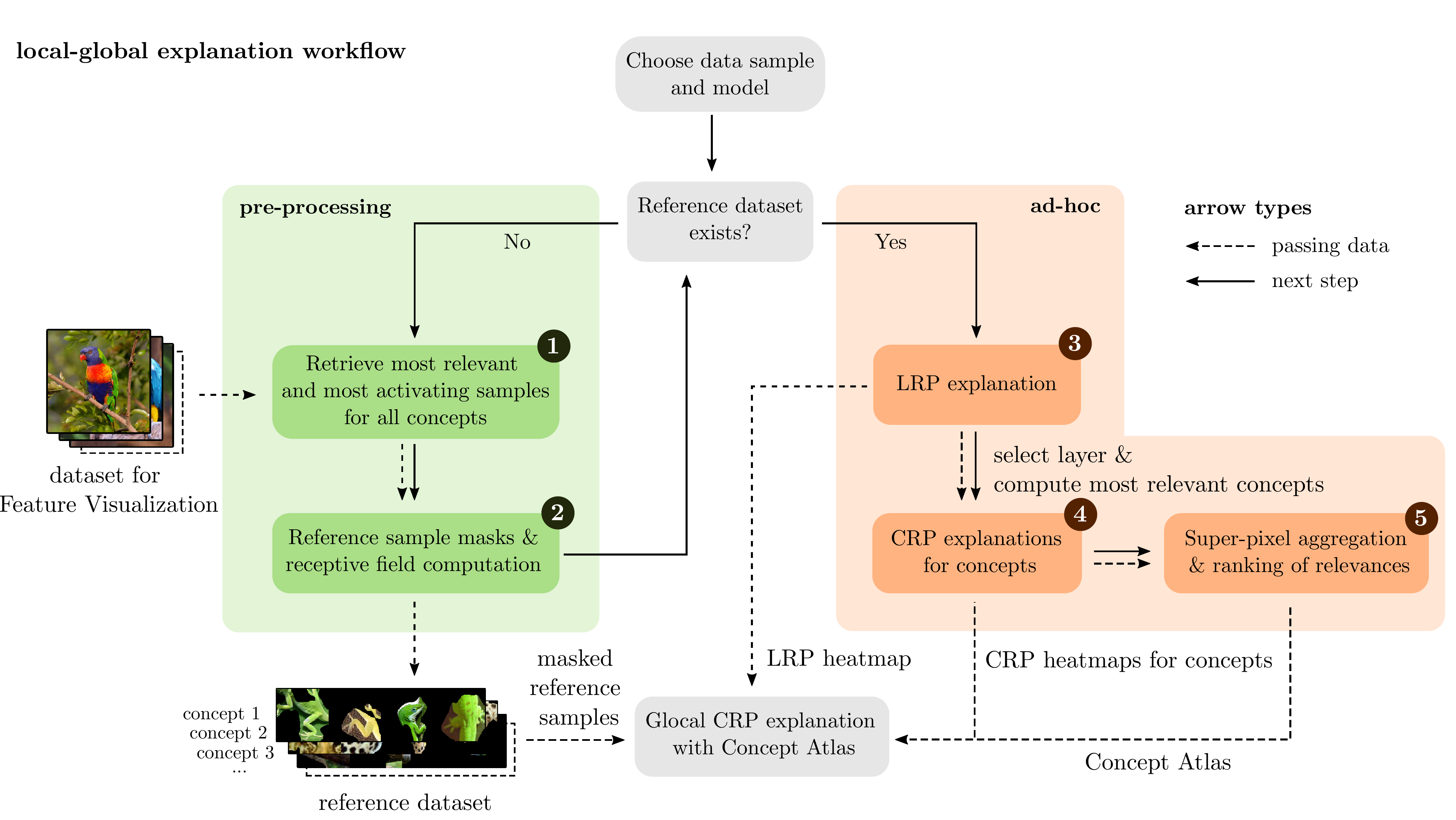}
    \caption{Workflow of generating global-local CRP explanations, as shown in our Concept Atlas visualization. Firstly, a data sample and model is chosen for which the prediction is to be performend and then post-hoc explained. If not existent, a reference dataset of concept visualizations (including masked reference samples for all concepts) has to be computed once (per model and dataset) as a pre-processing step. Regarding pre-processing,
    \textbf{(1)} the most relevant and/or activating samples for all concepts of a network are collected using samples from the dataset (\eg test data).
    Thereafter, \textbf{(2)} the masks for the previously collected reference samples (see Section~\ref{sec:appendix:methodsindetail:understanding:visualization}) and receptive fields of all neurons are computed.
    The data from steps (1) and (2) are stored in a reference dataset, which can then be used for concept visualization in CRP explanations.
    To compute an instance-based explanation for the given data point, \textbf{(3)} \gls{lrp} scores can be can be generated for all hidden- and input unites of the model in a single backward pass.
    Selecting a layer, the most relevant concepts can be extracted automatically using the hidden layer relevance scores from step (3).
    In step \textbf{(4)} by using \gls{crc}, the concept-conditional heatmaps can now be generated for the most relevant concepts.
    The \gls{crc} heatmaps are on the one hand displayed in the final explanation but are used in step \textbf{(5)}, where the concept atlas is computed by locally aggregating and ranking relevance scores.
    Finally,
    the glocal \gls{crc} explanation can be shown, by using the \gls{lrp} and \gls{crc} heatmaps, the concept atlas, and masked reference samples from the reference dataset.
    }
    \label{fig:appendix:workflow:workflow_overview}
\end{figure*}

\paragraph{Reference Dataset Preprocessing}
During pre-processing, the model is analyzed in a global (in case \gls{amax} is used), and potentially global-local (when using \gls{rmax}) fashion.

\textbf{(1)} In the first step, reference samples are sought for each concept encoded in the model. 
For highest granularity and practicability, we assume that each neuron (or filter in convolutional layers) encodes for a single concept, as discussed in~\ref{sec:appendix:methodsindetail:lrp:conceptual} and in the literature (see Section~\ref{sec:appendix:relatedwork:global}). In principle, a concept can also refer to a set of filters (cf. Section~\ref{sec:appendix:experiments:quantitative:clusters}), the directions described by several filters situated in the
same layer \cite{kim2018interpretability} or spanning a concept-defining subspace \cite{vielhaben2022sparse,chormai2022disentangled}. 
In general, \gls{crc} is applicable without restrictions in such a case, \eg, by adequately aggregating the relevance scores over the filters involved. 
Counting each neuron or filter as a concept encoding,
the most activating and relevant samples can be ranked collected for all units in every network layer.
Computationally,
for all samples in a dataset to be used for the purposed of Feature Visualization,
a forward and modified backward pass through the \gls{dnn} is necessary to compute the intermediate neuron activations and relevance scores.
Thereafter, it is a matter of aggregating and sorting of the activation and relevance values as well as sample indices (see Section~\ref{sec:appendix:methodsindetail:understanding:selecting_reference_samples:activation} and Section~\ref{sec:appendix:methodsindetail:understanding:selecting_reference_samples:relevance}).
Thus, the computational cost $\mathcal{O}(s)$ is proportional to the number of samples $s$ (and executed forward-backward-passes) used in this step.

\textbf{(2)} To further enhance interpretability, reference samples can be masked with the receptive field and their respective CRP heatmaps computed \wrt the latent concept-detecting units (see Section~\ref{sec:appendix:methodsindetail:understanding:scaling} and Section~\ref{sec:appendix:methodsindetail:understanding:visualization}). This step can optionally be pre-calculated during preprocessing or computed ad-hoc during the presentation of the concepts. The time complexity $\mathcal{O}(c\cdot r)$ is proportional to the number of concepts $c$ and reference samples $r$ per concept.

\paragraph{Computing Instance-based Post-hoc Explanations}
 
\textbf{(3)} When the user wishes to analyze a sample, a forward (for inference) and modified backward pass (for explaining) is performed conditioned on either the true, the predicted (if different) or any other arbitrarily chosen class to attribute relevance to all concepts in the model (see Section~\ref{sec:appendix:global_concept_importance}). The $N$ most relevant concepts, i.e. filters, are then selected and their corresponding (pre-computed and masked) reference samples displayed. 

\textbf{(4)} In order to localize individual concepts in input space, \gls{crc} is performed for each of the $k$ most relevant concepts, \ie, a forward and modified backward pass for a batch of units conditioned on a class and one or more concepts (see Supplementary Figure~\ref{fig:appendix:disentangle:conceptual}).

\textbf{(5)} For the Concept Atlas visualization, the input space may be divided, \eg, into super-pixels (in an image processing context) using a segmentation algorithm or time windows (in time series processing; see Section~\ref{sec:appendix:timeseries} for an example). Then, a local analysis is performed (see Section~\ref{sec:appendix:methodsindetail:local_conceptual_importance}) on each connected subset of input units. For this purpose, \gls{crc} is computed for the remaining least relevant concepts, that were not computed in step (4), and their relevance values in input space are locally aggregated. 

Computation times recorded on different single GPU devices for each step are listed in Supplementary Table~\ref{tab:appendix:workflow:runtime} for a single GPU. Parallel computation on HPCI clusters with multiple GPUs may reduce the pre-processing computation time up to linearly.
Using the MapReduce paradigm, sample sets may be split evenly across GPUs, where the map procedure performs the computationally intensive relevance attribution steps, and the reduce procedure computes a final ranking across all splits at a low cost.

\begin{table*}[]
\centering
\caption{
\edited{}{
Computation times for the steps described in this section and Supplementary Figure~\ref{fig:appendix:workflow:workflow_overview}, calculated with \edited{}{\texttt{zennit-crp 0.6.0}.}
The ImageNet validation dataset with 50,000 samples was analyzed in all 13 convolutional layers of a VGG-16 BN. Step (5) measures the computation times for \gls{crc} heatmaps of all 512 concepts in layer \texttt{features.40} utilized for the local analysis (see Section~\ref{sec:appendix:methodsindetail:local_conceptual_importance}). Note, that only a fraction of the 512 concepts would be required to be calculated for a visualization (see Section~\ref{sec:appendix:experiments:quantitative:whatif}) and that \texttt{zennit-crp} relies on not exhaustively optimized Python code.
}
}
\edited{}{
\begin{tabular}{@{}llllll@{}}
\toprule
step & GeForce GTX 1080 Ti (batch size 16) & Titan RTX (batch size 32)\\ \midrule
(1)  & $27$\,min  & $21$\,min\\
(2)  & $50$ min & 28 min\\
(3)-(4) & $\textless 1$ s & $\textless 1$ s\\ 
(5) & $11.6$\,s & $8$\,s\\ \bottomrule
\end{tabular}
}
\label{tab:appendix:workflow:runtime}
\end{table*}

\subsection{Interpretation Workflow}
\label{sec:appendix:workflow:interpret}

In the following, we briefly present how glocal \gls{crc} explanations can be read. 
Firstly,
the visualization of heatmaps and the corresponding colormap is discussed.
Thereafter,
an example walkthrough for glocal explanations with Concept Atlases is shown.
Lastly,
we discuss how to read localized explanations, to investigate concepts in specific input regions.

\paragraph{How to Read Heatmap Visualizations}
In order to interpret heatmaps,
it is to note that attributions are bound between [-1, 1] by normalizing all attribution scores of the representation layer of choice with the maximum of the absolute value over all dimensions. Thus, the normalized values can then be visualized using color maps, \ie, in our case the \texttt{bwr} (blue white red) colormap available from \texttt{matplotlib}, as visualized in Supplementary Figure~\ref{fig:appendix:color_map}. Accordingly, the potential maximum (positive) value of the color map is set to 1 and its minimum counterpart (negative) to -1. Zero-valued attributions, depicted as white color, signify elements irrelevant to the analyzed inference process. 

\begin{figure}[h]
    \centering
    \includegraphics[width=1\linewidth]{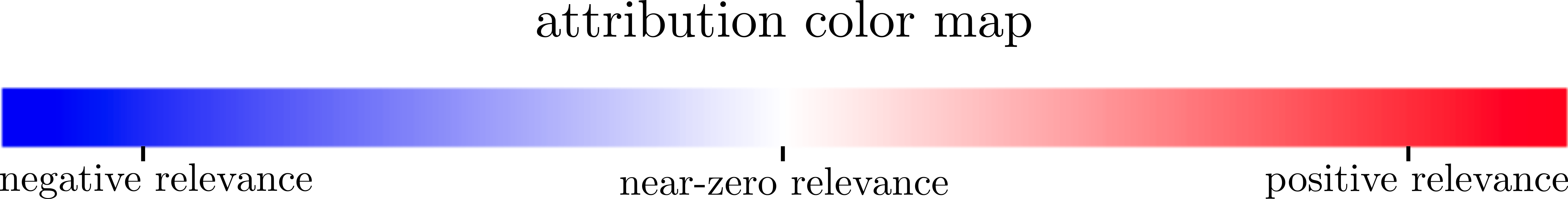}
    \caption{Color map \texttt{bwr (blue white red)} assigns blue color to negative values, white color to near-zero values and red color to positive values.}
    \label{fig:appendix:color_map}
\end{figure}

\paragraph{How to Read a Concept Atlas}
An example on how to read a Concept Atlas explanation is shown in Supplementary Figure~\ref{fig:appendix:workflow:how_to_read_explanations}.
Here,
a ``tabby cat'' is classified by a VGG-16 model with BatchNorm layers trained on ImageNet.
The user can choose a layer of interest (here \texttt{features.40}) and inspect the most relevant concepts.

\textbf{(1)} Starting with the measurably most relevant channel 281,
the channel-conditional heatmap can be investigated in a first step.
Conditional heatmaps indicate, where a concept is located in the input image. 
In this example, the concept of channel 281 dominantly points to the head and body of the cat. 

Next, \textbf{(2)} the masked reference samples can be inspected, as they visualize examples for which the model utilizes the concept preferably. By abstracting the shared nature of the images, the user could understand in this case that the concept encodes for cat-like features.

Additionally, \textbf{(3)}
the Concept Atlas can be viewed, visualizing where a concept is the first or second most relevant concept in input space.  The atlas shows, that the concept of channel 281 is most relevant on the position around the cat. 

Taking into account the results from steps (1)-(3) one can interpret how channel 281 is used \textbf{(4)}:
In this case, the model perceives a cat head and fur pattern using channel 281. It is an important concept with 5.8\,\% of total relevance, also indicated by being first or second most relevant in most parts of the input image.
Having gained understanding about the most relevant concept,
one can continue at this point with the second most relevant concept (channel 125).

\begin{figure*}[h]
    \centering
    \includegraphics[width=.9\textwidth]{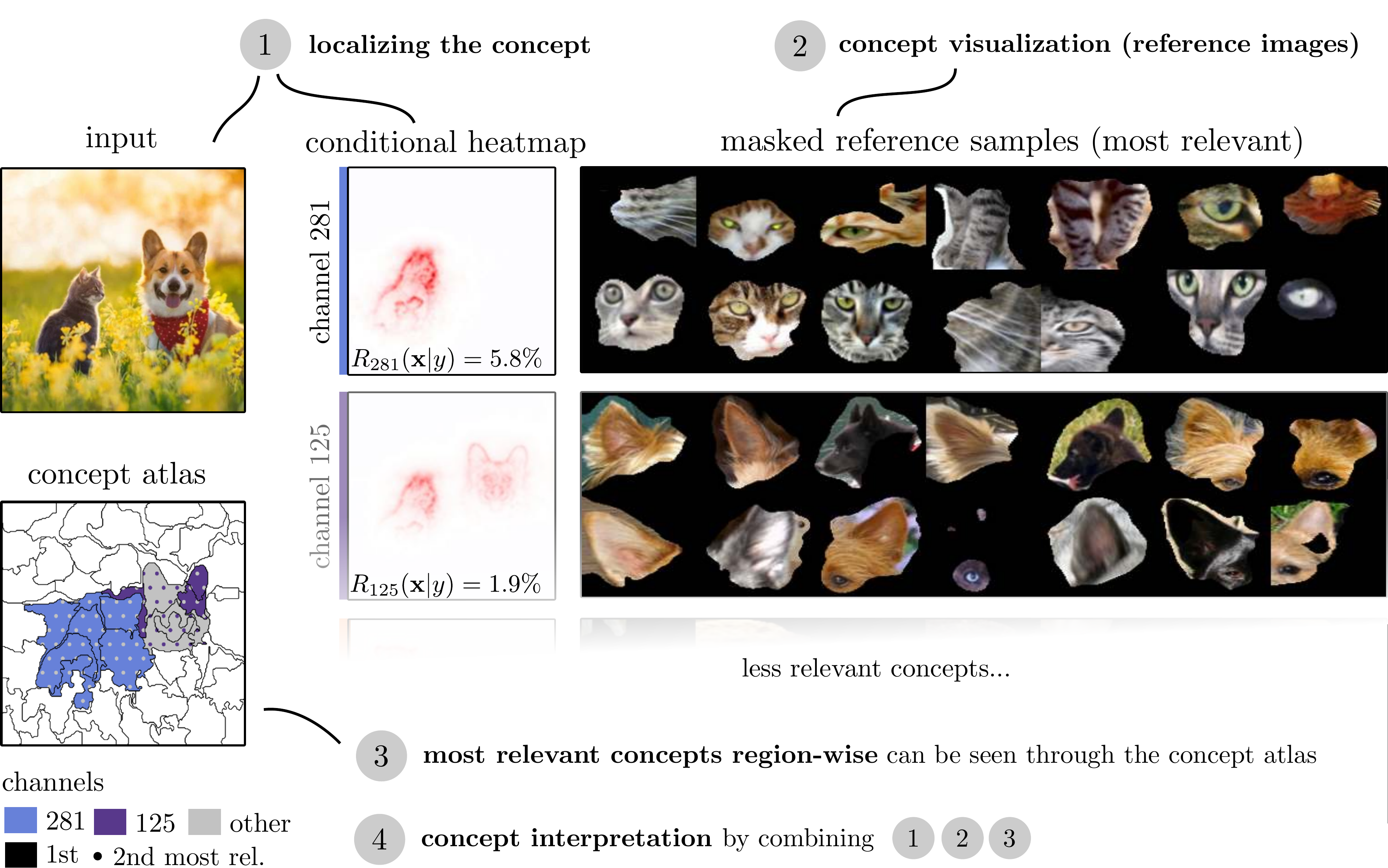}
    \caption{
    Workflow of reading glocal CRP explanations with a concept atlas.
    \textbf{(1)} In a first step of explanation reading, we pick the most relevant concept of a layer, and localize the concept using the conditional heatmap.
    \textbf{(2)} Thereafter,
    the masked reference samples allow for an understanding of the concept by visualizing the feature through example images.
    \textbf{(3)} Additionally, a Concept Atlas indicates, where a concept is most or (in this example only) second most relevant in specific regions of the input.
    Combining insights from (1) to (3),
    an interpretation of the concept and how it is used can be made.
    At this point, one can continue with the next less important concept. %
    }
    \label{fig:appendix:workflow:how_to_read_explanations}
\end{figure*}

\paragraph{How to Read Localized Explanations}
Complementary to the Concept Atlas visualizations of glocal \gls{crc}-based explanations discussed above, local analyses constitute a bottom-up counterpart for explaining a model decision, where the user can guide the analysis by, \eg, selecting a specific image region to investigate the therein recognized and utilized concepts.
Such an analysis workflow is illustrated in Supplementary Figure~\ref{fig:appendix:workflow:how_read_local} and discussed below.

\begin{figure*}[h]
    \centering
    \includegraphics[width=.65\textwidth]{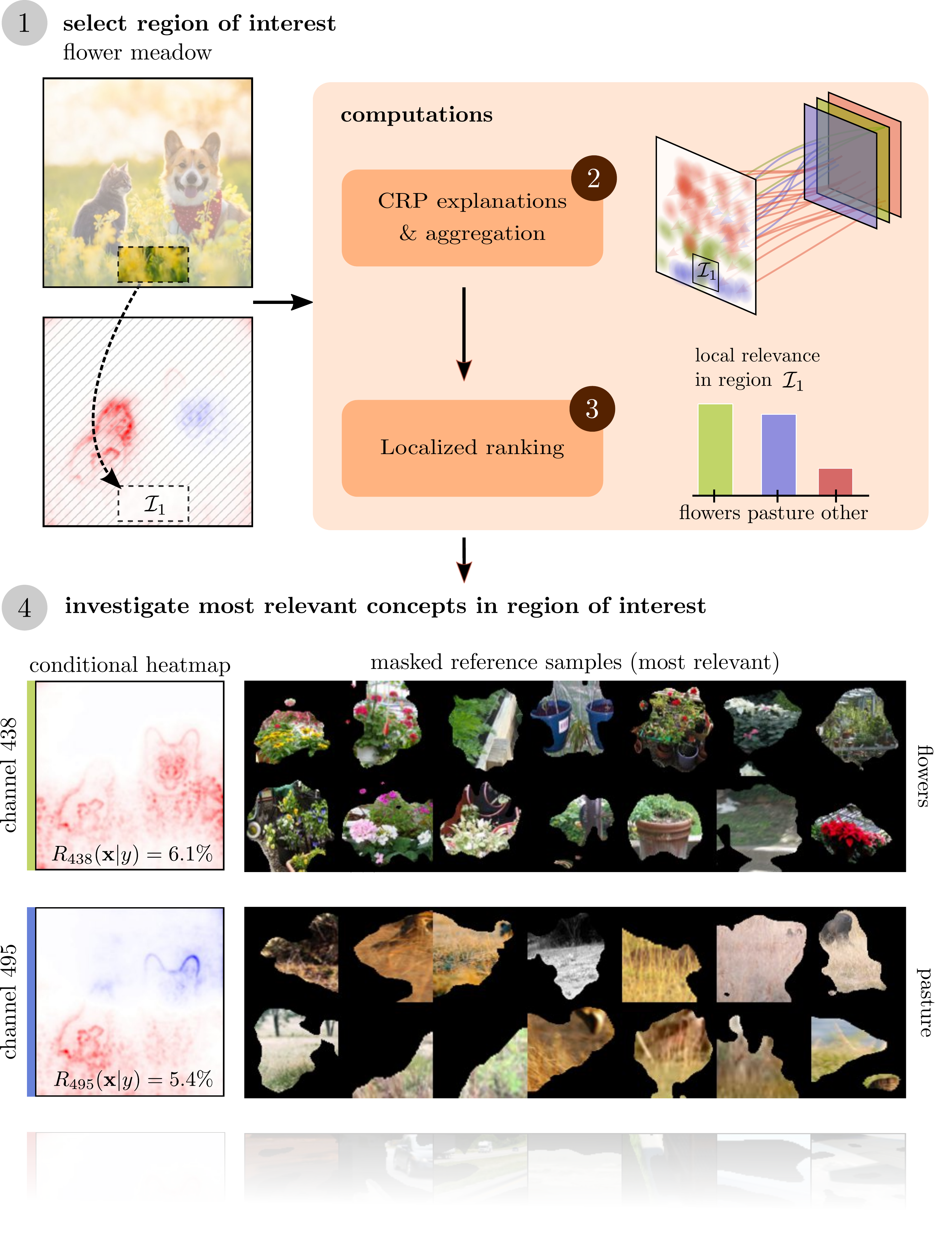}
    \caption{Workflow of reading and computation of local \gls{crc} explanations. The input image depicts a cat with a dog and the corresponding heatmap computed on a VGG16-BN model. The local conceptual importance is computed for layer \texttt{features.40}.
    \textbf{(1)} The user may select a region within the input sample to concentrate on for further analysis.
    \textbf{(2)} Relevance scores from \gls{crc} are then locally aggregated and analyzed, \eg via \textbf{(3)} ranking, in order to determine the locally most influential concepts.
    Finally \textbf{(4)} the locally most important concepts can be presented to the user for interpretation via conditional attribution maps and (masked) reference samples associated to the concept encodings. %
    } 
    \label{fig:appendix:workflow:how_read_local}
\end{figure*}

\textbf{(1)} First, the user identifies in the heatmap a region of interest. In the case of Supplementary Figure~\ref{fig:appendix:workflow:how_read_local}, a small amount of relevance is attributed on the flowers in the bottom part of the heatmap. However, no flower-like features seem to be recognized in the previous global analysis illustrated in Supplementary Figure~\ref{fig:appendix:workflow:how_to_read_explanations}. Thus, a closer examination of the model's perception in this region might be insightful.

\textbf{(2)} Selecting a \gls{dnn} layer of interest, \gls{crc} heatmaps are computed (or have been precomputed) $R(\x|\lbrace{y^L, c^l\rbrace})$  conditioned on the predictor outcome (class ``tabby cat'') in layer $L$ and each concept $c$ in layer $l$ (see Section~\ref{sec:appendix:methodsindetail:local_conceptual_importance}).
Taking advantage of the conservation principle of \gls{lrp} \eqref{eq:appendix:lrp_equality}, on which the \gls{crc} procedure is based on, the relative importance of each concept in the region of interest can be obtained by summing up the part of a concept's heatmap that is situated at the chosen region of interest.

\textbf{(3)} The aggregated relevance of each concept is sorted in descending order, and presented to the user accordingly. In this case, channels 438 and 495 attribute the most relevance with 6.1 \% and 5.4 \% of total relevance inside the region of interest respectively.

\textbf{(4)} The top-$k$ pre-computed and masked reference samples for channel 438 and 495 are displayed with their corresponding conditional \gls{crc} heatmaps. To elucidate the reference images, the user must uncover the common theme. In this case, channel 438 illustrates various flowerpots and channel 495 depicts a pasture foreground. The \gls{crc} heatmaps localize the concepts in the input space. Regarding channel 438, for example, the flowers are found at the bottom part of the image.

To conclude, the model perceives at the bottom of the image vegetation-like features that are to a lesser extend utilized to classify the image as a ``tabby cat'', though these features have nothing to do with cats from a human's point of view.

\cleardoublepage
\section{Additional Details: Human Evaluation Study}
\label{sec:appendix:study}
In the main manuscript in Section~\ref{sec:results:study}, we present a user study in which human subjects were asked to decide, based on explanations, whether a model's inference process given a data sample has been influenced by the presence of a particular and known data artifact: A black border, that is displayed around images.

For the study,
we fine-tuned two VGG-16 models trained on ImageNet (see Section \ref{sec:appendix:dataset_and_models} for more details about the training procedure). 
For one of the models, a data artifact ---a thick black border around the image--- was added to the training samples of a random subset of 30 classes to incentivize the model to learn a shortcut strategy on the data artifact.
Regarding the second model, the border was displayed around all images, \emph{irrespective} of the class, during training to lessen the impact of the artifact on predictions, in general. 
For both models, we then generate explanations for the same set of 30 randomly drawn classes and respective samples from their hold-out test sets, where the black image border artifact is applied to \emph{all} samples. The participants are then presented with these explanations.
In the shown samples for the impacted model,
cropping out the artifact is followed by a decrease of at least 20\,\% on the target output logit ($31.2\,\%$ decrease on average with a standard deviation of $6.7\,\%$).
For the second model,
the target output logit is changing less than 2\,\% in value ($0.3\,\%$ decrease on average with a standard deviation of $1.3\,\%$) when the border is cropped out.

The study was conducted via Amazon Mechanical Turk in a \emph{between-subject}s design,
\ie, each participant was only asked to assess visualized explanations for one of the compared methods to facilitate comparison without mutual influence \cite{edmonds2019between}.
We collected experimental data from 125 participants in total, with 25 participants per method.
After an instruction on how to read and interpret the respective explanations,
each worker is shown 14 randomly selected samples and explanations of \emph{one} \gls{xai} method (7 samples from each explanation set, respectively model).
In the primary task, the participants are first asked to assess, whether the black border impacts the model prediction according to the explanation (binary answer: yes or no).
We ask secondary questions on how confident they are in their answer (10-point scale from 0 to 9, where 9 indicates the highest possible level of confidence), and about the perceived clarity of the presented explanations (10-point scale, from 0-9, where 9 indicates the highest possible level of clarity).

In the following, we provide additional insights about the generation of the explanations and conclude with a summary of the study results.  
In order to compare the explainability value of different widely-used and popular local explanation methods to our approach, attribution maps are computed with
Integrated Gradients (IG) \cite{sundararajan2017axiomatic},
SHAP \cite{lundberg2017Shap},
Grad-CAM \cite{selvaraju2017grad} and LRP \cite{bach2015pixel, kohlbrenner2020towards}, next to our proposed \gls{crc} maps with \gls{rmax} examples. Integrated Gradients (using 100 integration steps), Grad-CAM and SHAP (using 100 SLIC super-pixels, 50 different permutations, and gray color as baseline) have been computed via \texttt{Captum}~\cite{kokhlikyan2020captum}.
LRP attributions have been computed with \texttt{zennit}~\cite{anders2021software} using the LRP$_{\varepsilon-{z^+}-\flat}$-rule as discussed in Section~\ref{sec:appendix:technical}.

\begin{figure*}[h]
    \centering
    \includegraphics[width=.75\textwidth]{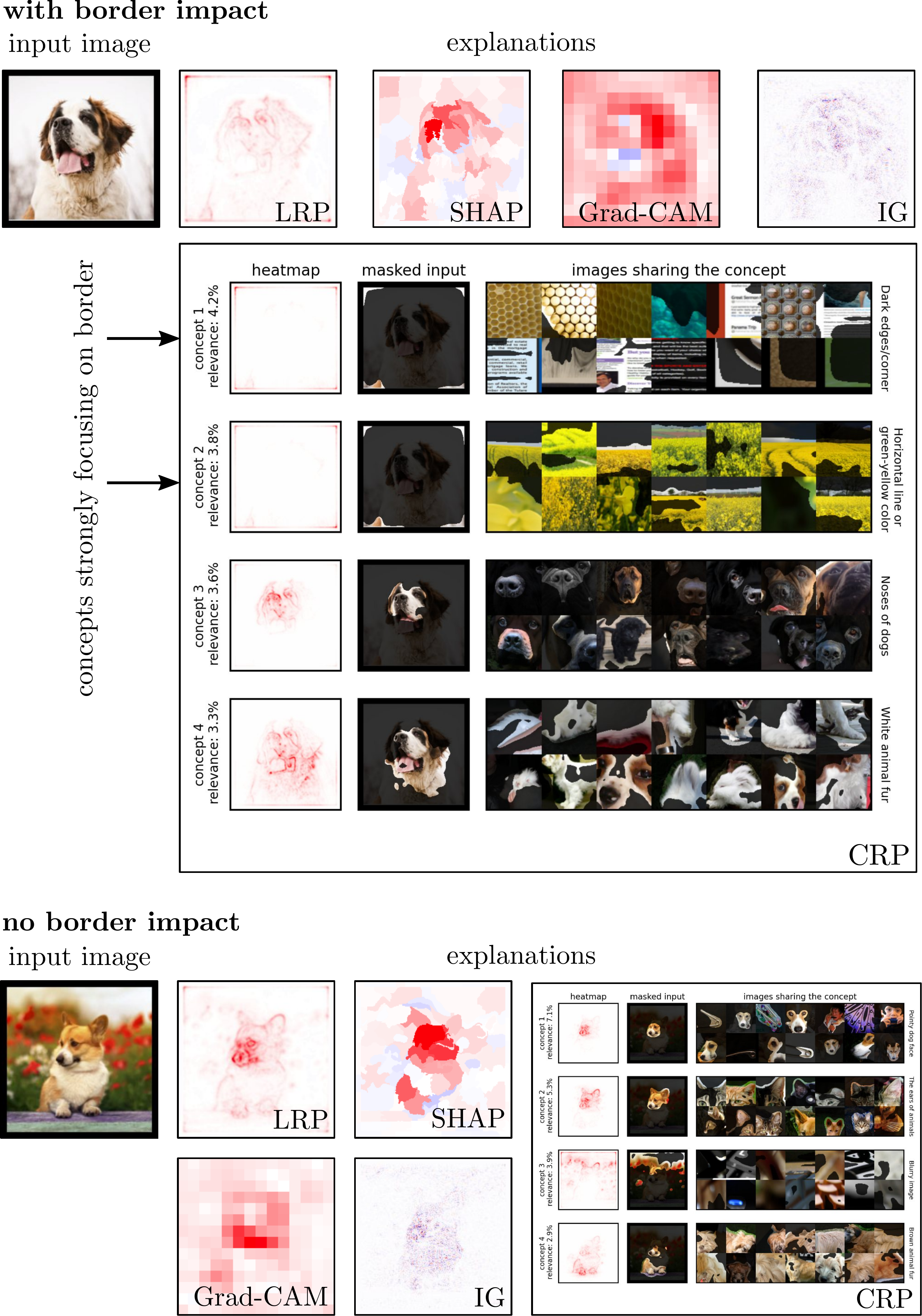}
    \caption{Example explanations from different attribution methods as presented to participants in the user study.
    \textbf{Top)}
    A sample predicted on by a model influenced by the black border artifact and corresponding attribution maps from \gls{lrp}, SHAP, Grad-CAM, Integrated Gradients (IG) and our proposed \gls{crc} and \gls{rmax} approaches.
    \textbf{Bottom)} A sample predicted on by a model not influenced by the border artifact, with corresponding explanations.
    The \gls{crc}- and \gls{rmax}-based explanations here show the top-4 most relevant concepts in layer \texttt{features.28}, their corresponding heatmaps, relevance-masked input images as well as masked reference samples.
    }
    \label{fig:appendix:study:study_explanation_example}
\end{figure*}

An example of resulting explanations is shown in Supplementary Figure~\ref{fig:appendix:study:study_explanation_example} for all five methods with respect to a sample with (\emph{top}) and without (\emph{bottom}) a strong \gls{ch}~\cite{lapuschkin2019unmasking, schramowski2020making, anders2022finding} impact on the target class. 
The black border width on the input sample is 9 pixels, while the overall image size was kept unchanged at 224\,px$\times$224\,px.

For \gls{lrp} and Grad-CAM heatmaps, a shift from high relevance densities in the heatmap center to the border is visible, when the border has indeed a strong impact. On occasion, we observe a disproportionate amount of relevance on the border region for the model which has not been trained to utilize that feature at all.
It is more difficult to see higher relevance on the border region for SHAP, as the border is part of a large number of super-pixels, where perturbations might not significantly impact the border artifact. Further, attributions from ig tend to be noisy, making it difficult to correctly interpret the explanations overall.
\gls{crc} heatmaps show the four most relevant concepts in layer \texttt{features.28} with corresponding heatmaps, relevance-masked input, masked reference images, and a suggested label for each concept. 
The label suggestions are generated in a first step using MILAN \cite{hernandez2022natural}, which automatically generates natural language descriptions of masked reference images, and are further refined by the authors when necessary.

\begin{table*}[]
\centering
\caption{
Results of the human study with the standard error of mean for predicting the impact of a \gls{ch} artifact on a prediction. 
CRP results in significantly higher accuracy and F1-score than all other methods. However, in terms of decision confidence and explanation clarity, IG and LRP, respectively, perform best.
}
\begin{tabular}{@{}llllll@{}}
\toprule
Method & Accuracy (\%) & F1-Score (\%) & Confidence (\%) & Clarity (\%)\\ \midrule
IG     & $51.7 \pm 1.9$            & $52.7 \pm 2.9$           & $\mathbf{77.0 \pm 1.6}$             & $70.7 \pm 1.6$\\
LRP    & $56.6 \pm 2.9$            & $61.6 \pm 2.4$           & $74.3 \pm 1.3$             & $\mathbf{71.8 \pm 1.7}$\\
SHAP   & $58.3 \pm 2.7$            & $62.2 \pm 2.4$           & $74.2 \pm 1.6$             & $67.7 \pm 1.8$\\
Grad-CAM    & $63.7 \pm 3.4$            & $67.4 \pm 2.3$       & $70.5 \pm 1.8$             & $64.9 \pm 1.9$\\
CRP (ours)    & $\mathbf{80.9 \pm 3.4}$   & $\mathbf{82.3 \pm 1.8}$  & $76.1 \pm 1.7$             & $64.1 \pm 1.6$\\ \bottomrule
\end{tabular}
\label{tab:appendix:study:results}
\end{table*}

The study participants were able to detect whether the prediction was impacted by the border most often using \gls{crc} explanations,
as shown in Supplementary Table~\ref{tab:appendix:study:results}. 
\gls{crc} results in the highest accuracy of $(80.9 \pm 3.4)\,\%$ and F1-score of $(82.3 \pm 1.8)\,\%$.
 Of the local methods Grad-CAM performs best with an accuracy of $(63.7 \pm 3.4)\,\%$ and F1-score of $(67.4 \pm 2.3)\,\%$,
and IG worst with $(51.7 \pm 1.9)\,\%$ accuracy and $(52.7\pm 2.9)\,\%$ F1-score. It is to note, that random guessing would correspond to values of 50\,\%, 
indicating that the insight of IG explanations is limited in this study.
Participants exposed to the IG explanations report the highest confidence
$(77\pm 1.6)\,\%$ — while at the same time performing worst in the primary
task — followed by CRP $(76.1\pm 1.7)\,\%$.
Regarding clarity of the explanations as perceived by the participants, the
fine-grained attribution maps of IG and LRP receive the highest scores. CRP
and RelMax interestingly result in the lowest reported clarity. These results are discussed in detail in the main manuscript in Section~\ref{sec:results:study}.

\cleardoublepage
\clearpage
\section{Detailed Analysis: Clever Hans Detection and Quantitative Results}
\label{sec:experiments:quantitative}
In the previous sections, we have qualitatively demonstrated the capabilities of our method for explaining machine learning predictions not only in terms of involved concepts, but also how those predictions can be broken down into corresponding concept-specific explanations again, and how this information may be visualized for exploring machine learning models and datasets. 

In the following, we will move forward by showcasing how our achieved level of explainability may be used to interact with the model, \eg, by second-guessing its decision based on the features it has used, as revealed by \glsdesc{crc} and its example-based explanations. First, starting with Section~\ref{sec:appendix:experiments:quantitative:filterperturbation} we give an answer to the question of \emph{how many} concepts need to be observed and understood in order to comprehend a particular model decision given the hundreds of filters potentially encoding unique abstractions\footnote{Fortunately, the answer is ``not that many''!}. This set of relevant concepts then allows for, \eg, entering a dialogue with the predictor and for asking \emph{``what-if''} questions regarding its decision-making strategies. In Section~\ref{sec:appendix:experiments:quantitative:whatif}, we then investigate how a model trained on the ISIC~2019 skin lesion analysis dataset~\cite{combalia2019bcn20000,codella2018skin,tschandl2018ham10000} reacts to the presence of clinically irrelevant concepts which have been discovered to be learned. In Section~\ref{sec:appendix:experiments:quantitative:imageretrieval} we show how our method can be used to first discover spuriously interesting (\glsdesc{ch}) concepts learned by the model, and then perform an explainability-based image retrieval in order to identify multiple classes affected by the same image artifact semantically, which however \emph{should} be unrelated to the target labels. Finally, in Section~\ref{sec:appendix:experiments:quantitative:clusters} we analyze the latent representations of relevance-based reference example sets and thus are able to identify sets of filters correlating in terms of activation, but differ in relevance for the purpose of encoding detailed differences between conceptually related ImageNet classes.

\subsection{Quantifying the Foundation of Decision Making}
\label{sec:appendix:experiments:quantitative:filterperturbation}
In the previous sections, we have so far shown only a very limited amount of the most relevant concepts and examples as identified via relevance scores. For one, this is due to the static nature of this manuscript and its associated limitations\footnote{Which arguably could be alleviated with interactive analysis tools, which are, at time of submission, under active development.}. More importantly, with the usually high number of layers within an analyzed network and the hundreds of filters per layer, an exhaustive presentation of all explainability data per inference might quickly become overburdening to the observer. With those practical limitations in mind, we want to address the question of \emph{how much} explanation is needed in order to sufficiently understand a particular inference of the model. Recent work on network compression and pruning has shown that contemporary neural network architectures often are highly over-parameterized and a large amount of filters can be removed without noticeable impact on the performance of the model~\cite{yu2018nisp,yeom2021pruning,becking2021ecq}. The explicit representation of prototypes in some neural network architectures~\cite{chen2019looks} has been exploited to remove spurious information from the decision making in a user-interactive fashion~\cite{gerstenberger2022but}. Based on this, we may assume that there exist neurons and filters which encode information irrelevant to a particular inference under analysis. To measure the extent of the (ir)relevance of explained and interpretable network elements, we perform an experiment based on the perturbation of a latent layer's filters in accordance to their relevance scores, similar to the evaluation based on input perturbation described in~\cite{samek2017evaluating}. We show that a large part of filters have indeed no impact on individual predictions, and only a small fraction of the available filters per layer do have a significant impact in favor or against the prediction made by the model, which need to be observed in order to understand the model's reasoning.

Specifically, for a particular prediction we start with the computation of relevance scores, which are assigned to all layers, filters, and neurons of the model. Within a layer to be evaluated, we first rank the filters according to their (spatially sum-aggregated) relevance. Then, according to this ranking, we successively deactivate the filters in descending order (\ie, deactivating the most relevant representations first) by setting their activations to zero, and re-evaluate the model output. Since the initial output logits differ for each data sample, values are measured relative to the initial confidence values of the observed and explained model output. We refer to this process as ``filter flipping'' in accordance to the designation of ``pixel flipping'' from~\cite{bach2015pixel}. This procedure measures the impact a set of perturbed filters has on the decision outcome, and whether their relevance-based ranking is indicative of the model's use of the therein encoded information during inference. The same procedure is repeated with an inverted filter ranking, \ie, perturbing the least relevant filters (potentially being relevant against the initial prediction outcome) first. Note that deactivating filters with negative relevance scores usually increases the prediction confidence, since negative relevance argues against the selected class, and removing them indirectly reinforces evidence for the selected class.

\begin{figure*}[t] 
  \centering
  \includegraphics[width=.9\linewidth]{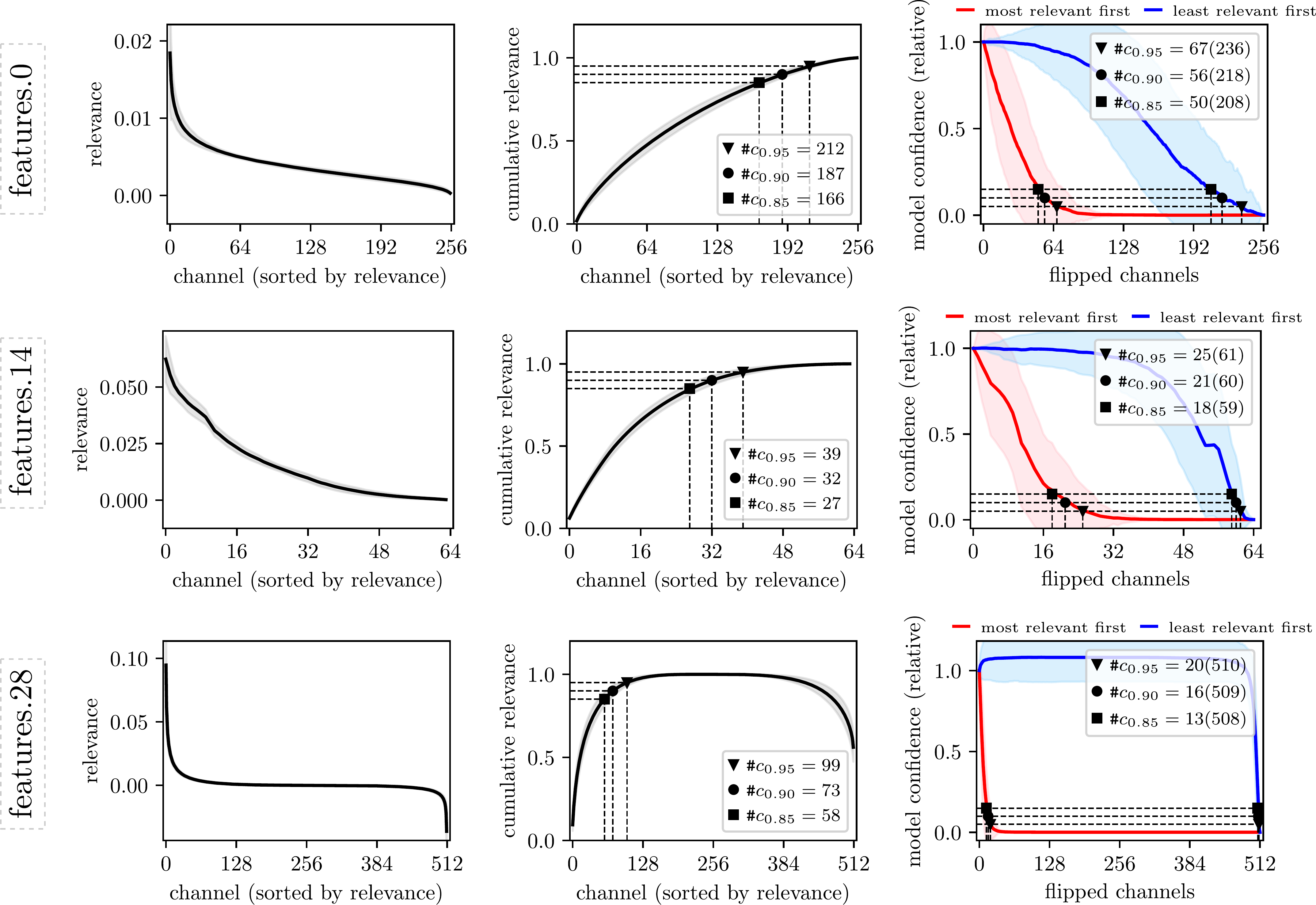}
  \caption{We evaluate the distribution of relevance and relevant filters in \texttt{features.28}, \texttt{features.14} and \texttt{features.0} of a VGG-16 model trained on ImageNet, averaged over 250 randomly drawn example images. Only a fraction of all 512 filter channels is relevant for classification predictions in \texttt{features.28} (high level concepts) and \texttt{features.0} (very basic concepts), with a wider utilization of latent features in \texttt{features.14} (concepts of intermediate abstraction level) . (\emph{Left}): Relevance (relatively scaled to the sum of positive relevances) is power-law distributed across the channels of the observed layer. (\emph{Center}): Cumulative relevance distribution with markers showing that, \eg, for \texttt{features.28},  on average 58, 74 and 100 channels are necessary to explain any prediction up to 85\,\%, 90\,\%, or 95\,\% in terms of cumulated relevance, respectively. (\emph{Right}): The activations of filters with the most positive (red line) and negative (blue line) attributed relevance are deactivated successively and the model is reevaluated. For \texttt{features.28}, after 14, 16, or 20 filters inactive, the output confidence (normalized to the initial output value) has decreased by 85\,\%, 90\,\%, or 95\,\%. Compared to the number of the for the prediction cumulatively relevant filters, a significantly smaller number of the most relevant filters needs removed in order to change the prediction outcome significantly. In contrast, flipping the least relevant channels first, the model remains confident even if the majority (up to 508, 509 or 510) of filters are removed. Only until the last four, three or two filters are flipped, the model confidence drops below 15\,\%, 10\,\% or 5\,\%, respectively. For all plots, the mean curve is displayed in opaque color and standard error of mean with transparent color. Similar observations can be made for \texttt{features.0} and \texttt{features.14}, which however do have an in general higher impact once removed. We ascribe this observation to the basic nature of concepts encoded by those low(er) level filters which are shared for highly specialized and class-specific abstractions in later layers. The flipping experiment is restricted to confident predictions (softmax probability above 50\,\%), so that neurons encode meaningful information.}
  \label{fig:appendix:experiments:vgg-flipping-features28}
\end{figure*}

Supplementary Figure~\ref{fig:appendix:experiments:vgg-flipping-features28}~(\emph{left}) shows the distribution of ranked relevance scores per filter (averaged over 250 randomly drawn samples) within \texttt{features.28} of a VGG-16 model, and Supplementary Figure~\ref{fig:appendix:experiments:vgg-flipping-features28}~(\emph{center}) the average cumulative sum of relevance over the rankings. Note that relevance scores attributed to filters tend to be power-law distributed, which holds true for both strong positive attributions, as well as strong negative attributions. This means that (on average) only a minority of filters are involved in per-sample inference, both in support and contradictory to the prediction outcome. In Supplementary Figure~\ref{fig:appendix:experiments:vgg-flipping-features28}~(\emph{center}) we can see that in order to observe 85\,\%, 90\,\%, or 95\,\% of the foundation of the model's reasoning \emph{speaking for} its prediction outcome (measured in terms of cumulative relevance), the user has to investigate and understand the concepts of the 58, 74 and 100 most important filters, respectively. Approximately 300 of the 512 filters in layer \texttt{features.28} are not involved in predictions on average.

While having to look at ``only'' approx.~20\% (which still are 100) of the layer's filters to understand 95\,\% of the most relevant filters might indicate a high user workload, the results of the filter flipping experiment in Supplementary Figure~\ref{fig:appendix:experiments:vgg-flipping-features28}~(\emph{right}) do showcase that effectively, an even smaller fraction of the most relevant filters is \emph{responsible} for the model's decision confidence: We measured that in order to decrease the output confidence of the model relatively by 85\,\%, 90\,\%, or 95\,\% it is on average sufficient to deactivate only the 14, 16 or 20 most relevant filters, respectively.

In general, investigating individual neurons and their concepts is easiest if a small proportion of neurons are sufficient to explain most of any prediction. Observing the results regarding \texttt{features.0} and \texttt{features.14} it can be seen, that the earlier layers require a higher fraction of channels to analyze --- compared to layer \texttt{features.28}. However, same as for layer \texttt{features.28},  flipping a few of the most relevant channels results in large drops regarding prediction confidence. The generally slower decay in confidence over most relevant flipped channels can be interpreted such that lower level concepts are not as class specific as more abstract representations later in the model, and as simpler features, such as edges or color gradients, are more likely to be shared among classes and exhibit a certain amount of redundancy and replacability in the lower layers. This is also supported by the earlier confidence decay when the least relevant channels are flipped. It is to note that for ResNet-like architectures, it should additionally be taken into account that relevance may be distribured unevenly among layers due to short-cut connections. ResNet models have high flexibility in how they allocate concepts within the model due to their short-cut connections. As a consequence, short-cut connections cause an additional non-uniform distribution of relevance between entire layers, not just neurons. This insight is for example exploited in neural network pruning~\cite{yeom2021pruning}. Note, that the conservation principle of \gls{lrp} (\cf~\cite{bach2015pixel}) inherited by \gls{crc} in theory guarantees that the total amount of relevance remains constant across parallel layers. The mean relevance flow throughout the ResNet architecture is shown in Supplementary Figure~\ref{fig:appendix:flipping:resnet34_flow_flipping}. Here, it is visible, that several layers do not contribute strongly to predictions, as they receive low attributions. This is further confirmed by a channel flipping experiment for a layer with high relevance (\texttt{layer3.0.conv2} with a mean of over 50\,\% relevance) and one layer with low relevance (\texttt{layer3.4.conv1} with a mean of less than 1\,\% relevance). The flipping of channels has a significant impact for \texttt{layer3.0.conv2} on the model confidence. For layer \texttt{layer3.4.conv1} on the other hand, the confidence is less affected, resulting in a mean confidence of still over 80\,\% after all channels are flipped. Investigating the flow of relevance through a model with, \eg, skip connections allows the user to distinguish which filters within a layer should be investigated, and whether this layer needs to be investigated at all, depending on whether the decision process has skipped the layer at hand.

\begin{figure*}[h]
    \centering
    \includegraphics[width=0.9\textwidth]{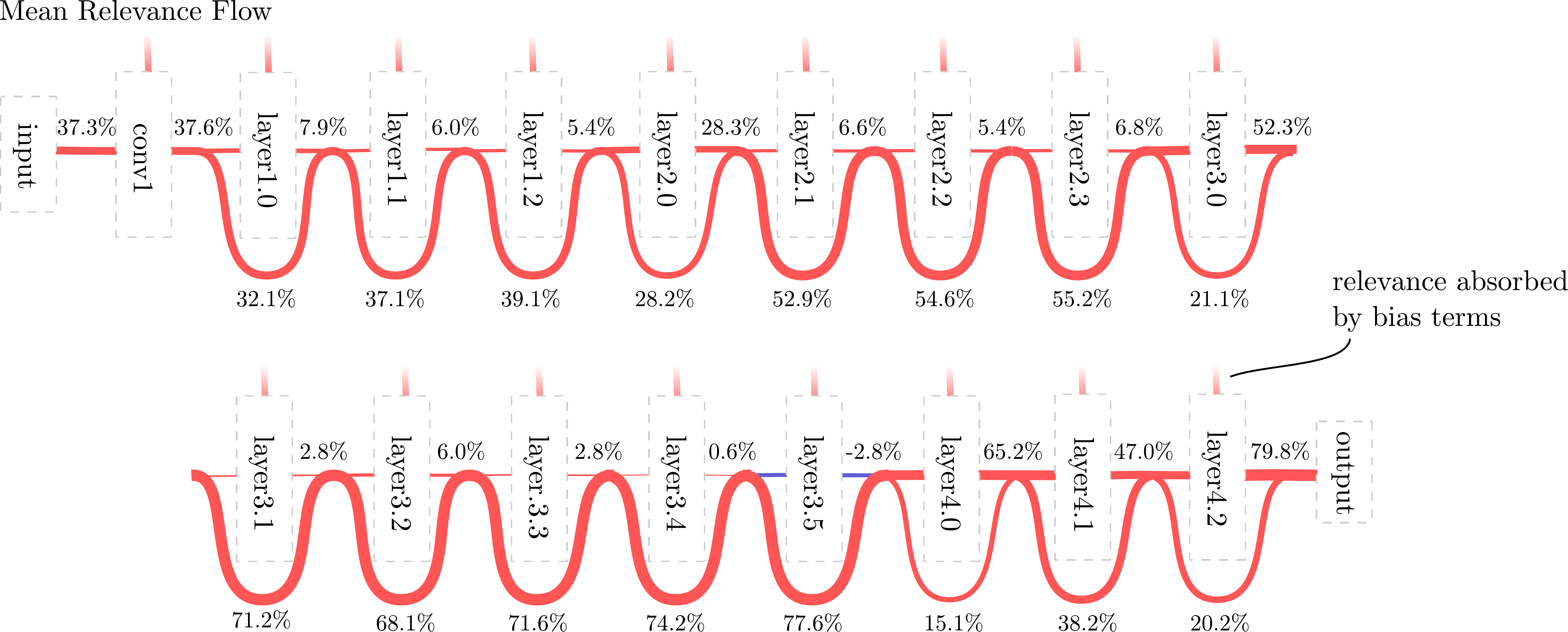}
    
    \vspace{5mm}
    
    \includegraphics[width=0.8\textwidth]{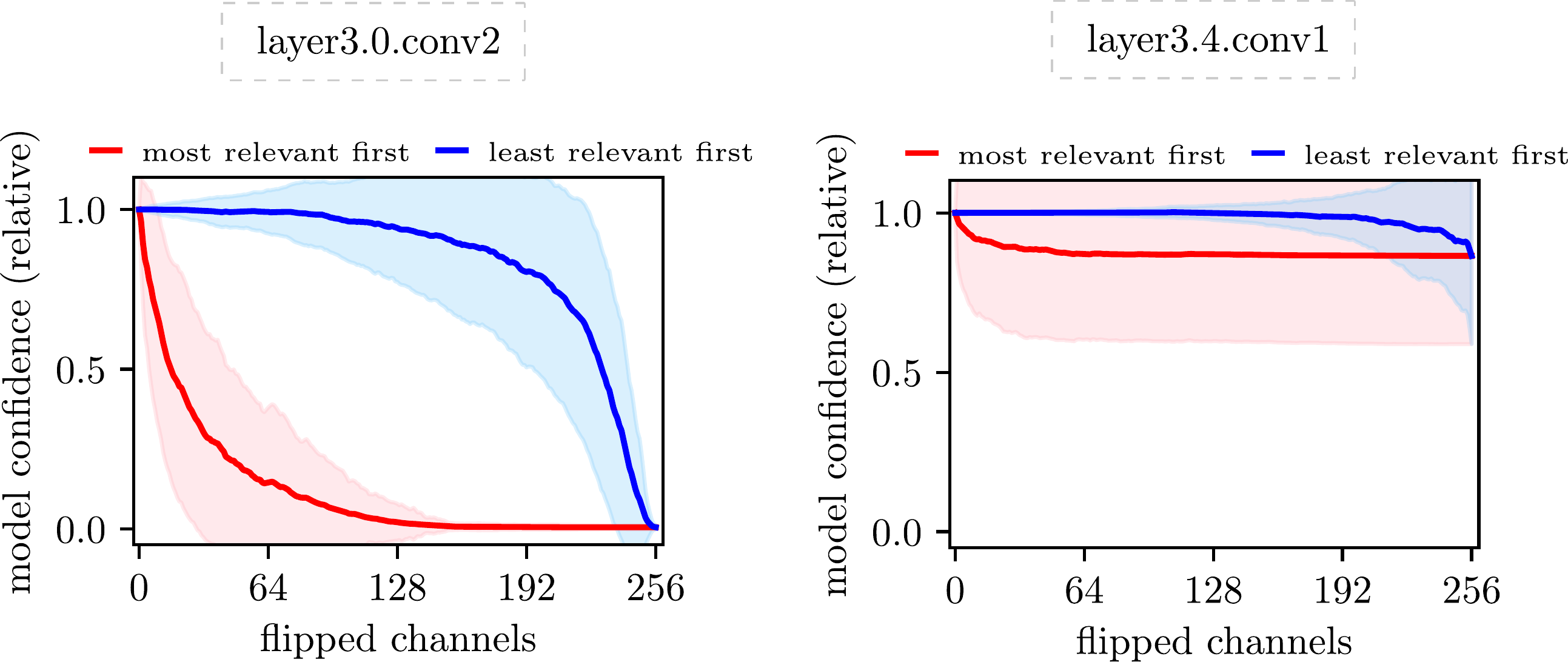}
    \caption{The ResNet34 model is a feed-forward model with shortcut connections between layers. As is depicted in the top, layers are receiving different amount of relevances during predictions. Most layers receive less than 7\,\% of total relevance. This result suggests, that these layers could be potentially pruned without large accuracy drops. This is further supported, by a pixel-flipping experiment (\emph{Bottom}), showing that flipping channels has no significant impact on the prediction confidence for \texttt{layer3.4.conv1} (\emph{Bottom Right}), which receives on average about 0.6\,\% of total relevance. Flipping channels of layer \texttt{layer3.0.conv2} (\emph{Bottom Left}) on the other hand has a stronger effect on the model confidence, as it also has a mean relevance of about 52\,\% for predictions. For the plots, the mean curve (of 250 samples) is displayed in opaque color and standard deviation with transparent color. The flipping experiment is restricted to confident predictions (soft-max probability above 50\,\%) in order to limit the variation in the plot.}
    \label{fig:appendix:flipping:resnet34_flow_flipping}
\end{figure*}

We conclude that it is therefore possible to understand (or at least obtain a thorough overview about) the reasoning of the model in context of a particular sample by considering only a small fraction of all filters, since only the most important filters have a noticeable influence on the outcome. Further, it is to note that for simplicity, we assume in this study, that each filter encodes a learned concept on its own. This assumption will in general not hold, with multiple filters redundantly or in combination encoding highly similar or compound concepts in practice. A meaningful grouping or clustering of filters (see Section~\ref{sec:experiments:quantitative:clusters} in the main manuscript and Section~\ref{sec:appendix:experiments:quantitative:clusters} for an outlook) can therefore further ease the task of interpreting machine learning explanations.

\subsection{
Investigating the Impact of (Clever Hans) Concepts
}
\label{sec:appendix:experiments:quantitative:whatif}
The methods proposed in this study allow to explore learned concepts and investigate their role in the decision process. Knowing which concepts exist allows to manipulate and test decisions. In the previous section,  concepts are suppressed without taking their semantic meaning into account, and by setting whole activation maps to zero. In the following, we first insert and secondly replace distinct concepts in order to manipulate decisions. Specifically, we present two examples for the ISIC lesion classification task~\cite{tschandl2018ham10000,codella2018skin,combalia2019bcn20000}. Here, different skin samples are classified into the diagnostic classes of melanoma (MEL), melanocytic nevus (NV), basal cell carcinoma (BCC), actinic keratosis (AK), benign keratosis (BKL), dermatofibroma (DF), vascular lesion (VASC), squamous cell carcinoma (SCC) and unknown (UNK). Using our approach, we are able to identify the filters responsible for encoding colorful band-aids polluting a significant fraction of the available samples~\cite{rieger2020interpretations}, which is known to impact the inference of models trained on the dataset~\cite{anders2022finding}. 
\begin{figure*}[t]
    \centering
    \includegraphics[width=0.8\textwidth]{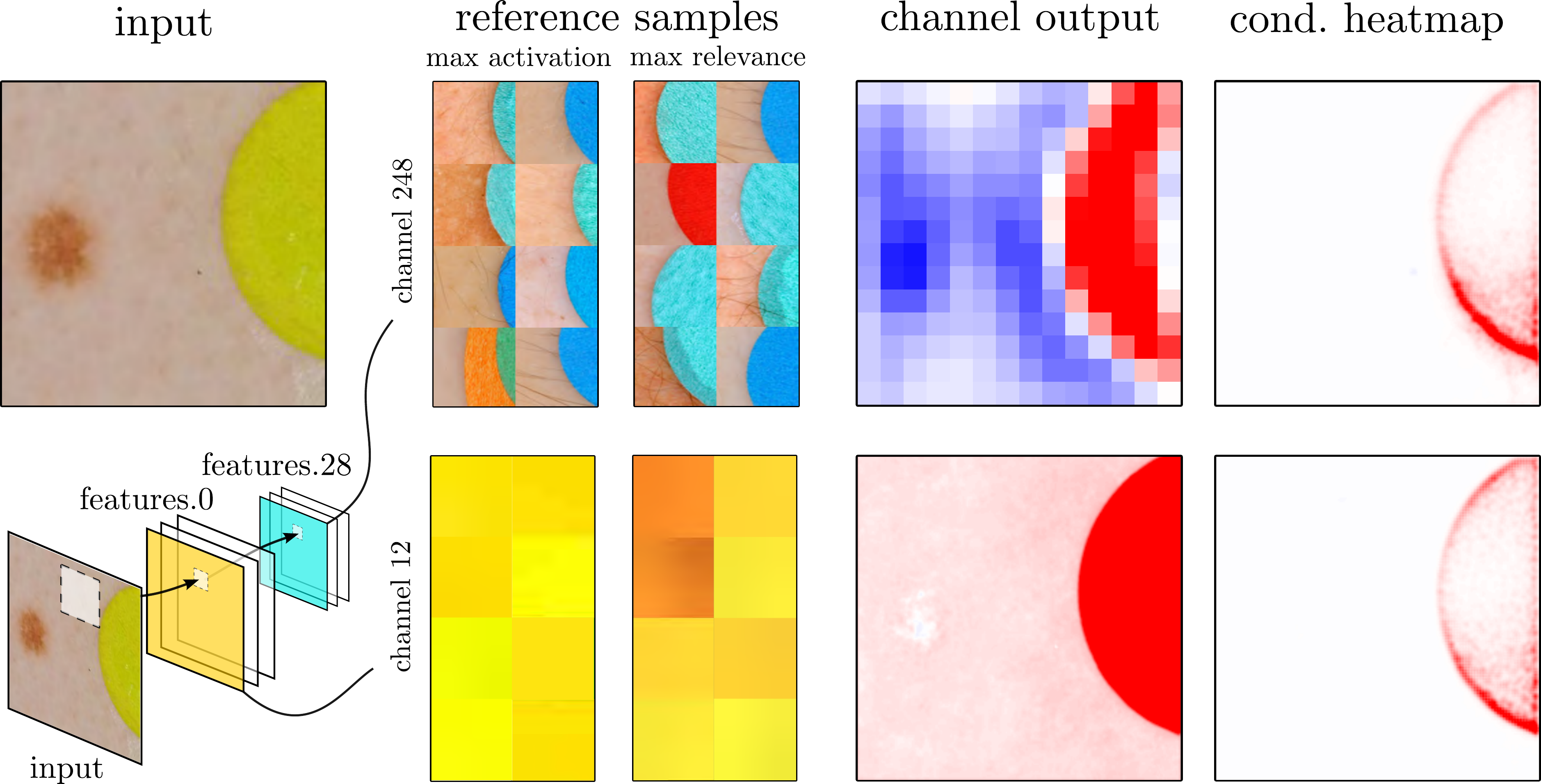}
    \caption{CRP-based analysis shows that concepts relevant for ``band-aid'' are used for lesion classification in the ISIC dataset. Specifically, a higher-level channel (index 248) for ``colorful band-aid''  in \texttt{features.28} (\emph{top}) and a lower-level channel (index 12) for ``yellow'' (\emph{bottom}) can be identified in \texttt{features.0}. The activation maps as well as channel relevance heatmaps of the shown sample support the finding locally. In this particular case, both channels are the most relevant channels of their respective layer.}
    \label{fig:appendix:experiments:isic-plaster}
\end{figure*}
A CRP-based analysis shows that a VGG-16 model trained on the ISIC dataset has learned to use a band-aid concept for lesion detection. Locally,  a higher-level channel (filter index 248 in \texttt{features.28}) for ``colorful band-aid'' and a lower-level channel (filter index 12 in \texttt{features.0})  for ``yellow'' can be identified and associated to the concepts, as is depicted in Supplementary Figure \ref{fig:appendix:experiments:isic-plaster}. The channel activation maps, as well as channel relevance heatmaps of the given sample support the finding, by consisting of large values in the spatial region of the band-aid. Further, these channels are most relevant for the decision, suggesting the identification of the source of Clever Hans behavior tied to these filters. 

\begin{figure*}[t]
    \centering
    \includegraphics[width=1\textwidth]{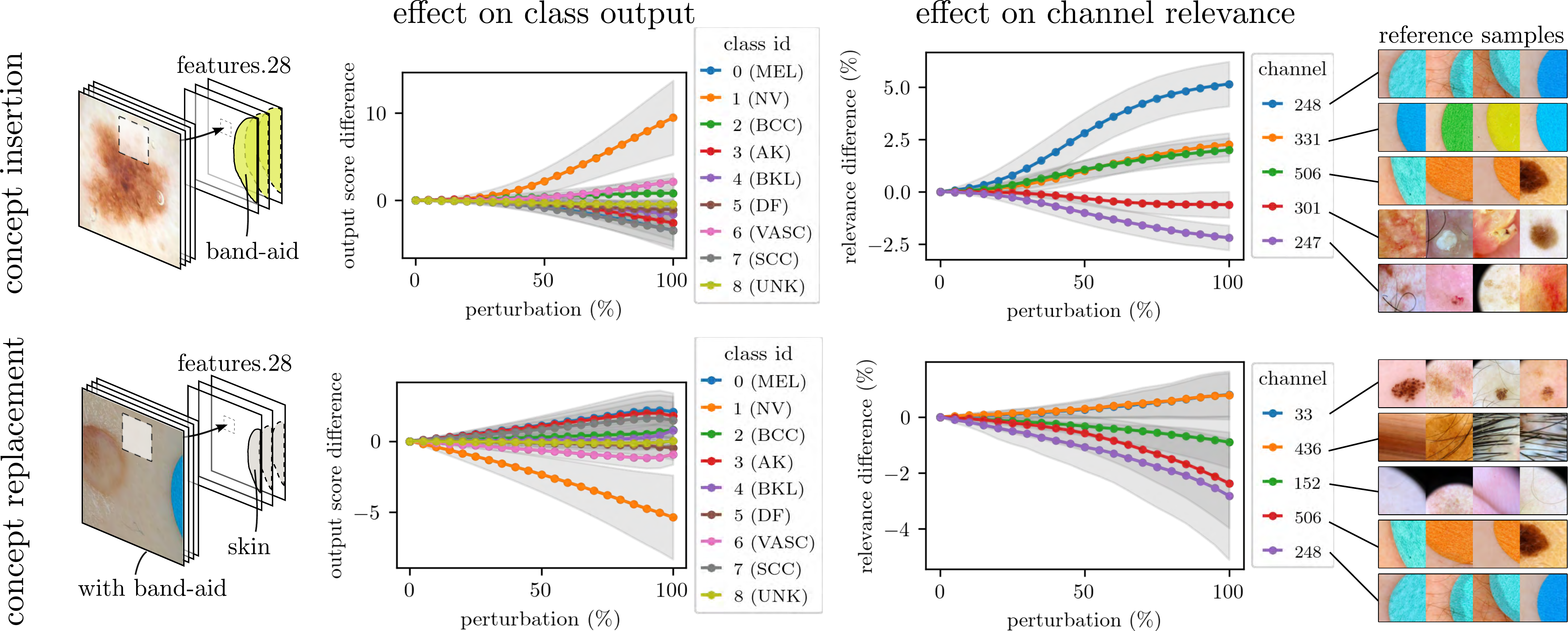}
    \caption{%
    A systematic evaluation of the impact of the ``band-aid'' concept to model behavior. The band-aid is inserted (\emph{top}) or replaced (\emph{bottom}) in the prediction process at layer \texttt{features.28}. The manipulation is performed by linearly alpha-blending the replacement features locally into the target image, and the mean change in class output as well as channel relevance is recorded. While inserting a ``band-aid'' concept in the latent space,  the output score for the benign class 1 (NV) increases most strongly (\emph{top left}). Further, relevance scores for channels supporting band-aid concepts increase (\emph{top right}). The gradual removal of the ``band-aid'' concept, results in an opposite effect: The output score of class 1 decreases (\emph{bottom left}) and channels supporting ``band-aid'' are decreasing in relevance as well  (\emph{bottom right}). In all plots, the standard deviation is highlighted in gray color.}
    \label{fig:experiments:isic-plaster-dataset} 
\end{figure*}

In the first experiment, the Clever Hans --- \ie, the ``band-aid'' --- concept, is inserted into 160 randomly drawn sample images which do not show band-aids. As shown in Supplementary Figure \ref{fig:experiments:isic-plaster-dataset} (\emph{top}), the ``band-aid'' expression  is inserted in the prediction process via the higher-level layer \texttt{features.28}. To this end, the activation tensors are copied from a sample with a band-aid features to a sample without band-aid features. In order to only copy activations describing the band-aid concept, all activations are masked to overlap with the band-aid in the spatial dimension and then transferred to the target sample. Specifically, the activations $\mathbf{Z}^{l}_{m} \in \mathbf{R}^{c\times w \times h}$ (located in layer $l$ consisting of $c$ channels with spatial dimensions of size $w$ and $h$) for prediction of sample $m$ without a band-aid are replaced with activations $\mathbf{Z}^{l}_{p}$ from sample $p$ with a band-aid as 
\begin{equation}
    \mathbf{Z}^{l}_{m} \rightarrow \mathbf{Z}^{l}_{m} \circ \overline{\mathbf{M}}_{p} + \mathbf{Z}^{l}_{p} \circ \mathbf{M}_{p}
\end{equation}
with the Hadamard product $\circ$ and the mask $\mathbf{M}_{p} = 1 - \overline{\mathbf{M}}_{p}$ that spatially localizes the band-aid with values of one and zero else. The manipulation of $\mathbf{Z}^{l}_{m}$ can also be done gradually, controlled with parameter $\alpha \in [0, 1]$:
\begin{equation}
    \mathbf{Z}^{l}_{m} (\alpha) = \mathbf{Z}^{l}_{m} + \alpha \mathbf{M}_{p} \circ \left(\mathbf{Z}^{l}_{p} - \mathbf{Z}^{l}_{m} \right)~,
\end{equation}
where $\alpha$ controls the degree of localizedly blending in the band-aid expression. Performing the perturbation linearly, the change in class output as well as relevance for the top-5 strongest changing channels is depicted in Supplementary Figure \ref{fig:experiments:isic-plaster-dataset} (\emph{top}). 
For each channel, the four most relevant reference samples are shown. As is visible in the plots, a mean increase in output score for class 1 can be seen for band-aid insertion. The output score for class 0 most strongly decreases. This can in fact lead to a change in prediction outcome, as is shown in Supplementary Figure~\ref{fig:appendix:isic-plaster-add}. Tracking relevance values of the individual channels shows, that channels with concepts supporting ``band-aid'' are growing in relevance with increasing perturbation. Specifically, channels with indices 248, 506 and 331 (activating for band-aids) receive more relevance on average when the band-aid concept is added.

\begin{figure*}[h]
    \centering
    \includegraphics[width=\textwidth]{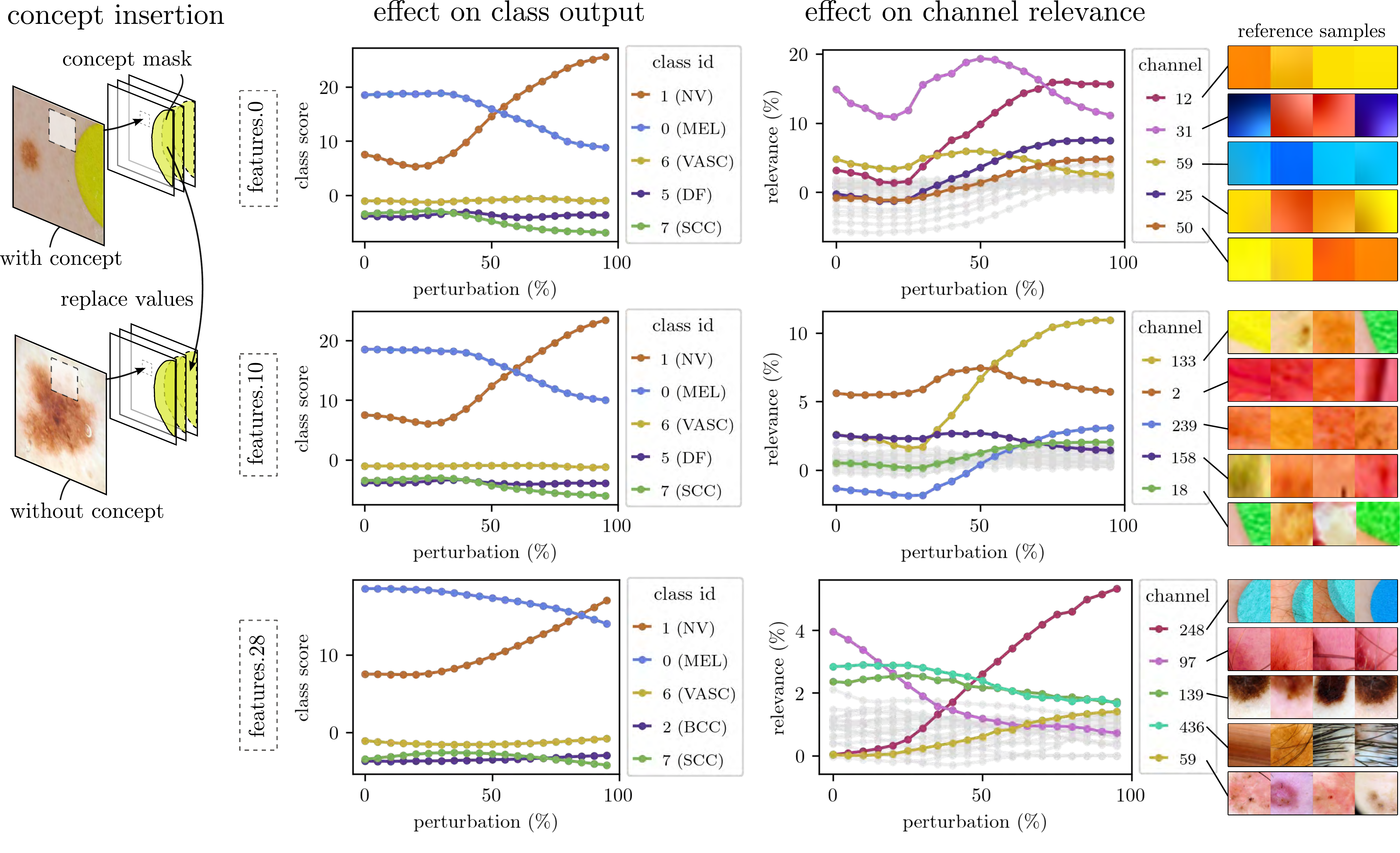}
    \caption{Manipulating decisions by inserting concept ``band-aid'' for one exemplary sample at several locations: First layer \texttt{features.0} (\emph{top}), intermediate layer \texttt{features.10} (\emph{middle}) and higher-level layer \texttt{features.28} (\emph{bottom}). For manipulation, the activation tensors are copied from a sample with a band-aid to a sample without band-aid. In order to only copy activations describing the band-aid concept, all activations are masked in the spatial dimension. The manipulation is performed linearly, and the change in class output as well as relevance for the top 5 decreasing/increasing channels is depicted. For each channel, the four most activating samples are shown. The closer the manipulation to the input, the stronger the change in class score.}
    \label{fig:appendix:isic-plaster-add} 
\end{figure*}

\begin{figure*}[h]
    \centering
    \includegraphics[width=\textwidth]{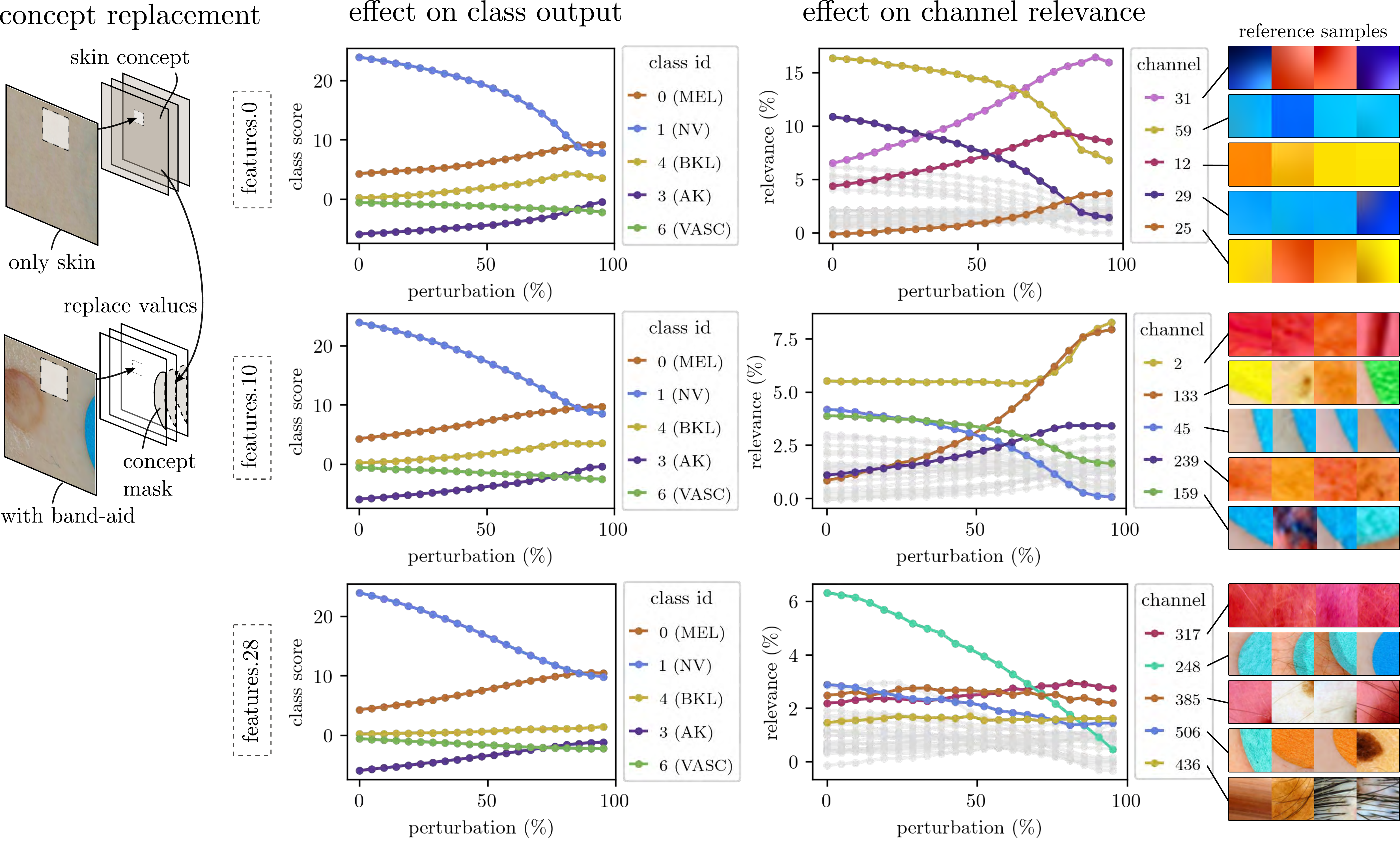}
    \caption{Manipulating decisions by replacing concept ``band-aid'' with ``skin'' at several locations: first layer \texttt{features.0} (\emph{top}), intermediate layer \texttt{features.10} (\emph{middle}) and higher-level layer \texttt{features.28} (\emph{bottom}). Therefore, the activation tensors are copied from a sample only consisting of skin to a sample with a band-aid. For the skin concept, activations are averaged over the spatial dimension and replace activations defined by a mask covering the band-aid. The manipulation is performed linearly, and the change in class output as well as relevance for the top 5 strongest changing channels is depicted. For each channel, the four most activating samples are shown.}
    \label{fig:appendix:isic-plaster-replace}
\end{figure*}

In the second example, the ``band-aid'' concept is replaced in 25 band-aid-containing samples  by a ``skin'' concept from a sample solely depicting skin. The manipulation is again performed in the higher-level layer \texttt{features.28}, as shown in Supplementary Figure \ref{fig:experiments:isic-plaster-dataset} (\emph{bottom}). Please refer to Supplementary Figure \ref{fig:appendix:isic-plaster-replace} for applications of the process to different layers. Regarding the skin concept,  activations are averaged over the spatial dimension for the skin-only input image. Thereafter, these activations replace the latent space features during predictions defined by a mask covering the band-aid. 
Specifically, the activations $\mathbf{Z}^{l}_{p}$ for the sample $p$ with a band-aid are gradually replaced with spatially averaged activations $\mathbf{\bar Z}^{l}_{s}$ from a sample $s$ consisting of skin concept using parameter $\alpha$ as
\begin{equation}
    \mathbf{Z}^{l}_{p} (\alpha) = \mathbf{Z}^{l}_{p} + \alpha \mathbf{M}_{p} \circ \left(\mathbf{\bar Z}^{l}_{s} - \mathbf{Z}^{l}_{p} \right)
\end{equation}
where $\alpha$ controls the degree of  blending in the skin expression and mask $\mathbf{M}_{p}$ localizing the band-aid spatially with values of ones and zeros else. As is visible in the plots of Supplementary Figure \ref{fig:experiments:isic-plaster-dataset} (\emph{bottom}), the replacement of the band-aid concept leads to a decrease in prediction output for class 1. This manipulation again can lead to a change in prediction outcome, as is shown in Supplementary Figure \ref{fig:appendix:isic-plaster-replace}. A decrease in the score for class 1 is expected, as in the first experiment the opposite happened when concept ``band-aid'' is inserted. Investigating the change in relevance of individual channels, we can see, that channels supporting a ``band-aid'' concept are decreasing in relevance. Channels such as (\eg, 248, 506 and 152), that previously gained relevance when ``band-aid'' was added, now lose relevance when the band-aid concept is replaced.

\begin{figure*}[h]
    \centering
    \includegraphics[width=\textwidth]{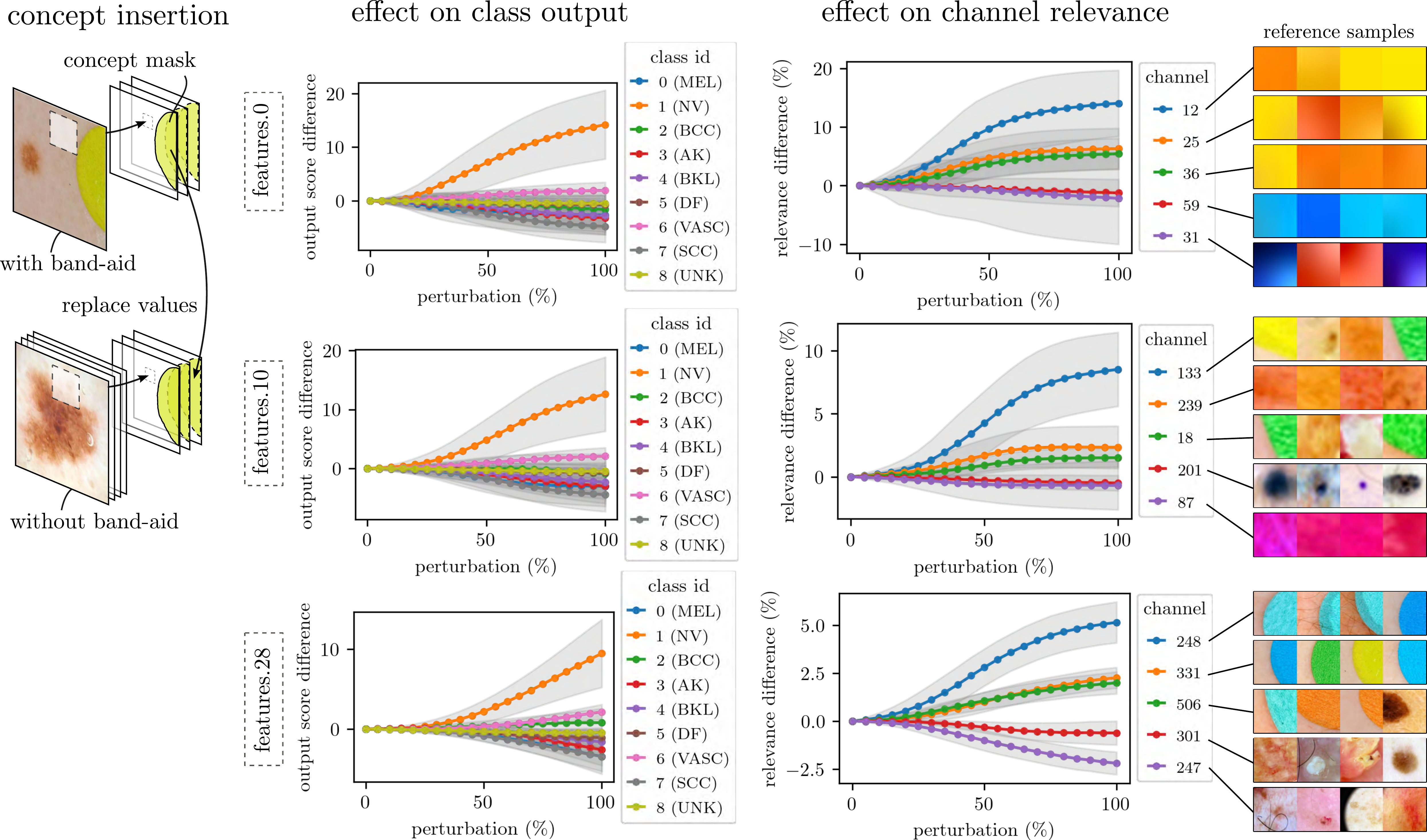}
    \caption{Manipulating decisions by inserting concept ``band-aid'' into 160 randomly drawn samples. The band-aid concept is therefore inserted in the prediction process at several locations: first layer \texttt{features.0} (\emph{top}), middle layer \texttt{features.10} (\emph{middle}) and higher-layer \texttt{features.28} (\emph{bottom}). Therefore, the activation tensors are copied from a sample with a band-aid to a sample without band-aid. In order to only copy activations describing the band-aid concept, all activations are masked in the spatial dimension. The manipulation/perturbation is done linearly, and the change in class output for the top 8 classes is depicted on the right. Adding a band-aid generally has the effect of increasing the output score of class 1. Regarding the concepts, it can be seen that channels corresponding to band-aid (\eg in color for earlier layers or color and shape for later layers) grow in terms of relevance. The standard deviation is highlighted in light-gray color. This figure extends Supplementary Figure~\ref{fig:experiments:isic-plaster-dataset}.}
    \label{fig:appendix:isic-plaster-add-dataset} 
\end{figure*}

Besides showing additional individual manipulated samples, we show in Supplementary Figure~\ref{fig:appendix:isic-plaster-add-dataset} the band-aid insertion experiment for 160 randomly drawn samples for earlier layers, \ie, \texttt{features.0} and \texttt{features.10}. Here, a similar effect as for \texttt{features.28} can be seen: Adding the band-aid concept generally has the effect of increasing the output score of class 1. Further, channels corresponding to band-aid (\eg in color for earlier layers or color and shape for later layers) grow in terms of relevance.

\paragraph{An ImageNet Use Case}
In Supplementary Figure~\ref{fig:appendix:localize-concepts:concept-atlas:swimming-trunk}, we have seen, that the model uses several concepts that do not focus directly on the swimming trunks when detecting the respective class.
\begin{figure*}
    \centering
    \includegraphics[width=1\textwidth]{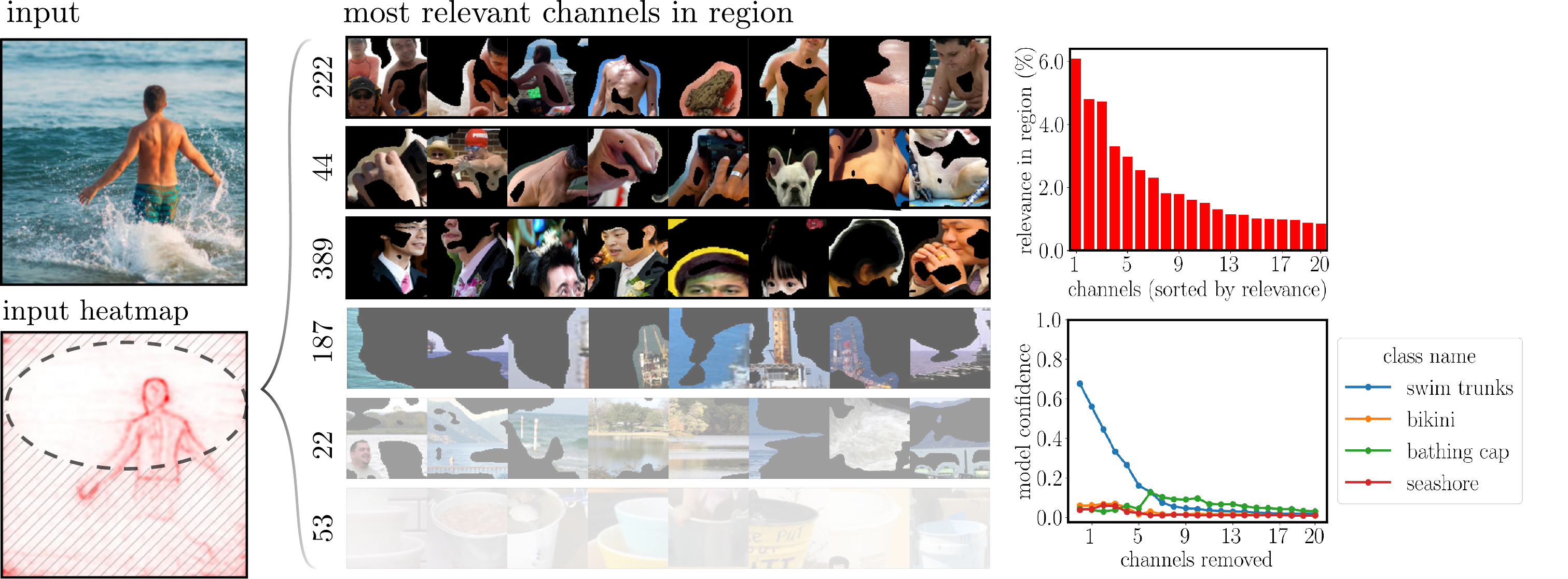}
    \caption[Feature ablation experiment for the swimming trunk classification shown in Supplementary Figure~\ref{fig:appendix:localize-concepts:concept-atlas:swimming-trunk}.]{Feature ablation experiment for the swimming trunk classification shown in Supplementary Figure~\ref{fig:appendix:localize-concepts:concept-atlas:swimming-trunk}. Local analysis of the attribution map reveals several channels (222, \edited{}{44, 389, 187, 22, 53} and more) in layer \texttt{features.28} of a VGG-16 model pretrained on ImageNet that encode for several \glsdesc{ch} features exploited by the model to detect the ``swimming trunks'' class. (\textit{Left}): Input image and heatmap. (\textit{Center}): Reference samples ${\mathcal{X}^{*}_{8}}^{\text{rel}}_{\text{sum}}$ for the 6 most relevant channels in the selected region in descending order of their relevance contribution. (\textit{Right}): Relevance contribution of 20 most relevant filters inside the region. These filters  are successively set to zero and the change in prediction confidence of different classes is recorded.}
    \label{fig:appendix:surfer_hitl}
\end{figure*}
Supplementary Figure~\ref{fig:appendix:surfer_hitl} illustrates how the model confidence changes when setting these \glsdesc{ch} artifacts to zero. The filter indices of potential \glsdesc{ch} artifacts are found by applying a local analysis on the skin of the surfer. The progressive masking or deactivation of whole latent filter channels encoding bare skin quickly decreases the model's confidence in predicting class ``swimming trunks'', verifying the \glsdesc{ch} behavior and its cause.

\subsection{Explanation-based Image Retrieval for Clever Hans Assessment Beyond Class Boundaries }
\label{sec:appendix:experiments:quantitative:imageretrieval}
In this section, we demonstrate how \glsdesc{crc} can be leveraged as a \gls{hitl} solution for dataset analysis. In a first step, we will uncover another \gls{ch} artifact, suppress it by selectively eliminating the most relevant concepts in order to assess its decisiveness for the recognition of the correct class of a particular data sample, similarly as in the previous section. Then, we utilize class-conditional reference sampling (\cf~Section~\ref{sec:appendix:experiments:qualitative:class_conditional}) to perform an inverse search to identify multiple classes making use of the filter encoding associated concept, both in a benign and a \glsdesc{ch} sense.

\begin{figure*}[t]
    \centering
    \includegraphics[width=1\textwidth]{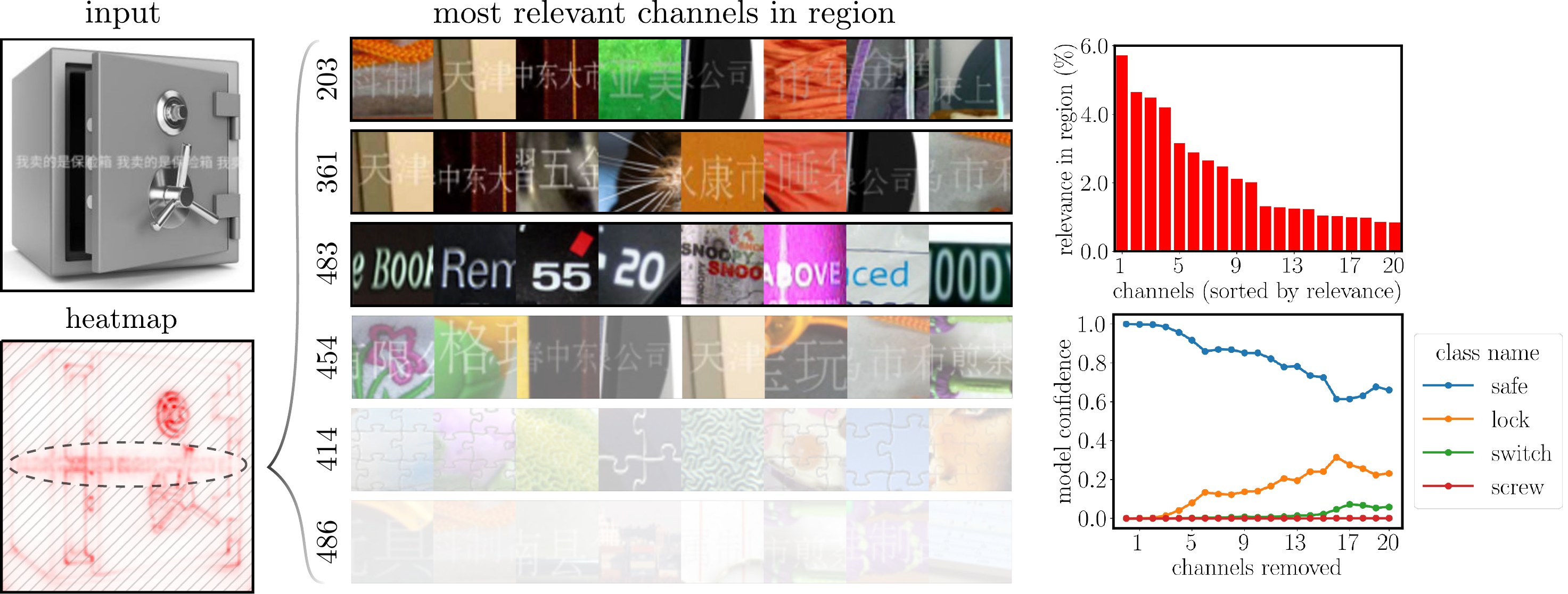}
    \caption[\glsdesc{ch} analysis on a sample of "safe" class]{Local analysis on the attribution map reveals several channels \edited{}{(203, 361, 483, 454,} 414, 486 and more) in layer \texttt{features.30} of a VGG-16 BN model pretrained on ImageNet that encode for a \glsdesc{ch} feature exploited by the model to detect the \text{safe} class. (\textit{Left}): Input image and heatmap. (\textit{Center}): Reference samples ${\mathcal{X}^{*}_{8}}^{\text{rel}}_{\text{sum}}$ for the 6 most relevant channels in the selected region in descending order of their relevance contribution. (\textit{Right}): Relevance contribution of 20 most relevant filters inside the region. These filters  are successively set to zero and the change in prediction confidence of different classes is recorded.}
    \label{fig:appendix:hitl:safe}
\end{figure*}

In Supplementary Figure~\ref{fig:appendix:hitl:safe} we analyze a sample of the ``safe'' class of ImageNet in a pretrained VGG-16 BN model. Initially, we obtain an input attribution map highlighting a centered horizontal band of the image, where a watermark is located. If we take a closer look at layer \texttt{features.30} and perform a local analysis (\cf~Section~\ref{sec:appendix:methodsindetail:local_conceptual_importance}) on the watermark, we notice that the five most relevant filters are 454, 361, 203, 414 and 486. Visualizing them using \gls{amax} as illustrated in Supplementary Figure~\ref{fig:appendix:act_max_hitl}, we conclude that they approximately encode for white strokes. Using \gls{rmax} instead (as also shown in Supplementary Figure~\ref{fig:appendix:hitl:safe}), we gain a deeper insight into the model's preferred usage and discover that the model utilizes the filters to detect white strokes in \emph{written characters}.

\begin{figure*}[h]
    \centering
    \includegraphics[width=.95\textwidth]{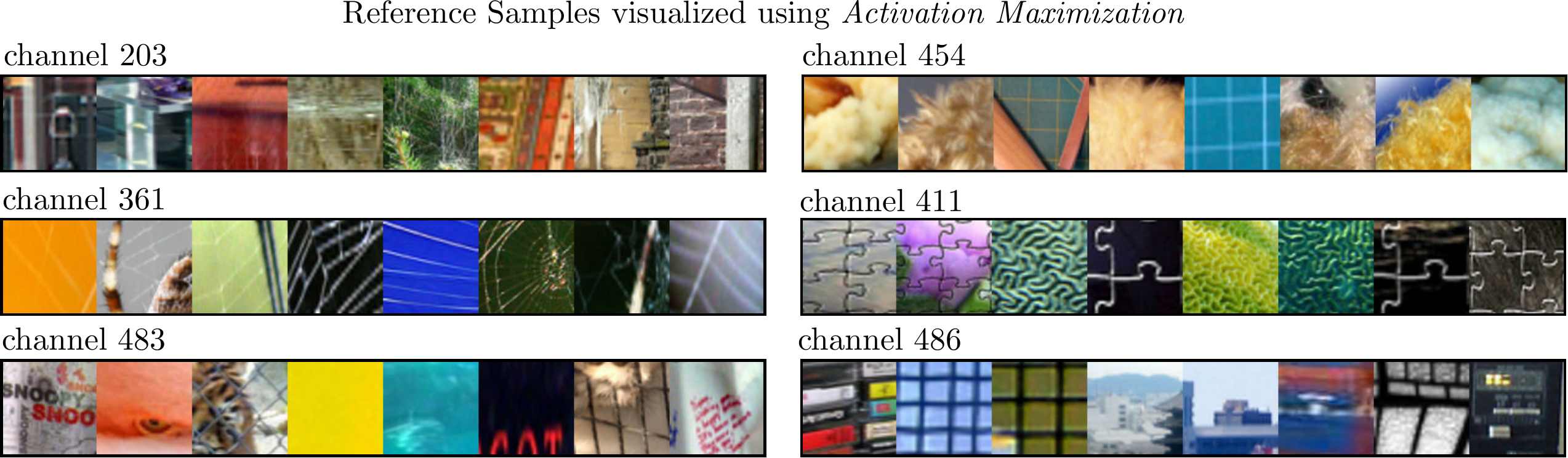}
    \caption{Reference samples ${\mathcal{X}^{*}_{8}}^{\text{act}}_{\text{sum}}$ for the six most important filters activating on the watermark of the \textit{safe} class discussed in Supplementary Figure~\ref{fig:appendix:hitl:safe}. It is to note, that regarding the most activating samples, no indication of a behavior regarding watermark signs is visible.}
    \label{fig:appendix:act_max_hitl}
\end{figure*}

To test the robustness of the model against the \glsdesc{ch} artifact, we successively set the activation output map of the 20 most relevant filters activating on the watermark to zero. In Supplementary Figure~\ref{fig:appendix:hitl:safe} (right), we record the decline of classification confidence of four classes with highest prediction confidence for this sample. From the graph, it can be inferred that the \glsdesc{ch} filters help the model in prediction, but they are not decisive for correct classification. Thus, the model relies on other potential non-\glsdesc{ch} features to detect the safe, verifying the correct functioning of the model in cases of samples without watermarks.

In an inverse search, we can now explore for which samples and classes these filters also generate high relevance. This allows us to understand the behavior of the filter in more detail and to find other possible contaminated classes. In Supplementary Figure~\ref{fig:hitl:inverse_search} are the seven most relevant classes for filter 361 illustrated. Surprisingly, many classes including ``whistle'', ``mop'', ``screw'', ``mosquito net'', ``can opener'' \text{and} ``safe'' (among others) in the ImageNet Challenge 2014 data are contaminated with similar watermarks encoded via filter 361 of \texttt{features.30} which is used for the correct prediction of samples from those classes. To verify our finding, we locate via \gls{crc} the source of the filters' relevance \wrt\ the true classes in input space, and confirm that these filters indeed activate on the characters. This implies that the model has learned a shared  \glsdesc{ch} artifact spanning over multiple classes to achieve higher accuracy in classification. The high number of contamination of samples with the identified artifactual feature could be explained by the fact, that watermarks are sometimes difficult to see with the naked eye (location marked with a black arrow) and thus slip any quality ensuring data inspection. The impact of this image characteristic can, however, be clearly marked using the \gls{crc} heatmap. Although the filter is mainly used to detect characters, there are also valid use cases for the model, such as for the puma's whiskers or the spider's web. This suggests that the complete removal of \glsdesc{ch} concepts through pruning may harm the model in its ability to predict other classes which make valid use of the filter, and that a class-specific correction~\cite{anders2022finding} might be more appropriate.

\begin{figure*}[t]
    \centering
    \includegraphics[width=1\textwidth]{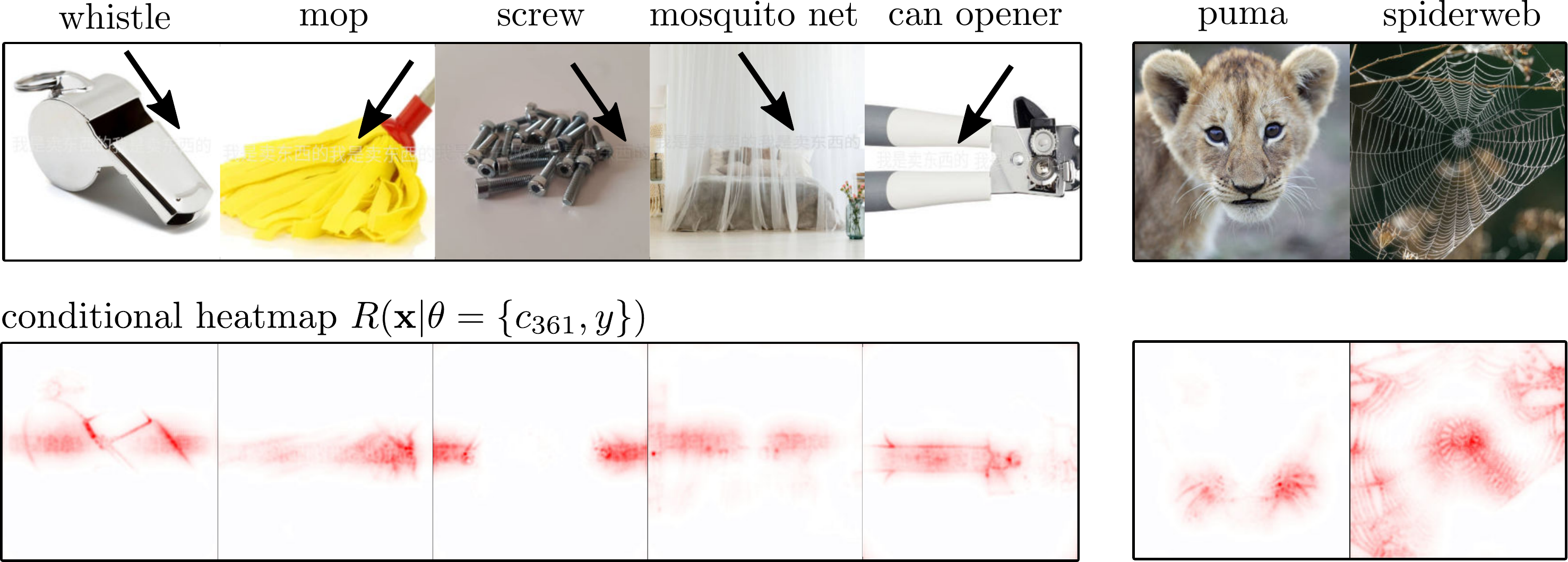}
    \caption{The \glsdesc{ch} filter 361 in layer \texttt{features.30} of a VGG-16 model with BatchNorm pretrained on ImageNet activates on samples of different classes (most relevant reference samples shown). Here, black arrows point to the location of a \gls{ch} artifact, \ie, a white, delicate font overlaid on images (best to be seen in a digital format). In the case of class ``puma'' or ``spiderweb'', the channel activates on the puma's whiskers or the web itself, respectively. Below the reference samples, the \gls{crc} heatmaps conditioned on filter 361 illustrate, which part of the attribution map results from filter 361. %
    }
    \label{fig:hitl:inverse_search}
\end{figure*}

\subsection{Assessing Concept Similarity in Latent Space}
\label{sec:appendix:experiments:quantitative:clusters}
So far throughout our experiments, we have treated single filters as functions assumed to (fully) encode learned  concepts. Consequently, we have visualized examples and quantified effects based on per-filter granularity. While previous work and our own experiments suggest that individual neurons or filters \emph{mostly} encode for a single human comprehensible concepts~\cite{zhou2015object, olah2017feature, radford2017learning, bau2020understanding, cammarata2020thread, goh2021multimodal}\footnote{The cited literature concludes that the occurrence of poly-semantic units is rare for models trained on ImageNet.}, it can generally be assumed, that concepts are encoded by sets of filters: The learned weights of potentially multiple filters might correlate and thus redundantly encode the same concept, or the directions described by several filters situated in the same layer might span a concept-defining subspace~\cite{vielhaben2022sparse}. Our proposed \glsdesc{crc} readily supports the definition of concepts as sets of multiple features.

In this section, we now aim to investigate the encodings of filters of a given neural network layer for similarities in terms of activation and use within the model. For this purpose we present a simple but qualitatively effective method for grouping similar concepts in \gls{cnn} layers: Based on the notation in previous sections, ${\mathcal{X}^{*}_{(k, q)}}$ denotes a set of $k$ reference images for a channel $q$ in layer $l$ and $\textbf{z}_q^l(\textbf{W},\textbf{x}_m)$ the ReLU-activated outputs of channel $q$ in layer $l$ for a given input sample $\textbf{x}_m$ and all required network parameters $\textbf{W}$ for its computation. Specifically, for each channel $q$ and its associated full-sized (\ie~\emph{not} cropped to the channels' filters' receptive fields) reference samples $\textbf{x}_m \in {\mathcal{X}^{\star}_{40, q}}^{\text{rel}}_{\text{sum}}$ we compute $\textbf{z}^q_m = \textbf{z}_q^l(\textbf{W},\textbf{x}_m)$, as well as $\textbf{z}^p_m = \textbf{z}_p^l(\textbf{W},\textbf{x}_m)$ for all other channels $p \neq q$, by executing the forward pass, yielding activation values for all spatial neurons for the channels\footnote{Here, $k=40$ is an arbitrary choice, resulting from the common minimal size of available pre-computed reference sample sets for all filters.}. We then define the averaged cosine similarity $\rho_{qp}$ between two channels $q$ and $p$ in same layer $l$ as
\begin{align}
\rho_{qp} & = \frac{1}{2} \left( \cos(\phi)_{qp} + \cos(\phi)_{pq} \right) \label{eq:appendix:rho}\\
 \text{with} \qquad \cos(\phi)_{qp} & =  \frac{1}{k} \sum_{\textbf{x}_m \in {{\mathcal{X}^{\star}_{(k, q)}}^{\text{rel}}_{\text{sum}}}} \frac{\textbf{z}^q_m \cdot \textbf{z}^p_m}
{||\textbf{z}^q_m|| \cdot ||\textbf{z}^p_m||}~. \label{eq:appendix:cos}
\end{align}
Note that we symmetrize $\rho_{qp}$ in Equation~\eqref{eq:appendix:rho} as the cosine similarities $\cos(\phi)_{qp}$ and $\cos(\phi)_{pq}$ are in general not identical, due to the potential dissimilarities in the reference sample sets ${\mathcal{X}^{*}_{(k, q)}}$  and ${\mathcal{X}^{*}_{(k, p)}}$. Thus, $\cos(\phi)_{qp}$ measures the cosine similarity between filter $q$ and filter $p$ \wrt the reference samples representing filter $q$. The from Equation~\eqref{eq:appendix:rho} resulting symmetric similarity measures $\rho_{qp} = \rho_{pq} \in [0,1]$ can now be clustered, and visualized via a transformation into a distance measure\footnote{Note that normally, the output value of the cosine distance covers the interval $[-1,1]$, where for $-1$ the two measured vectors are exactly opposite to one another, for 1 they are identical and for 0 they are orthogonal. In case output channels of dense layers are analyzed, \ie scalar values, the range of output values reduces to the set $\lbrace -1,0,1 \rbrace$, as both values are either of same or different signs, or at least one of the values is zero. Since we are processing layer activations \emph{after} the ReLU nonlinearities of the layer, which yields only positive values for $\textbf{z}^q_m$ and $\textbf{z}^p_m$. This results in $ \rho_{pq} \in [0,1]$, and a conversion to a canonical distance measure $d_{qp} \in [0,1]$.} $d_{qp} = 1 - \rho_{qp}$ serving as an input to t-SNE \cite{van2008visualizing} which clusters similar filters together in, in our case, $\mathbb{R}^2$.

\begin{figure*}[t]
    \centering
    \includegraphics[width=1\textwidth]{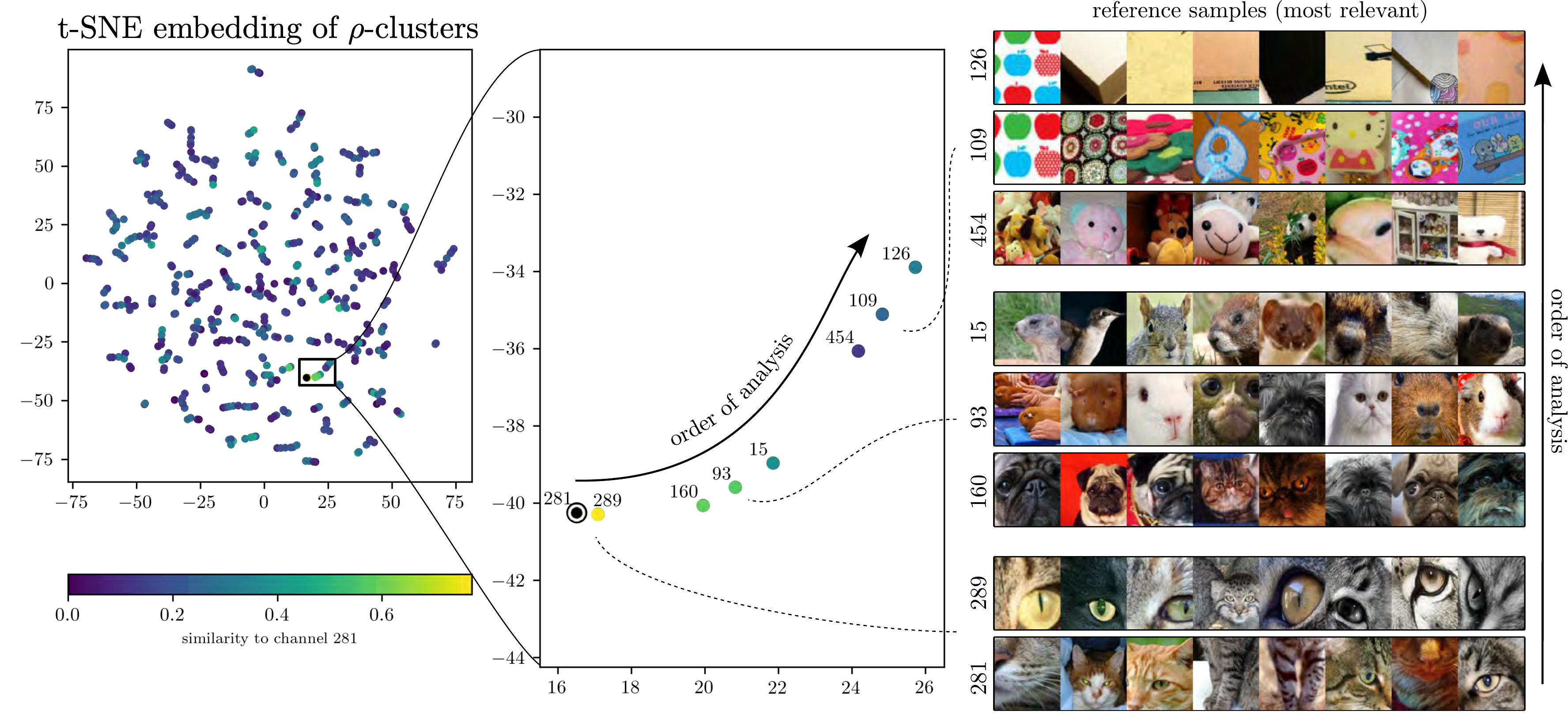}
    \caption
    {\emph{(Left and center)}: Filter channels from \texttt{features.40} of a VGG-16 with BatchNorm, clustered and embedded according to $\rho$-similarity. Markers are colored according to their $\rho$-similarity to filter 281. \emph{(Right)}: Reference examples are selected based on \gls{rmax} summation. Our analysis has identified a cluster of related filters which, when observed according to decreased similarity to filter 281, describes a transition of simplification from cat faces and eyes over other animals with flat faces and dark eyes to simple colored shapes with dots.}
    \label{fig:appendix:experiments:tsne_squishy}
\end{figure*}

Supplementary Figure~\ref{fig:appendix:experiments:tsne_squishy} shows an analysis result focusing on a cluster around filter $q = 281$ from \texttt{features.40} of a VGG-16 network with BatchNorm layers trained on ImageNet, representing facial features and fur patterns of house cats,  with the most similar neighbor in terms of activatory behavior being filter 289, representing zoomed-in representations and facial features of tabby cats. Investigating the other neighboring filter of the cluster by following the spatial embedding and thus roughly their $\rho$-similarity to filter 281, we can for one observe clear conceptual similarities of the filters as represented by their ${\mathcal{X}^{\star}_{8}}^{\text{rel}}_{\text{sum}}$ reference sample sets. Furthermore, we can observe a \emph{change} in the nature of represented concepts: While filters 281 and 289 represent cat faces and fur pattern, as well as cat eyes in close-up view, the filters 160, 93 and 16 pick up on the concept of big eyes and respectively seem to represent large-eyed and flat nosed Persian cats and pug dogs in frontal view (filter 160), flat-faced animals with large black eyes and brown to white fur patterns (filter 93), and faces of animals of brown to gray fur with small black eyes (filter 15). After a larger step in embedding space, filters 454, 109 and 126 seem to further simplify the progression of animal face representations to black or large dots on uniformly colored round shapes (filter 454), to colorful shapes of lower complexity with dots (filter 109) to ultimately almost uniformly colored objects (filter 126), seemingly completing a transition to simpler and more abstract concepts starting at filter 281.

We discuss another analysis in Supplementary Figure~\ref{fig:appendix:experiments:tsne_buttons} focussing on filter $q = 446$ in \texttt{features.40} of the same model, showing various types of typewriter and rectangular laptop keyboard buttons and roofing shingles photographed in oblique perspective, as well as round buttons of typewriters, remote controls for televisions, telephone keys and round turnable dials of various devices and machinery. Other than the concepts discovered in context of Supplementary Figure~\ref{fig:appendix:experiments:tsne_squishy} and filter 281, the filters around filter 446 seem to cover different aspects of a shared ``button'' or ``small tile'' concept. The filters located in this cluster have been identified as similar due to their similar activations over the combined reference sample sets. Assuming redundancy based on the filter channels' apparently similar activation behavior, the user could merge or cluster them to one encompassing concept, thereby simplifying interpretation by reducing the number of filters in the model. We therefore further investigate the filters 7, 94, 446 and 357 (all showing buttons or keys) in order to find out whether they encode a concept collaboratively, whether they are partly redundant, or whether the cluster serves some discriminative purpose.

\begin{figure*}[t]
    \centering
    \includegraphics[width=1\textwidth]{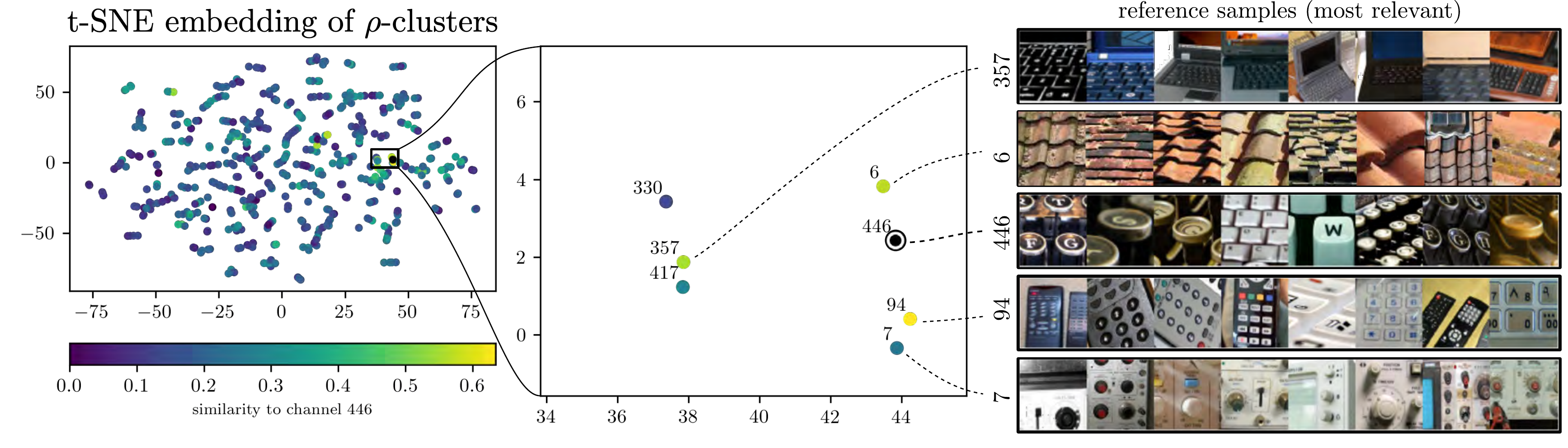}
    \caption
    {\emph{(Left and center)}: Filter channels from \texttt{features.40} of a VGG-16 with BatchNorm, clustered and embedded according to $\rho$-similarity. Markers are colored according to their $\rho$-similarity to filter 446. \emph{(Right)}: Reference examples are selected based on \gls{rmax} summation. (\emph{Center and Right}): One particular cluster around channel 446 is shown in more detail with five similarly activating channels and their reference images obtained via \gls{rmax}. As per the reference images, the over-all concept of the cluster seems to be related to rectangular roofing shingles and keyboard keys, as well as round buttons.} 
    \label{fig:appendix:experiments:tsne_buttons}
\end{figure*}

We observe that the filters are most relevant for classes ``laptop computer'' and ``remote control''. Consequently, we compute neuron activations and attribution scores for samples from both classes and observe the attributed relevance to those features when performing \gls{crc} \wrt~the ground truth class output. See Supplementary Figure~\ref{fig:appendix:experiments:cluster_buttons} for visualizations. Regardless of whether an instance from class ``laptop'' or ``remote control'' is chosen as input, the activation map across the observed channels is similar for each instance. The per-channel \gls{crc} attribution map however reveals that while all filters react to similar stimuli in terms of activations, the model seems to use the subtle differences among the observed concepts to distinguish between both classes they are most relevant for: For both class ``laptop'' and ``remote control'', buttons are striking and defining features, and all observed filters activate for button features. However, when computing relevance scores for class ``laptop'', the activating filters representing round buttons (filters 7, 94, and 446) dominantly receive negatively signed attribution scores, while filter 357 clearly representing typical keyboard button layouts receives high positive relevance cores. For samples of class ``remote control'', the computation of relevance scores \wrt their true class\footnote{or even the computation of \gls{crc} for samples from class ``laptop'' \wrt~the class output for ``remote control''} yields almost exactly opposite attributions, indicating that filters encoding round buttons and dials (filters 94 and 7) provide evidence for class ``remote control'', while the activation of channel 357 clearly speaks against the analyzed class visible in relevance attribution. In both relevance analyses, however, filter 446 receives no attributions, presumably as it represents a particular expression of both round and angular buttons which fits (or contradicts) neither of the compared classes particularly well. In fact, filter 446 is highly relevant for class ``typewriter keyboard'' instead.

In conclusion, we report that although several filters may show signs of correlation in terms of output activation, they are not necessarily encoding redundant information or are serving the same purpose. Conversely, using our proposed \glsdesc{crc} in combination with the \glsdesc{rmax}-based process for selecting reference examples representing seemingly correlating filters, we are able to discover and  understand the subtleties a neural network has learned to encode in its latent representations 
\begin{figure*}[t]
    \centering
    \includegraphics[width=1\textwidth]{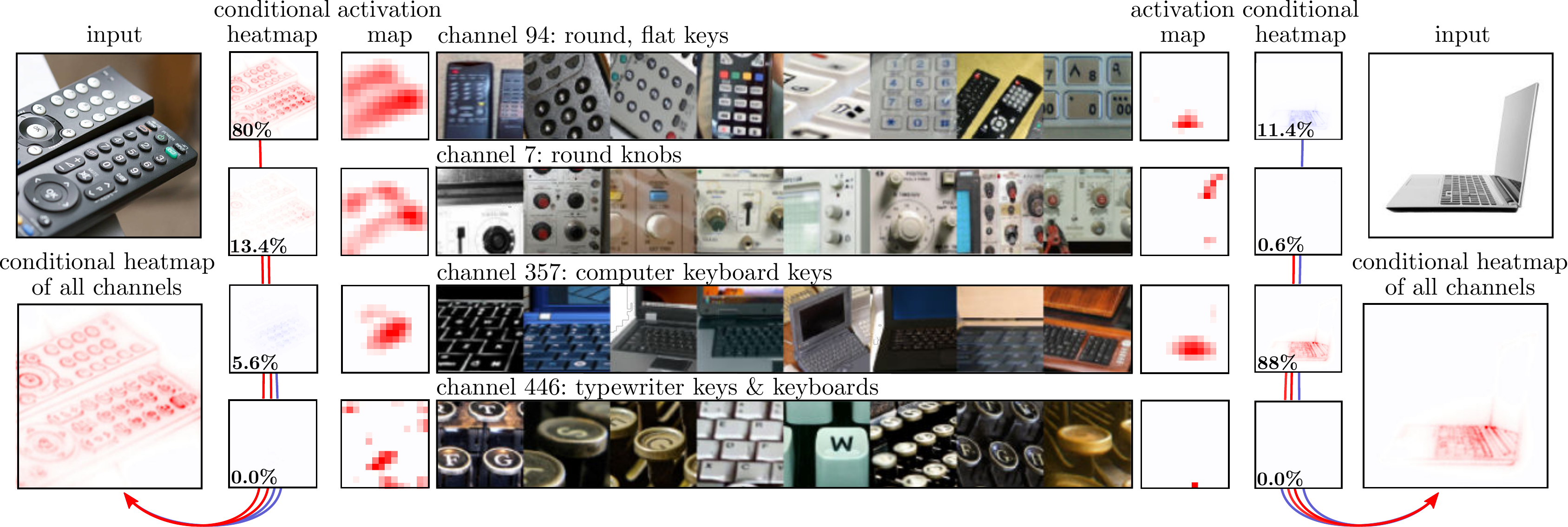}
    \caption{Relevance-based investigation of similarly activating channels identified in context of Supplementary Figure~\ref{fig:appendix:experiments:tsne_buttons}. \textit{(Center)}: Reference examples for the analyzed channels 7, 04, 357 and 446 from \texttt{features.40} from a VGG-16 with BatchNorm layers. \textit{(Left)}: Exemplary input from class ``remote control'' with per-channel activation maps and respective ground truth conditional relevance maps, as well as their aggregation $\theta=\{y, c_{94}, c_{7}, c_{357}, c_{446} \}$ (bottom left). \textit{(Right)}: Exemplary input from class ``laptop computer''  with per-channel activation maps and respective ground truth conditional relevance maps, as well as their aggregation. Conditional relevance attributions $R(x|\theta)$ are normalized \wrt~the common maximum amplitude. Similarly \emph{activating} channels do not necessarily encode redundant information, but might be used by the model for making fine-grained distinctions, which can be observed from the attributed relevance scores.}
    \label{fig:appendix:experiments:cluster_buttons}
\end{figure*}

We present some additional interesting findings below. An additional t-SNE cluster example is shown in Supplementary Figure~\ref{fig:appendix:tsne_water}, visualizing filters spanning a conceptual subspace of water- , sea floor- and ground surfaces. Specifically, a total six channels forming a chain in the embedding space are displayed with their most relevant reference samples. Starting with channel 23 relevant for waves on the water surface seen from above, the next similar channel is channel 18 activating for a similar concept with differently colored water. Channel 48 corresponds to turquoise water above and below the surface. Thereafter, channel 99 and 325 are characterized by concepts describing corals and stones underwater. Eventually, channel 481 is used for grass and sand textures. This example illustrates how a semantic transition in terms of concepts can be seen discovered when analyzing the activations of latent neurons based on \gls{rmax} reference examples. 
\begin{figure*}[h]
    \centering
    \includegraphics[width=1\textwidth]{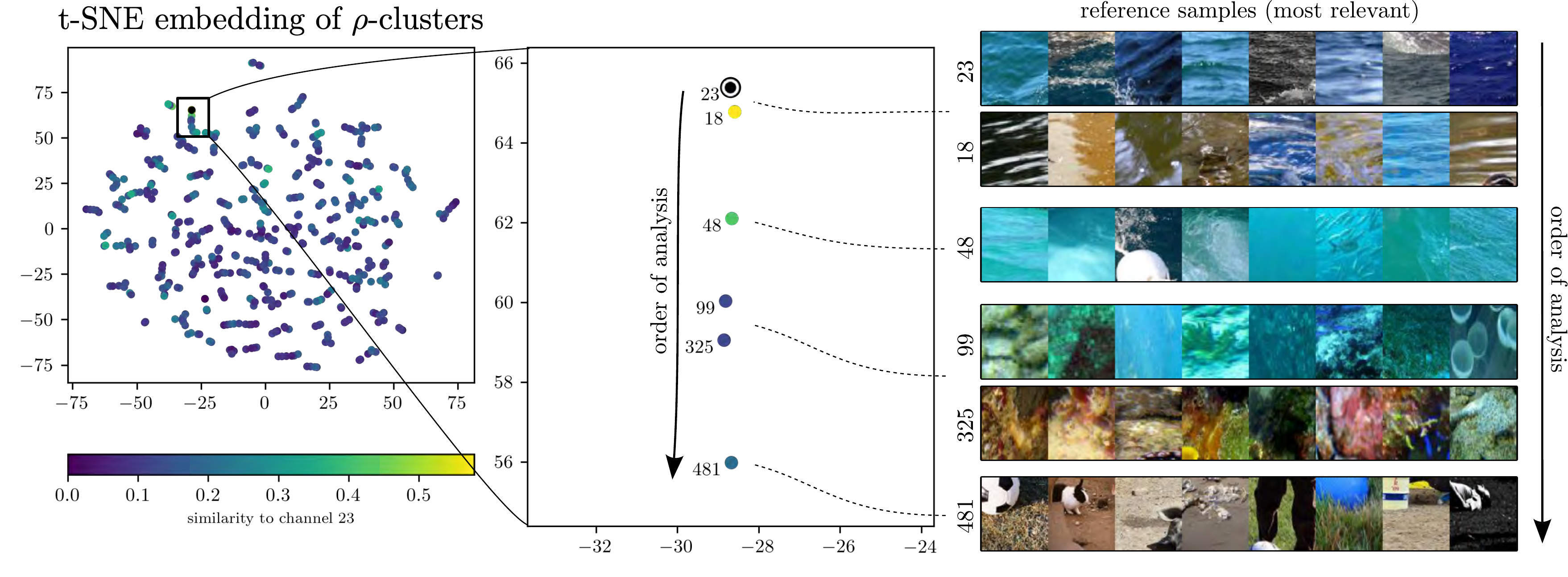}
    \caption{Channels can be clustered using the symmetric cosine similarity, and visualized using the t-SNE algorithm. (\emph{Left}): Several clusters can be found when layer \texttt{features.40} of a VGG-16 with BatchNorm layers trained on ImageNet is analyzed, indicating many distinct groups of simultaneously firing sets of neurons. (\emph{Right}): One particular cluster forming a line in the embedded space is shown in more detail with a total of six channels and their reference images obtained via \gls{rmax}. As is visible from the reference images, the over-all concept of the cluster is related to \textit{water}. Starting with channels 23 and 18, which are relevant for waves on the water surface. The next similar channel is channel 48 activating for turquoise water above and below the surface. Thereafter, channel 99 and 325 are characterized by concepts describing corals and stones underwater. Eventually, channel 481 is used for grass and sand textures. This example illustrates how a similarly activating filters may span semantically related topics, and semantic transitions are seemingly possible in related filter sets.}
    \label{fig:appendix:tsne_water}
\end{figure*}
\begin{figure*}[h]
    \centering
    \includegraphics[width=1\textwidth]{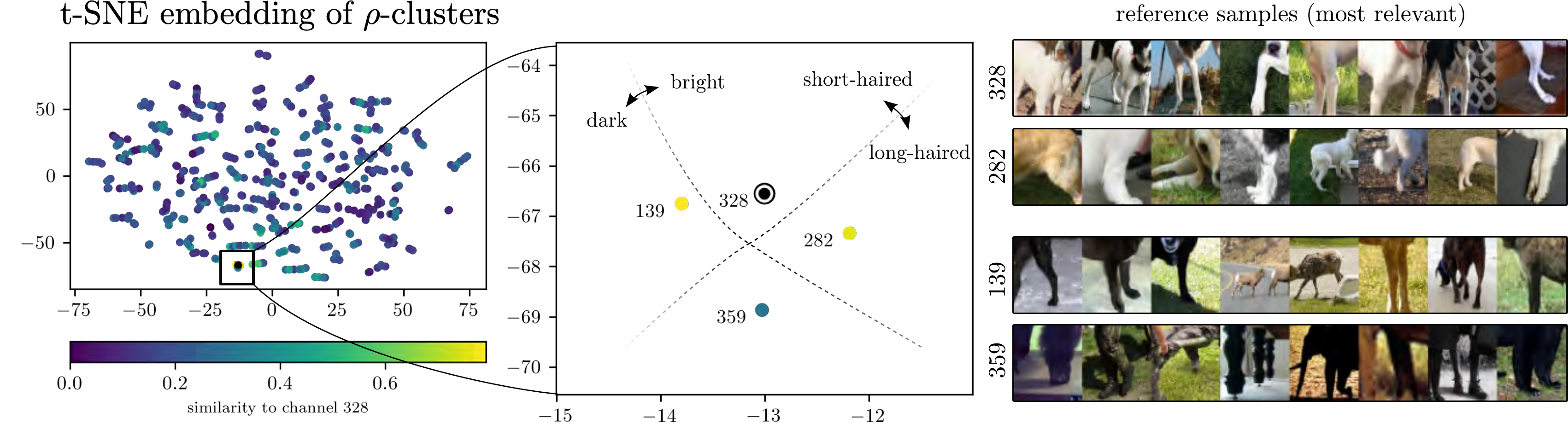}
    \caption{Channels can be clustered based on their symmetrized cosine similarity and correspondingly embedded using the t-SNE algorithm.. (\emph{Left}): Several clusters can be found when layer \texttt{features.40} of a VGG-16 with BatchNorm layers trained on ImageNet is analyzed. (\emph{Right}): One particular cluster around channel 328 is shown in more detail with a total of four channels and their reference images obtained via \gls{rmax}. As is visible from the reference images, the over-all concept of the cluster is related to  ``dog legs''. Here, the embedding allows splitting the space into several subspaces: Channels 328 and 282 are relevant for ``bright-colored fur'', whereas channels 139 and 359 for ``darker-colored fur''. On the other hand, channels 139 and 328 react for ``long-haired fur'' and channels 359 and 282 for ``short-haired fur''. }
    \label{fig:appendix:tsne_dogs}
\end{figure*}
Another example of finding channels with similar concepts using t-SNE is shown in Supplementary Figure~\ref{fig:appendix:tsne_dogs}, where four channels can be found that are relevant for the recognition of legs of dogs and other animals (and objects). Here, the embedding allows splitting the space into several subspaces: Channels 328 and 282 are relevant for ``bright-colored fur'', whereas channels 139 and 359 for ``darker-colored fur''. On the other hand, channels 139 and 328 react for ``long-haired fur'' and channels 359 and 282 for ``short-haired fur''. We might have identified a group of channels encoding the common concept of ``legs'' with degrees of freedom with regards to brightness and ``leg volume''.
\begin{figure*}[h]
    \centering
    \includegraphics[width=1\textwidth]{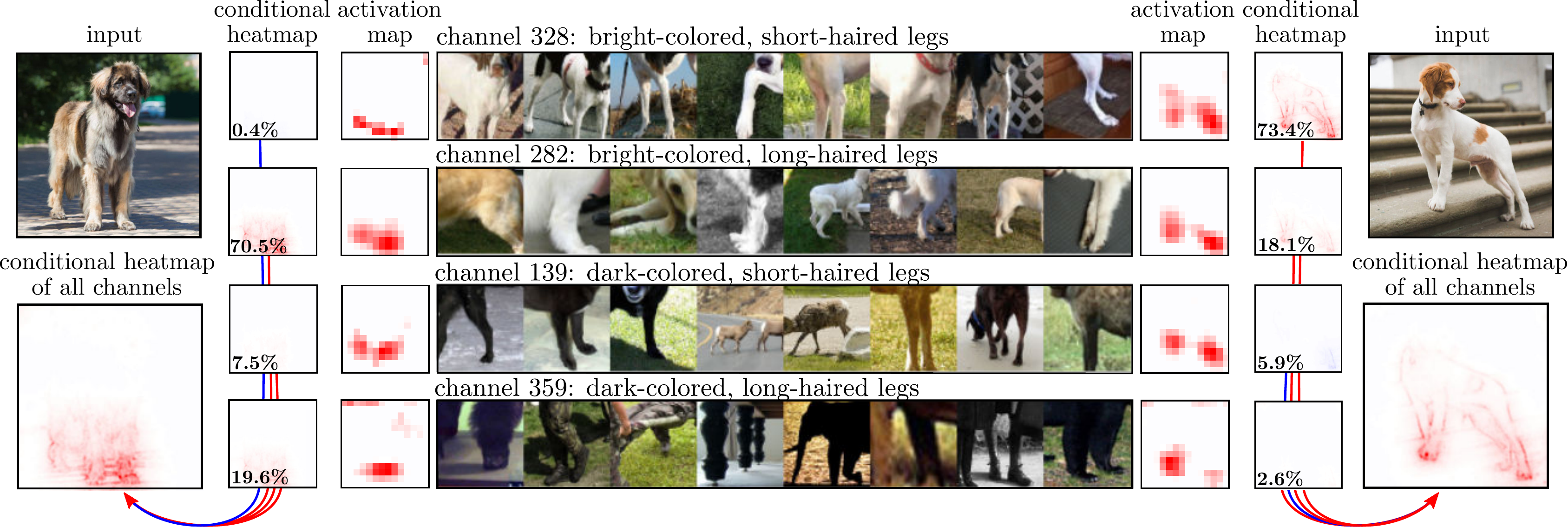} 
    \caption{Several similar filters in layer \texttt{features.40} of a VGG-16 BN model trained on \textit{ImageNet} have high symmetrized cosine similarity (see Supplementary Figure~\ref{fig:appendix:tsne_dogs}), but are used differently by the model during predictions. These presented filters commonly activate for different kind of dog legs, as is visible on the conditional heatmaps as well as channel maps. Their different function is illustrated by the prediction of two different dog samples. (\emph{Left}): Channels 282 and 359 are strongly used to predict a brown-colored and long-haired dog, as relative relevance scores show. (\emph{Right}): For the other example, channel 328 and 282 is strongly used for classifying a short-haired, bright-colored dog.}
    \label{fig:appendix:cluster:legs}
\end{figure*}
How these similar channels are differently used in predictions is shown in Supplementary Figure~\ref{fig:appendix:cluster:legs}.

\cleardoublepage
\clearpage
\section{
Additional Experiment: Towards XAI for Fairness in ML
}
\label{sec:appendix:fairness}

So far, we have demonstrated qualitatively how the reasoning process of a \gls{dnn} can be understood using \gls{crc} and \gls{rmax}. 
Through a user study (see Section~\ref{sec:results:study} in the main manuscript) we have further been able to quantify that 
our proposed set of novel techniques can be leveraged effectively to reveal shortcuts in a model prediction that exploit \gls{ch} artefacts,  on a conceptual and human-understandable level.

In this section we aim to demonstrate how our proposed techniques can be harnessed in probable real world applications, \eg,  to expose discriminatory behavior against demographic groups in high-stake decisions potentially inflicting harm on individuals. 
This is specially relevant for applications with ethical and judicial implications, where decisions must adhere to moral and objective standards, as requirements set in governmental regulatory frameworks~\cite{goodman2017european}. 
Examples for such discriminatory behavior include a recruiting tool for STEM job openings conjecturing that females are less qualified than men~\cite{dastin2018amazon} or a software for rating a defendant’s risk of future crime systematically estimating African Americans having a higher risk than White American inmates~\cite{angwin2016machine}. The following experiment is merely intended as a brief excursion to the topic of \gls{xai}-supported fairness in \gls{ml}. A detailed treatment is subject to future work.

\begin{figure*}[h]
    \centering
    \includegraphics[width=1\textwidth]{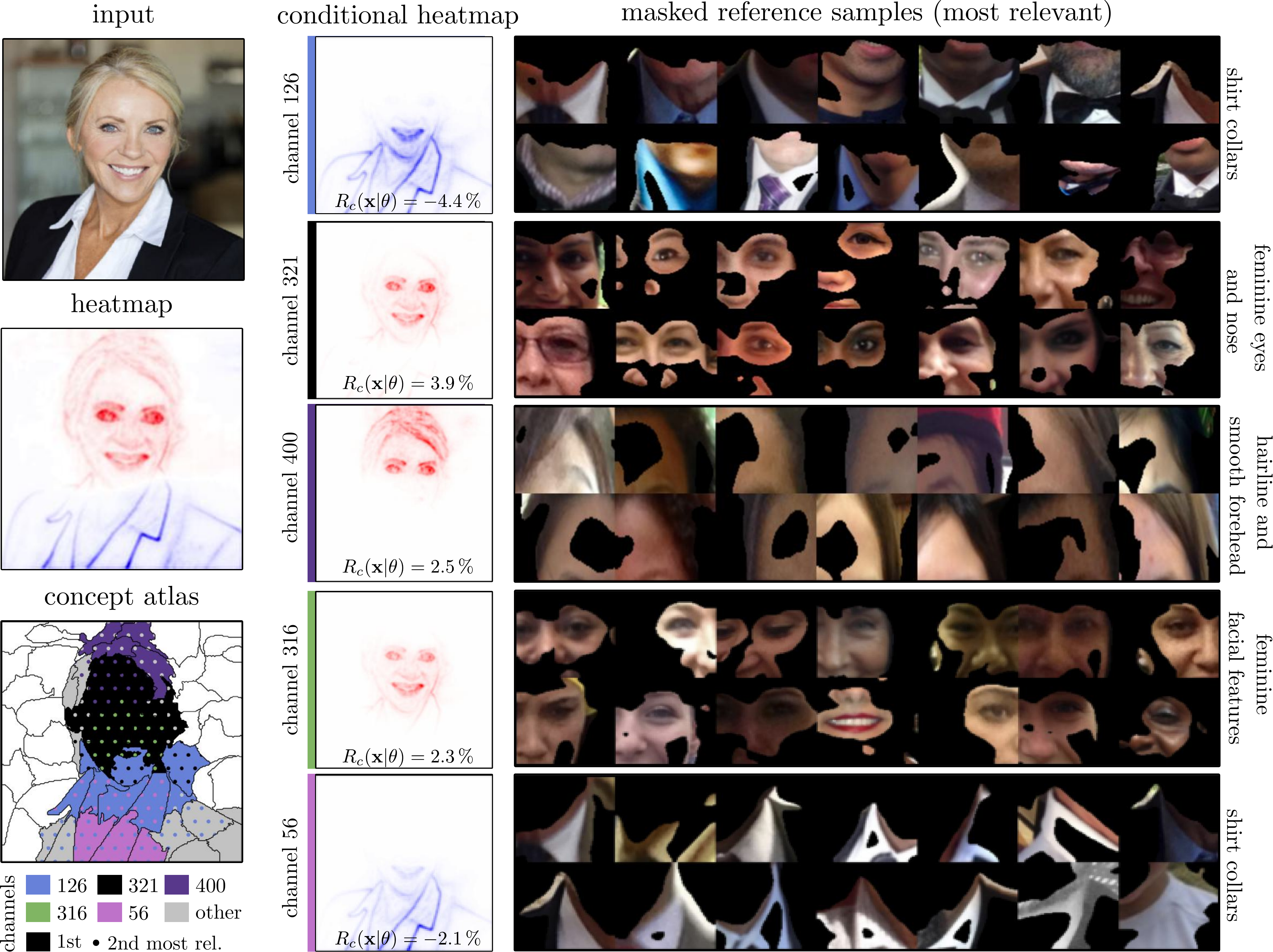} 
    \caption{
    The Concept Atlas explanation for a sample with ground truth label ``female'', which has been predicted correctly. The explanation is provided \wrt the outcome of class ``female'' using a VGG-16 model pre-trained on \textit{ImageNet} and fine-tuned on the \textit{Adience} dataset and the male vs.\ female face classification task.
    The most relevant (absolute value) channels of layer \texttt{features.26} can be identified with concepts encoding for ``feminine eyes and nose'' (channel 321), ``hairline and smooth forehead'' (channel 400), ``feminine facial features'' (channel 316) and `` shirt collars'' (channel 126 and 56).
    }
    \label{fig:appendix:fairness}
\end{figure*}

For the purpose of this demonstration, we trained a VGG-16 model on the openly available Adience dataset~\cite{eidinger2014age} for male vs.\ female face classification (details on training are in  Section~\ref{ap:adience}). Supplementary Figure~\ref{fig:appendix:fairness} illustrates a concept atlas for a sample of class “female” that was correctly predicted.
However, at the bottom part of the heatmap visualization to the left an aggregation of negative relevance scores is recognizable.
Especially the collar of the person's clothing receives negative relevance, indicating some visual evidence against the categorizaton of this sample into the output category ``female''.
If only the single heatmap visualization were to be considered for interpretation, a user might infer that the striking yellow color induces an unexpected model behavior.
However, examining the first and fifth-most influential concept as shown in the Concept Atlas (channel 126 and 56 in Supplementary Figure~\ref{fig:appendix:fairness})
clarifies that the presence of the shirt collar as a particular clothing item, often related to the presence of a (bow) tie and male-looking chings, visual feature recognized and assessed as contradictory by the model. 

Given this singular instance of inference and explanation, we are able to recognize that the
model has learned a feature systematically,
that encodes for collars of suits and shirts, which is responsible for the attribution of negative relevance due to its recognition as a non-female feature.
Thus, conversely, the model has been exposed to over-associate the attribute ``clothes with shirt collars'' with the possible ``male'' outcome in this binary prediction problem, eventually causing discriminatory behavior in settings where an incorrect predictions due to a bias related to particular clothing items might have a negative impact on the well-being of individuals.

\begin{figure*}[h]
    \centering
    \includegraphics[width=1\textwidth]{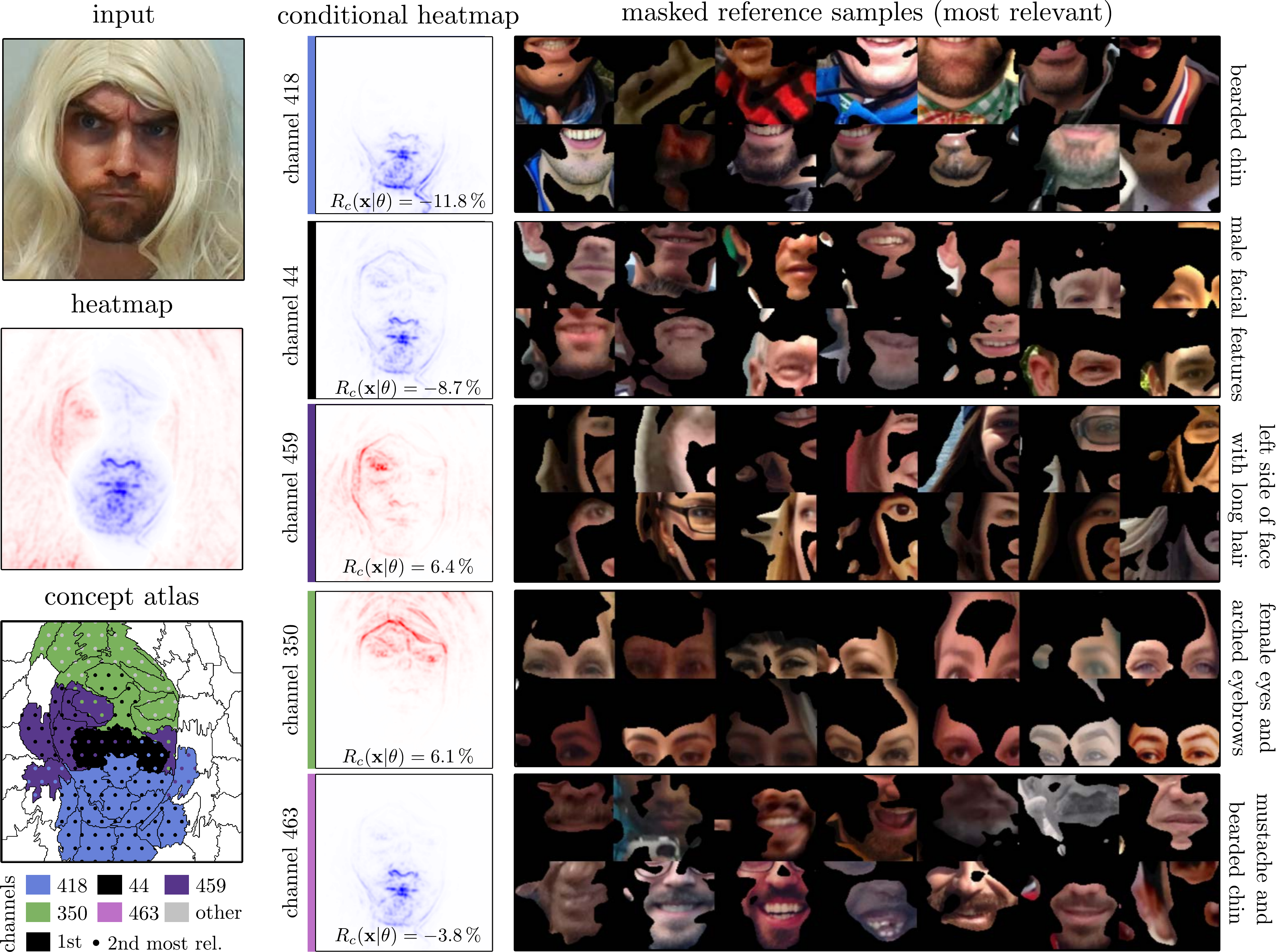} 
    \caption{
    The Concept Atlas explanation for a image of one of the male authors, wearing a blonde hairpiece, predicted ``female'' by the model and explained as such.
    The here used VGG-16 model was pre-trained, the input image classified and the prediction explained \wrt “female” using a VGG-16 model pre-trained on \textit{ImageNet} and fine-tuned on the sex recognition task of the \textit{Adience} dataset.
    The most relevant (absolute value) channels of layer \texttt{features.28} can be identified with concepts encoding for beards (channel 418, 44 and 463), indicating striking feature against the model's prediction of class ``female'', 
    which have been outweighed by features indicating the presence of (for the model) prototypically female attributes, such as
    ``left side of head with long hair and female eyes'' (channel 459) and ``female eyes and eyeborws'' (channel 350), with other concepts not shown in the image.
    }
    \label{fig:appendix:adience_concept_atlas}
\end{figure*}

Supplementary Figure~\ref{fig:appendix:adience_concept_atlas} shows another instance of a male face with long hair, which has been mis-predicted by the model as ``female'' based on a learned stereotype related to hairstyle.
Provided explanatory information as shown in the figure, developers of machine learning models and datasets do have access to detailed information about systematic dataset biases as learned as well as abused by the analyzed model.
One next reasonable step for developers could therefore be extension and diversification, as well as a more careful curation of the training data the model has been trained on, based on the insights from our \gls{xai} approach.

\cleardoublepage
\clearpage
\section{Additional Experiment: Concept Analysis on Time Series Data}
\label{sec:appendix:timeseries}
The application of explanation methods is not constrained to the image domain and has in the past seen utilization in numerous other data domains~\cite{samek2021explaining}. Also as long as the data domain can be meaningfully visualized in an interpretable way, obtaining concept-based explanations with \gls{crc} is possible. 

Here, we present a use-case on how \gls{crc} can be utilized in a \gls{cnn} model trained on Electrocardiography (ECG) arrhythmia data to verify the concepts used during inference and to open up the door to potential use-cases for scientific discovery in the future.  A decisive factor for interpretability is choosing the right representation for the input domain. For example in sex classification on voice recordings a frequency representation is more advantageous in terms of \gls{xai} interpretability since here,  pitches are easier to distinguish for the naked eye~\cite{becker2018interpreting}, than in time domain. For ECG data, the traditional time series representation is much more comprehensible for physicians, because they are familiar with this visualization on which characteristic markers have been defined in literature~\cite{macfarlane2010comprehensive}. The original training data is provided by PhysioNet MIT-BIH Arrhythmia~\cite{moody2001impact} and consists of ECG recordings from 47 different subjects, where each beat was annotated by at least two cardiologists. The work~\cite{kachuee2018ecg} reordered the dataset by isolating the ECG lead II data, splitting and padding beats into a fixed size of 150 at a sampling rate of 125 Hz and finally grouping all annotations into five different categories in accordance with the \textit{Association for the Advancement of Medical Instrumentation} (AAMI) EC57 standard~\cite{a1998testing}. We train a model containing multiple layers of 1D-convolutional and densely connected layers. Details on the architecture and training procedure can be found in Appendix~\ref{ap:ecg}. Assuming concepts are confined to single channels, we visualize again only single filters using \gls{rmax}. The attribution maps are projected to the input data domain and visualized as line plots. Similar to the image domain setting in Section~\ref{sec:appendix:methodsindetail:understanding:scaling}, we infer the receptive field at the hidden layers in order to identify and crop the most representative signal patterns for individual filters. In the future, this visual representation could be augmented by symbolic regression or other more precise mathematical descriptors of the concepts, such as the Romhilt-Estes score system \cite{estes2015romhilt}.

\begin{figure*}[t!]
    \centering
    \includegraphics[width=1\textwidth]{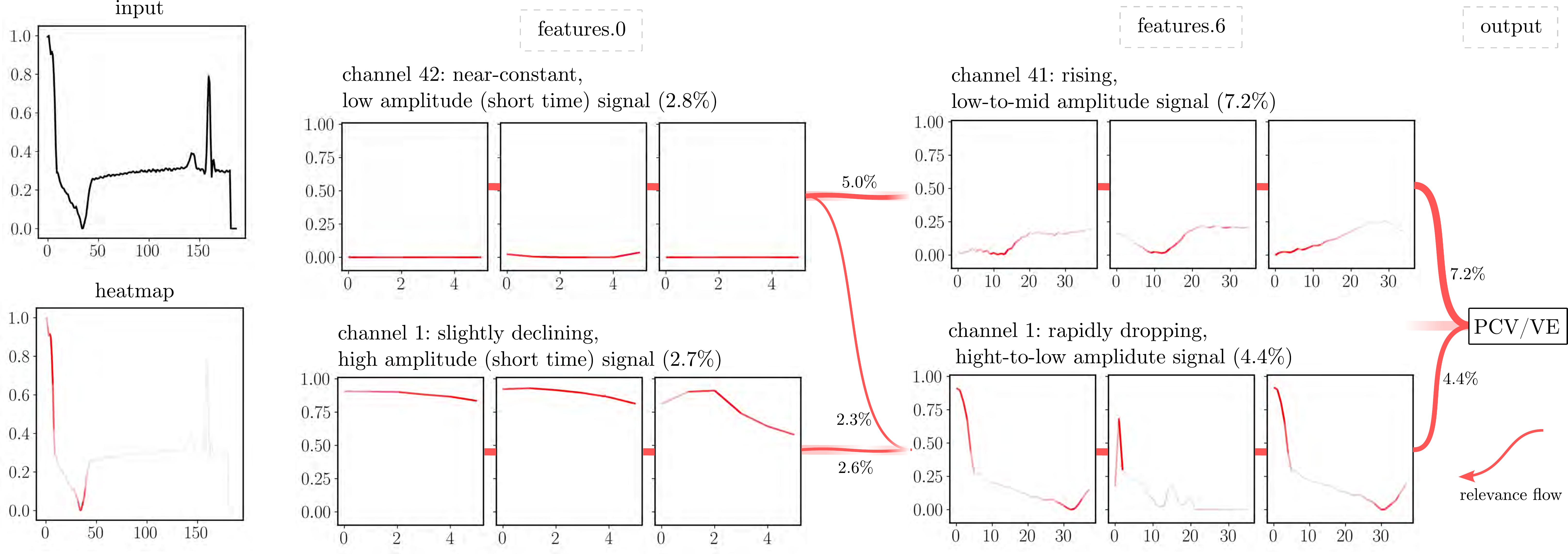}
    \caption{Glocal analysis of ECG time series data. The heatmap of the signal indicates, that the strong decline in the beginning and slight rebound in signal thereafter are the main elements the model is focusing on. Investigating individual layers with their channels, specific relevant units can be identified with concepts corresponding either to the strong drop or rebound. Specifically, in the lower layer \texttt{features.0} channel 42 encodes for low signal amplitudes, whereas channel 1 represents high amplitude signals. Further, each channel is relevant for channels with concepts ``low, rising signal'' (41) and ``highly peaking signal'' (1) in the higher level layer \texttt{features.6}. The concept composition is illustrated by the relevance flow in the form of an attribution graph.}
    \label{fig:appendix:experiments:ecg}
\end{figure*}

The input sequence in Supplementary Figure~\ref{fig:appendix:experiments:ecg} depicts the end of a premature ventricular contraction (PVC) or ventricular escape beat that was successfully detected by the model. A PVC is characterized by a wide QRS complex accompanied by a a left bundle-branch block pattern. In this case, a tall R wave and negative S wave, respectively, a high amplitude value followed by a valley is indicative~\cite{haist2002clinician, akdemir2016premature}. Investigating the concepts most relevant for classification, we perform two local analyses on two regions of interest. We can see that the filters used by the model correspond to the input pattern at the chosen region. The attribution graph, depicting the relevance flow from higher layer concepts in layer \texttt{features.6} to lower level concepts in layer \texttt{features.0}, illustrates how filter 1 is composed of two concepts corresponding at one hand to the upper decline and secondly to the low voltage valley. Channel 41 in \texttt{features.6}, responsible for detecting a high amplitude followed by a valley, is most relevant for the PVC category on average, indicating that the model correctly identified the distinctiveness of a PVC\footnote{Time dependency between beats plays also a role that is not apparent from the shown visualization.}. In the same way, \gls{xai} has the potential to be used for scientific discovery, \eg, for revealing previously unknown characteristics of known phenomena \cite{davies2021advancing, preuer2019interpretable, toms2020physically}. However, the filter corresponds only to 4.4 \% of total absolute relevance representing a fraction of the decision process. 

In summary, we demonstrated the applicability to data in non-image-domains and we have taken the first steps towards testing models for plausibility and scientific discovery.

\cleardoublepage
\clearpage

\section{Challenges and Future Work}
\label{sec:appendix:limitations}
While we qualitatively show in the paper, and quantitatively verify in the conducted user study (Section~\ref{sec:results:study} in the main manuscript), that our \gls{crc}- and \gls{rmax}-based explanations are able to deliver more informative explanations compared to their attribution-only counterparts, there are still certain difficulties to overcome in future work.
While some of those difficulties are related to the choices a user can (and has to) make when using our approaches, others may be solved algorithmically in the future.

\subsection{Interpreting Reference Samples}
As it is common practice with Feature Visualization methods in the literature (see Section~\ref{sec:appendix:relatedwork:global}), reference samples are typically intended to be interpreted by the human user.
We address certain shortcomings of the in related work widely used \gls{amax} approach  for selecting or generating reference examples, \eg the task-independency of activation scores or activations ``overshooting'' beyond for the model useful levels.
While our proposed \gls{rmax} measure addresses some of these concerns, it might still by very challenging to uncover the ``gist'' of a learned latent encoding as represented by a limited set of either not very, or even too varied reference examples (see Section~\ref{sec:appendix:experiments:qualitative:variety}).
The shown examples might at times simply be too abstract to be palpable with human intuition (see Section~\ref{sec:appendix:difficult}),
as there is no guarantee that a neural network will categorize the world and abstract its concepts the way we humans prefer.
With \gls{rmax} we complement
the widely used concept ``stimulation'' technique embodied by the many variations of \gls{amax} with a measure for concept ``utility''.
Nonetheless, we argue that an incorporation of constraints towards ``image simplicity''~\cite{mai2012detecting} or measures related to concept ``clarity''~\cite{nguyen2016multifaceted} might be beneficial to the human interpretability of the models' internal representations.

In general, the degree of human-interpretability can vary depending on the model architecture, training procedure (training steps, dataset size, etc), or from layer to layer.
In our experience, concepts are more distinct in models with higher accuracy as well as in VGG-like models compared to, \eg, ResNet models.
Some models may rely on more neurons than others to encode similar concepts, potentially increasing robustness and redundancy.
Such an effect can occur with, \eg,  dropout layers \cite{zunino2021excitation} used during training.
With multiple filters redundantly or in combination encoding highly similar or compound concepts, a meaningful grouping or clustering of filters
(see Section~\ref{sec:experiments:quantitative:clusters} in the main manuscript and Section~\ref{sec:appendix:experiments:quantitative:clusters} for an outlook) may
therefore further ease the task of interpreting machine learning explanations.

Contemporary work such as Concept Whitening~\cite{chen2020concept} tries to increase the clarity and distinctness of concepts with a specialized training procedure by aligning neuron activations to concepts predefined by humans.
MILAN~\cite{hernandez2022natural} on the other hand, approaches the interpretation of latent encodings in a data-driven way by training an image-to-text model applicable to reference images in order to provide suggestions for language descriptors, as an option to  provide more objective interpretations of concepts, instead of potentially per-user unique readings.

Since \gls{crc} and \gls{rmax} are out-of-the-box applicable to (almost) any model without imposing constraints on the training process, the data and label availability, or the model architecture, the aforementioned techniques or other future concept enhancing methods can be integrated within the \gls{crc} framework.

\subsection{Degrees of Freedom in Glocal Analyses and the Effects User Workload}
One important point to consider when it comes to the application of \gls{xai} is the workload imposed on the user.
With the exponential amount of parameters to consider and information to process with glocal explanations compared to traditional local approaches, we see this as a major issue when it comes to the effective utilization of this kind of \gls{xai} feedback.
In the following we will briefly highlight some of the encountered issues with glocal \gls{xai}, and how to potentially mitigate them in future work.

\paragraph{Where to start a glocal analysis?}
Compared to local methods such as \gls{lrp} or IG, for instance, \gls{crc} explanations offer a much higher potential for model understanding by providing the possibility to interpret how concepts are used in all layers of the model architecture.
However, the large number of layers and neurons to investigate can be overwhelming at first, which, \eg, reflects in the relatively low ``clarity'' scores of \gls{crc} \& \gls{rmax} in our user study (see Supplementary Note~\ref{sec:appendix:study}), despite the methods' leading success in informing the user to solve the study's primary task.

We find that a good starting point for the investigation of concept-based explanations usually are higher level layers, as they typically are comprised of more specialized and class-specific, as well as abstract concepts.
From then on, a hierarchical analysis of concept composition might be followed, as it has been briefly introduced in this work (see Supplementary Figure~\ref{fig:experiments:localize-concepts:concept-atlas}d in Section~\ref{sec:experiments:local:analysis} of the main manuscript), yet not explored to the full extent.
Still, with the expansive amount of potential interactions of encodings, manual analysis might become infeasible quickly, and an algorithmic approach to present a condensed explanatory report (at least as an initial summary) to the user might be required.

\paragraph{The pain of choosing the right attribution variant}
In theory, \gls{lrp} --- and therefore also \gls{crc} --- can be applied to any model which can be presented in form of a directed acyclic graph,  which includes all common \gls{dnn} architectures. However, since the computational steps in different architecture types may change significantly, purposed implementations of \gls{lrp}-rules and model canonizations (see Section~\ref{sec:appendix:technical}) may have a positive impact on the quality of the explanations obtainable~\cite{kohlbrenner2020towards,motzkus2022measurably,hedstrom2022quantus},
and very well-tuned parameterizations for \gls{lrp} and in extension \gls{crc} might be desired in order to optimize the expressiveness of the glocal explanations as much possible.
Hence, Arras et al. \cite{arras2017explaining} specifically adapt \gls{lrp} to analyse LSTMs, while Chereda et al. \cite{chereda2021explaining} and Schnake et al. \cite{schnake2021higher} modify \gls{lrp} for Graph CNNs. Despite research endeavors such as \texttt{zennit}~\cite{anders2021software}, at the time of publication, no all-encompassing library exists that implements all \gls{lrp} variants exhaustively. Alternatively, other backpropagation-based attribution methods can easily be utilized as well within the broader scope of the \gls{crc} framework, yielding different measurements on a model's inference process and thus extending the option space of the user,
but also the agony of choice. %

\paragraph{What to investigate on a concept level?}
Next to the framework-specific choice to make \emph{where} (within the model) to start the analyses, and \emph{how} to assess feature importance via the choice of an attribution method, there is the parameter-dependent choice of \emph{what} to analyze.
We have proposed the \gls{rmax} quantity measuring ``utility'' complementarily to ``feature stimulation'' as measured by \gls{amax}.
While alternative measures such as a potential Sensivity~\cite{morch1995visualization,baehrens2010explain} Maximization measuring the volatility of the model around a given reference example, the wealth of (the product space of) choice here again might become overwhelming quickly, even if one considers that each measure usable for procuring reference examples may generate an output representing a single distinct aspect of an \gls{ml} model's behavior.
That even the restricted glocal explanations we have shown to the participants of our user study can be demanding already is apparent from the study results:
even though the presented explanations based on \gls{crc} and \gls{rmax} have been demonstrated to be most effective for the goal of understanding model behavior,
they perform worst in terms of user-reported explanation clarity.

In the \gls{xai} research domain, there is an emerging understanding on the criticality of objectively measuring different aspects of explanation quality quantitatively~\cite{hedstrom2022quantus, agarwal2022openxai}. 
While this is currently still difficult terrain, at times under-specified and overall not well-explored even when dealing with explanations on a single-heatmap-level, quantitative measures to algorithmically tune glocal explanations might
be an option to optimize the user-presented information for optimal technical parameters, to be at the same time as informative concise as possible.

\cleardoublepage
\clearpage
\section{Datasets, Models and Implementation} %
\label{sec:appendix:dataset_and_models}
In the following, we provide a brief overview over the various datasets and models used throughout our experiments. Furthermore, we provide technical details and recommendations regarding our Concept Relevance Propagation approach.

\subsection{Datasets and Models}
\paragraph{ImageNet}
\label{ap:vgg16}
We utilized models pretrained on ImageNet \cite{russakovsky2015imagenet} including the VGG-16 \cite{simonyan2015very} architecture with and without BatchNorm layers provided by the PyTorch \cite{paszke2017automatic} library.

\paragraph{Caltech-UCSD Birds 200}
\label{ap:resnet34}
We fine-tuned a ResNet34 \cite{he2016deep} pretrained on ImageNet on the Caltech-UCSD Birds 200 \cite{welinder2010caltech} dataset. Images are normalized with ImageNet values\footnote{Subtraction of mean values [0.485, 0.456, 0.406], followed by a normalization of standard deviation values [0.229, 0.224, 0.225] over the red, green and blue color channels respectively, as per specifications at \url{https://pytorch.org/vision/stable/models.html}.}. Further,  Gaussian noise (zero mean, standard deviation of 0.05), random horizontal flips (probability of 0.5), random rotation (up to 10 degrees), random translation (up to 20\,\% in all directions) as well as a random scaling (between 80 and 120\,\%) is applied for data augmentation. After 100 epochs of training using Stochastic Gradient Descent (SGD) with a value for momentum of 0.9 and weight decay of 1e-4, we achieved 76 \% top-1-accuracy on the test set.

\paragraph{ISIC 2019}
We employ the VGG-16 model with pretrained weights on ImageNet from the PyTorch model zoo to train on the ISIC 2019 dataset. The model is trained over 100 epochs, using an SGD optimizer with a learning rate of 0.001, momentum 0.9 and batch size of 32. ISIC 2019 does not offer a pre-defined labeled test set. Therefore, 10\,\% of the original training set is split off in order to evaluate the model's performance. The model achieves a final test accuracy of 82.15\,\%.

\paragraph{Adience}
\label{ap:adience}
We fine-tuned two VGG-16 models  without BatchNorm layers pretrained on ImageNet on the Adience dataset \cite{eidinger2014age} for age estimation as well as sex prediction. For training, folds 0,1,2 and 3 were selected and fold 4 was used as as a test set. Images are normalized with ImageNet values and linear transformations were applied as augmentation techniques. After training for 10 epochs with the Adam optimizer and a batch size of 32 with a learning rate of $2\cdot 10^{-4}$ as well as standard hyperparameter values in the PyTorch library, we achieved a top-1-accuracy of  51.0\,\% (age estimation) and 93.6\,\% (sex prediction) on the test set.

\paragraph{PhysioNet MIT-BIH Arrhythmia}
\label{ap:ecg}
The original training data is provided by PhysioNet MIT-BIH Arrhythmia \cite{moody2001impact} and consists of ECG recordings from 47 different subjects, where each beat was annotated by at least two cardiologists. The work \cite{kachuee2018ecg} reordered the dataset by isolating the ECG lead II data, splitting and padding beats into a fixed size of 150 at a sampling rate of 125 Hz and finally grouping all annotations into five different categories in accordance with the \textit{Association for the Advancement of Medical Instrumentation} (AAMI) EC57 standard \cite{a1998testing}. Preprocessing steps include subtracting the mean value of 0.2115 and standardizing by dividing with 0.2462. As optimizer, RMSprop \cite{hinton2012neural} with a learning rate of 1e-4 and weight decay of 1e-6 were chosen. Since the data set is unbalanced, we used cross entropy loss with class balancing. We trained the model for 25 epochs and achieved on the test set an average accuracy of 89 \%. Deep Residual CNN models achieve in comparison 93.4 \% average accuracy \cite{kachuee2018ecg}. The model consists of three feature extraction blocks, each composed of a 1D-Convolutional layer with 64 channels and kernel size 7, followed by a ReLU non-linearity and a 1D-Max-Pooling layer with kernel size 3 and stride 2. After feature extraction, a 1D-Average-Pooling layer with kernel size 7 and a flatten operation direct the activations into a regression head. The regression head consists of two linear layers with 64 and 32 neurons as well as a ReLU non-linearity. Finally, a linear output layer with 5 neurons concludes the architecture.

\paragraph{Fashion-MNIST}
\label{ap:lenet}
A LeNet-5 \cite{lecun1998gradient} was trained on the Fashion-MNIST \cite{xiao2017fashion} dataset. Preprocessing included only normalizing the input interval into the range of [0,1]. As optimizer, Adam was chosen with standard PyTorch parameters. After training for 5 epochs, a top-1-accuracy of 88 \% on the test set was achieved.

\paragraph{Human Study ImageNet Models}
In the human study presented in Section~\ref{sec:results:study},
two fine-tuned VGG-16 models pretrained on ImageNet are used.
Both models are trained for one epoch using a learning rate of $5\cdot 10^{-6}$ with the Adam optimizer and a batch size of 32.
Images during training are resized to 256\,px for the smaller edge, followed by a center crop to 224\,px$\times$224\,px size and normalization with means of [0.485, 0.456, 0.406] and standard deviations of [0.229, 0.224, 0.225] over the red, green and blue color channels respectively.
Finally,
for the first model,
all images are padded with a black border of 10\,px width and again re-scaled to 224\,px$\times$224\,px size.
For the second model,
only images from 30 randomly chosen classes are displayed with the border artifact during training.

\edited{}{
\paragraph{Image Licenses}
The license to re-use and reproduce have been granted for the images shown in the figures of this supplementary material and the main manuscript to the authors by the respective copyright holders by iStock, Shutterstock, Pixabay and Pexels. 

}
\subsection{Implementation Details for Concept Relevance Propagation}
\label{sec:appendix:technical}
\paragraph{Distribution Rules}
In this work, we compute \gls{crc} attribution scores for ResNets and VGG-16 models using the recommended~\cite{kohlbrenner2020towards} composite LRP$_{\varepsilon-{z^+}-\flat}$-rule\footnote{read: "epsilon-plus-flat" rule} after model canonization steps applied~\cite{hui2019batchnorm,guillemot2020breaking,motzkus2022measurably} as implemented in the \texttt{zennit} \cite{anders2021software} package for PyTorch \cite{paszke2019pytorch}. The \gls{lrp}$-\varepsilon$-rule is defined as
\begin{align}
R^{(l,\: l+1)}_{i \leftarrow j} & =  \frac{z_{ij}}{\preact_j + \varepsilon \cdot \text{sign}(\preact_j)}R_j^{l+1}
\end{align}
where $z_{ij} = w_{ij} x_i$ with inputs $x_i$ from the lower layer and weights $w_{ij}$ of the linear layer with outputs $\preact_j = \sum_i w_{ij} x_i$. It ensures that each neuron receives the attribution value, that it contributed to the output. The added $\varepsilon \in \mathbb{R}^+$ stabilizes the division of the denominator. Note that for this purpose, the definition of the sign function is altered such that $\text{sign}(0) = 1$. The $\varepsilon$-rule is in ReLU networks highly similar to the multiplication of the input times its gradient~\wrt the output~\cite{kohlbrenner2020towards}, but not identical, leading to different attribution maps~\cite{motzkus2022measurably}. These attribution values for different predictions can now, in principle, be compared.  However, due to its similarity to gradient-based attribution computation, the \gls{lrp}$-\varepsilon$-rule may result in noisy attributions in very deep models, where gradient shattering and noisy gradients appear\cite{balduzzi2017shattered}. Thus, the computation of the \gls{lrp}$-\varepsilon$-rule may not result in meaningful attributions in models with a high number of layers and ReLU activations. An alternative approach is to combine multiple propagation rules, so-called composite rules, which result in attributions that are less influenced by a noisy gradient: the LRP$_{\varepsilon-{z^+}-\flat}$-rule is an established best practice \cite{kohlbrenner2020towards, montavon2019layer} to keep LRP attribution maps informative, readable and representative while combating gradient shattering effects. Here, applying the LRP-$\flat$-rule to the first convolutional layer smoothes the attribution map in input space and yields an invariance against input normalization effects. The LRP-$z^+$-rule, on the other hand, operates on the remaining convolutional layers and \gls{lrp}$-\varepsilon$-rule is utilized in standard MLP layers. The LRP-$\flat$-rule is defined as
\begin{align} 
R^{(l,\: l+1)}_{i \leftarrow j} & =  \frac{1}{\sum_i 1}R_j^{l+1}
\end{align}
Using the $\flat$-rule, relevance of upper-level neuron $j$ is equally distributed to all connected lower-level neurons disregarding any influence of learned weights or input features. The LRP-$z^+$-rule is given as
\begin{align} \label{eq:appendix:zplus-rule}
R^{(l,\: l+1)}_{i \leftarrow j} & =  \frac{(w_{ij} x_i)^+}{\preact_j^+}R_j^{l+1}
\end{align}
by only taking into account positive contributions $\preact_j^+ = \sum_i (w_{ij} x_i)^+$ with $(\cdot)^+ =\max(0, \cdot)$.

\paragraph{Initialization}
The calculation of an attribution map starts with an initial relevance value at the output neuron. A meaningful and canonical choice is the model output activation of the neuron representing the class of interest to be explained. All intermediate relevance values are then proportional to the initial relevance value, since LRP is a layer-wise linear distribution process, \cf Equation~\eqref{eq:appendix:lrp_basic}. We recommend to start propagation with the output values as they are, \ie, $R^L_j = f_j(\x)$, for an output (set) $j$ of choice. Thus, the relevance of all neurons is proportional to the confidence of the model prediction. Note that a selection of more than one, output neuron to initialize the $R^L_j$ might lead to unclear attribution-based explanations, as shown in Supplementary Figure~\ref{fig:appendix:methods:lrp:entangle}. As a result, when ranking the samples of a dataset for representing the concept of a filter, \gls{rmax} will give preference to samples that the model predicted with highest confidence. This behavior is advantageous when calculating class-conditional reference samples or when the original dataset is \emph{not} expanded with additional datasets as demonstrated in (see Section~\ref{sec:appendix:experiments:qualitative:variety}).
\begin{figure*}[t]
    \centering
    \includegraphics[width=1\linewidth]{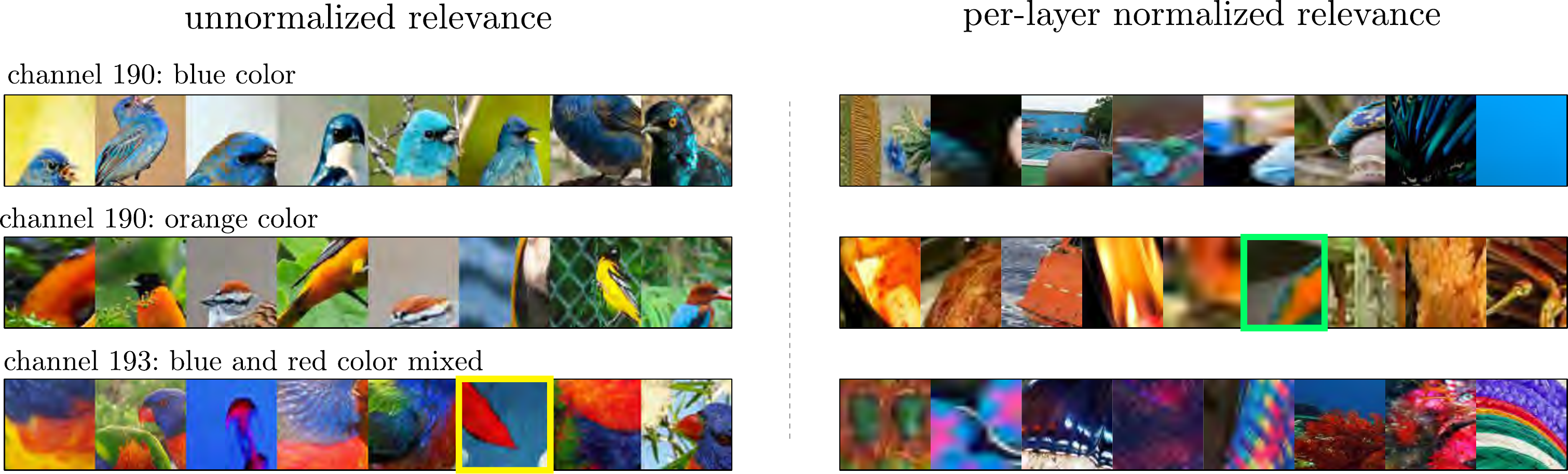}
    \caption{Reference samples selected using \gls{rmax} with different normalization techniques. Depicted are channel 190, 117 and 193 in layer \texttt{layer3.0.conv1} of a ResNet34 trained on Caltech-UCSD Birds 200. The ${\mathcal{X}^{\star}_{8}}^{\text{rel}}_{\text{max}}$  are drawn from the Birds dataset, merged with ImageNet data. \emph{(Left)}: Reference Samples selection without layerwise normalization favors samples that the model predicted with high confidence. Samples from ImageNet dataset are marked with a yellow border. \emph{(Right)}: Using per-layer relevance normalization, samples outside the training distribution are taken into account more often. Samples from Caltech-UCSD Birds 200 dataset are marked with a green border.}
    \label{fig:appendix:normalization_effect}
\end{figure*}
\paragraph{Canonization}
Since rule-based attributions, such as \gls{lrp} and \gls{crc}, are not implementation invariant, canonization restructures a model into a functionally identical equivalent of alternative design to which established attribution rules can be applied in an improved manner. Canonization efforts usually concentrate on replacing the BatchNorm layer \cite{hui2019batchnorm, guillemot2020breaking} or the handling of skip connections~\cite{binder2020notes}. By merging BatchNorm layers into a neighboring linear layer, significant improvements in terms of explanations can be achieved \cite{motzkus2022measurably}.

\paragraph{Per-layer Relevance Normalization}
We generally recommend starting the relevance backpropagation process with the output of a network element. However, when reference datasets are extended in order to gain additional perspectives on the concepts to be understood, out-of-distribution data (or data the model has not sufficiently fit to) will produce low output confidences which serve as low initial relevance values. This will result in the added reference data from alternative source to not be ranked highly as meaningful representations for latent concepts. In order to be in this case independent of the model confidence of individual predictions, the relevance attributions $R^{(l)}(\x|\theta_{c})$ with condition $\theta_c$ can be normalized at each layer $l$ as
\begin{equation}
R^{(l)}_{\text{norm}}(\x|\theta_{c}) 
=
\frac{R^{(l)}(\x|\theta_{c})}{\sum_k |R^{(l)}_k(\x|\theta_{c})|}
\end{equation}
Thus, individual attributions are first divided by the absolute sum of relevances at layer $l$ to ensure that each attribution is bounded in $[-1, 1]$. The concept importance becomes independent of the output of the model, while maintaining class-specificity. This procedure facilitates the comparison of concept importance for different samples during \gls{rmax}, if the original dataset is expanded with additional, external datasets.
In our case, ImageNet images are fed into a model trained for bird categorization only. These ImageNet images will then be considered more frequently as representations for latent concepts with per-layer normalization than without layerwise normalization. It is to note, that class conditional relevance computation \emph{without ground truth labels} has to be applied with caution, because the natural filter functionality, that the model confidence provides, is lost via per-layer relevance normalization. However, if (matching) ground truth labels are available, class conditional relevance could be computed for true labels only instead.

The qualitative difference of reference samples selected using \gls{rmax} with and without per-layer relevance normalization is illustrated in Supplementary Figure~\ref{fig:appendix:normalization_effect}. Here, layer \texttt{layer.3.0.conv1} of a pretrained ResNet34 model trained on Caltech-UCSD Birds 200 is analyzed. The dataset is extended with additional ImageNet samples to increase reference sample variety. Supplementary Figure~\ref{fig:appendix:normalization_effect} \emph{(left)} depicts reference samples selected without per-layer relevance  normalization, \ie, favoring samples that achieve highest model confidence. As a consequence, samples that belong to the original training data are visualized more often, illustrating how the filter is \emph{actually} utilized. On the other hand, Supplementary Figure~\ref{fig:appendix:normalization_effect} \emph{(right)} depicts reference samples selected with per-layer relevance normalization, \ie selecting samples independent of model confidence. As a result, samples outside the training data are shown more often, providing examples more focused on the concept itself and with less distracting information increasing model output confidence in context of other concepts.
The choice of applying layer-wise relevance normalization provides the option for an additional perspective on the visualized concept representations

\paragraph{Practical Implementation}
The library \texttt{zennit} \cite{anders2021software} was used to perform the experiments. It takes advantage of PyTorch's autograd module to compute gradients in computational graphs. \texttt{zennit} modifies the backpropagated gradient in such a way that they are replaced by attribution values for  \gls{crc} instead. We provide a \texttt{zennit}-based open source %
package for \gls{crc} and \gls{rmax} computations for PyTorch and Python at \url{https://github.com/rachtibat/zennit-crp}.

\end{document}